\relax
\documentclass[letterpaper]{article} % DO NOT CHANGE THIS
\usepackage{aaai22}  % DO NOT CHANGE THIS
\usepackage{times}  % DO NOT CHANGE THIS
\usepackage{helvet}  % DO NOT CHANGE THIS
\usepackage{courier}  % DO NOT CHANGE THIS
\usepackage[hyphens]{url}  % DO NOT CHANGE THIS
\usepackage{graphicx} % DO NOT CHANGE THIS
\usepackage{natbib}  % DO NOT CHANGE THIS AND DO NOT ADD ANY OPTIONS TO IT
\usepackage{caption} % DO NOT CHANGE THIS AND DO NOT ADD ANY OPTIONS TO IT
\DeclareCaptionStyle{ruled}{labelfont=normalfont,labelsep=colon,strut=off} % DO NOT CHANGE THIS
\usepackage{amsmath}
\usepackage{booktabs}
\usepackage{multirow}
\usepackage{xcolor}
\usepackage[
 font=normalsize % or small or scriptsize
 ]{subfig}
\usepackage{float}
\usepackage{array}
\usepackage[export]{adjustbox}
\usepackage{multicol}
\usepackage{enumitem}
\title{Semantic-guided Automatic Natural Image Matting with Trimap Generation Network and Light-weight Non-local Attention}
\author {
    Yuhongze Zhou,\textsuperscript{\rm 1}
    Liguang Zhou, \textsuperscript{\rm 2,3}
    Tin Lun Lam, \textsuperscript{\rm 2,3} \thanks{Corresponding Author}
    Yangsheng Xu \textsuperscript{\rm 2,3}
}
\affiliations {
    \textsuperscript{\rm 1} McGill University\\
    \textsuperscript{\rm 2} The Chinese University of Hong Kong, Shenzhen\\
    \textsuperscript{\rm 3} Shenzhen Institute of Artificial Intelligence and Robotics for Society\\
    yuhongze.zhou@mail.mcgill.ca, \{liguangzhou@link., tllam@, ysxu@\}cuhk.edu.cn
}

\begin{document}

\maketitle

\begin{abstract}
Natural image matting aims to precisely separate foreground objects from background using alpha matte. Fully automatic natural image matting without external annotation is challenging. Well-performed matting methods usually require accurate labor-intensive handcrafted trimap as extra input, while the performance of automatic trimap generation method of dilating foreground segmentation fluctuates with segmentation quality. Therefore, we argue that how to handle trade-off of additional information input is a major issue in automatic matting. This paper presents a semantic-guided automatic natural image matting pipeline with Trimap Generation Network and light-weight non-local attention, which does not need trimap and background as input. Specifically, guided by foreground segmentation, Trimap Generation Network estimates accurate trimap. Then, with estimated trimap as guidance, our light-weight Non-local Matting Network with Refinement produces final alpha matte, whose trimap-guided global aggregation attention block is equipped with stride downsampling convolution, reducing computation complexity and promoting performance. Experimental results show that our matting algorithm has competitive performance with state-of-the-art methods in both trimap-free and trimap-needed aspects.
\end{abstract}
\section{Introduction}
\label{sec:intro}
Image matting is a popular image editing task which attempts to extract perfect foreground object mask, i.e. alpha matte, from background. Matting problem can be formulated in a general mathematical manner. An image $I$ can be defined as a combination weight of alpha matte $\alpha$, foreground $F$, and background $B$ image as follows:
\begin{equation}
    I=\alpha F + (1-\alpha) B,
\end{equation}
where the RGB color $I$ is known, but $F, B$ and $\alpha$ are unknown. That is to say, matting attempts to solve 7 unknown variables with only 3 variables provided. Therefore, most compelling matting methods usually require a handcrafted trimap for region constrain to reduce complexity and assist matte estimation, which makes fully-automatic natural image matting such an appealing task to explore.

\begin{figure}[thpb]
\setlength\tabcolsep{0pt}
\renewcommand{\arraystretch}{0.25}
\begin{center}
% \resizebox{\columnwidth}{!}{
\begin{tabular}{cccc}
\includegraphics[width=2cm]{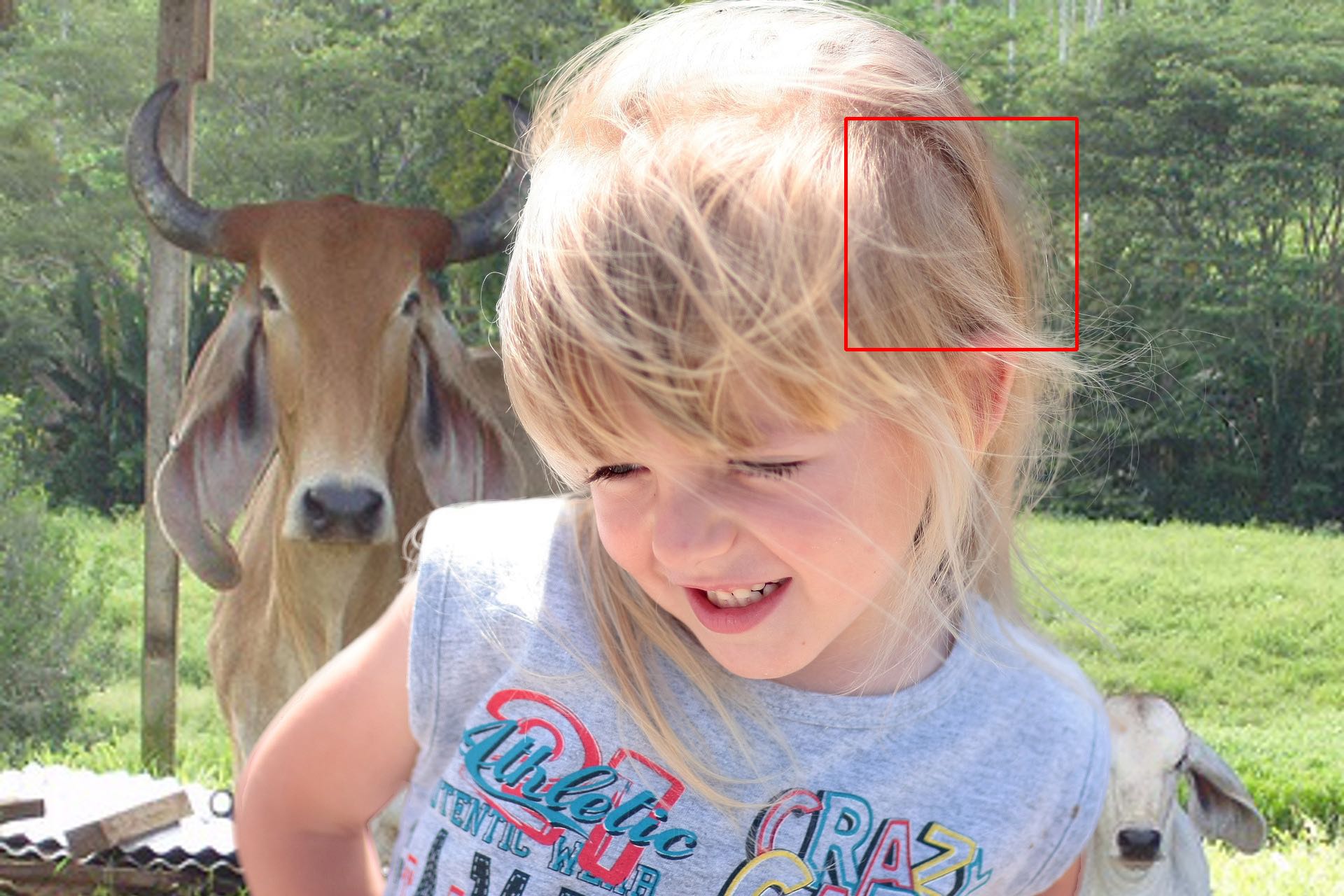}&\includegraphics[width=2cm]{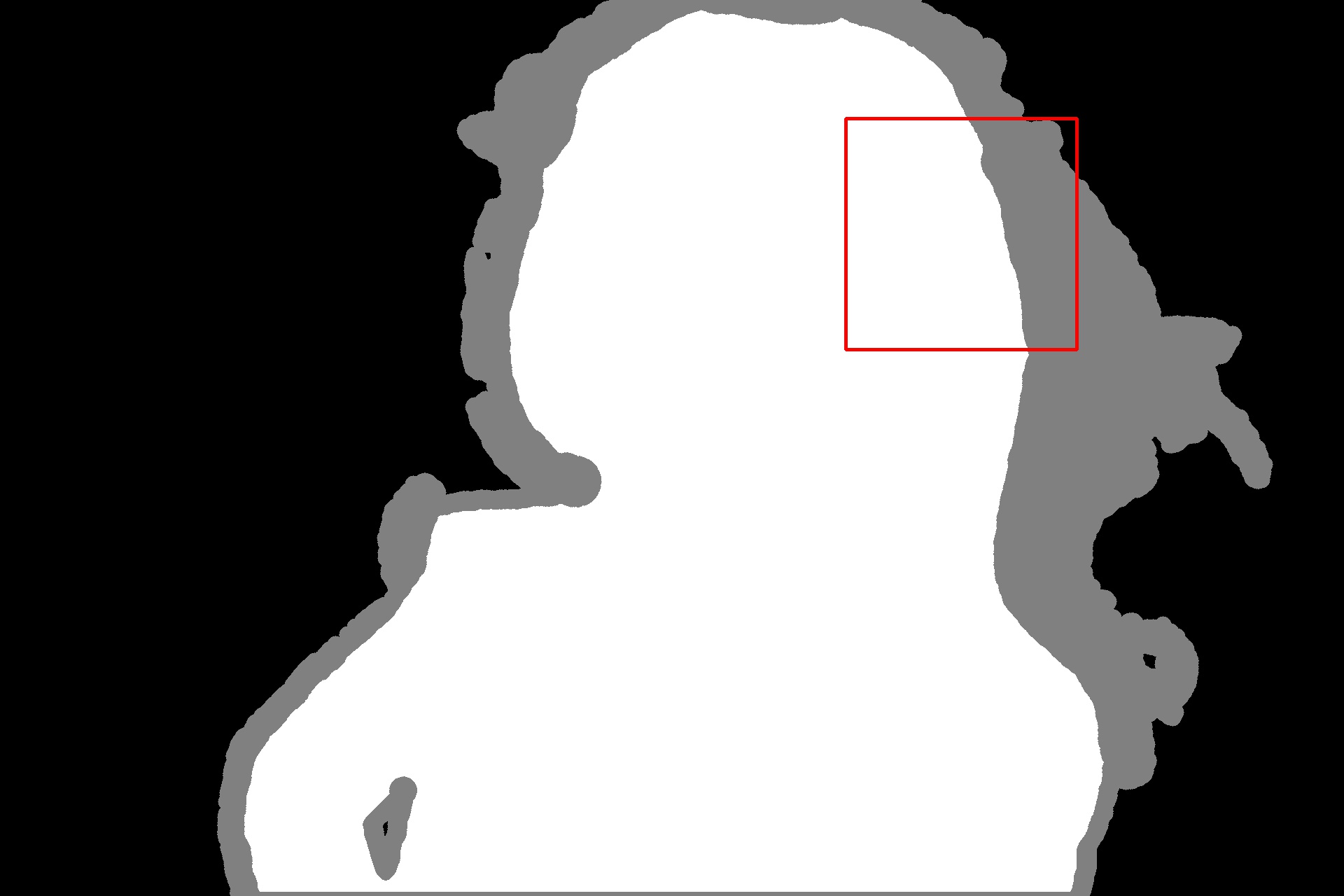}&\includegraphics[width=2cm]{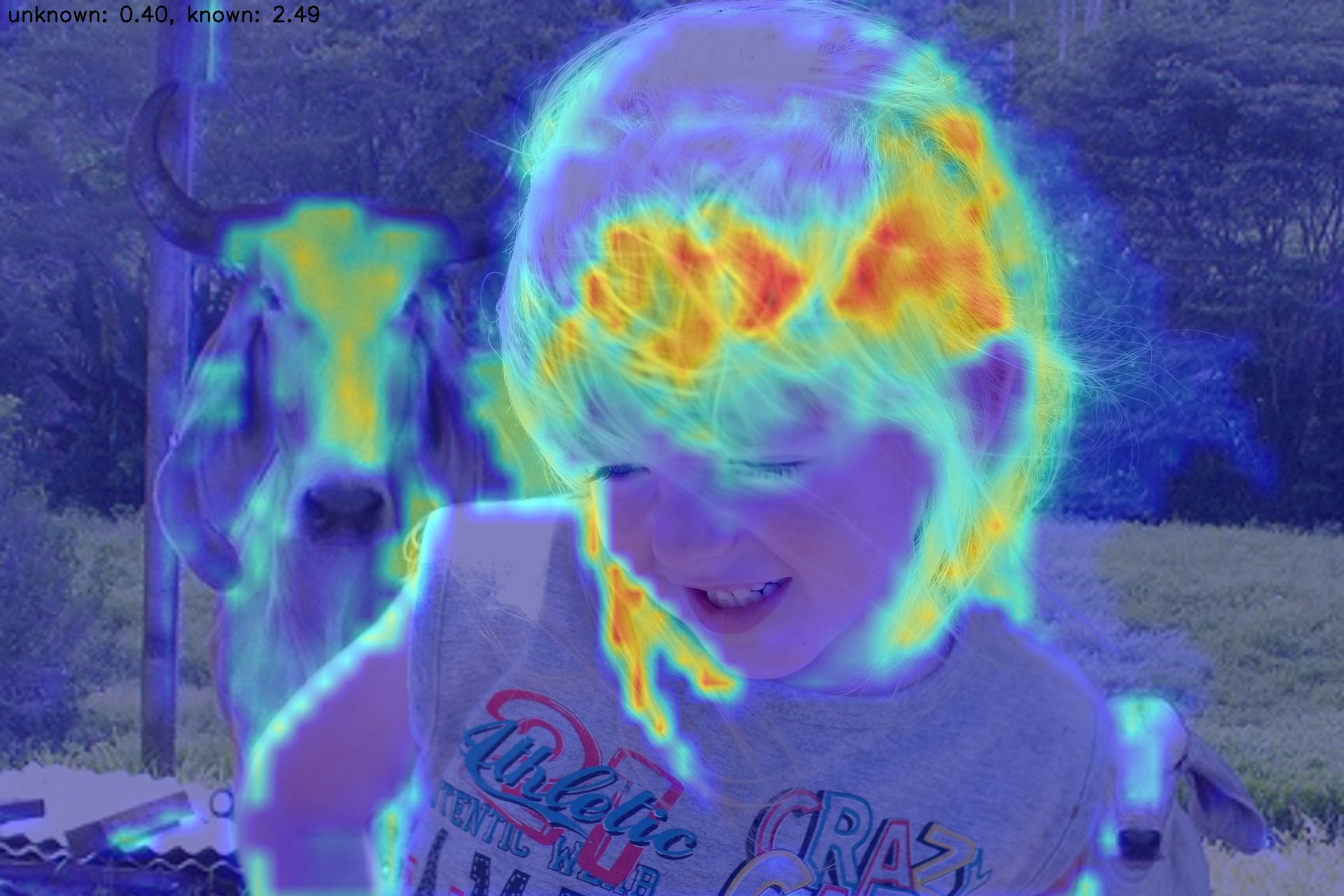}&\includegraphics[width=2cm]{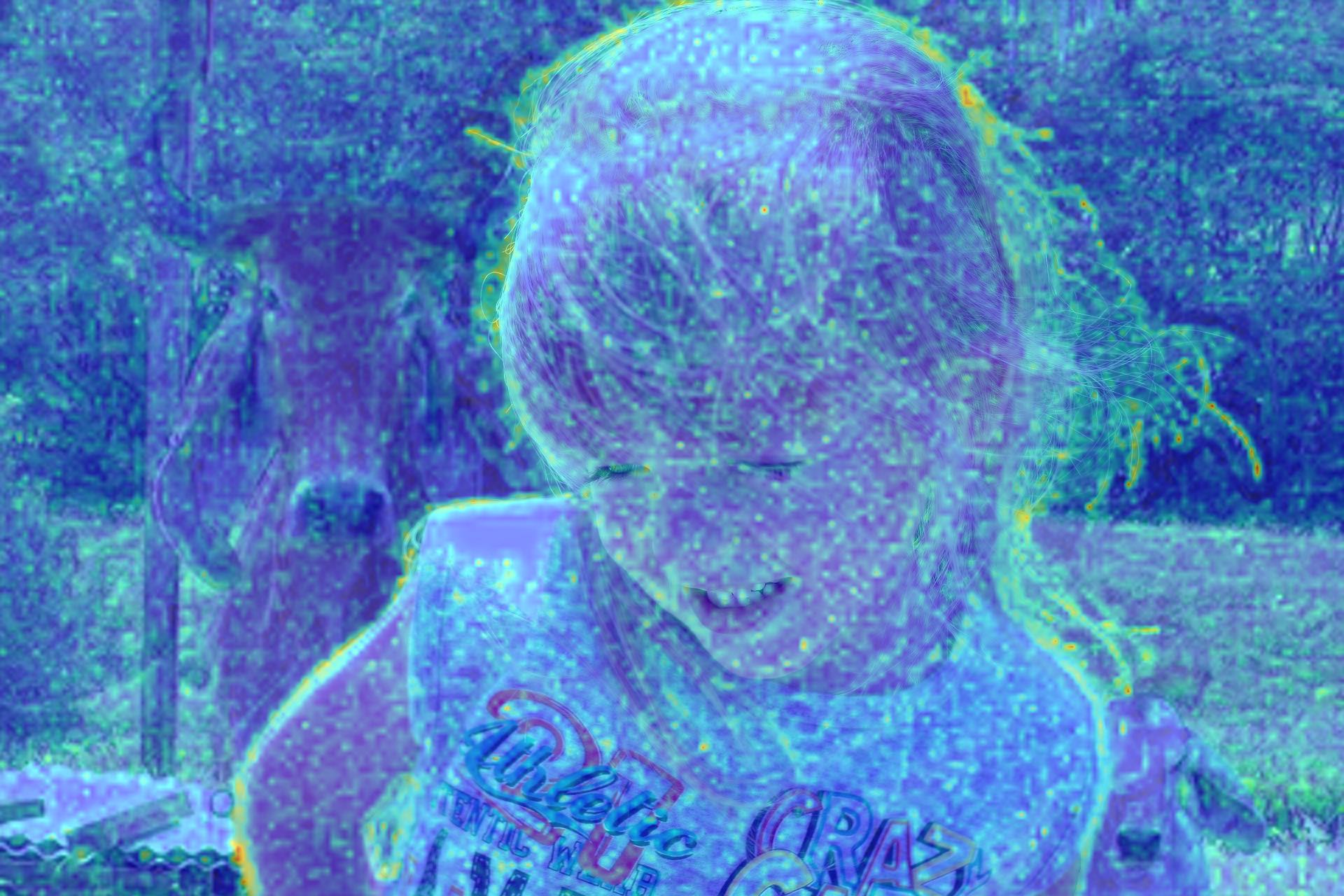}\\
\includegraphics[width=2cm]{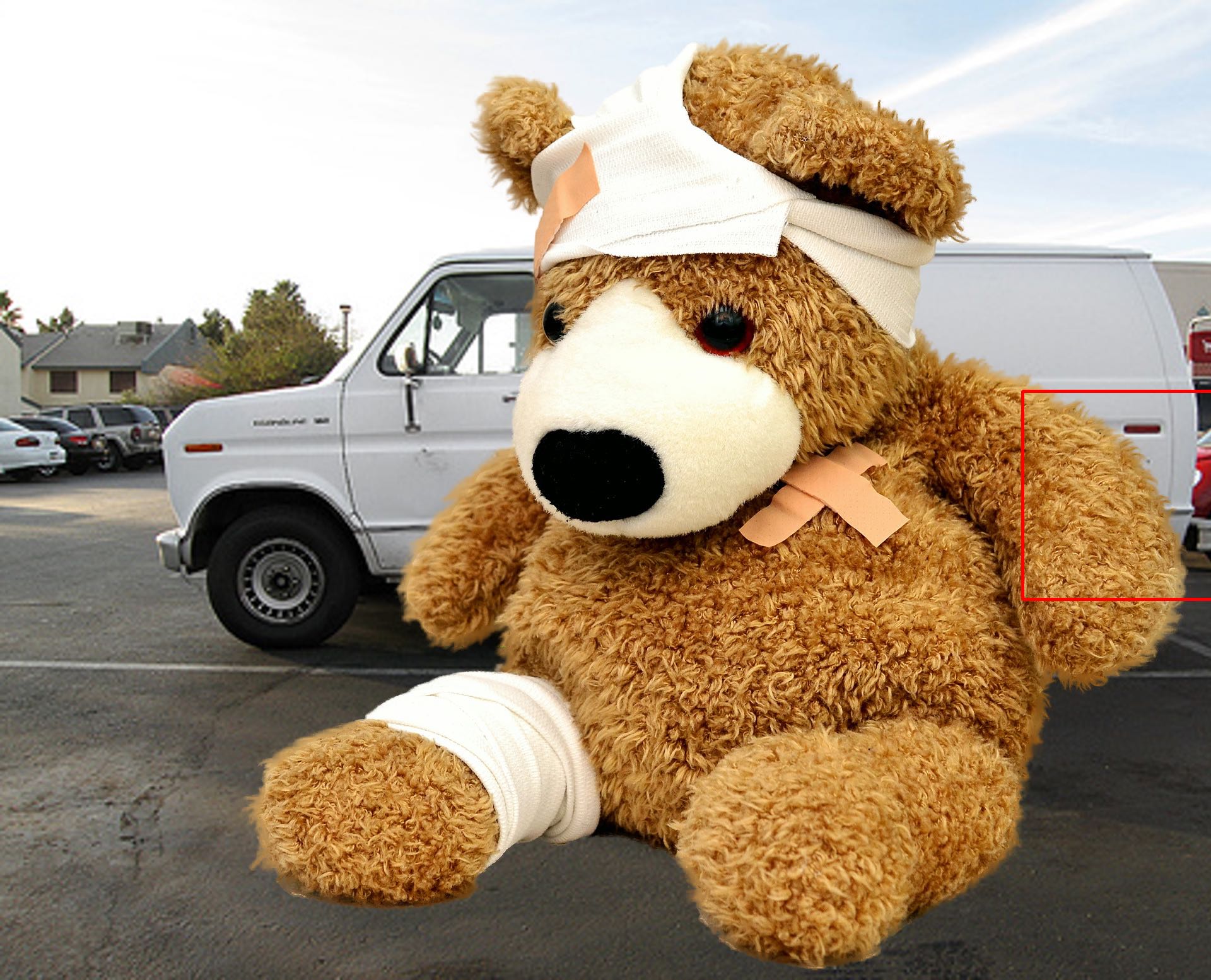}&\includegraphics[width=2cm]{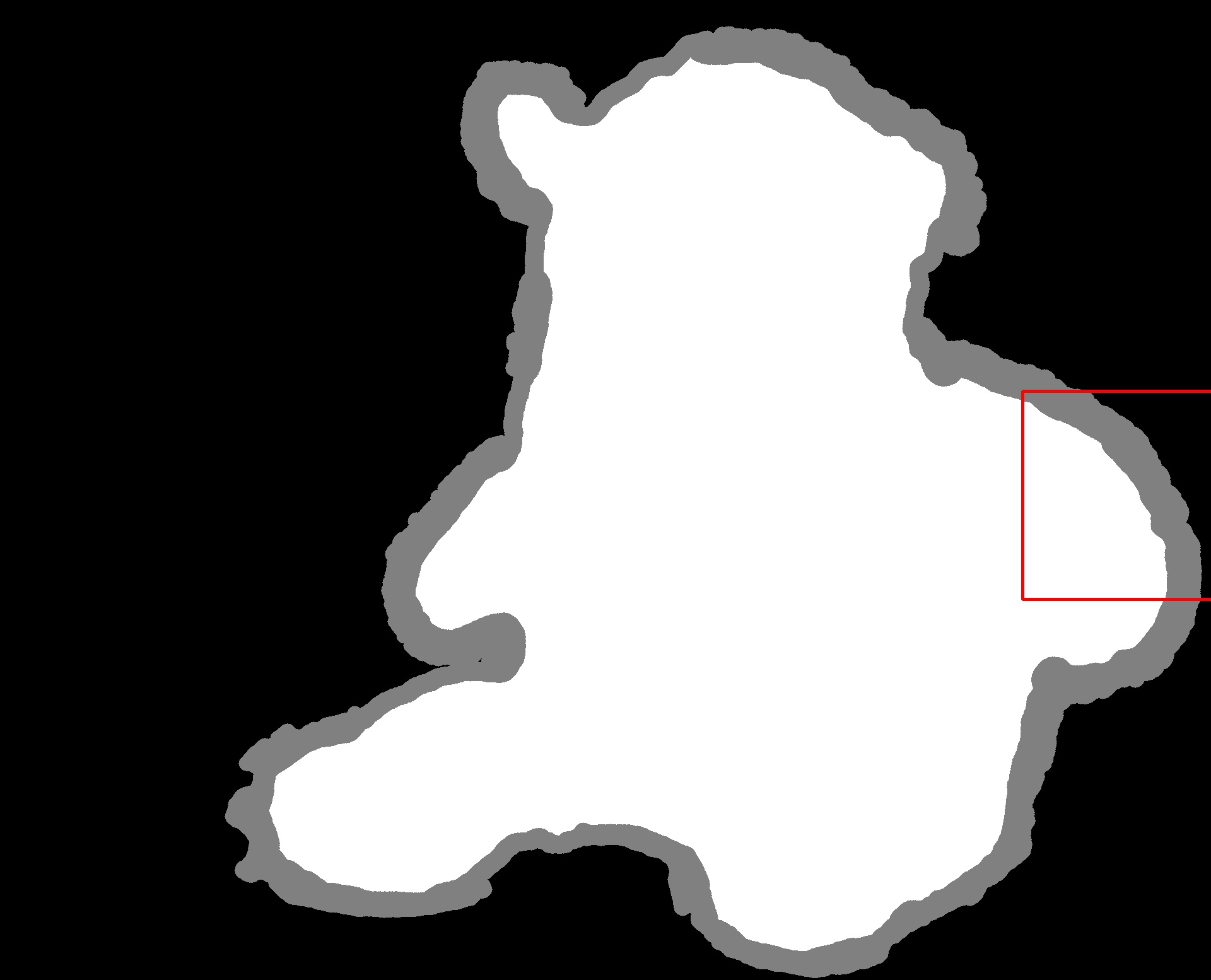}&\includegraphics[width=2cm]{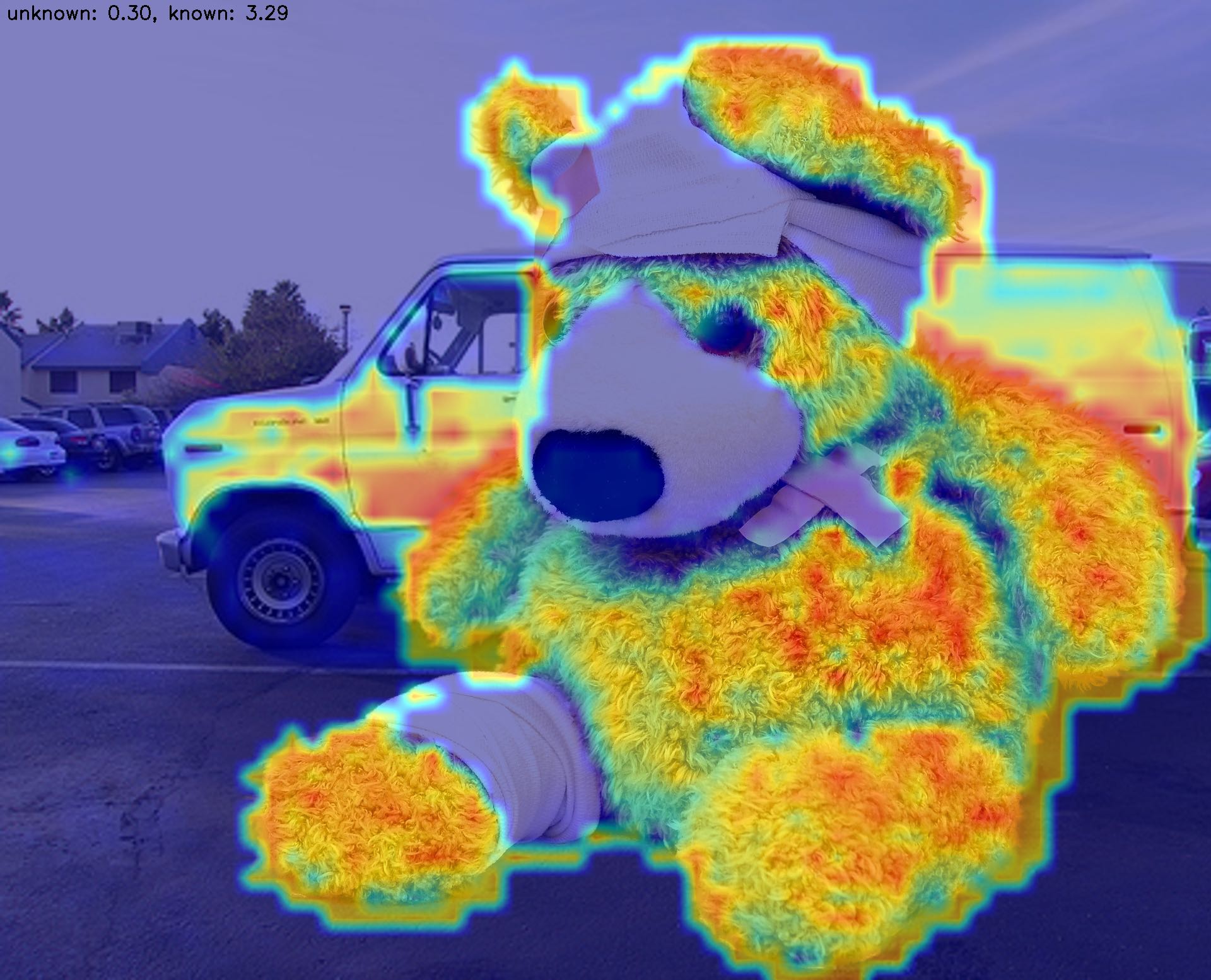}&\includegraphics[width=2cm]{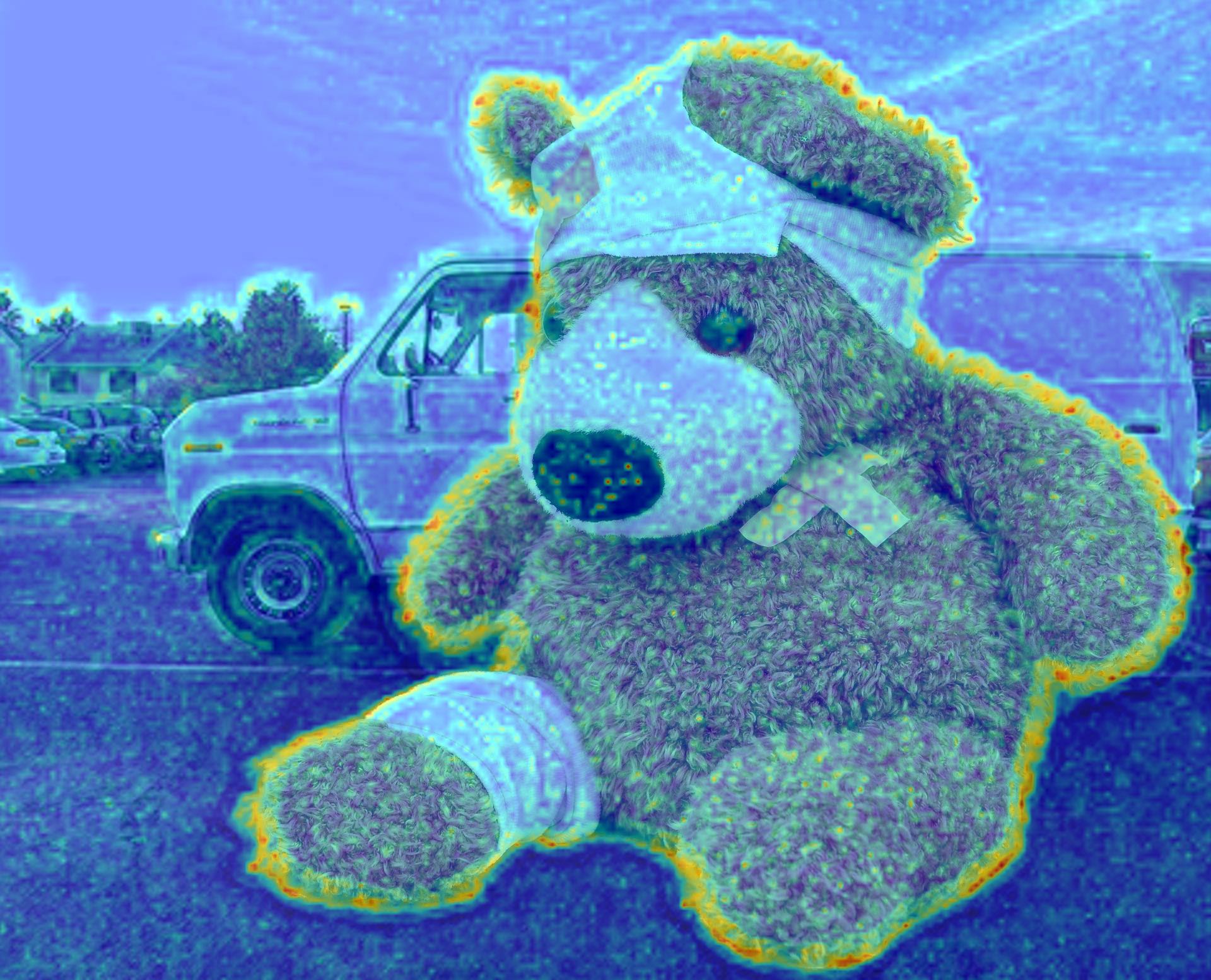}\\
% \includegraphics[width=2.5cm]{./images/real_data/images/VID_20201109_141618_0459_img_blurred.jpg}&\includegraphics[width=2.5cm]{./images/real_data/cam1/VID_20201109_141618_0459.jpg}&\includegraphics[width=2.5cm]{./images/real_data/ours/VID_20201109_141618_0459.jpg}\\
% Image&CAM-1&Net-T-M-Real\\
\end{tabular}
\caption{The visualization of our attention block. From left to right, image, trimap, global attention weight map of given query patch marked by red box, and reconstructed alpha feature of attention block.} 
\label{fig:att_vis}
\end{center}
\end{figure}

Let us recap recent learning-based image matting approaches and their pros and cons. Learning-based image matting can be divided into three primary categories, i.e. background-required~\cite{qian1999video,sengupta2020background}, only-image~\cite{qiao2020attention,zhang2019late,wei2021improved} input, and trimap-needed~\cite{cho2016natural,xu2017deep,lutz2018alphagan,lu2019indices,tang2019learning,li2020natural,sun2021semantic,dai2021learning}.

Recently, novel background matting~\cite{sengupta2020background,lin2021real} is proposed, but it cannot resist interference of shadows or complex light condition. For algorithms requiring only single image~\cite{qiao2020attention,zhang2019late}, results on generic objects are far from practical expectations. Although human matting, one branch of trimap-free matting, has achieved impressive performance~\cite{shen2016deep,chen2018semantic,liu2020boosting,ke2020green}, human is regarded as such a specific domain and salient object that it is easy for network to capture foreground/background discrepancy. So does animal image matting~\cite{li2020end}.

% that makes relation modeling between known and unknown areas possible. 
For trimap-needed matting, its accuracy is paramount, which gives the credit to auxiliary trimap. The trimap provides deterministic foreground, unknown, and background regions of image, which narrows down matte estimation to unknown region and reset pixel values of known region. Besides, the manual creation of trimap is painstaking, which diminishes its application potential. Hence, trimap quality is one significant factor that can affect matting performance. One possible workaround is a general automatic trimap generation method, that is, target foreground items are roughly extracted by semantic segmentation and then processed by image dilation/erosion. Regarding this way, semantic segmentation quality has a dominant influence on corresponding trimap, similar to what trimap is to matte. %the quality of 
% as do dilating contour of foreground segmentation for trimap generation and segmentation quality. 

From the above-mentioned problems, it is obvious that \textbf{trimap-needed matting (resp., mentioned trimap generation method) has a trimap (resp., foreground segmentation) quandary}. Although previous methods attempt to solve these puzzles, they mainly focus on single-category matting by exploring refining trimap to boost matting results~\cite{shen2016deep,cai2019disentangled}, employing implicit trimap to assist human/animal matting without trimap input~\cite{chen2018semantic,zhang2019late,li2020end}, and coupling coarse annotated data with fined one to promote human matting~\cite{liu2020boosting}, which \textbf{are hard to generalize to comprehensive data and usually require salient single-category object}.

Considering forementioned issues, we argue that how to find a balance between extra knowledge and matte accuracy is a critical cornerstone for automating natural image matting. Therefore, we disentangle automatic matting into trimap and alpha estimation subtasks as workaround. Different from all past attempts, we aims to generalize trimap-free matting more properly to comprehensive data. We propose a semantic-guided trimap-free matting approach, which consists of Trimap Generation Network and light-weight Non-local Matting Network with Refinement Module. Coarse foreground segmentation provides additional semantic information and can help network capture rough location and shape of target object, which can be easily obtained by salient/segmentation models. Then, Trimap Generation Network employs coarse foreground segmentation as guidance to estimate a proper trimap. This estimated trimap serves as guidance for matting network and also buffer for negative trimap quality chain reaction. Our proposed light-weight non-local attention block utilizes stride downsampling convolution to reduce similarity computation cost and rearranges alpha feature by propagating the global pixel-to-pixel relationship of image feature on an explicit fashion. Refinement module with fusion techniques bridges two main components together to produce high-quality alpha matte without trimap and background as input. Extensive experiments show that our matting pipeline has superior performance and comparable with other state-of-the-art methods on the Composition-1k testset, alphamatting.com benchmark, and Distinctions-646 testset. To demonstrate real-world application capability of our pipeline, we conduct real data adaption by finetuning our framework on real imagery data and user study for verification. The sufficient ablation analysis also justifies Trimap Generation Network to be segmentation fault-tolerance and qualified for automatic matting task.
% We apply image processing to synthesize coarse foreground semantic segmentation from alpha to train our model in synthesized matting datasets.

The main contribution of this work is threefold. First, we propose a novel two-stage trimap-free automatic natural image matting approach boosted by coarse foreground segmentation and our light-weight non-local attention, which is on a par with the state-of-the-art matting in trimap-needed and trimap-free respects. Our proposed method finds a trade-off between additional information and performance. \textbf{We believe that this matting architecture should be more rational trimap-free matting framework, which has comprehensive integration ability with other semantic segmentation/salient object detection/matting approaches to better solve automatic natural image matting.} Second, we propose Trimap Generation Network to predict the possibility of each pixel belonging to foreground/background/unknown areas, which can not only better capture semantic information, but also provide accurate trimap with defective segmentation as input. Third, our light-weight attention layer not only reduces computational complexity but also maintains effective performance.
% Sampling-based approaches~\cite{chuang2001bayesian,wang2005iterative,wang2007optimized,gastal2010shared,he2011global,shahrian2013improving,aksoy2017designing} initially sample colors from foreground and background pixels for each unknown region marked by trimap, and then select the best foreground-background color pair according to quantitative metrics. Propagation-based methods~\cite{chen2013knn,grady2005random,lee2011nonlocal,levin2007closed,levin2008spectral,sun2004poisson} work by propagating alpha value from known pixels to unknown ones based on some similarity measurements.
\section{Related Works}
\label{sec:related}
\textbf{Natural Image Matting}: Traditional image matting can be roughly classified into sampling-based~\cite{chuang2001bayesian,wang2005iterative,wang2007optimized,gastal2010shared,he2011global,shahrian2013improving,aksoy2017designing} and propagation-based~\cite{chen2013knn,grady2005random,lee2011nonlocal,levin2007closed,levin2008spectral,sun2004poisson} approaches, which usually require trimap as additional input. Recently, matting techniques using deep learning have shown increasingly prominent performance, which can be categorized into trimap-needed~\cite{cho2016natural,xu2017deep,lutz2018alphagan,lu2019indices,tang2019learning,li2020natural,yu2020high,sun2021semantic,dai2021learning}, background-required~\cite{qian1999video,sengupta2020background,lin2021real} and only-image~\cite{aksoy2018semantic,zhang2019late,qiao2020attention,wei2021improved} input. After Cho et al.~\cite{cho2016natural} introduce deep neural networks into image matting task, Xu et al.~\cite{xu2017deep} propose a deep neural network matting solution with a comprehensive matting database which has promoted research progress significantly. Lutz et al.~
\cite{lutz2018alphagan} explore matting task with a generative adversarial framework. Then, appealing matting results are achieved by Lu et al.~\cite{lu2019indices} and Tang et al.~\cite{tang2019learning}. Subsequently, a state-of-the-art matting method with guided contextual attention is proposed, which not only simulates information flow of affinity-based methods but also models matting in a view of image inpainting~\cite{li2020natural}. Given an image, background, and soft segmentation, background matting~\cite{sengupta2020background} is adapted to real human data and obtains appealing estimation but is not robust to images with shadow or under complex light conditions. Since trimap-needed matting algorithm usually requires high-quality time-consuming handmade trimap, a few works are attempting to take only image as input and produce matte~\cite{zhang2019late,qiao2020attention}. However, these trimap-free methods are not capable of producing comparable quality matte.

%bahdanau2014neural,luong2015effective,britz2017massive,
% Attention mechanism has been widely utilized in deep learning tasks, e.g. machine translation~\cite{vaswani2017attention,tang2018self}, image classification~\cite{mnih2014recurrent,jetley2018learn}, image generation~\cite{deng2018r3net,parmar2018image}, object detection~\cite{ba2014multiple}, image synthesis~\cite{zhang2019self}, video classification~\cite{wang2018non}, and semantic segmentation~\cite{fu2019dual,huang2019ccnet}. 
\textbf{Attention Mechanism}: Attention mechanism has been widely utilized in deep learning tasks like machine translation~\cite{vaswani2017attention,tang2018self}, image classification~\cite{mnih2014recurrent,jetley2018learn}, video classification~\cite{wang2018non}, and semantic segmentation~\cite{fu2019dual,huang2019ccnet}. The self-attention block is introduced to transformer and contributes to each position of output by referring to every position of input~\cite{vaswani2017attention}. Similarly, Wang et al.~\cite{wang2018non} propose non-local attention to acquire long-range contextual information and promote video classification tasks. Instead of capturing long-term dependencies in sequences, attention has shown its superiority in image matting~\cite{li2020natural,qiao2020attention,yu2020high}, by making matting network capture structural pixel-to-pixel dependencies of image that can deepen semantic understanding of network. Li et al.~\cite{li2020natural} simulate non-local attention by convolution and deconvolution to enhance alpha matte estimation. Qiao et al.~\cite{qiao2020attention} use channel/spatial-wise attention to filter out noise from hierarchical appearance cues and boost alpha mattes. Yu et al.~\cite{yu2020high} introduce three non-local attentions to propagate each trimap region of context patches to corresponding region of query patches. However, our approach leverages only one attention layer and has better matting performance. % and achieve high-resolution image matting with cross-patch modeling

%singh2013automatic
\textbf{Trimap Generation}: Automatic trimap generation, which is popular in traditional matting~\cite{wang2007automatic,hsieh2013automatic,gupta2016automatic,chen2020interactive}, usually contains two steps: binary segmentation for foreground/background separation and image erosion/dilation. These methods mainly differentiate in how to obtain segmentaion. For example, Wang et al.~\cite{wang2007automatic} leverage depth information to compute segmentation; Gupta et al.~\cite{gupta2016automatic} combine salient object detecion with superpixel analysis for segmentation; Hsieh et al.~\cite{hsieh2013automatic} use graph cuts for foreground extraction; Chen et al.~\cite{chen2020interactive} require user to indicate foreground/background by a few clicks and apply one-shot learning for binary mask prediction. Recently, neural networks have been utilized to generate implicit trimap for human matting automation~\cite{shen2016deep,chen2018semantic}.% It is obviously noted that .
% Besides, Al-Kabbany et al.~\cite{al2015novel} first introduce the Gestalt laws of grouping to matting that assists more robust trimap generation. Cho et al.~\cite{cho2016automatic} integrate depth map with color distribution based processing for foreground/background separation by graph cut followed by unknown region detection for trimap genneration. 
\section{Approach}
% \subsection{Alpha Matting}
We decompose our trimap-free matting approach into \textbf{Trimap Generation Network (Net-T)} and \textbf{light-weight Non-local Matting with Refinement (Net-M)}. With rough foreground segmentation~\cite{yang2019parsing,chen2017rethinking} as additional indicator, Net-T understands target object shape and its relation with surroundings to perform accurate pixel-wise classification among foreground/background/unknown regions. Net-M utilizes RGB image and output of Net-T to estimate alpha matte. The overview of our proposed matting framework is illustrated in Fig.~\ref{netm}.

\begin{figure}[thpb]
\centering
\includegraphics[scale=0.25]{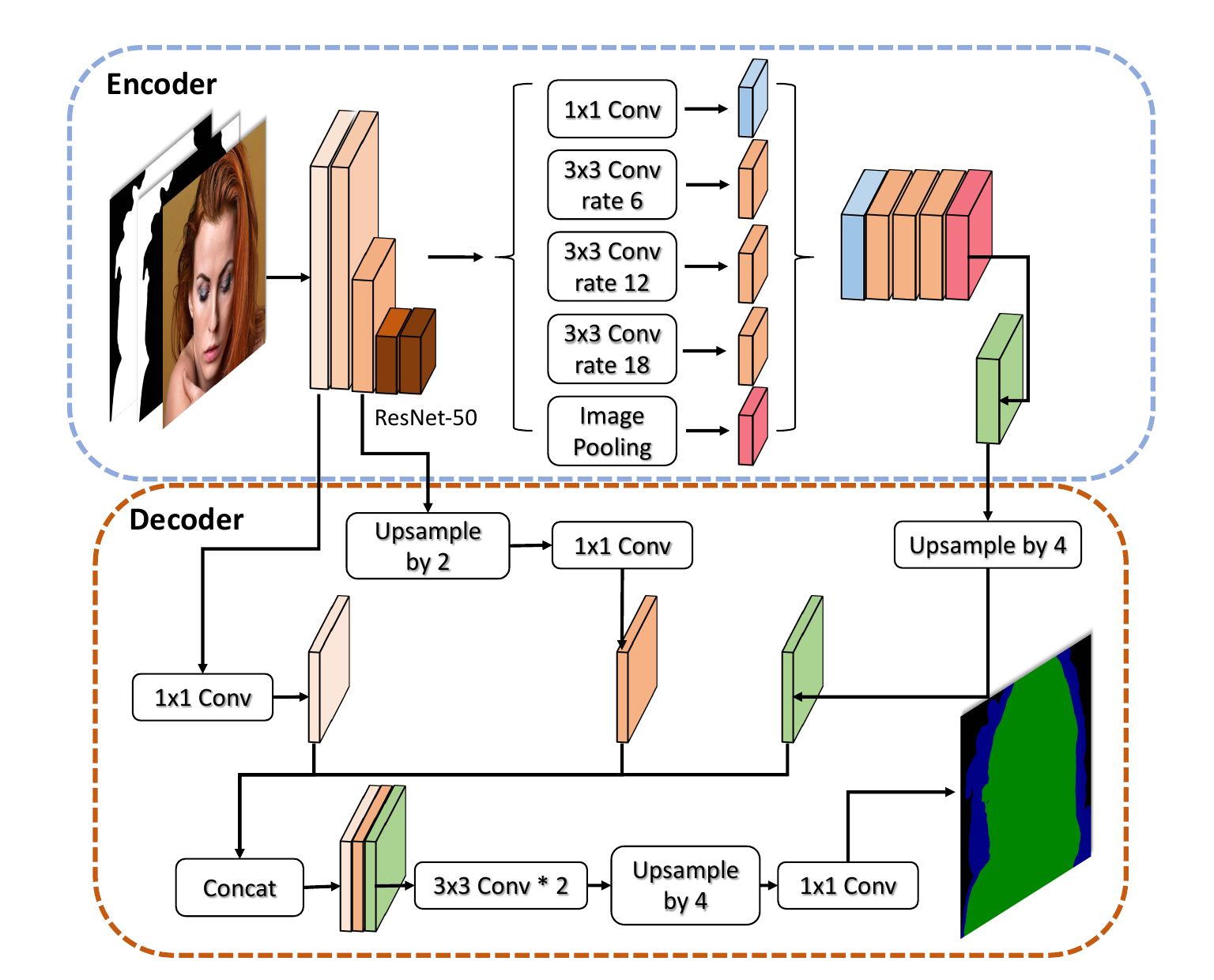}
\caption{Trimap Generation Network (Net-T)}
\label{nett}
\end{figure}

\subsection{Trimap Generation Network (Net-T)}
Net-T conducts a 3-class semantic segmentation task, where the input of Net-T is the concatenation of a cropped RGB image and a 2-channel one-hot soft foreground segmentation, and the output of Net-T is a 3-channel feature map indicating the possibility that each pixel is assigned to each of 3 classes.
% As shown in Fig~\ref{nett}, the output of Net-T is a 3-class semantic map, which is represented by three colors, i.e. green, black and blue, corresponding to foreground, background, and unknown regions.
We utilize a modified Deeplabv3~\cite{chen2017rethinking} as the encoder by adjusting the input channel size and taking the first two channel weights of $conv\_1$ of the pretrained ResNet-50 as the weight for 2-channel segmentation.
% \textcolor{red}{Instead of listing the details of the network design, the authors should explicitly explain the reason why the network was designed in this way. The paper writing is mainly about the spreading of the scientific ideas.}
Instead of using its original decoder, we propose our own decoder to reconstruct semantic information, as illustrated in Fig.~\ref{nett}. The dropout layer by 0.5 factor is first applied to high-level encoder feature, followed by a four-time bilinearly upsampling. Then, instead of directly adopting high-level encoder feature for final classification, we aggregate it with low and middle-level features of encoder to enrich the decoding process and make network pay attention to image appearance and less dependent on segmentation input. And the swapped order of upsampling and final convolution makes classification more fine-grained. Due to these careful designs, Net-T can be tolerant for inaccuracy of foreground segmentation.
\subsection{Non-local Matting with Refinement (Net-M)}
\begin{figure*}[thpb]
\centering
\includegraphics[scale=0.25]{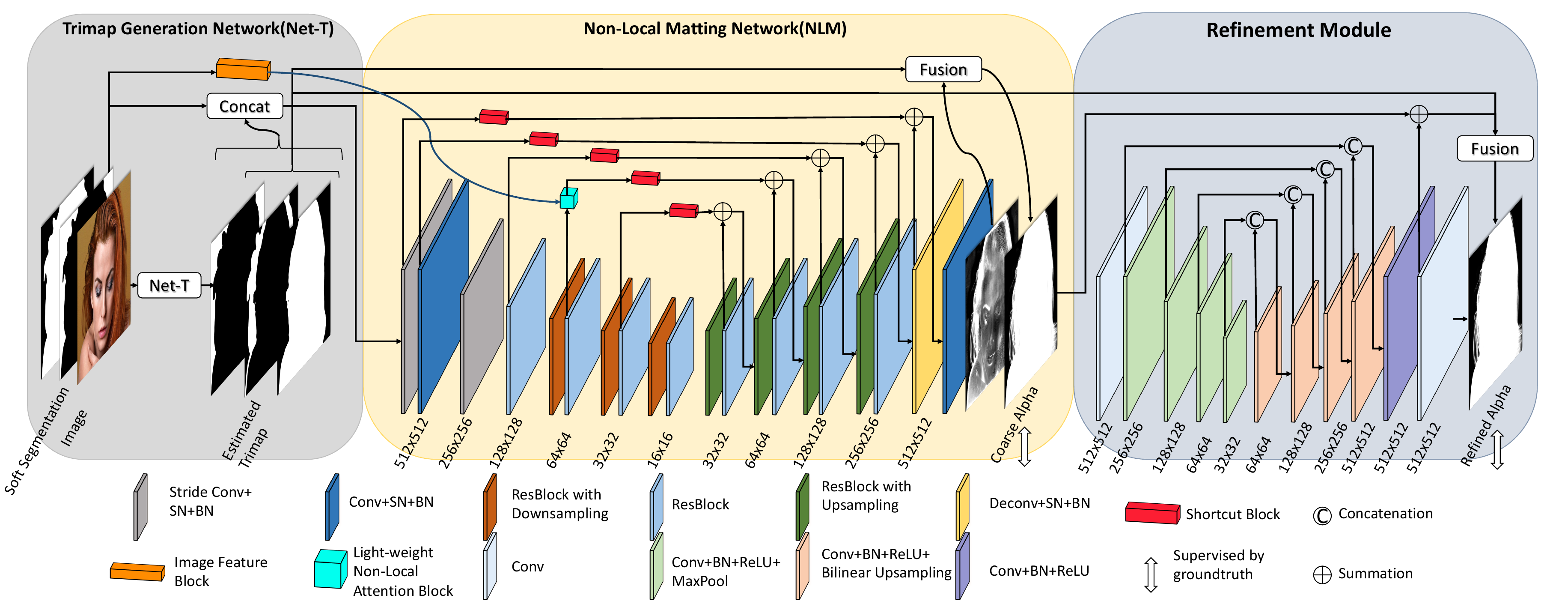}
\caption{The Overview of our Matting Pipeline. Non-local Matting Network with Refinement Module is Net-M. Net-T and Net-M collaborative testing is Joint Inference. (SN is Spectral Normalization, while BN is Batch Normalization.)}
\label{netm}
\end{figure*}

We adopt popular U-Net structure~\cite{ronneberger2015u,li2020natural} as the main architecture, illustrated in Fig.~\ref{netm}.

\begin{figure}[thpb]
\centering
\includegraphics[scale=0.25]{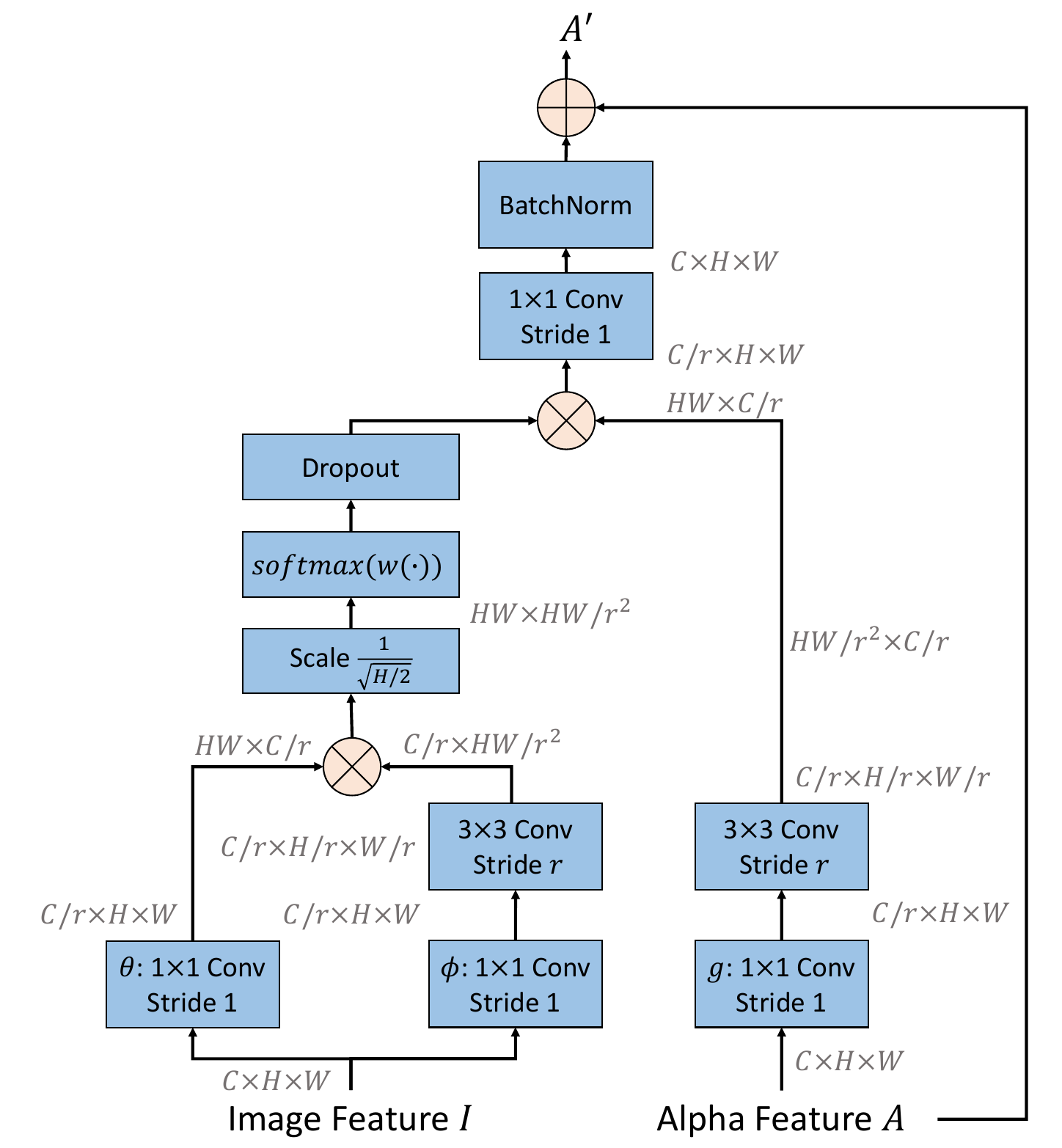}
\caption{Light-weight Non-local Attention Block. The size of feature maps are shown. ``$\bigoplus$'' means element-wise sum, ``$\bigotimes$'' denotes matrix multiplication and $r$ is the downsample ratio. The softmax operation is performed on each row.}
\label{nonlocal}
\end{figure}

\textbf{light-weight Non-local Matting (NLM)}:%vaswani2017attention,wang2018non,yu2018generative,liu2019coherent,
Traditional matting methods usually estimate unknown pixels by investigating color similarity of unknown/known areas, and formulating transition pixels as weight combination of relevant foreground/background pixels based on certain criteria. Inspired by startling capability of attention~\cite{li2020natural}, we design a light-weight non-local attention block to model classic matting in order to promote alpha feature learning and speed up training. Fig.~\ref{nonlocal} shows details of our light-weight Non-local attention block. Instead of simulating similarity computation by a convolution between unknown region and kernels reshaped from image feature~\cite{li2020natural}, we explicitly calculate similarity between each pixel and the rest via embedded dot-product in the transformed image feature space with scaled softmax normalization. And then this pixel-to-pixel relation is employed to reconstruct the original alpha feature $A$ as $A^{\prime}$:
\begin{equation}
    A_{x,y}^{\prime}=A_{x,y}+\mathbf{W}(\sum_{\forall{x^{\prime},y^{\prime}}}f(I_{x,y},I_{x^{\prime},y^{\prime}})g^{\prime}(A_{x^{\prime},y^{\prime}})),
\end{equation}
\begin{equation}
    f(I_{x,y},I_{x^{\prime},y^{\prime}})=softmax(w(U,K,x,y)\frac{\theta(I_{x,y})^{T}\phi^{\prime}(I_{x^{\prime},y^{\prime}})}{\sqrt{d/2}}),
\end{equation}
% \begin{equation}
%     w(U,K,x,y)=
%     \begin{cases}
%     clip(\sqrt{\frac{\left|U\right|}{\left|K\right|}}) \qquad I_{x,y}\in U, \\
%     clip(\sqrt{\frac{\left|K\right|}{\left|U\right|}}) \qquad I_{x,y}\in K
%     \end{cases},
% \end{equation}
\begin{equation}
w(U,K,x,y)=
\begin{cases}     
clip(\sqrt{{\left|U\right|}/{\left|K\right|}}) \qquad I_{x,y}\in U,\\
clip(\sqrt{{\left|K\right|}/{\left|U\right|}}) \qquad I_{x,y}\in K,
\end{cases},
\end{equation}
\begin{equation}
    clip(x)=\min{(\max{(x,0.1)},10)},
\end{equation} % \textcolor{red}{What does I(Input image?) stands for?}
where $I$ refers to image feature map, $I_{x,y}$ is pixel value at the position $(x,y)$ of $I$, $g^{\prime}(\cdot)$ and $\phi^{\prime}(\cdot)$ denote embedded linear transformation and downscale operation, $d$ is the dimension of original feature map, $U$ is the unknown region, $K=I-U$ and $\mathbf{W}$ is the learnable weight matrix. Considering the barrier of high computational cost of dot-product calculation, we choose stride convolutions to downscale features which can not only maintain less information loss than pooling or interpolation does, but also speed up training and prevent gradient explosion/vanishing. The downscale ratio of $r$ is set to 4 in our experiments. Dropout is applied to prevent overfitting, and the residual summation is to stable the training. We assume that the attention block escorts the encoder to understand unknown areas and summation connections between encoder and decoder open the information communication gate to assist transition reconstruction in the decoder.

\textbf{Refinement Module}:
Refinement technique has been applied to salient object detection and semantic segmentation~\cite{deng2018r3net,amirul2018revisiting}, and shown impressive performance. There exists similarity between salient object detection, which can be considered as coarse alpha matte estimation using only RGB image, and image matting, a regression version of 2-class semantic segmentation.
% \textcolor{red}{What the relationship between the first part of sentence and second part?}.
Based on this insight, we introduce refinement module to refine the predicted coarse alpha $\alpha_{coarse}$ by learning the residuals $\alpha_{residual}$ between the coarse alpha and ground truth as $\alpha_{refined} = \alpha_{coarse} + \alpha_{residual}$.
% \begin{equation}
%     \alpha_{refined} = \alpha_{coarse} + \alpha_{residual}.
% \end{equation}
% $\alpha_{refined} = \alpha_{coarse} + \alpha_{residual}$.

\begin{figure}[thpb]
\setlength\tabcolsep{0pt}
\renewcommand{\arraystretch}{0.25}
\begin{center}
\resizebox{\columnwidth}{!}{%
\begin{tabular}{ccccccc}
% \small{Image}&
\includegraphics[width=2cm]{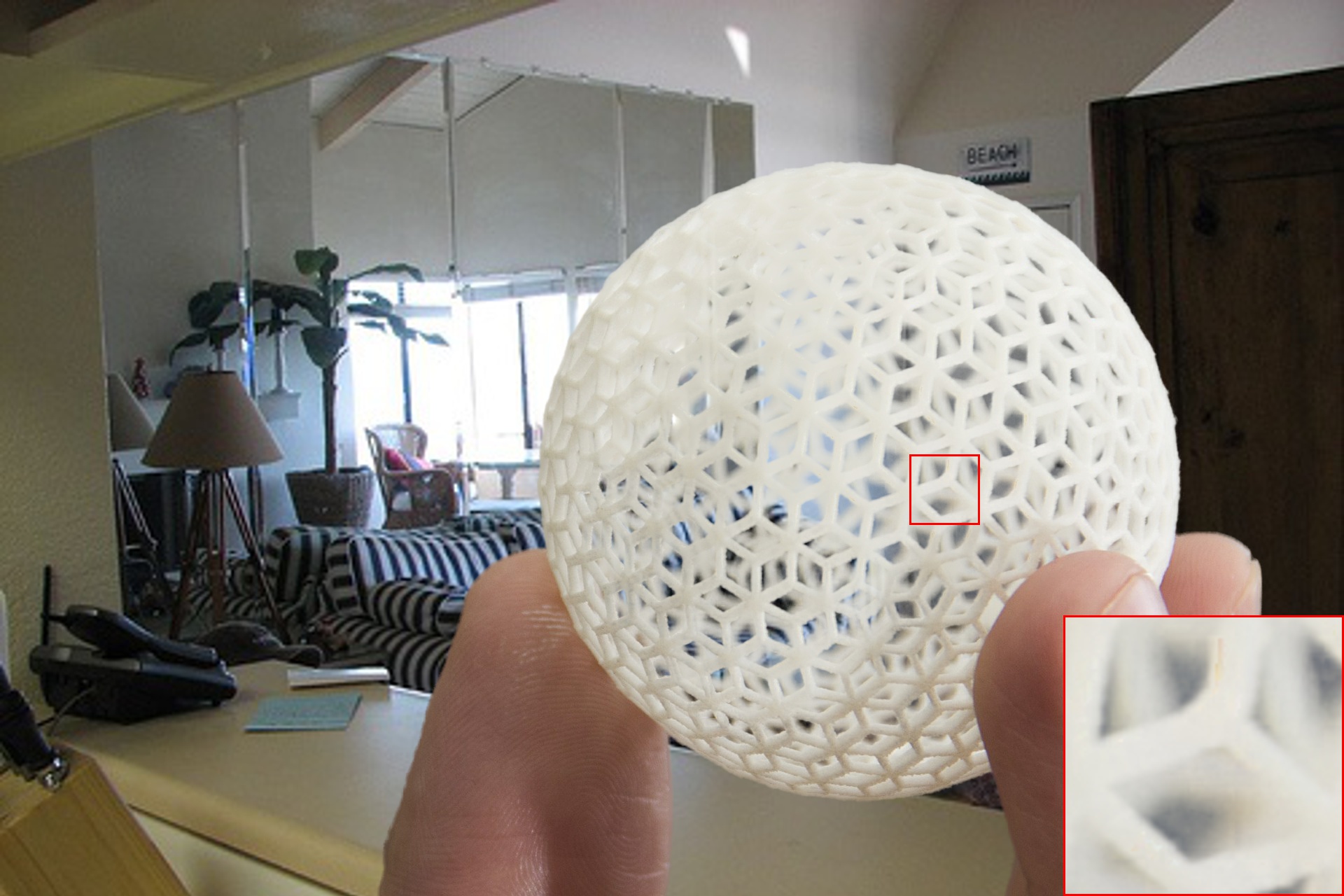}&\includegraphics[width=2cm]{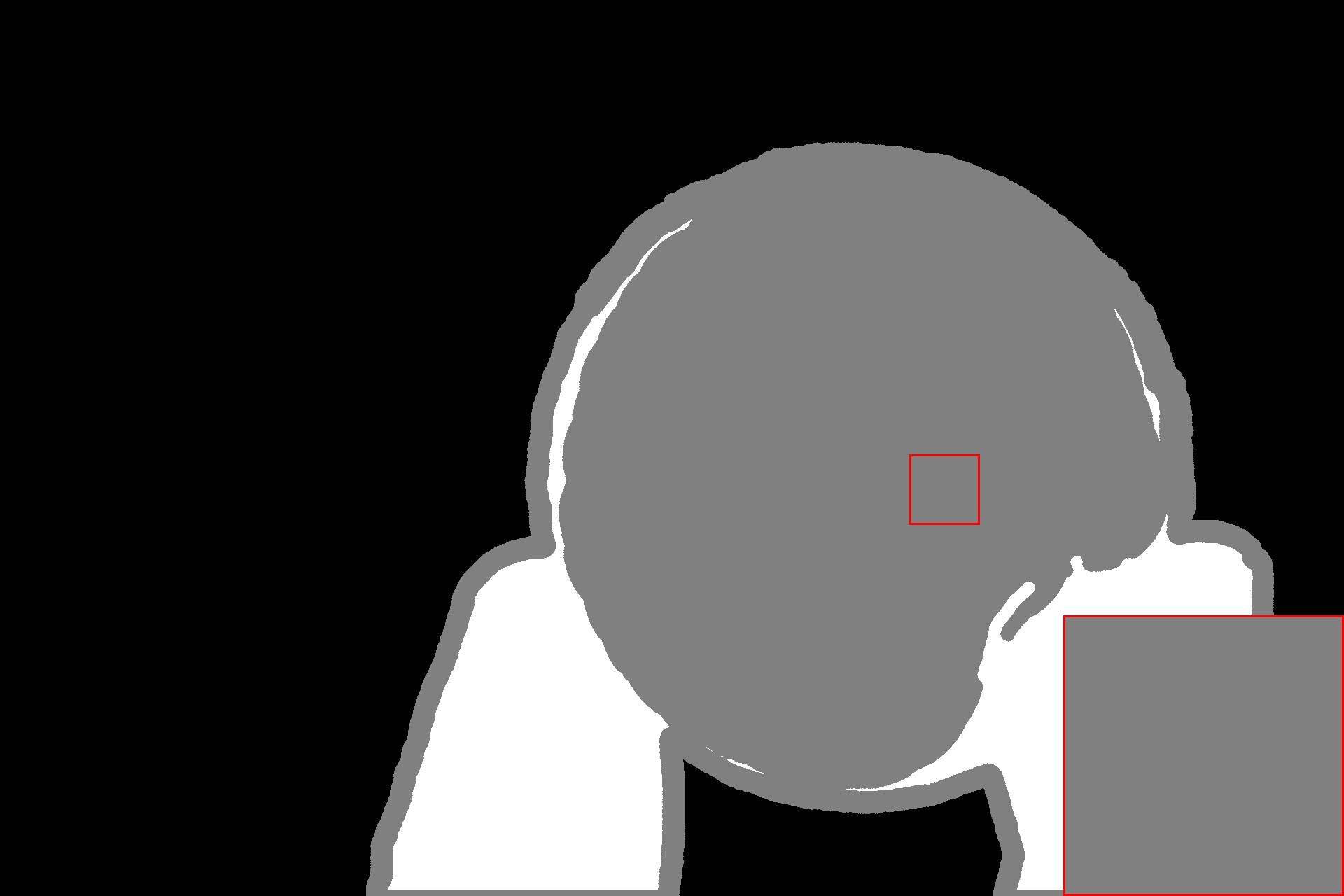}&\includegraphics[width=2cm]{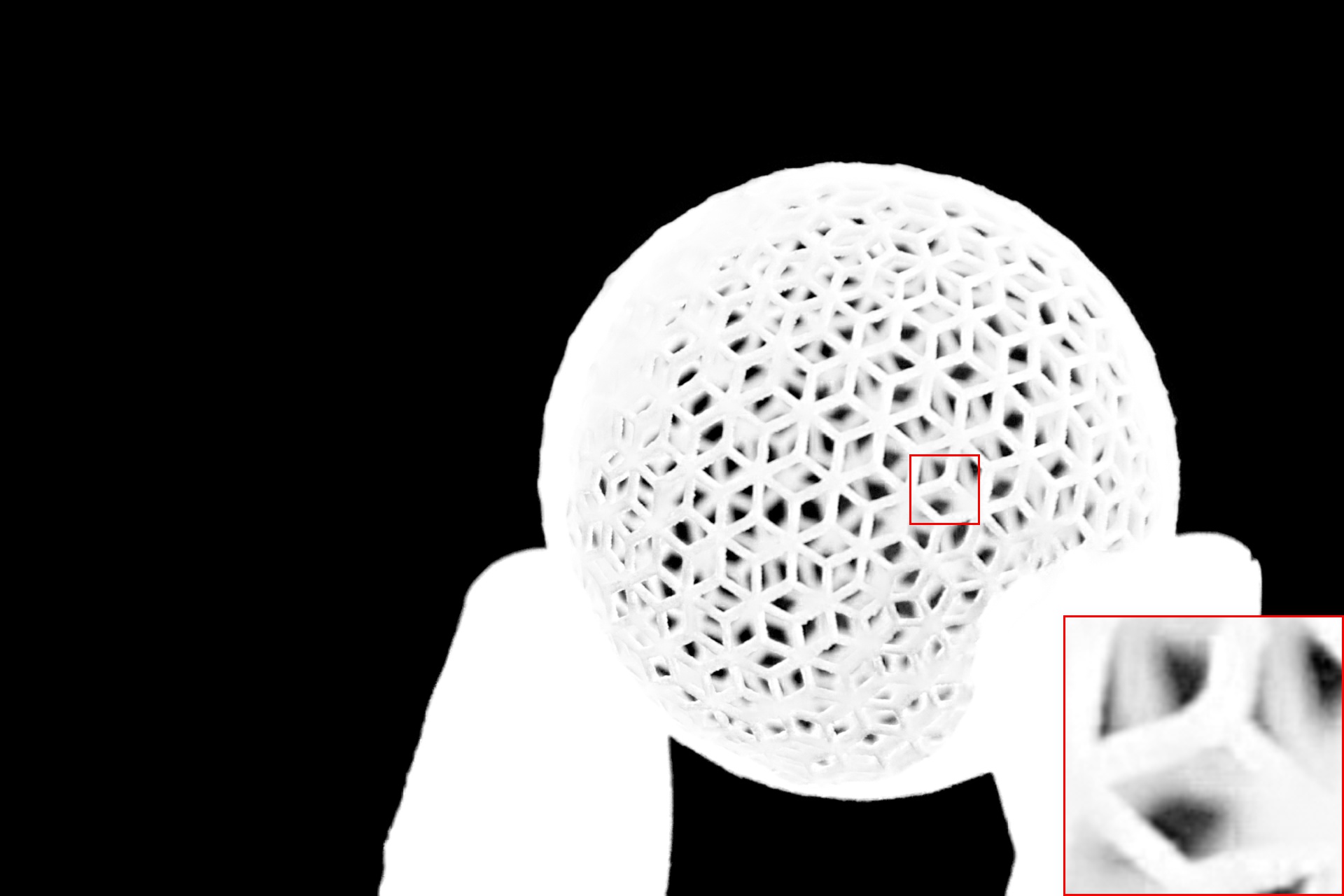}&\includegraphics[width=2cm]{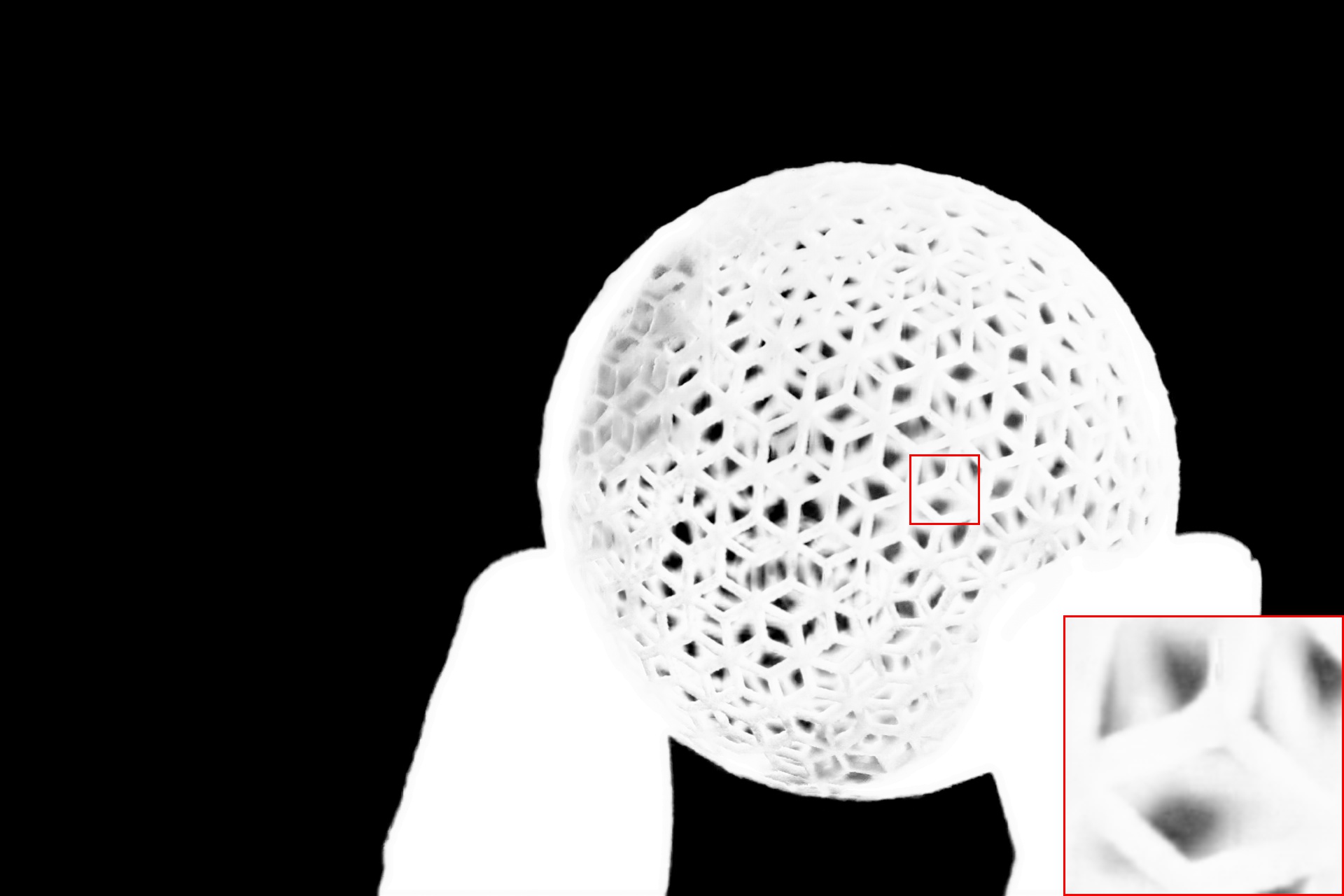}&\includegraphics[width=2cm]{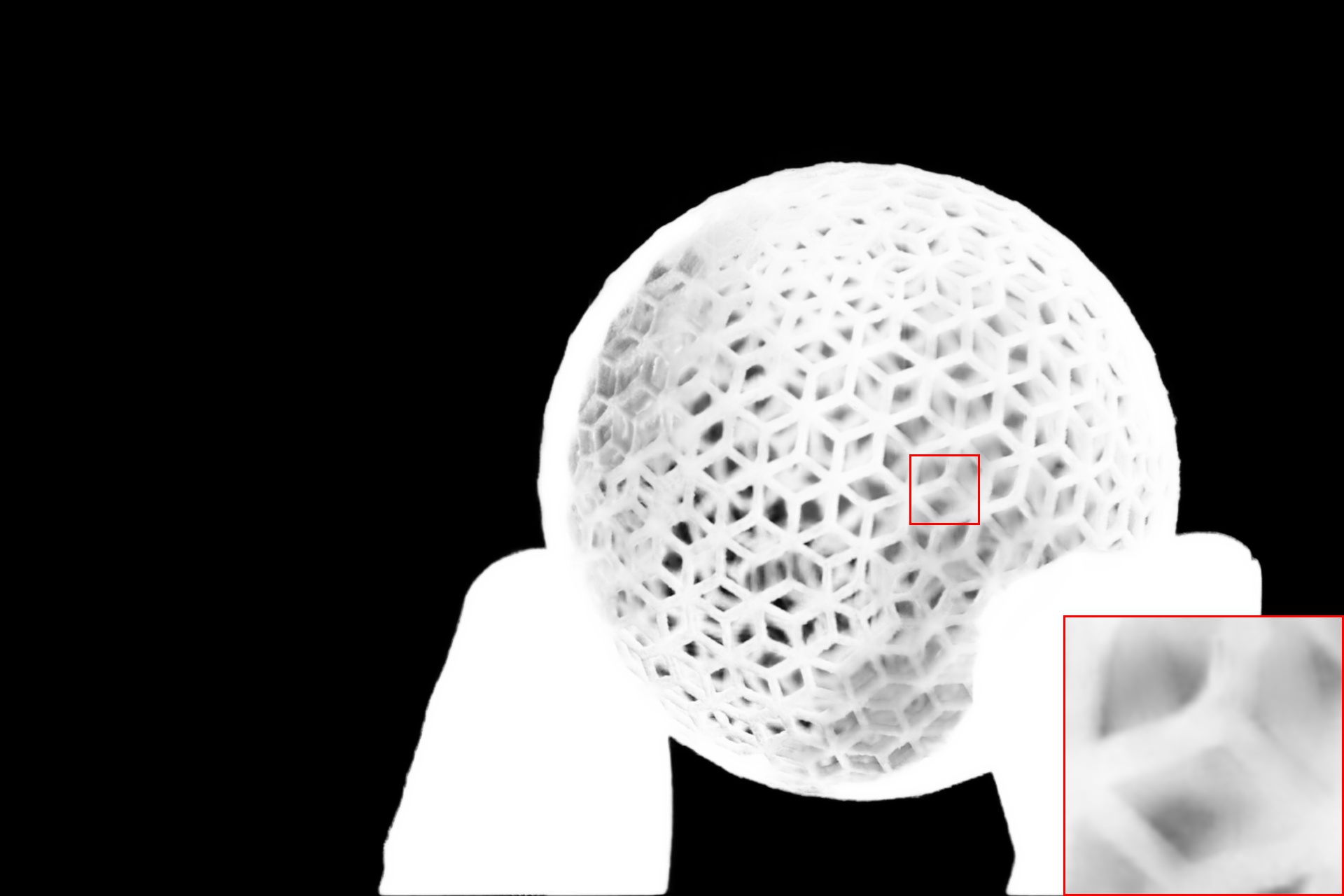}&\includegraphics[width=2cm]{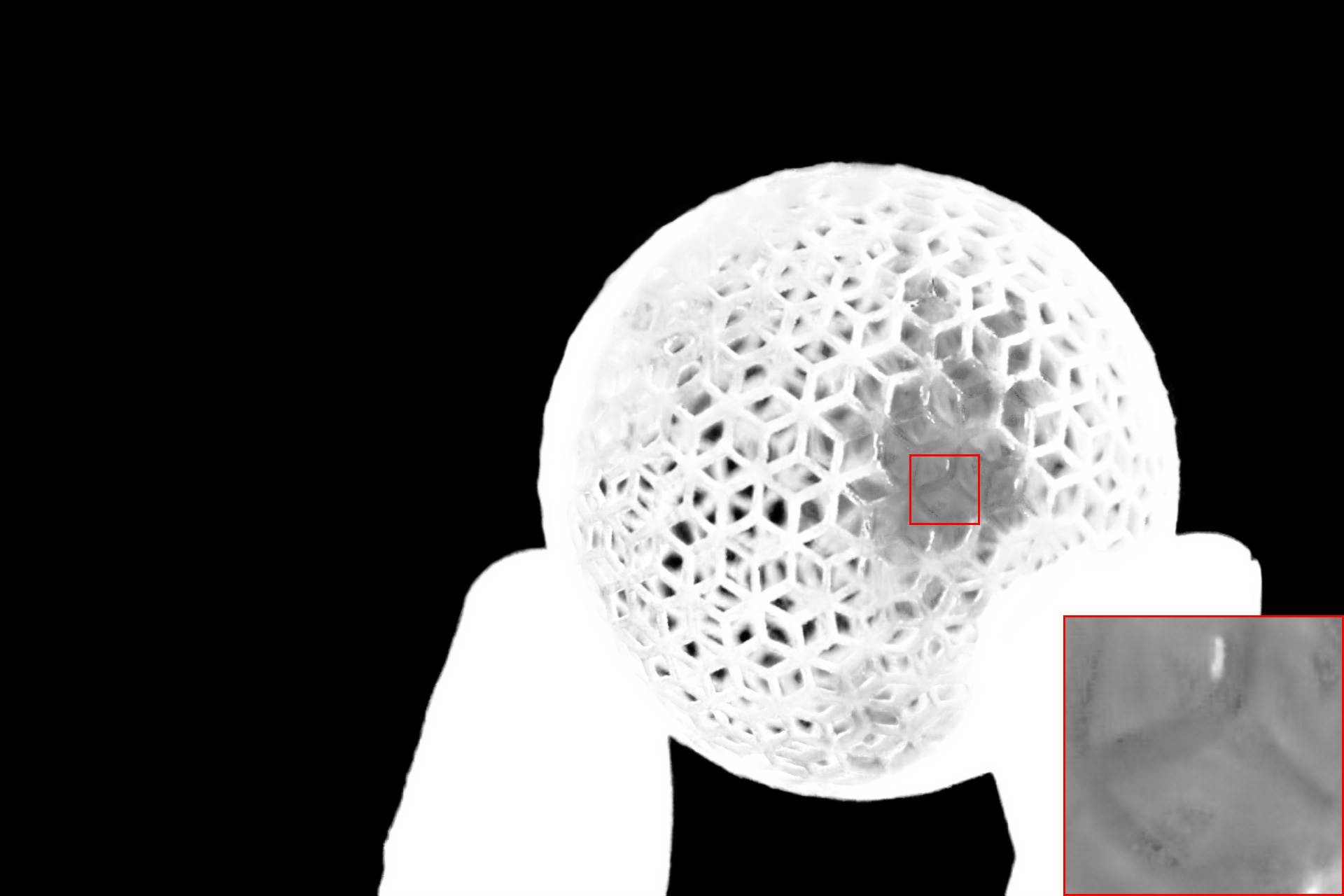}&\includegraphics[width=2cm]{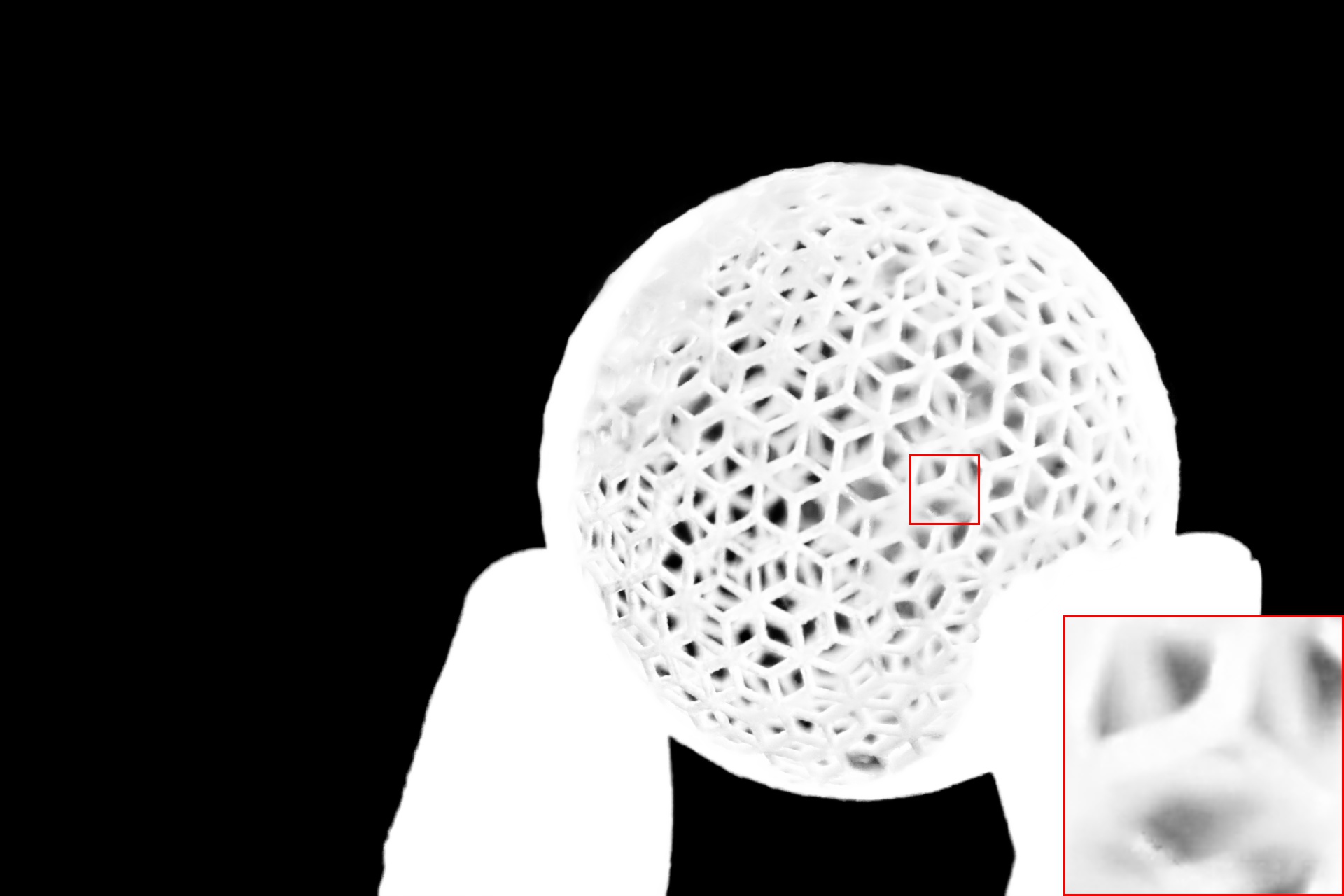}\\

\includegraphics[width=2cm]{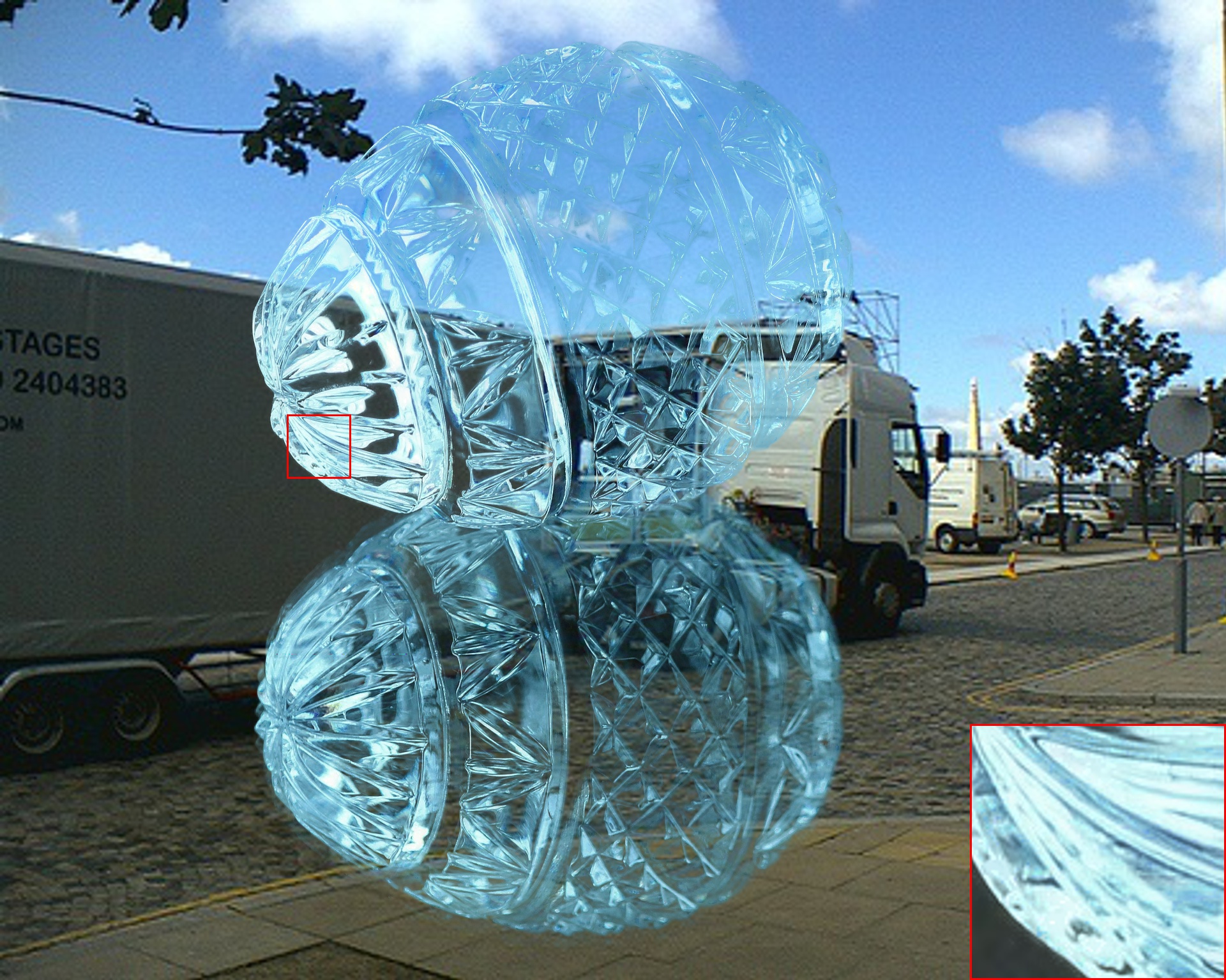}&\includegraphics[width=2cm]{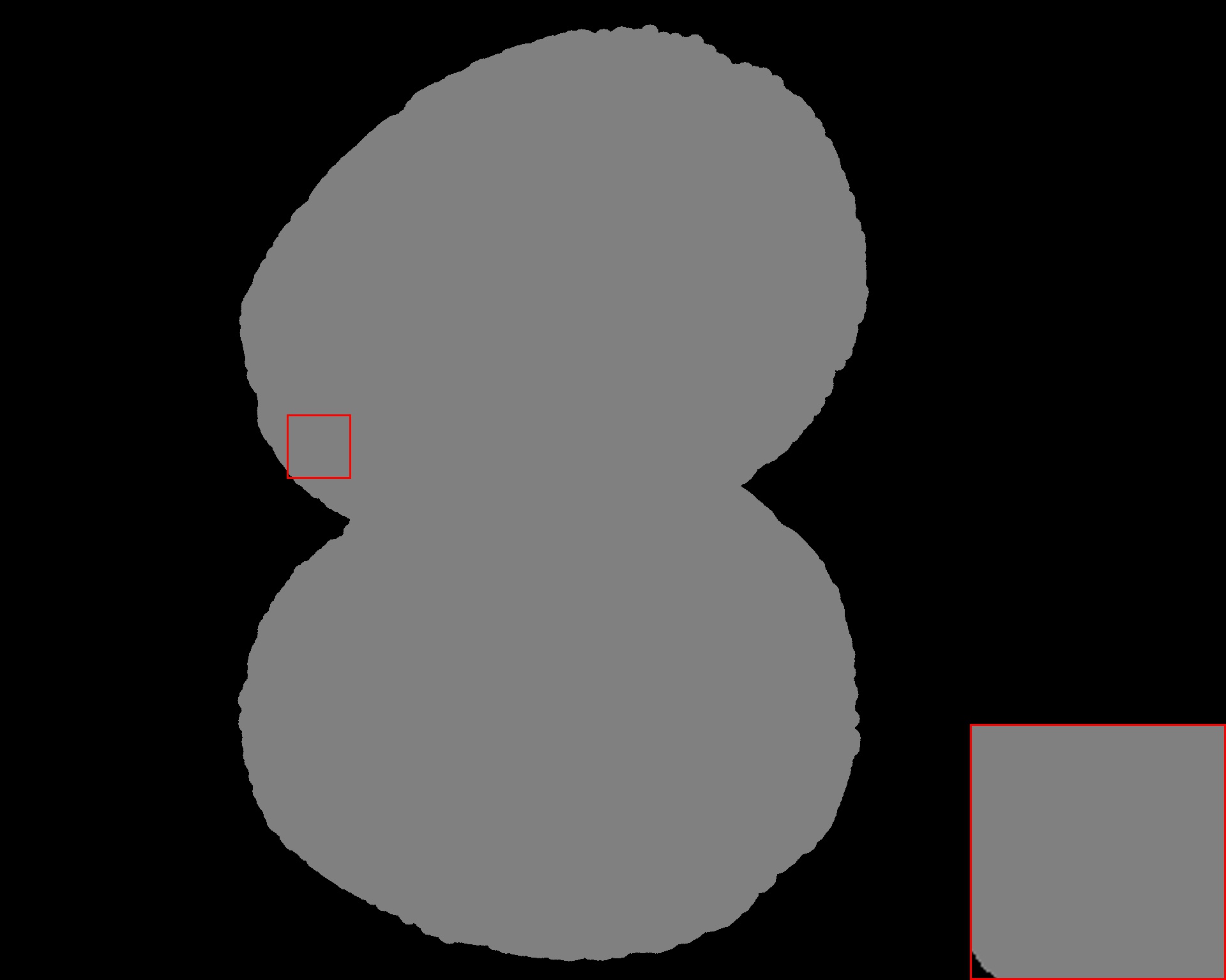}&\includegraphics[width=2cm]{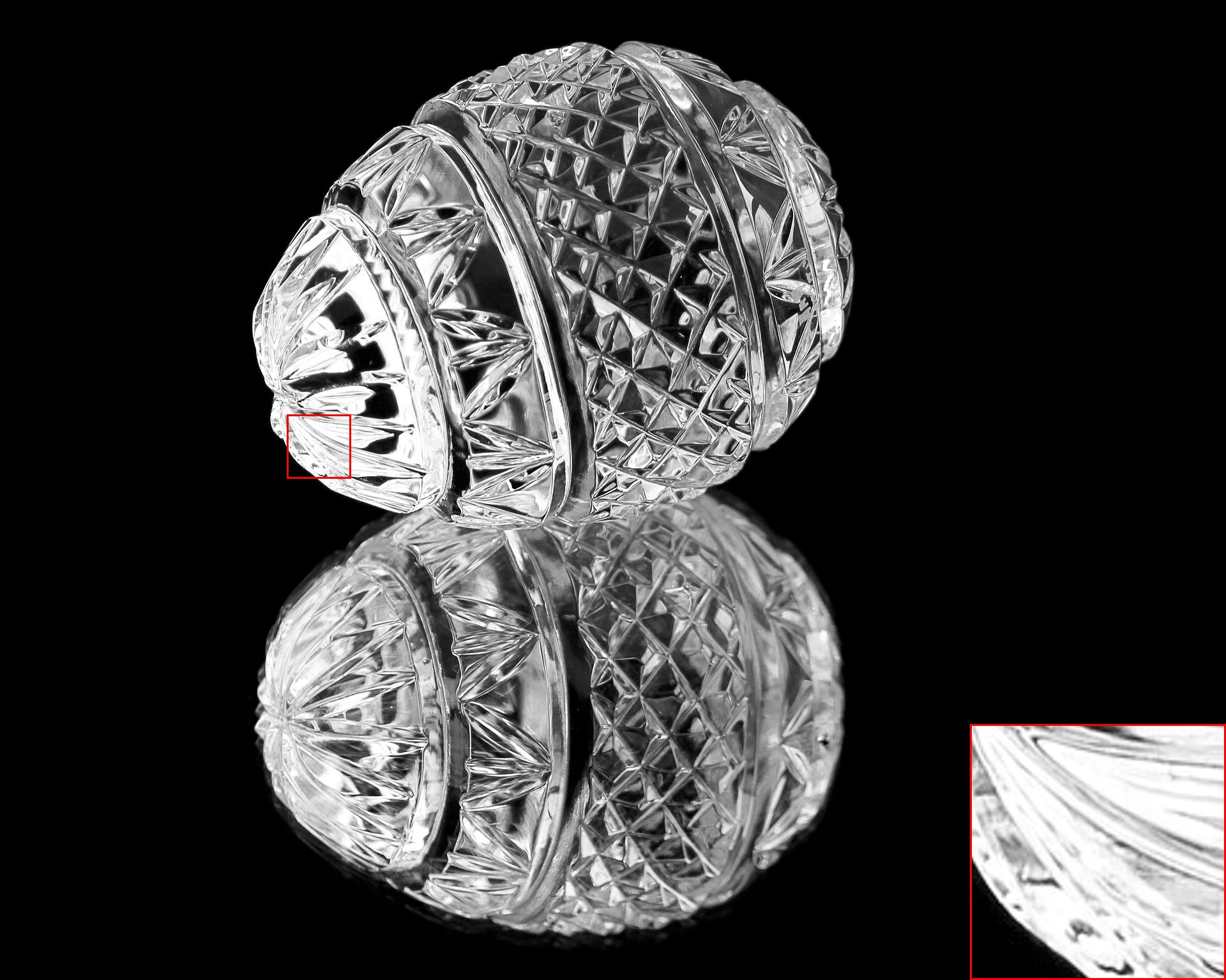}&\includegraphics[width=2cm]{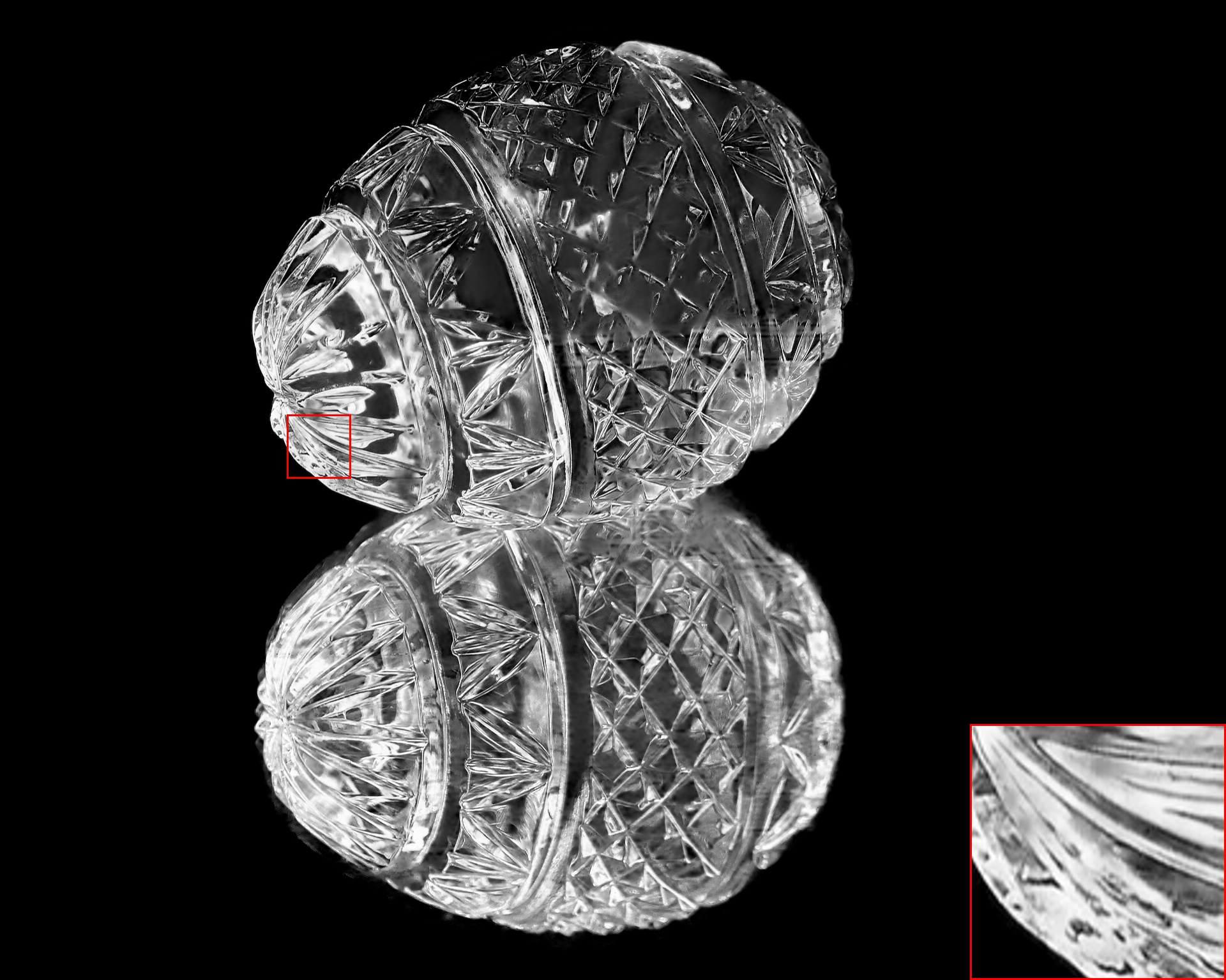}&\includegraphics[width=2cm]{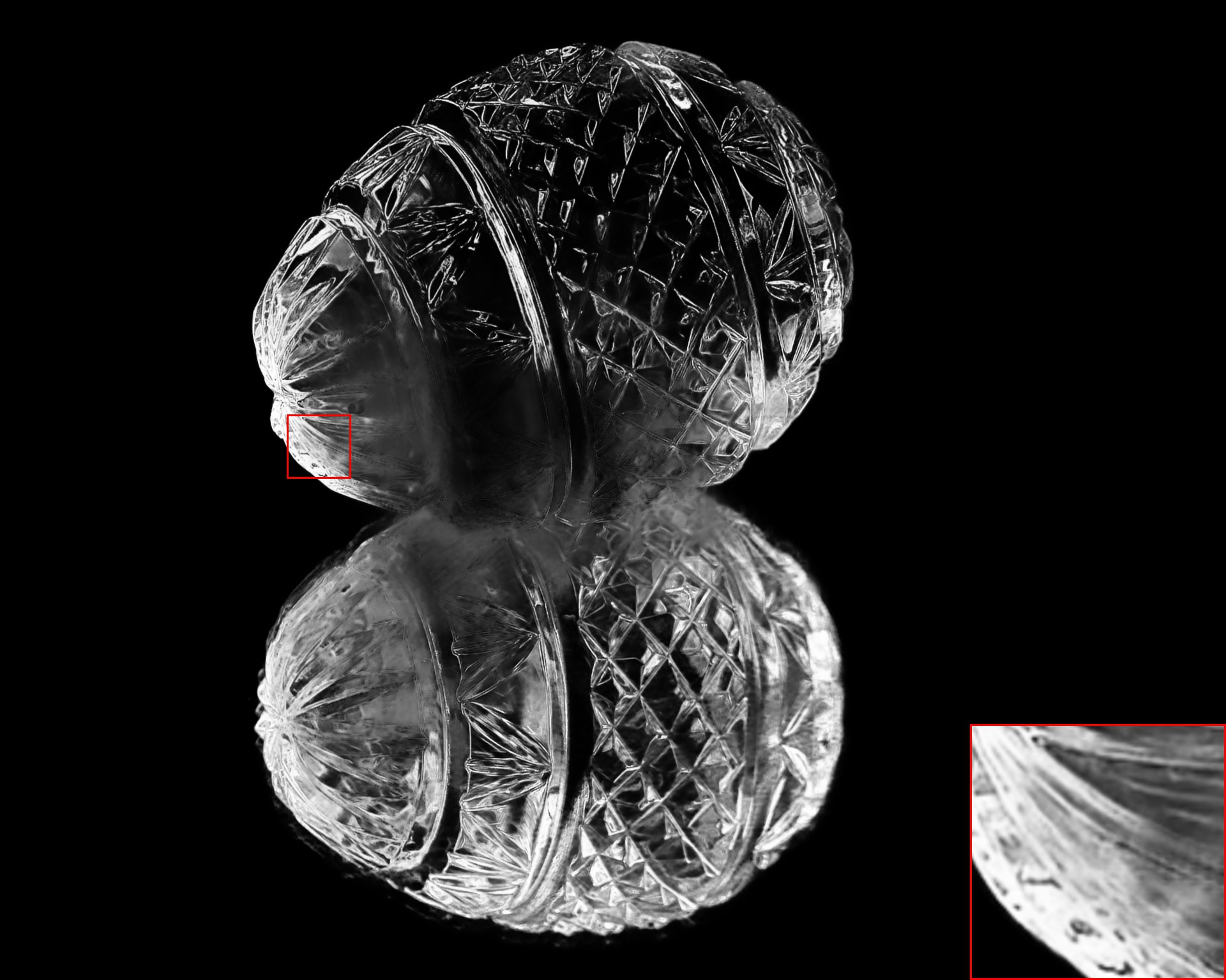}&\includegraphics[width=2cm]{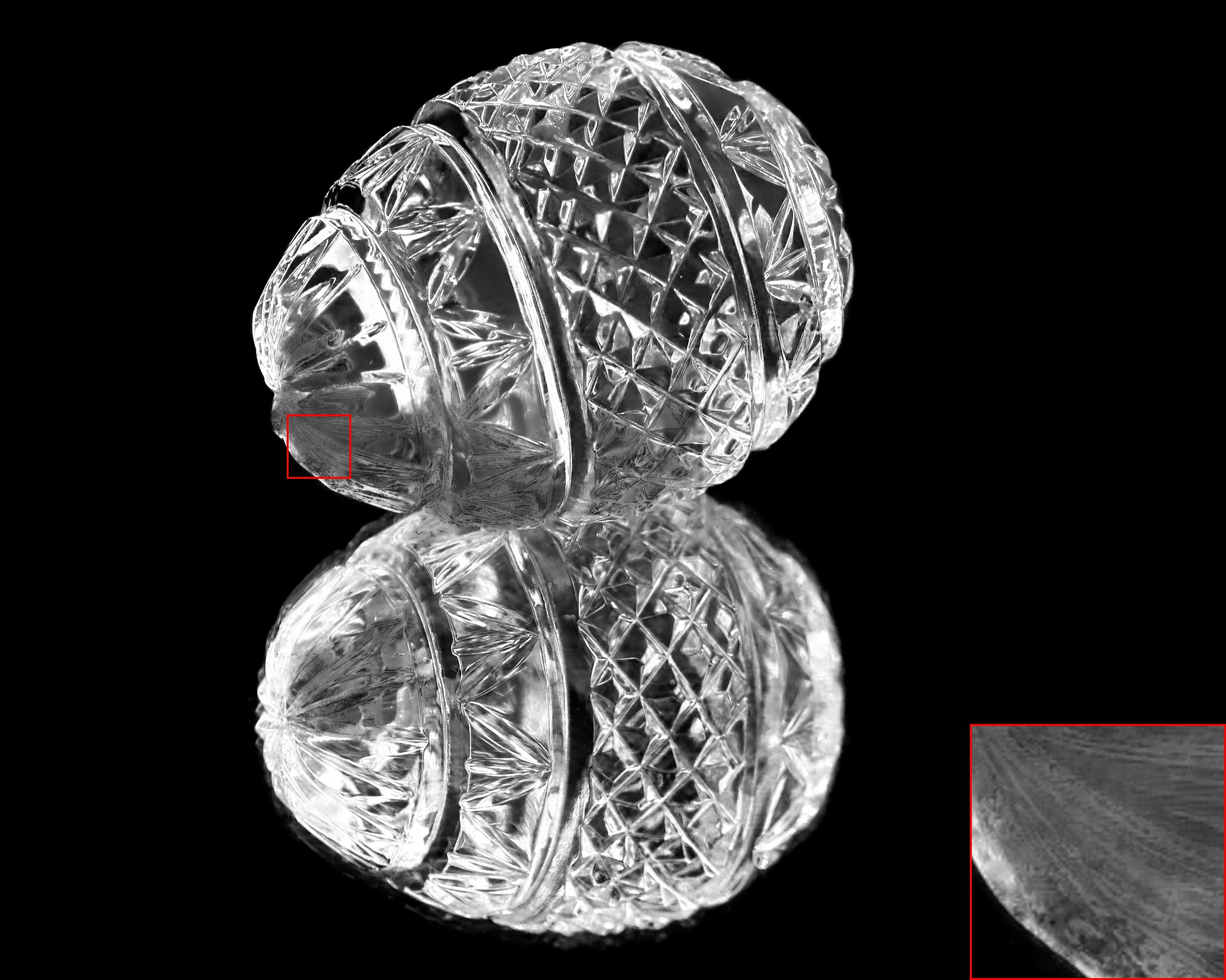}&\includegraphics[width=2cm]{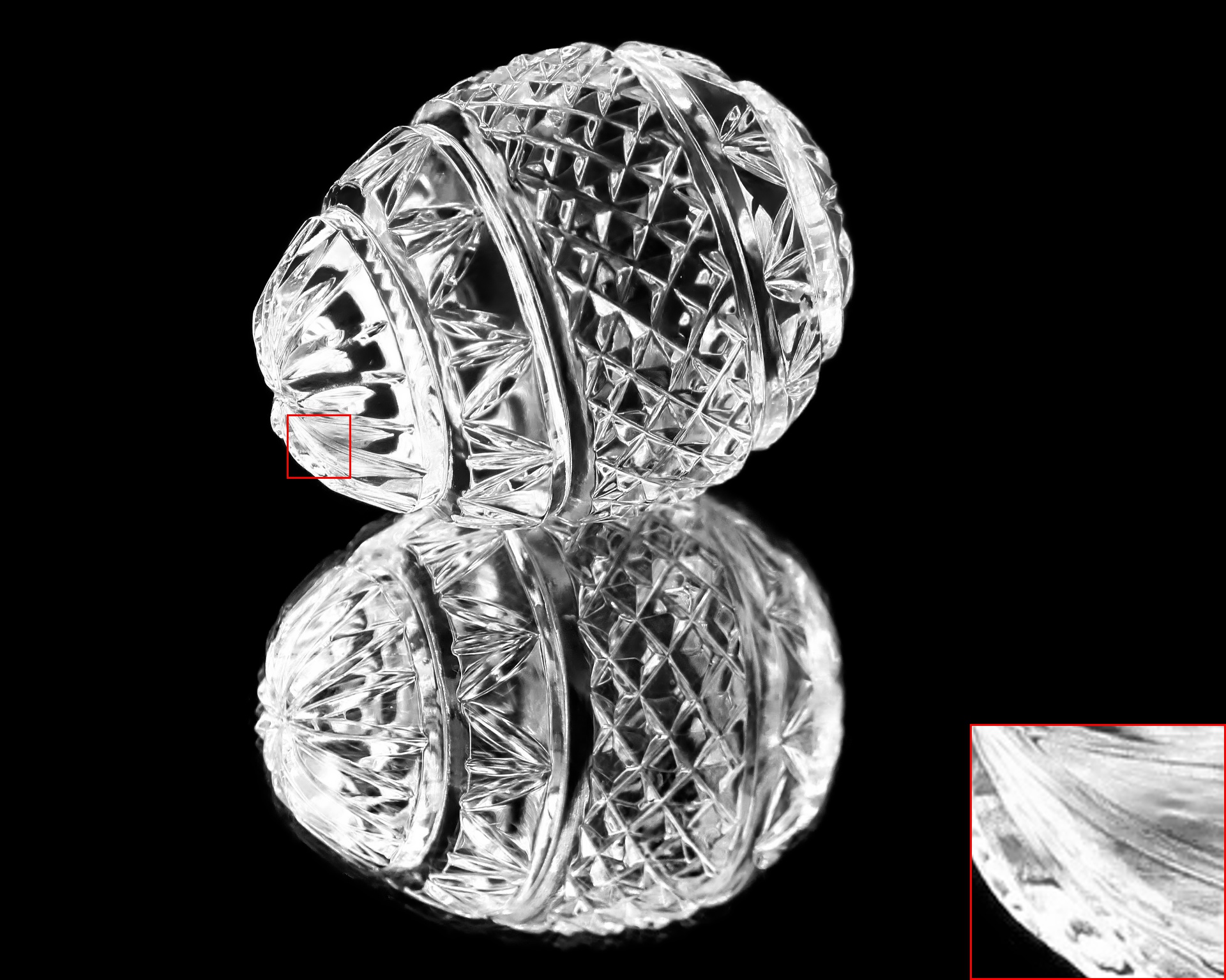}\\

\includegraphics[width=2cm]{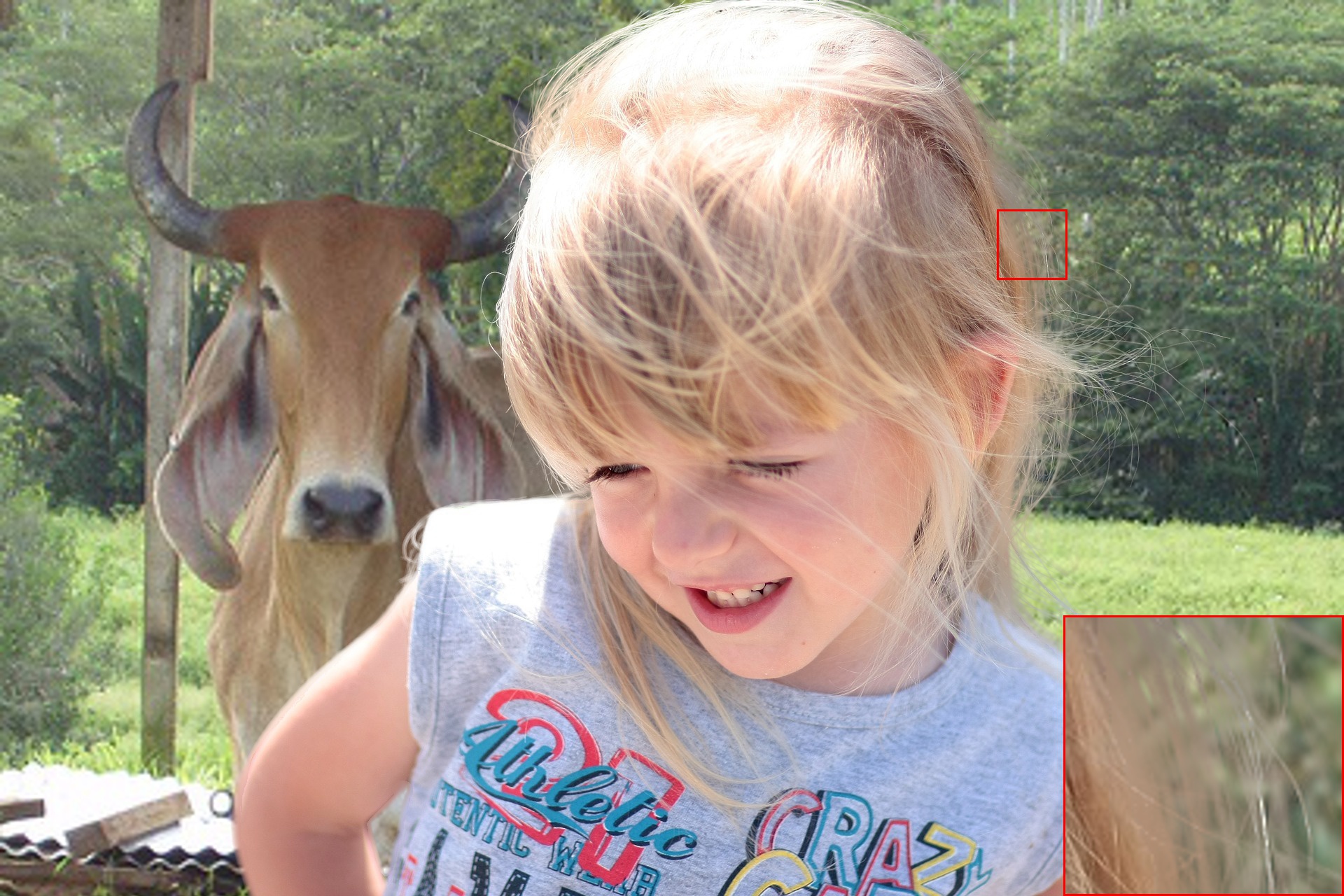}&\includegraphics[width=2cm]{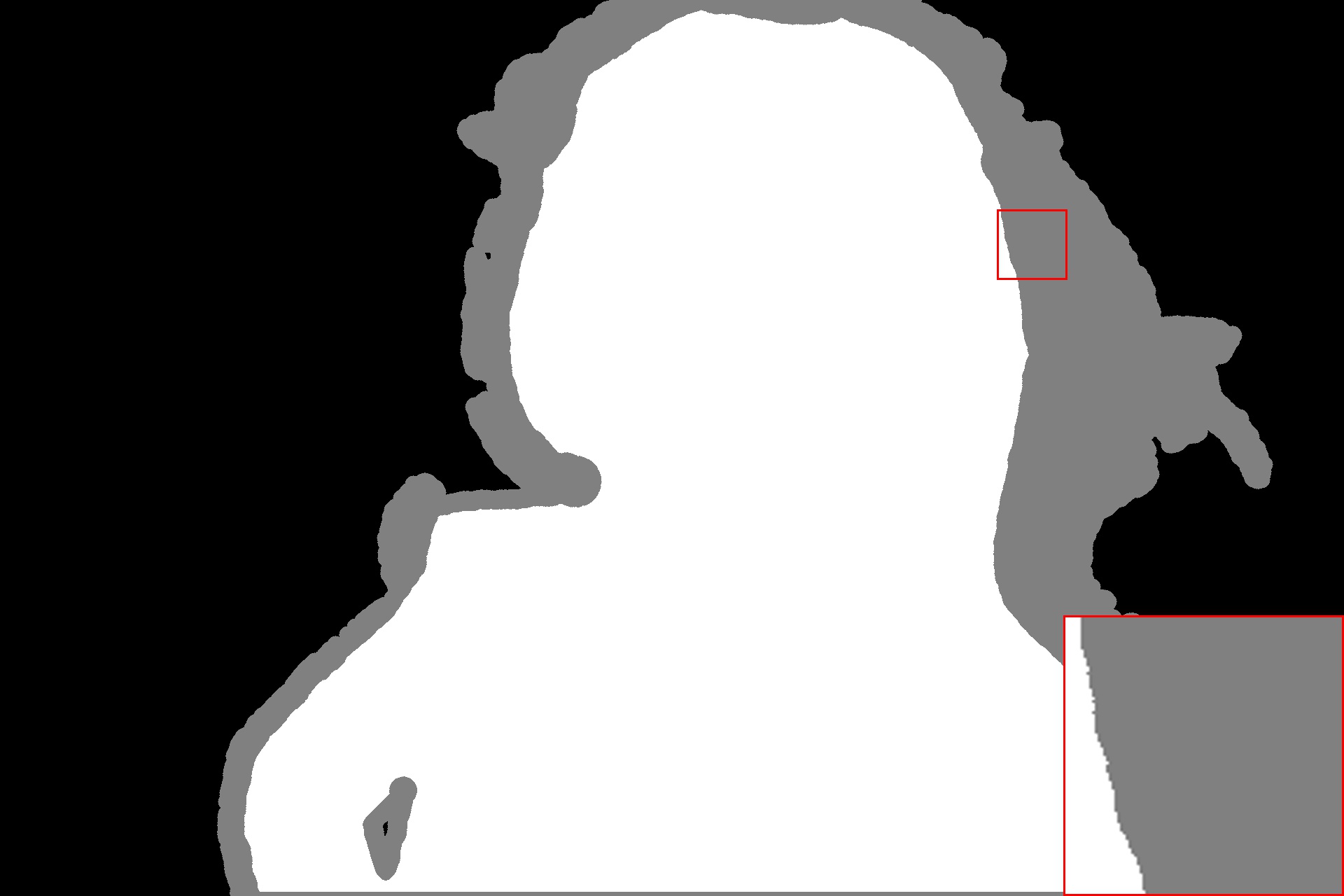}&\includegraphics[width=2cm]{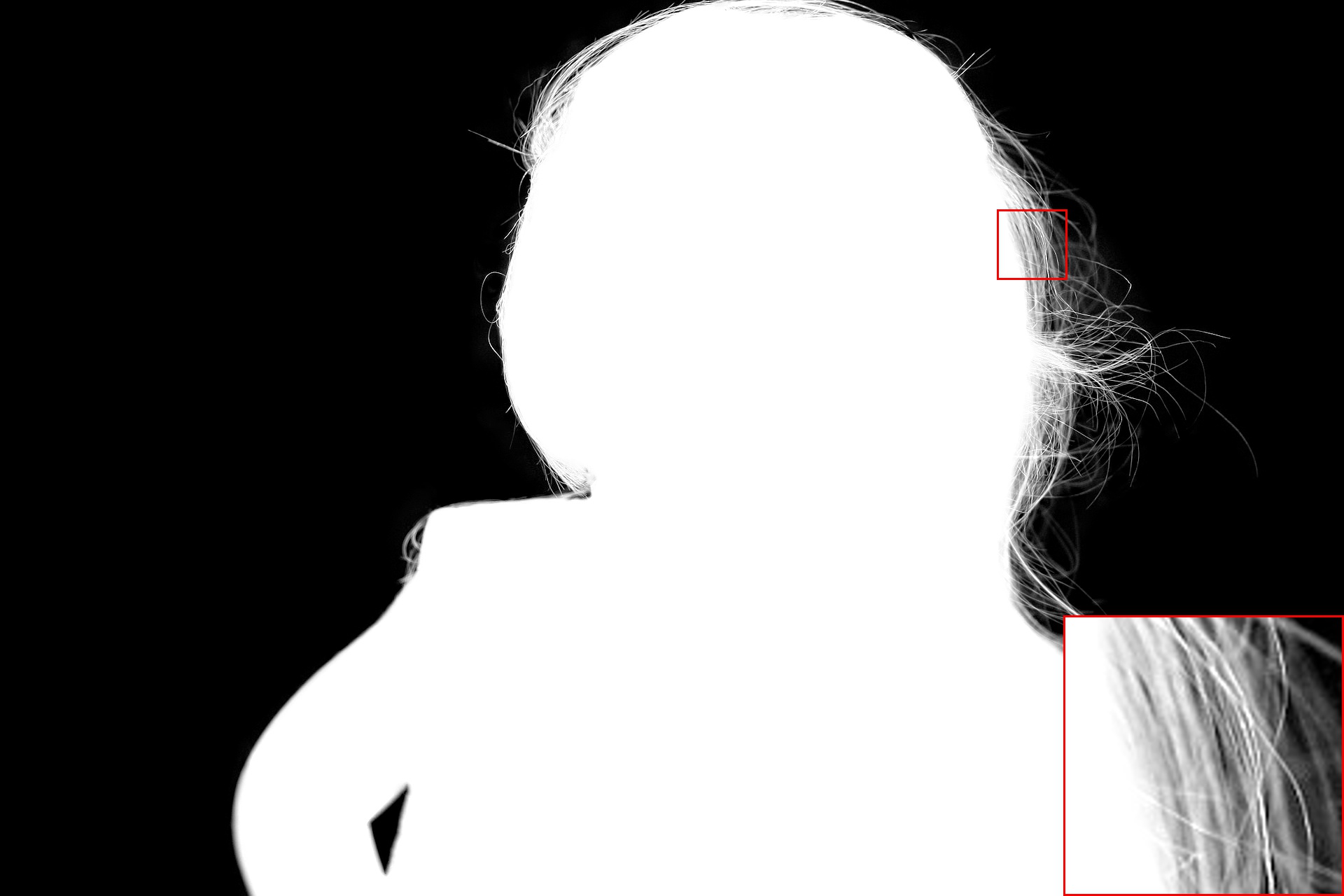}&\includegraphics[width=2cm]{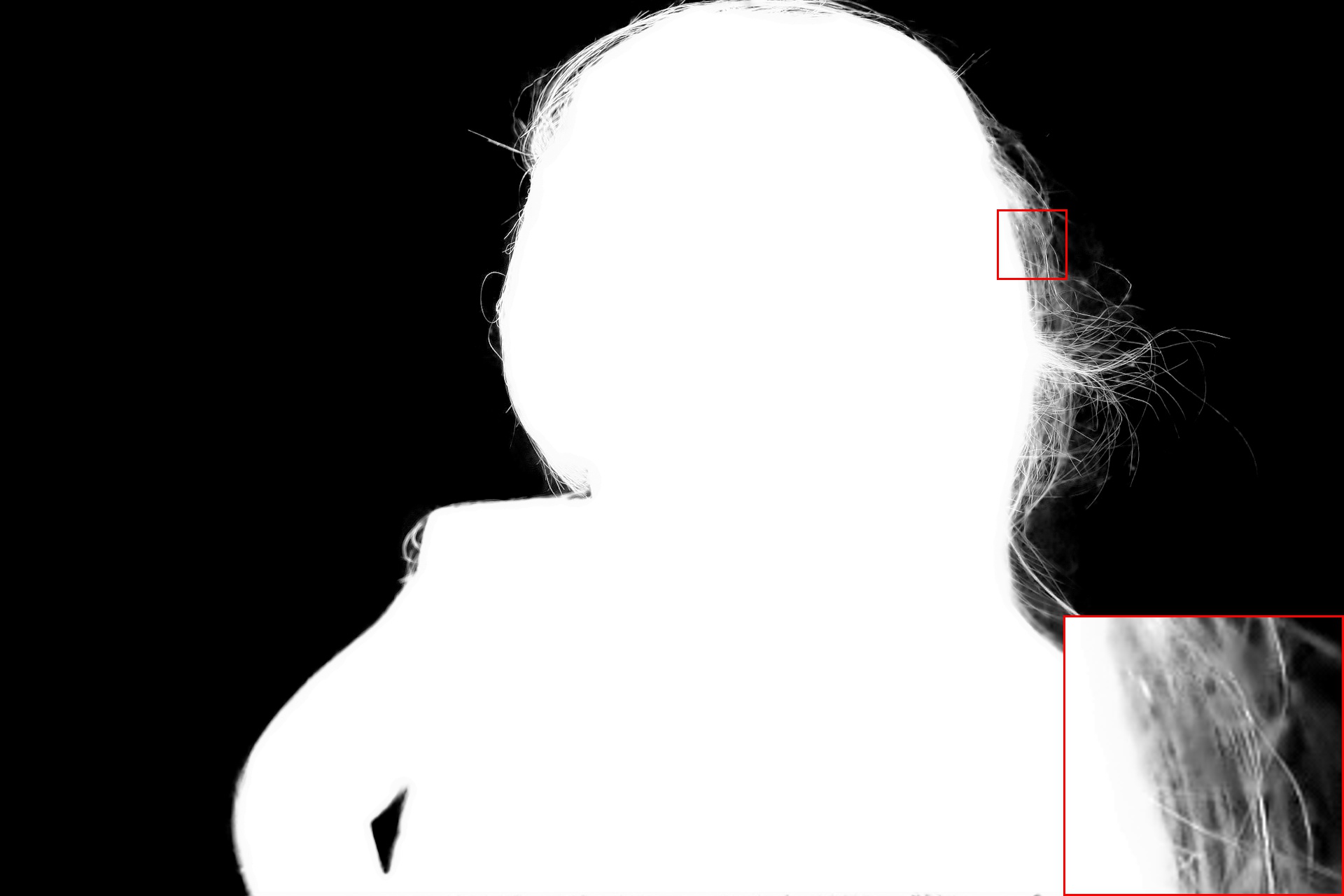}&\includegraphics[width=2cm]{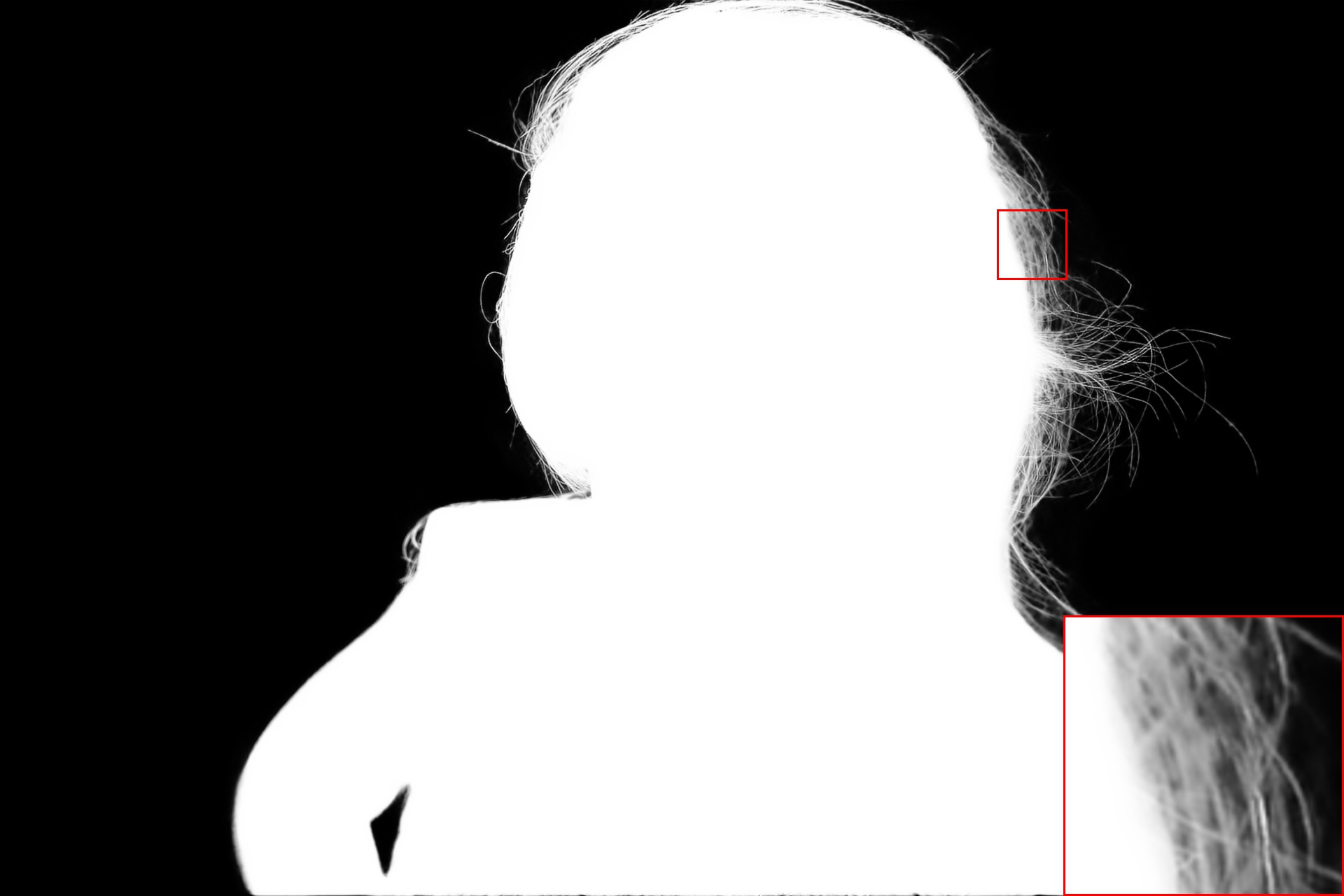}&\includegraphics[width=2cm]{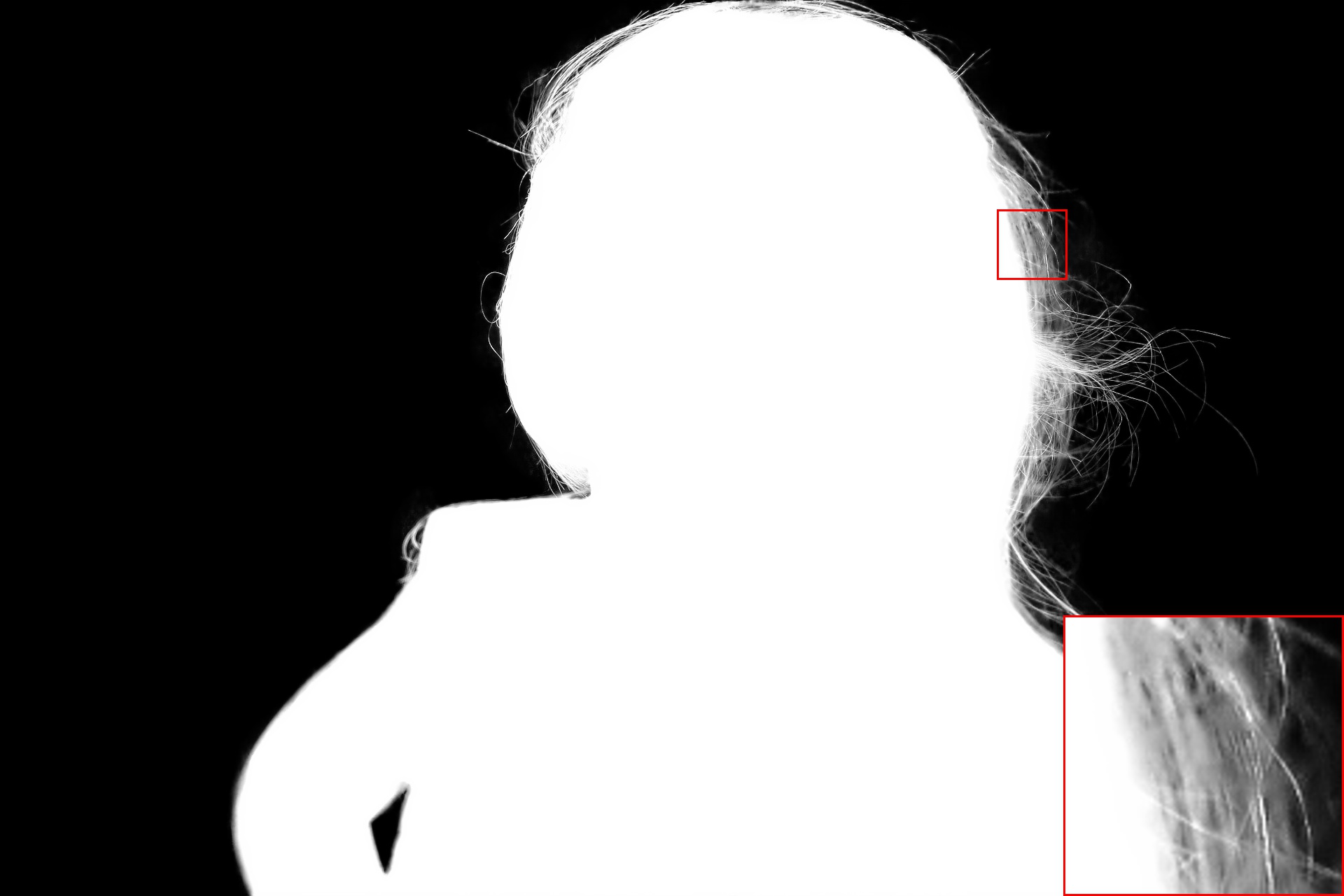}&\includegraphics[width=2cm]{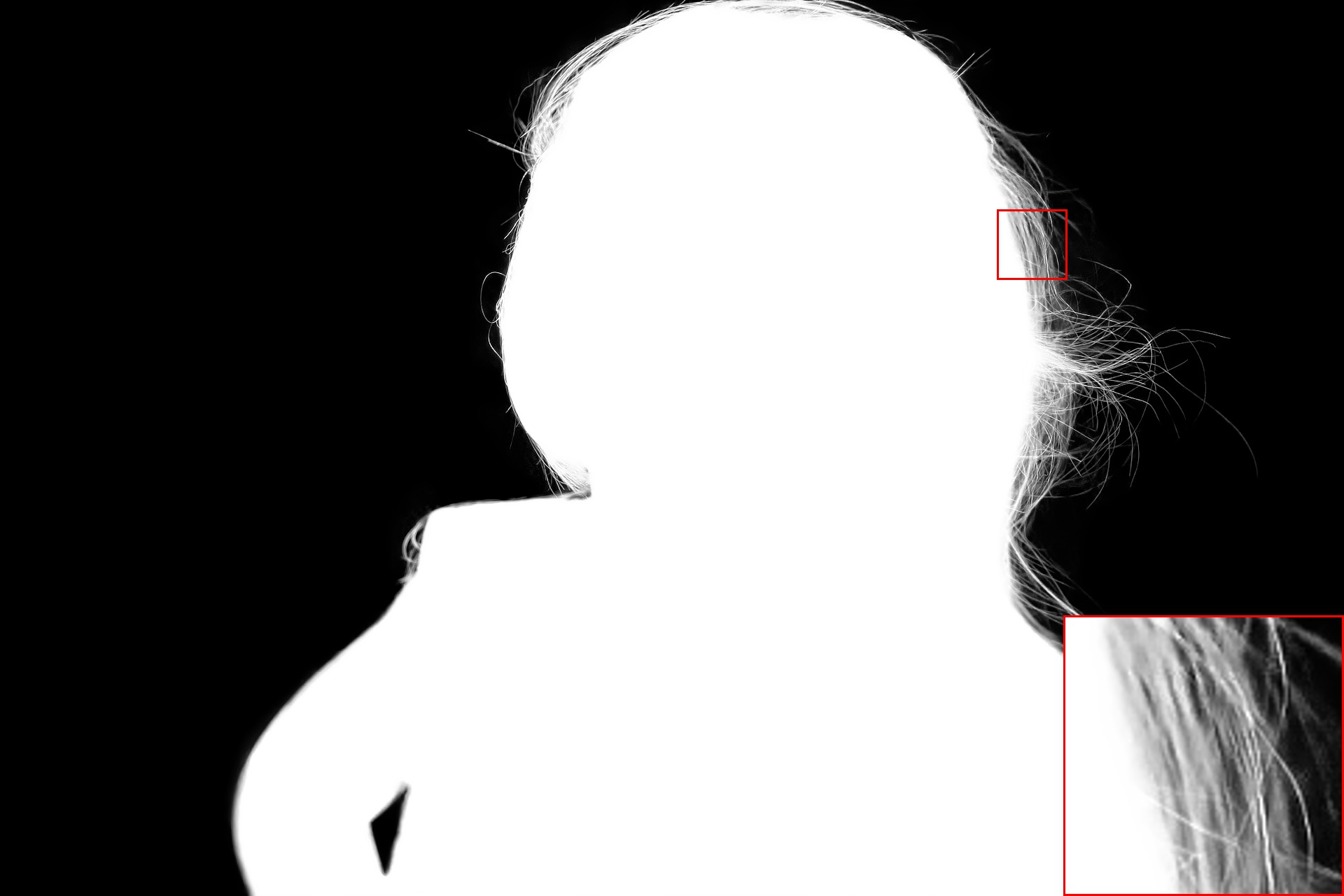}\\
\end{tabular}}
\caption{Comparison of visual results of the Composition-1k test dataset of different methods. From left to right, the original image, dataset-provided trimap, GT alpha matte, IM~\cite{lu2019indices}, CAM~\cite{hou2019context}, GCA~\cite{li2020natural}, Net-M}
\label{fig:matting}
\end{center}
\end{figure}%-3\protect\footnotemark~\cite{hou2019context}
% \footnotetext

\subsubsection{Loss Function}
The cross-entropy loss is for 3-class classification in Net-T. For Net-M, its training loss $L$ is defined as the summation over losses of coarse and refined alpha estimations, $L=L_{coarse\_\alpha}+L_{refined\_\alpha}$.
% \begin{equation}
%     L=L_{coarse\_\alpha}+L_{refined\_\alpha}.
% \end{equation}
% $L=L_{coarse\_\alpha}+L_{refined\_\alpha}$.
To obtain high-quality alpha matte, we employ the summation of alpha prediction loss $L_{alpha}$ and alpha hard mining loss $L_{hard}$ for both coarse and refined alpha prediction. The alpha prediction loss is defined as the absolute difference between the ground truth and predicted alpha, $L_{alpha}=\frac{1}{\left| \mathcal{M}\right|}\sum_{i\in \mathcal{M}}\left|\hat{\alpha}_{i}-\alpha_i\right|$, where $\mathcal{M}$ refers to the unknown region.
% \begin{equation}
%     L_{alpha}=\frac{1}{\left| \mathcal{M}\right|}\sum_{i\in \mathcal{M}}\left|\hat{\alpha}_{i}-\alpha_i\right|,
% \end{equation}
% $L_{alpha}=\frac{1}{\left| \mathcal{M}\right|}\sum_{i\in \mathcal{M}}\left|\hat{\alpha}_{i}-\alpha_i\right|$
%inspired by online hard example mining in object detection~\cite{} and hard flow example mining in video inpainting, 
Since difficulty degree of each alpha pixel for the network to learn varies a lot and indiscriminately calculating loss on unknown region might misguide training process, we introduce hard mining loss~\cite{shrivastava2016training,xu2019deep}. The hard mining loss calculates the absolute difference between the predicted and groundtruth alpha, sort all pixels, and pick the top $p$ percent of the largest error pixels as hard samples to automatically guide model to focus more on hard alpha region. The hard mining loss is defined as $L_{hard}=\frac{1}{\left| \mathcal{HM}\right|}\sum_{i\in \mathcal{HM}}\left|\hat{\alpha}_{i}-\alpha_i\right|$,
% \begin{equation}
%     L_{hard}=\frac{1}{\left| \mathcal{HM}\right|}\sum_{i\in \mathcal{HM}}\left|\hat{\alpha}_{i}-\alpha_i\right|,
% \end{equation}
where $\mathcal{HM}$ means the region that contains hard samples, $\hat{\alpha}_i,{\alpha}_i$ indicate the predicted and ground-truth alpha at position $i$ and $p=50$ in our experiments.

% \footnotetext{CAM with Matting Encoder(ME)+Context Encoder(CE)+Laplacian loss(lap)+feature loss(fea)+color loss(color)+Data Augmentation(DA)}
\section{Experiment Settings and Results}
\subsection{Experiment Settings}
% \subsubsection{Implementation Details}
We train Net-T and Net-M separately and then test the whole net jointly with fusion techniques~\cite{chen2018semantic} (Joint Inference).

\textbf{Net-T}:
Here is random soft foreground segmentation input generation process.\footnote{Please refer to supplementary material for code snapshot.} Random trimap is first produced by random erosion on both foreground and background of alpha ranging from 1 to 29 pixels~\cite{li2020natural}. The unknown and foreground areas of random trimap is considered as foreground of random \textit{initial segmentation}. Then, random soft segmentation is generated by erosing and dilating random \textit{initial segmentation} sequentially with random number of pixels ranging from 1 to 59 and followed by a random Gaussian Blur~\cite{sengupta2020background}. To obtain \textit{synthesized ground-truth trimap} for supervision, we apply 15-pixel erosion on both foreground and background of alpha. Image patches are randomly cropped from input images and then resized to 512$\times$512. We train Net-T for 129,300 iterations with 10 batch size. The learning rate is initialized to 0.001 and adjusted every iteration.

% AIM
% [01-21 01:07:26] INFO: mIoU:       0.9040147034615457
% [01-21 01:07:26] INFO: TEST NUM:      1000
% [01-21 01:07:26] INFO: mACC:       0.9640897331498012
% [01-21 01:07:26] INFO: mIoUs:      [0.93530564 0.82781042 0.70635222]
% [01-21 01:07:26] INFO: mIoU:       0.823156090830527
% Dis
% [01-21 03:16:54] INFO: TEST NUM:      1000
% [01-21 03:16:54] INFO: mACC:       0.9574376130933773
% [01-21 03:16:54] INFO: mIoUs:      [0.96535713 0.80828916 0.66724178]
% [01-21 03:16:54] INFO: mIoU:       0.813629355808835
%AIM*
% [01-21 03:37:15] INFO: TEST NUM:      1000
% [01-21 03:37:15] INFO: mACC:       0.9600102418344837
% [01-21 03:37:15] INFO: mIoUs:      [0.92754117 0.81395112 0.69563214]
% [01-21 03:37:15] INFO: mIoU:       0.812374812108391

\textbf{Net-M}: 
We follow the same data processing and augmentation procedure as GCA-Matting~\cite{li2020natural}. Net-M is trained for 400,000 iterations with 20 batch size and $L_{coarse\_\alpha}+L_{refined\_\alpha}$, where unknown region of trimap is $\mathcal{M}$. The adam optimizer with $\beta_1 = 0.5$ and $\beta_2 = 0.999$ is adopted with initialized learning rate, $4\times10^{-4}$, plus warmup and cosine decay techniques.

\textbf{Joint Inference (JI)}: Joint Inference is Net-T and Net-M collaborative testing by adopting Net-T-predicted trimap as one input of Net-M. Since 1) categories of foreground objects of synthesized matting datasets may not match segmentation datasets; 2) background objects may have the same category as foreground and can also be salient, salient/segmentation models may not be suitable for coarse segmentation generation. Therefore, JI setting here is for synthesized datasets (For real data, we provide transfer approach in Real World Human Data part of section 4.2.2.). The soft segmentation input for testing is generated by erosion on \textit{initial segmentation} derived from \textit{synthesized ground-truth trimap} with 20 pixels and followed by a Gaussian Blur. In Fig.~\ref{netm}, raw alpha estimation of Net-M is far away from impressive except unknown region. Hence, we propose two fusion methods to solve this. One is probability-based soft fusion, another is region-based hard fusion. For soft fusion, we use region probabilities estimated by Net-T to reconstruct final alpha $\alpha_r$ from predicted alpha $\alpha_p$~\cite{chen2018semantic} as $\alpha_{r}=(1-U_p)\frac{F_p}{F_p+B_p}+U_{p}\alpha_p$, $U_p=1-(F_p+B_p)$ and $\alpha_{r}=F_p+U_{p}\alpha_p$, where $U_p$, $F_p$, and $B_p$ are probabilities of each pixel belonging to unknown/foreground/background regions severally. For hard fusion, we reset trimap-predicted foreground (background) to 255 (0) on predicted alpha.

% \begin{table}[h]
%   \begin{center}
%     \resizebox{0.7\columnwidth}{!}{%
%     \begin{tabular}{l|c|c|c|c}
%       \toprule % <-- Toprule here
%       \multirow{2}{*}{\textbf{Methods}} & \multicolumn{4}{c}{\textbf{Evaluation metrics}}\\
%     %   \cmidrule{2-5}\\
%       & SAD & MSE &Grad & Conn\\
%       \midrule % <-- Midrule here
%       IndexNet Matting\cite{lu2019indices} & 14.02 & 0.0094 & 6.08 & 12.03\\
%       CAM-1~\tablefootnote{CAM with ME+CE+lap}\cite{hou2019context} & 10.92 & 0.0062 & 5.00  & 9.10 \\
%       CAM-2~\tablefootnote{CAM with ME+CE+lap+fea+DA}\cite{hou2019context} & 16.18 & 0.0136 & 9.56  & 14.95 \\
%       CAM-3~\tablefootnote{CAM with ME+CE+lap+fea+color+DA}\cite{hou2019context} & 16.72 & 0.0139 & 9.69 & 15.39\\
%       GCA-Matting\cite{li2020natural} & 11.45 & 0.0068 & 4.18 & 9.98\\
%     %   Background Matting(GT)$^{\dagger}$\cite{sengupta2020background} & 35.28 & 0.0091 & 16.92 & 32.53\\
%       \midrule % <-- Midrule here
%       Net-M & \textbf{9.00} & \textbf{0.0046} & \textbf{3.13}  &\textbf{7.31}\\
%       Net-M-nh&9.82&0.0054&3.50&8.22\\
%       NLM & 9.42 & 0.0048 & 3.17 & 7.64\\
%       %Joint Inference with Fusion$^{\dagger}$&\\
%       \bottomrule % <-- Bottomrule here
%     \end{tabular}}
%     \caption{The quantitative results on Composition-240 test dataset. %($^{\dagger}$ means that all metrics are calculated on the whole image.)
%     }
%     \label{tab:humantest}
%   \end{center}
% \end{table}

\begin{table}[thpb]
\huge
% \LARGE
% \Large
  \begin{center}
    \resizebox{0.75\columnwidth}{!}{%
    \begin{tabular}{c|ccccc}
      \toprule % <-- Toprule here
      \multicolumn{6}{c}{\textbf{AIM}}\\
     % \bottomrule % <-- Bottomrule here
    %   \multirow{1}{*}{\textbf{Methods}} &\multicolumn{5}{c}{\textbf{Evaluation metrics}}\\
      \cmidrule{1-6}
      Methods & pixAcc & mIoU Bg  & mIoU Unk  & mIoU Fg  & mIoU\\
      \midrule % <-- Midrule here
      Deeplabv2&93.58&89.81&76.25&63.09&76.38\\
      Deeplabv3&95.83&91.59&80.25&\textbf{70.66}&80.83\\
      Net-T&\textbf{96.41}&\textbf{93.53}&\textbf{82.78}&70.64&\textbf{82.32}\\
      % Deeplabv3+&0.9549&0.9364&0.8325&0.6935&0.8208\\
      % AIM$^{\bigstar}$&0.9600&0.9275&0.8140&0.6956&0.8124\\
      % Distinctions-646&0.9574&0.9654&0.8083&0.6672&0.8136\\
      \bottomrule % <-- Bottomrule here
       \multicolumn{6}{c}{\textbf{Distinctions-646}}\\
    %   \bottomrule % <-- Bottomrule here
    %   \multirow{2}{*}{\textbf{Methods}} & \multicolumn{5}{c}{\textbf{Evaluation metrics}}\\
      \cmidrule{1-6}
      Methods & pixAcc & mIoU Bg  & mIoU Unk & mIoU Fg  & mIoU\\
      \midrule % <-- Midrule here
      % AIM&0.9309&0.8717&0.6663&0.6351&0.7244\\
      % AIM$^{\bigstar}$&0.3452&1.0612e-05&3.4229e-04&2.6485e-04&0.0002059\\
      Deeplabv2&92.58&94.40&72.97&53.82&73.73\\
      Deeplabv3&95.66&95.11&79.65&65.52&80.09\\
      Net-T&\textbf{95.74}&\textbf{96.54}&\textbf{80.83}&\textbf{66.72}&\textbf{81.36}\\
      % Deeplabv3+&-&-&-&-&-\\
      \bottomrule % <-- Bottomrule here
    %    \multicolumn{6}{c}{Deeplabv3+}\\
    %    \midrule
    %   \multirow{2}{*}{\textbf{Datasets}} & \multicolumn{5}{c}{\textbf{Evaluation metrics}}\\
    % %   \cmidrule{2-5}\\
    %   & Accuracy & mean Bg IoU &mean Unknown IoU &mean Fg IoU & mIoU\\
    %   \midrule % <-- Midrule here
    %   AIM&0.9549&0.9364&0.8325&0.6935&0.8208\\
    %   AIM$^{\bigstar}$&0.9515&0.9270&0.8172&0.6821&0.8088\\
    %   Distinctions-646&-&-&-&-&-\\
    %   \bottomrule % <-- Bottomrule here
    \end{tabular}}
    \caption{The quantitative comparison of Net-T with adapted Deeplabv3 and Deeplabv2 on Adobe Image Benchmark testset and Distinctions-646 testset. Bold numbers represent the best scores.}
    \label{tab:nett}
  \end{center}
\end{table}

\subsection{Experiment Results}
% \wzx{I think the this part should focus on group photo synthesis, and give some discussions, e.g., ablation study, failure cases,}
\subsubsection{Evaluation on trimap}
We evaluate Net-T and other popular adapted semantic segmentation methods, e.g. Deeplabv3 and Deeplabv2, on Composite-1k and Distinctions-646 test set, by using pixel classification accuracy (pixAcc) and mean IoU (mIoU) metrics of background(Bg)/unknown(Unk)/foreground(Fg), and three-region-involved. Results in Table~\ref{tab:nett} shows that trimap segmentation accuracy and mIoU metrics of Net-T are superior to other methods. Considering quantitative results in Table~\ref{tab:nett} and visual examples presented in Fig.~\ref{fig:joint}, our trimap estimation is competent to mentor Net-M.

\subsubsection{Evaluation on alpha matte}
% \textcolor{red}{Motivation?}
We follow common evaluation metrics, i.e. Sum of Absolute Differences (SAD), Mean Squared Error (MSE), Gradient error (Grad), and Connectivity error (Conn) to evaluate our approaches on popular matting datasets, including Composition-1k, alphamatting.com\footnote{Please see both quantitative and qualitative results of alphamatting.com benchmark on supplementary material.}, and Distinctions-646 benchmarks. To validate unlabeled real-world adaption ability of our approach, we choose human as a typical illustration case and conduct a user study for evaluation.

\begin{table}[ht!]
\huge
  \begin{center}
    \resizebox{0.75\columnwidth}{!}{%
    \begin{tabular}{c|cccc}
    \toprule % <-- Toprule here
      \multicolumn{5}{c}{\textbf{Trimap-needed Evaluation}}\\
     % \midrule % <-- Midrule here
    %   \multirow{2}{*}{\textbf{Methods}} & \multicolumn{4}{c}{\textbf{Evaluation metrics}}\\
      \cmidrule{1-5}
      Methods & SAD $\downarrow$ & MSE$\downarrow$ &Grad$\downarrow$ & Conn$\downarrow$ \\
      \midrule % <-- Midrule here
      % \cmidrule{1-2}\cmidrule{3-6}
      AlphaGAN& 52.40 & 0.0300 & 38.00 & 53.00\\
      Deep Image Matting(DIM)& 50.40 & 0.0140 & 30.00 & 50.80\\
      IndexNet Matting(IM) & 45.80 & 0.0130 & 25.90 & 43.70\\
      AdaMatting & 41.70 & 0.0100 & 16.80 & -\\
      Learning Based Sampling & 40.35 & 0.0099 & -  & -\\
      Context-Aware Matting(CAM) & 35.80 & 0.0082 & 17.30  & 33.20 \\
      GCA-Matting(GCA) & 35.28 & 0.0091 & 16.92 & 32.53\\
      HDMatt&33.50 &0.0073 &14.50 &29.90\\
      \midrule
      Net-M & \textbf{29.43}& \textbf{0.0060} & 12.27 & \textbf{25.80}\\
      Net-M w/o $L_{hard}$(Net-M-nh)&30.75&0.0063&13.12&27.31\\
      Non-local Matting(NLM) & 29.77&0.0061&\textbf{11.96}&26.10\\
      % \midrule
      % JIS&37.96&0.01283&18.23&35.42\\
      % JIS w/o refinement(JIS-c)&37.87&0.01256&17.97&35.18\\
      % JIH&\textbf{37.06}&\textbf{0.01253}&\textbf{17.81}&\textbf{34.47}\\
      % JIH w/o refinement(JIH-c)&37.51&0.01262&18.04&34.88\\
      \bottomrule % <-- Bottomrule here
      \multicolumn{5}{c}{\textbf{Trimap-free Evaluation}}\\
    %   \midrule % <-- Midrule here
    %   \multirow{2}{*}{\textbf{Methods}} & \multicolumn{4}{c}{\textbf{Evaluation metrics}}\\
      \cmidrule{1-5}
      Methods & SAD$\downarrow$ & MSE$\downarrow$ &Grad $\downarrow$& Conn$\downarrow$\\
      \midrule % <-- Midrule here
      Late Fusion$^{\dagger}$&58.34&0.011&41.63&59.74\\
      HAttMatting$^{\dagger}$&44.01&0.007&29.26&46.41\\
      \midrule % <-- Midrule here
      JIS w/o refinement(JIS-c)$^{\dagger}$&47.43&0.0055&18.61&41.43\\
      JIS$^{\dagger}$&43.91&\textbf{0.0054}&18.64&39.82\\
      JIH w/o refinement(JIH-c)$^{\dagger}$&42.39&  0.0056&18.59&39.87\\
      JIH$^{\dagger}$&\textbf{41.85}&0.0055&\textbf{18.34}&\textbf{39.42}\\
      \bottomrule % <-- Bottomrule here
    \end{tabular}}
    \caption{The quantitative results on Composition-1k testset. (- indicates not given in the original paper. $^{\dagger}$ means that all metrics are calculated on the whole image. JIS and JIH means JI with soft and hard fusion respectively. JIS, JIH, -, and $^{\dagger}$ express forementioned meanings through the entire paper.)}
    \label{tab:1ktest}
  \end{center}
\end{table}

\begin{table}[thpb]
\huge
  \begin{center}
    \label{tab:distinction}
    \resizebox{0.75\columnwidth}{!}{%
    \begin{tabular}{c|cccc}
      \toprule % <-- Toprule here
    %   \multirow{2}{*}{\textbf{Methods}} & \multicolumn{4}{c}{\textbf{Evaluation metrics}}\\
      % \cmidrule{1-5}
      Methods & SAD$\downarrow$ & MSE$\downarrow$ &Grad$\downarrow$ & Conn$\downarrow$\\
      \midrule % <-- Midrule here
      Deep Image Matting(DIM)$^{\dagger}$&47.56&0.009&43.29&55.90\\
      HAttMatting$^{\dagger}$&48.98&0.009&41.57&49.93\\
      \midrule % <-- Midrule here
      JIS-c$^{\dagger}$&46.14&\textbf{0.0079}&\textbf{36.74}&40.13\\
      JIS$^{\dagger}$&41.91 & 0.0080 & 38.38&39.46\\
      JIH-c$^{\dagger}$&39.33&0.0081&38.48&38.76\\
      JIH$^{\dagger}$&\textbf{38.52}&0.0080&38.23&\textbf{38.28}\\
      \bottomrule % <-- Bottomrule here
    \end{tabular}}
    \caption{The quantitative results on Distinctions-646 testset.}
    \label{tab:distinction}
  \end{center}
\end{table}

\begin{figure}[thpb]
\setlength\tabcolsep{0pt}
\Large
\renewcommand{\arraystretch}{0.25}
\begin{center}
% \subfloat[Distinctions-646 benchmark]{
\resizebox{\columnwidth}{!}{
\begin{tabular}{ccccccc}
\includegraphics[width=1.25cm]{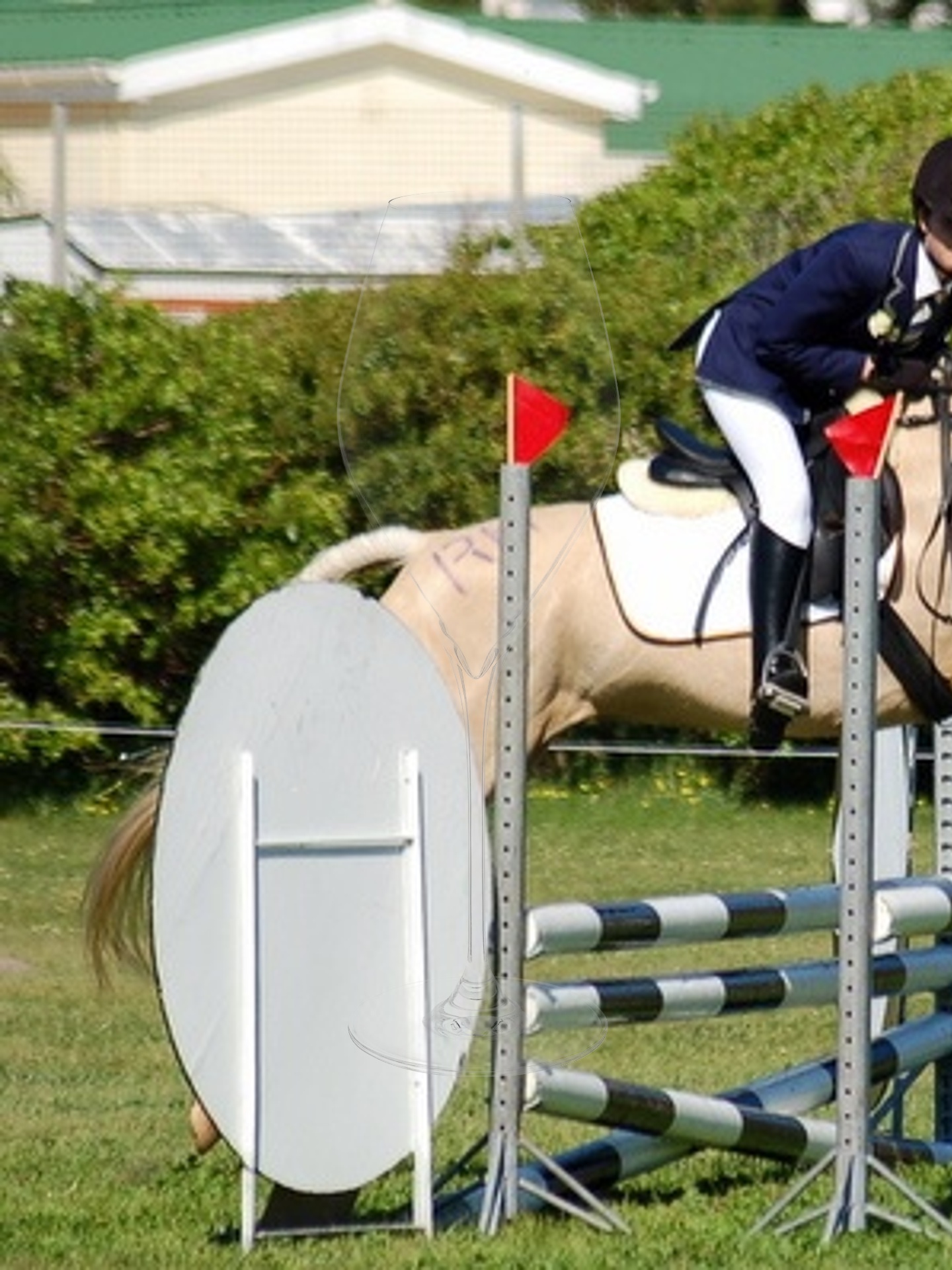}&\includegraphics[width=1.25cm]{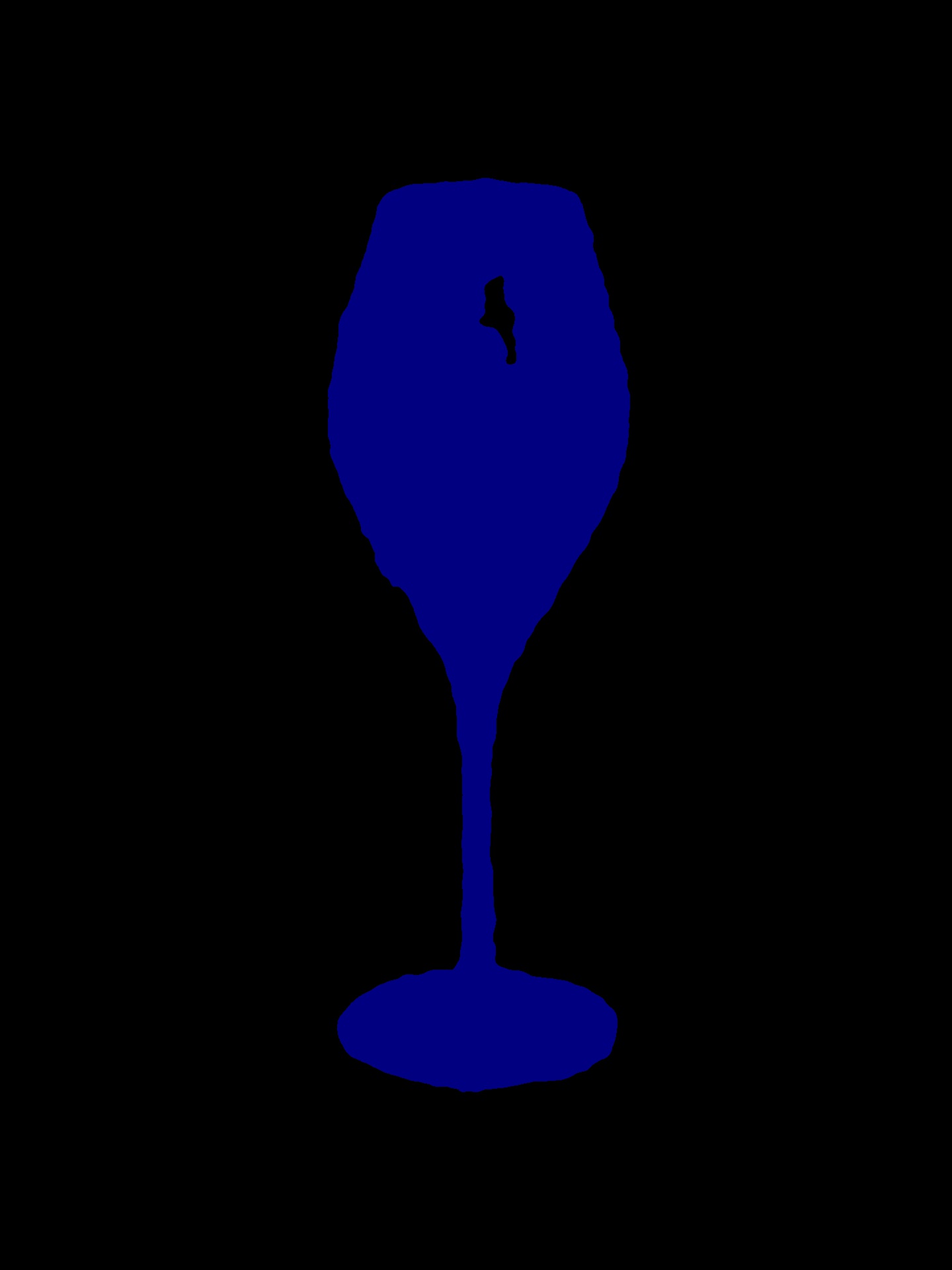}&\includegraphics[width=1.25cm]{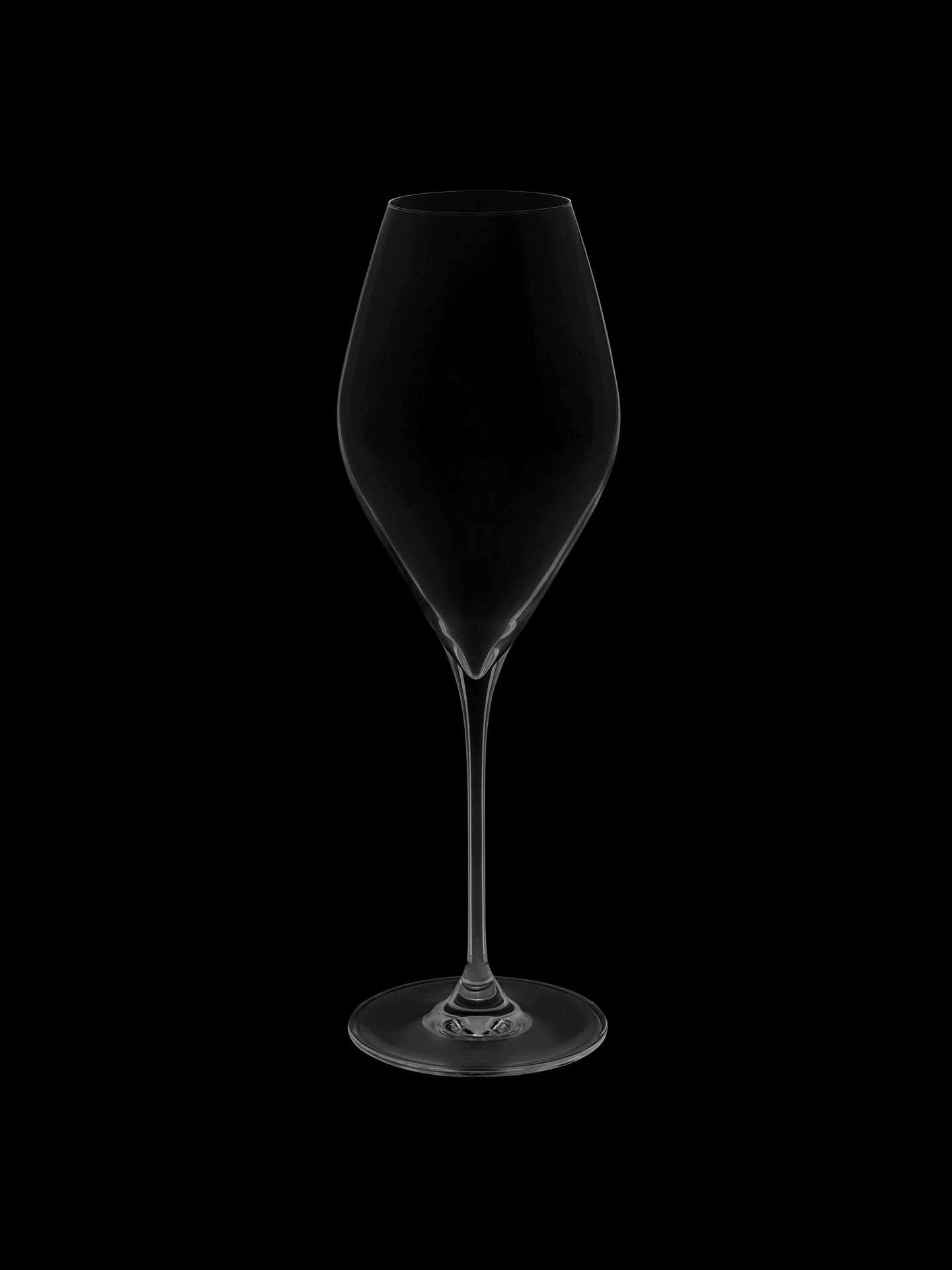}&\includegraphics[width=1.25cm]{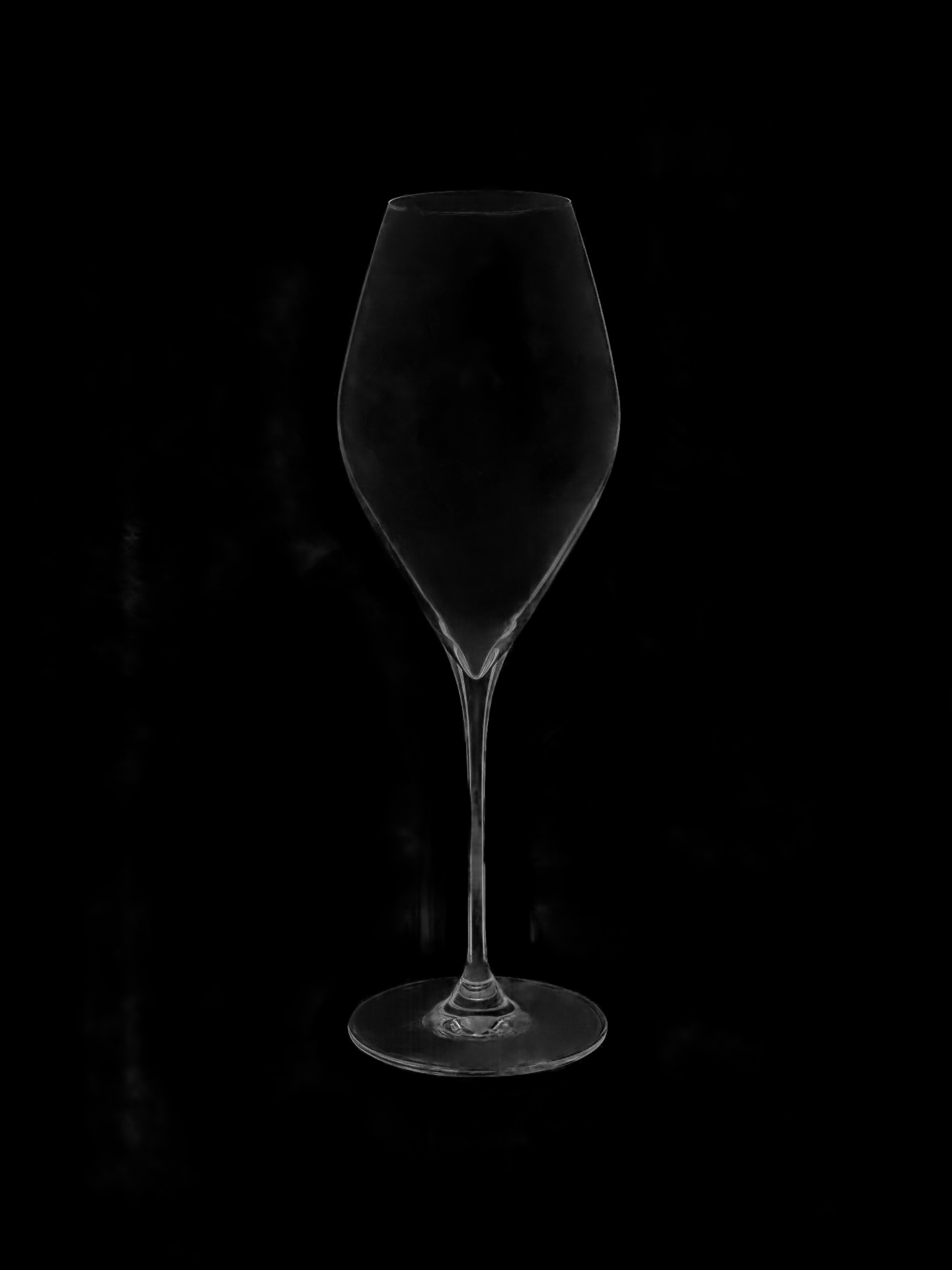}&\includegraphics[width=1.25cm]{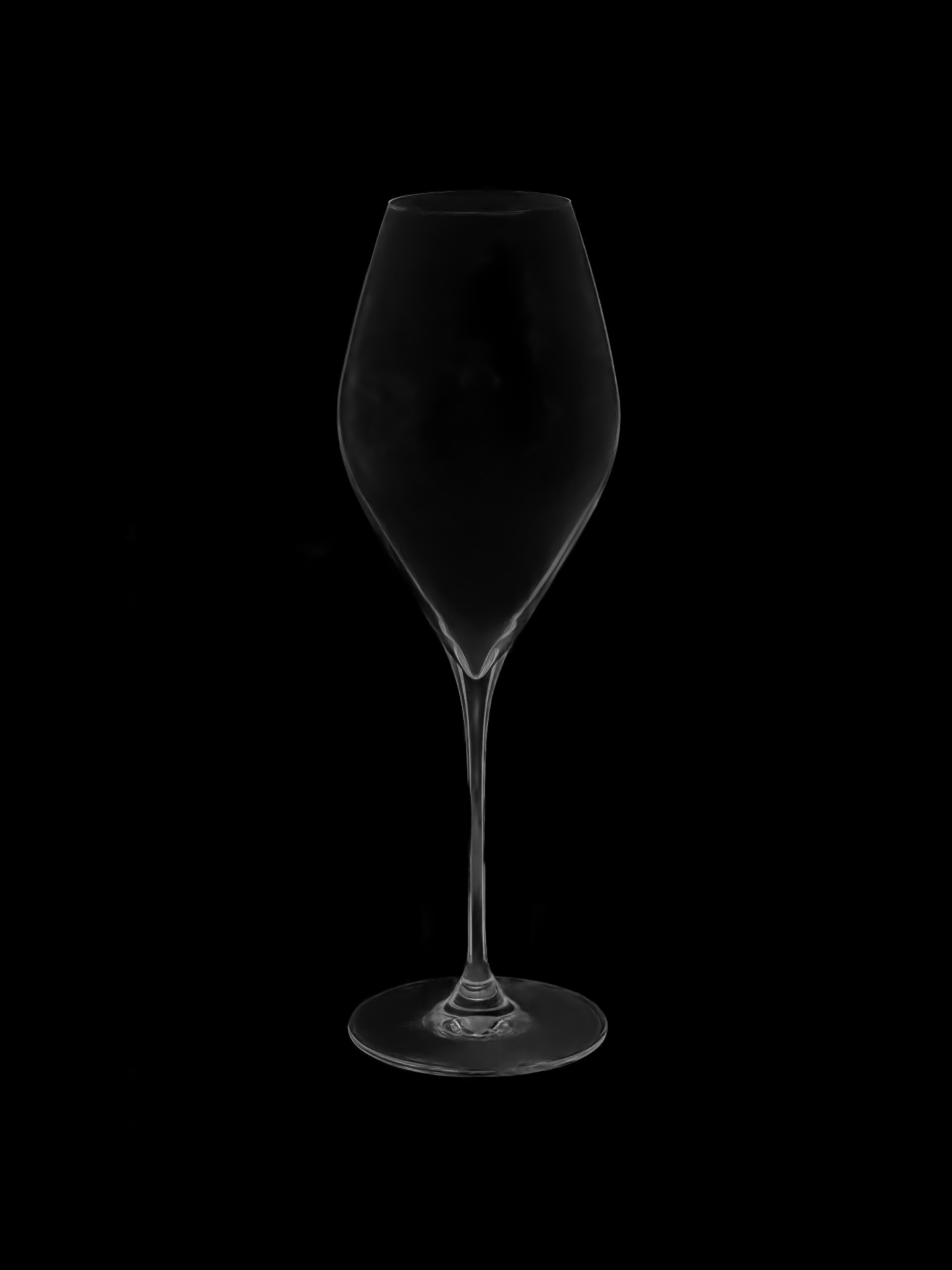}&\includegraphics[width=1.25cm]{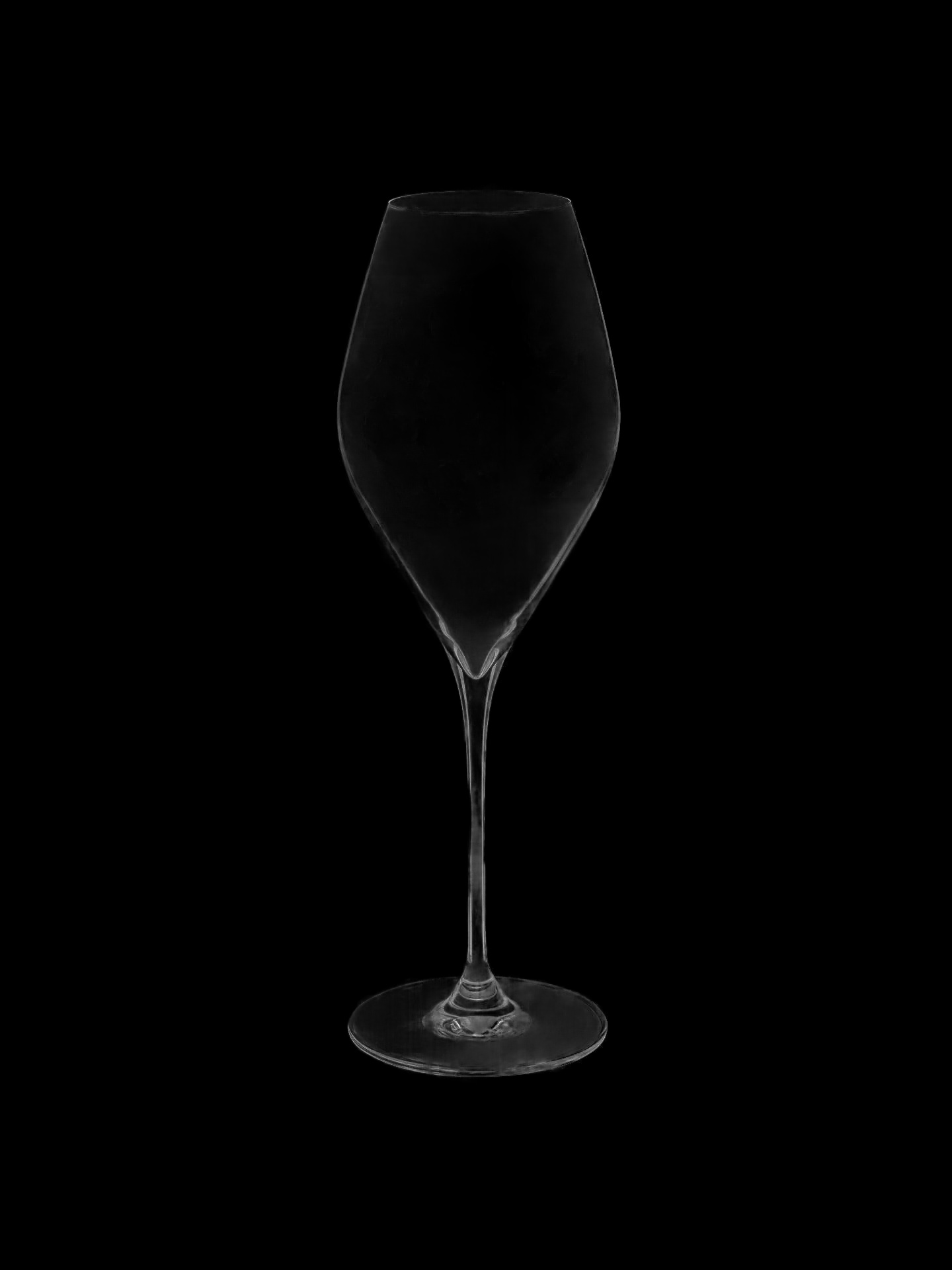}&\includegraphics[width=1.25cm]{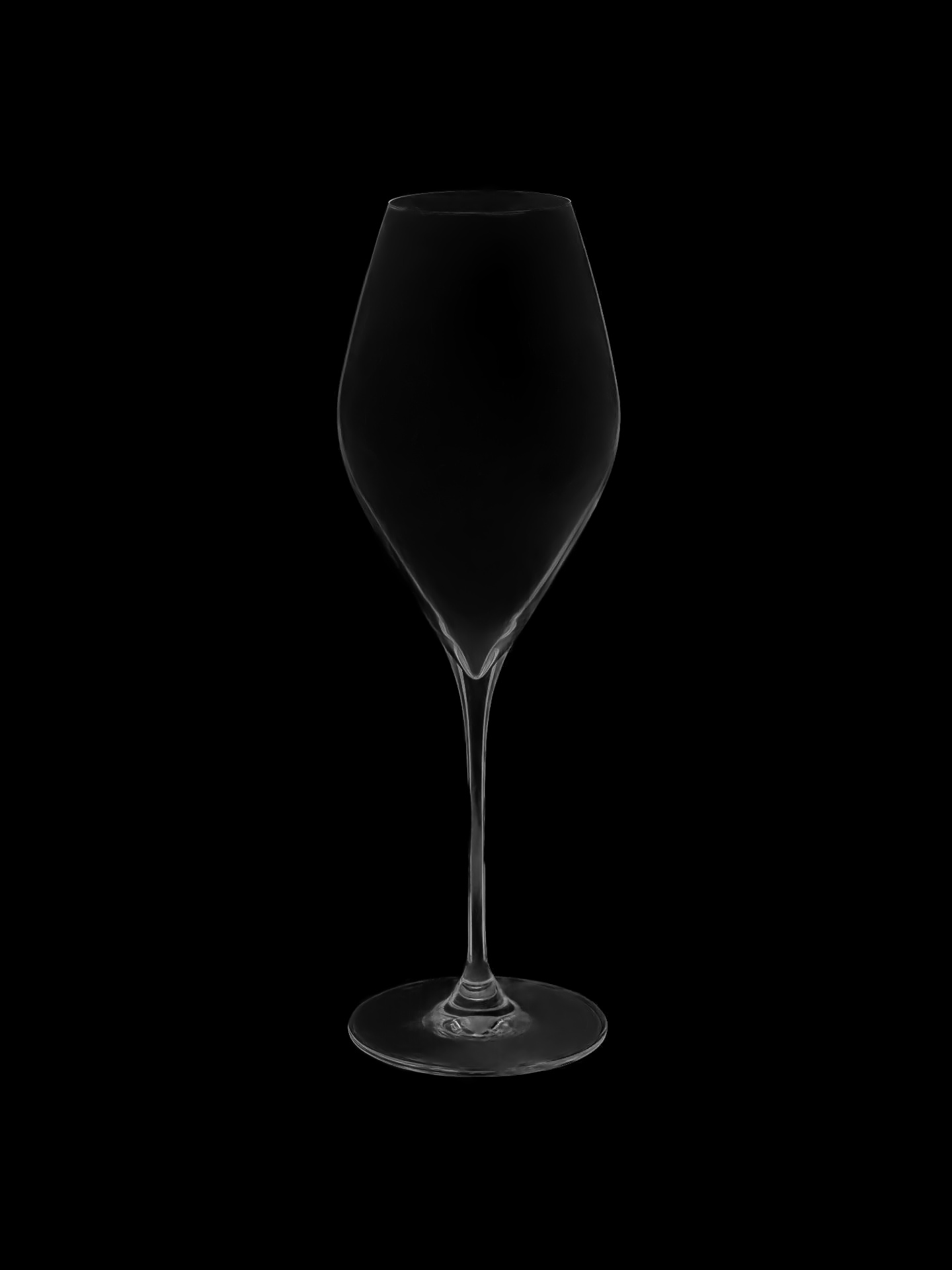}\\

\includegraphics[width=1.25cm]{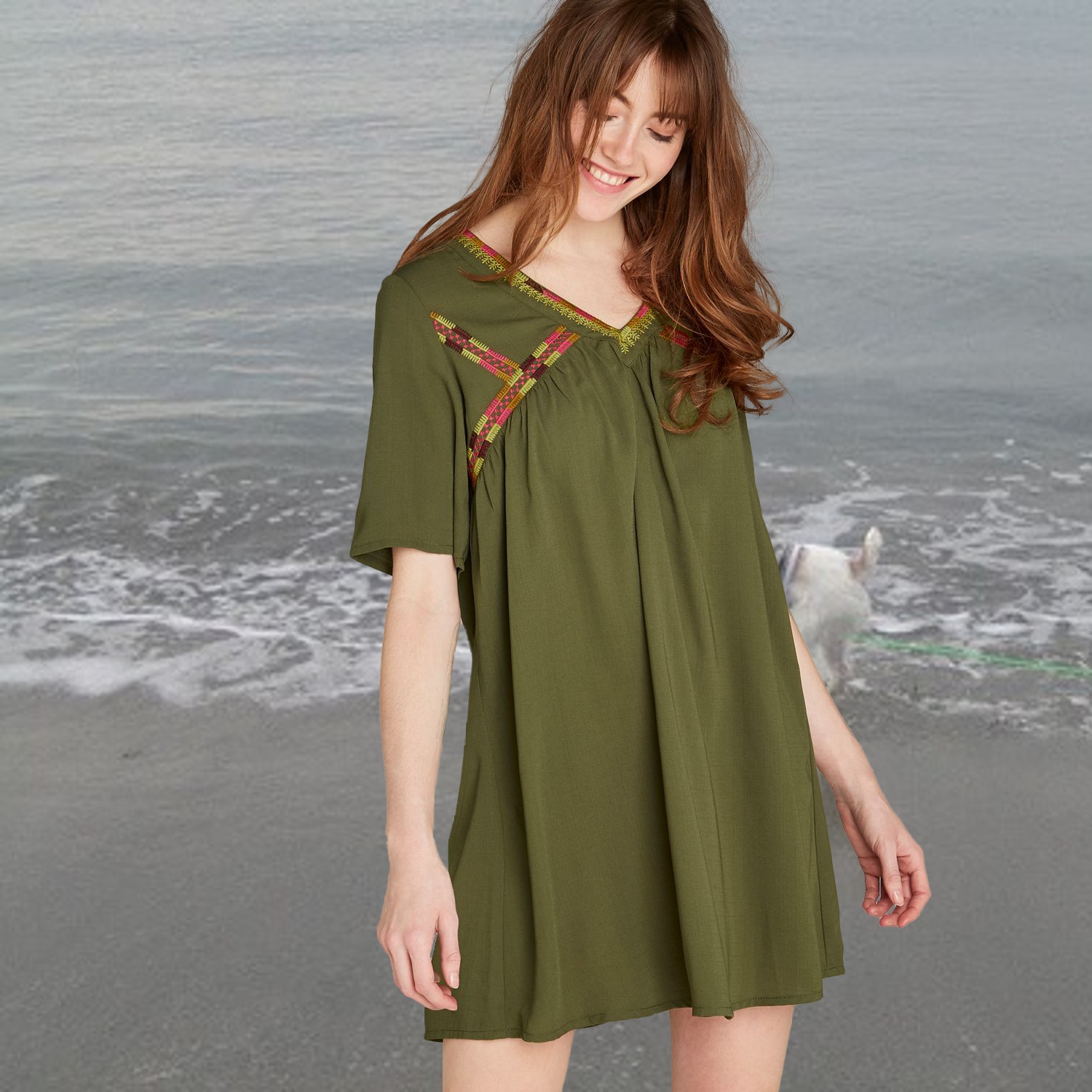}&\includegraphics[width=1.25cm]{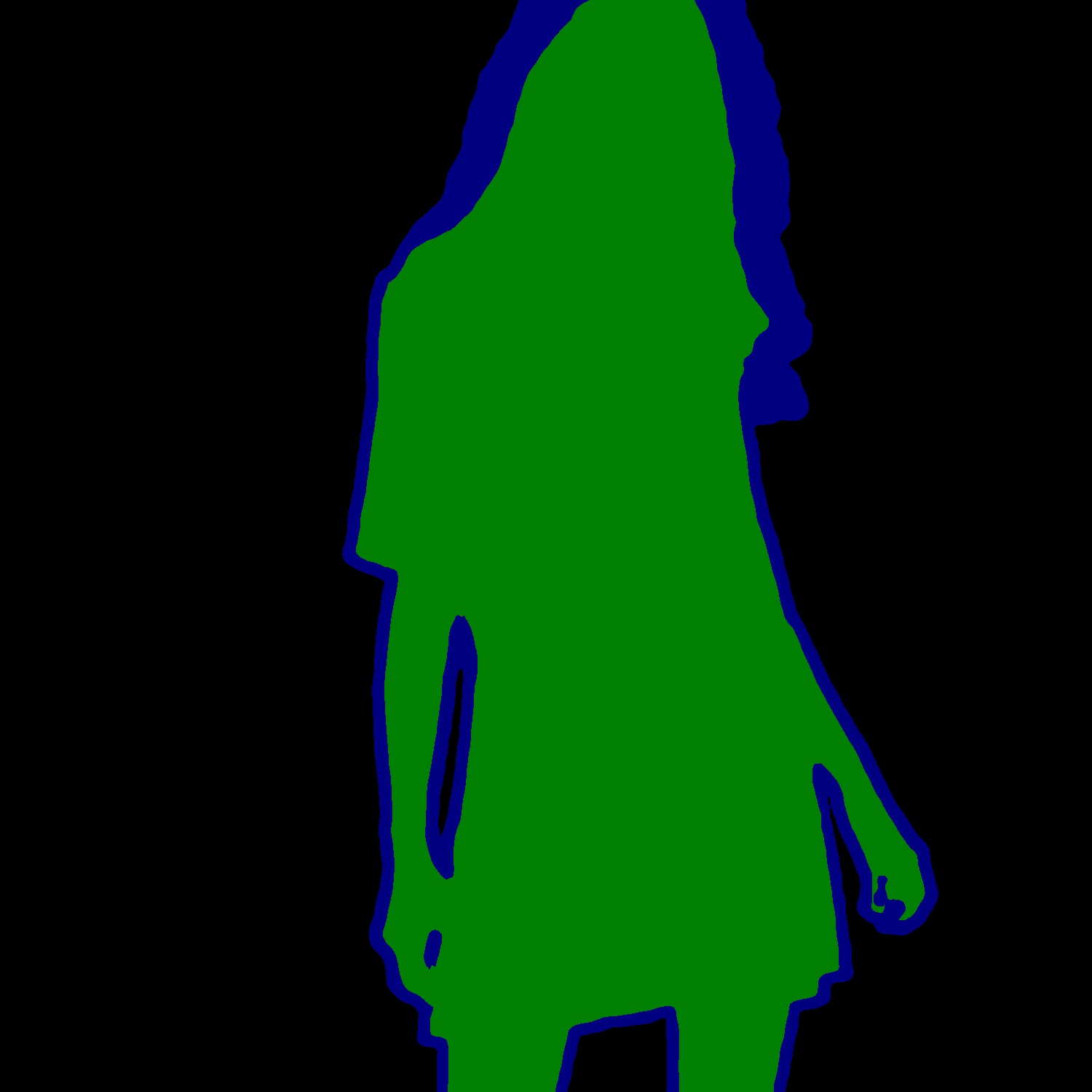}&\includegraphics[width=1.25cm]{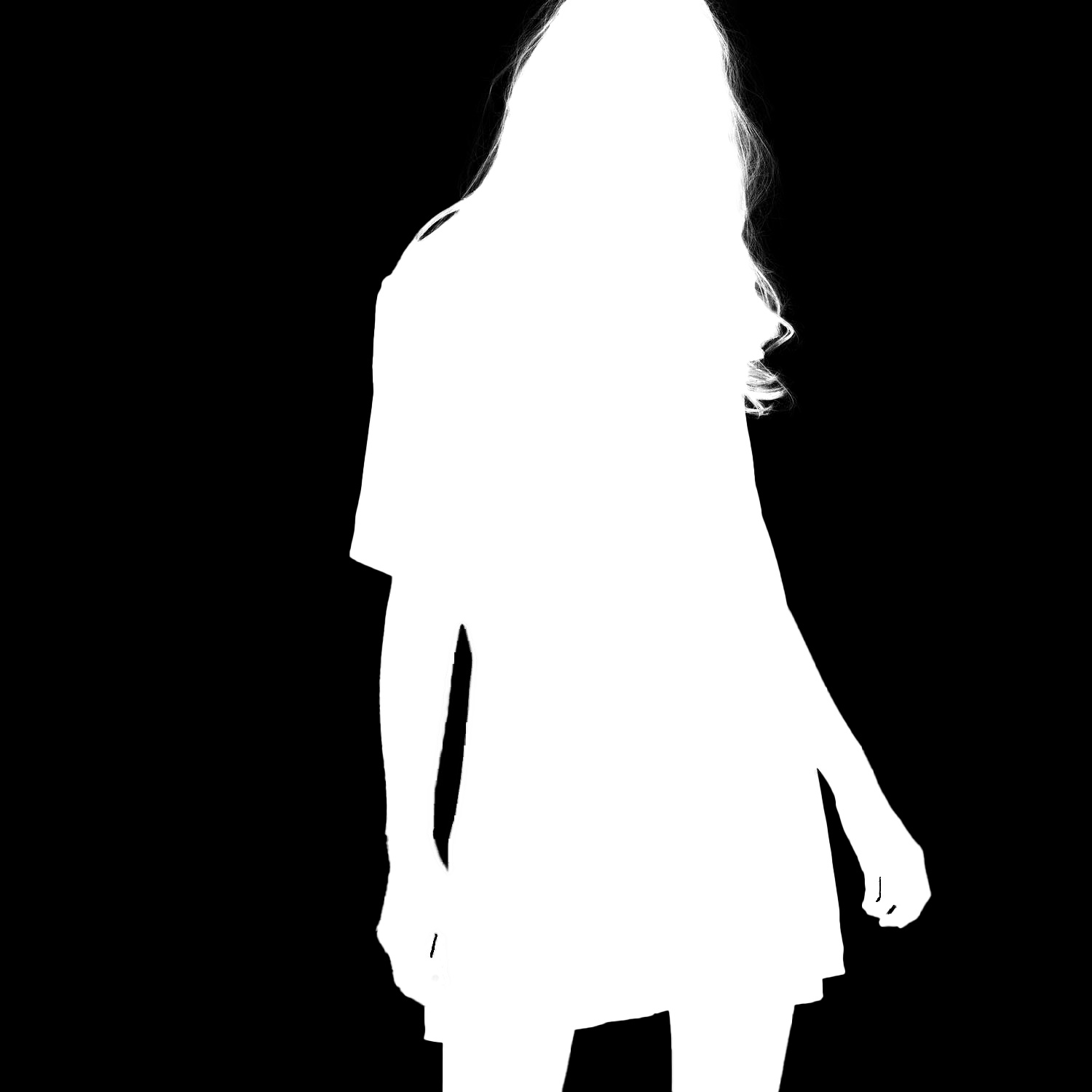}&\includegraphics[width=1.25cm]{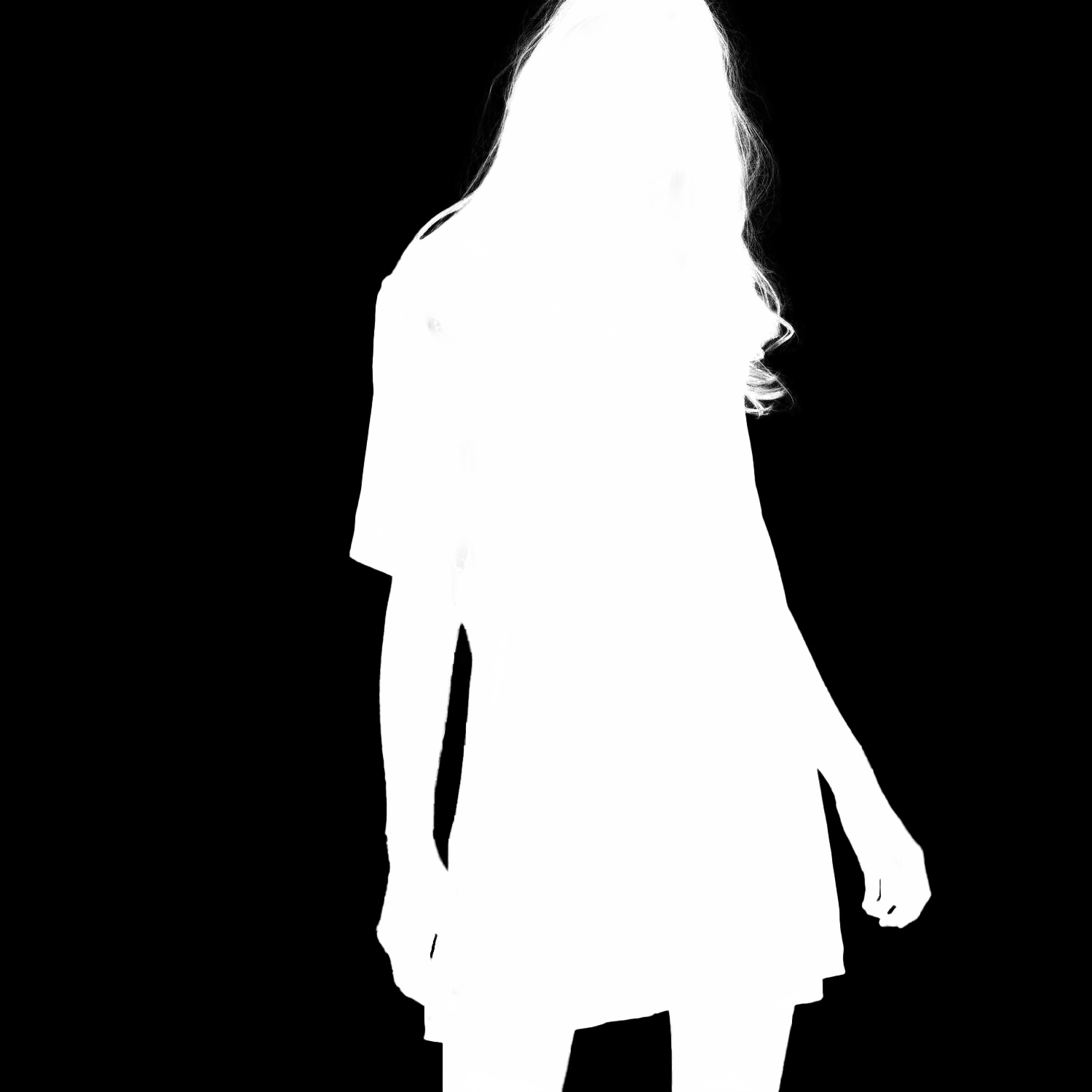}&\includegraphics[width=1.25cm]{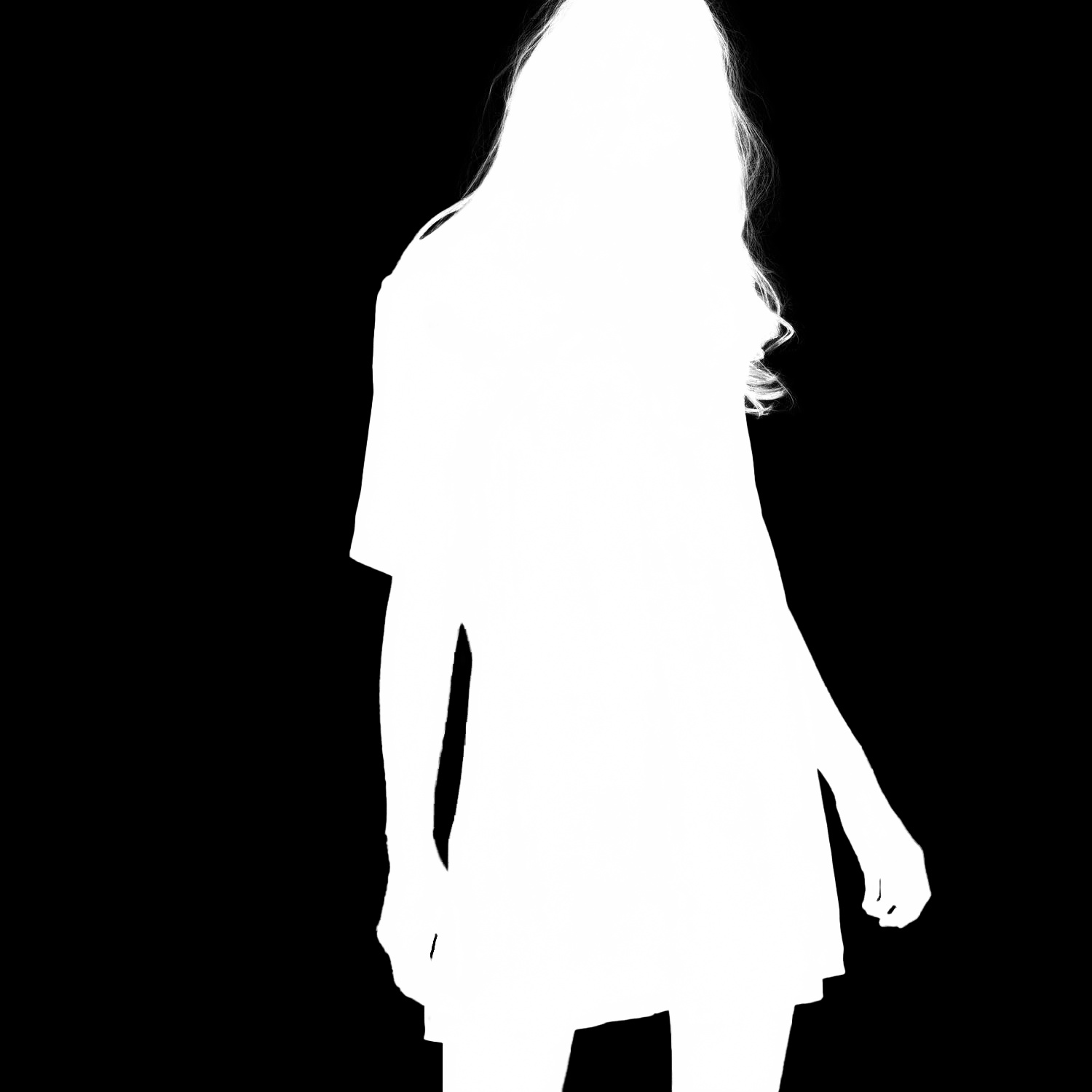}&\includegraphics[width=1.25cm]{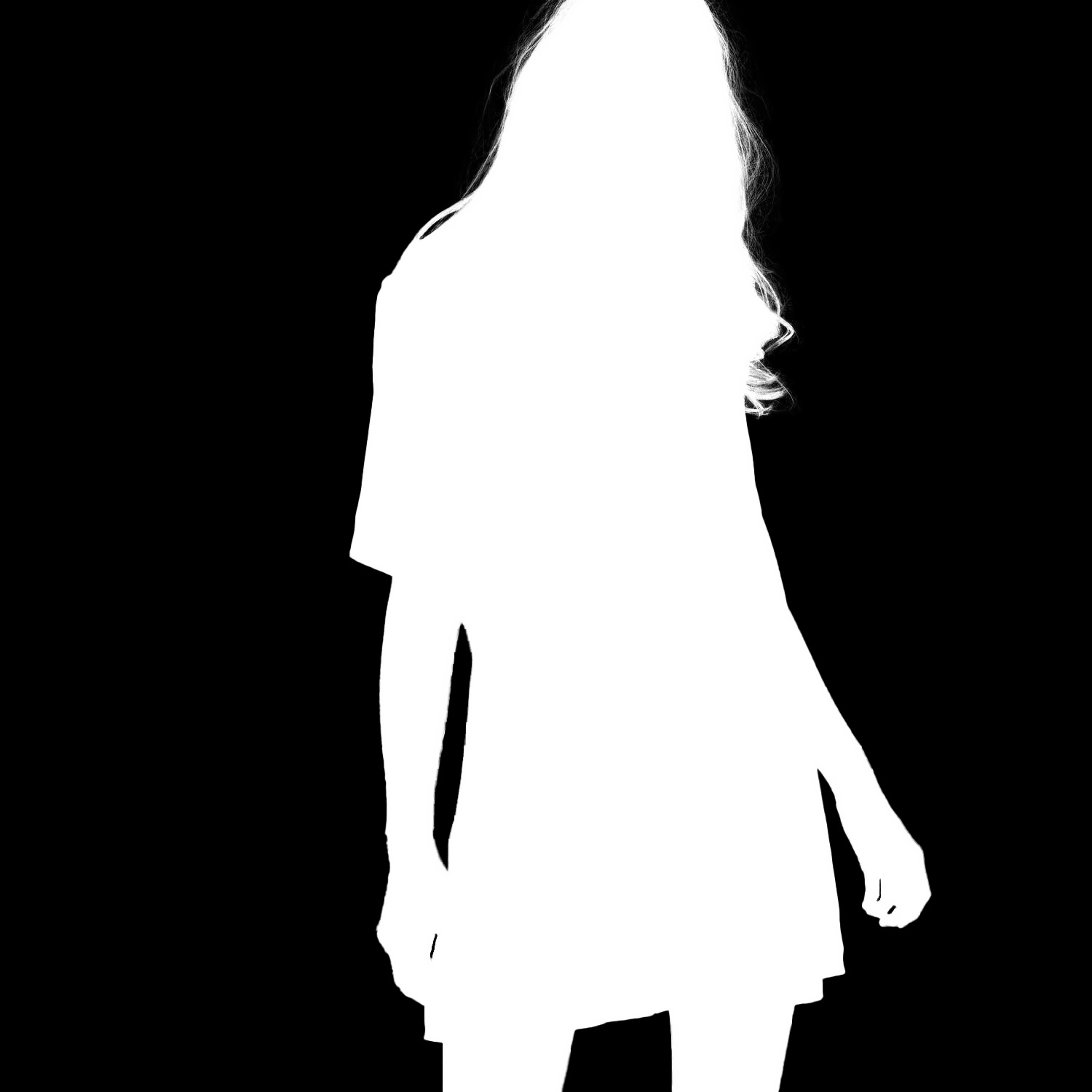}&\includegraphics[width=1.25cm]{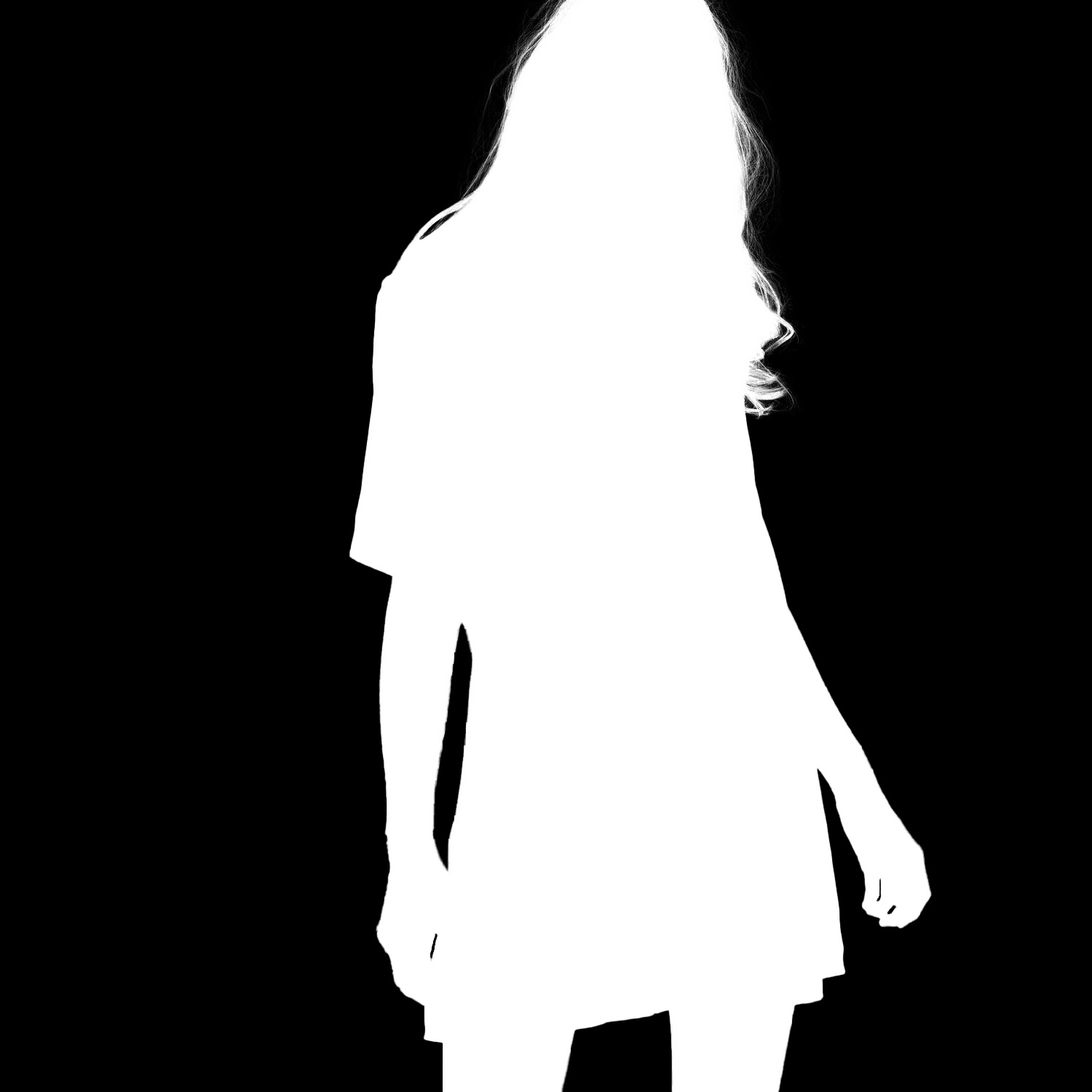}\\

\includegraphics[width=1.25cm]{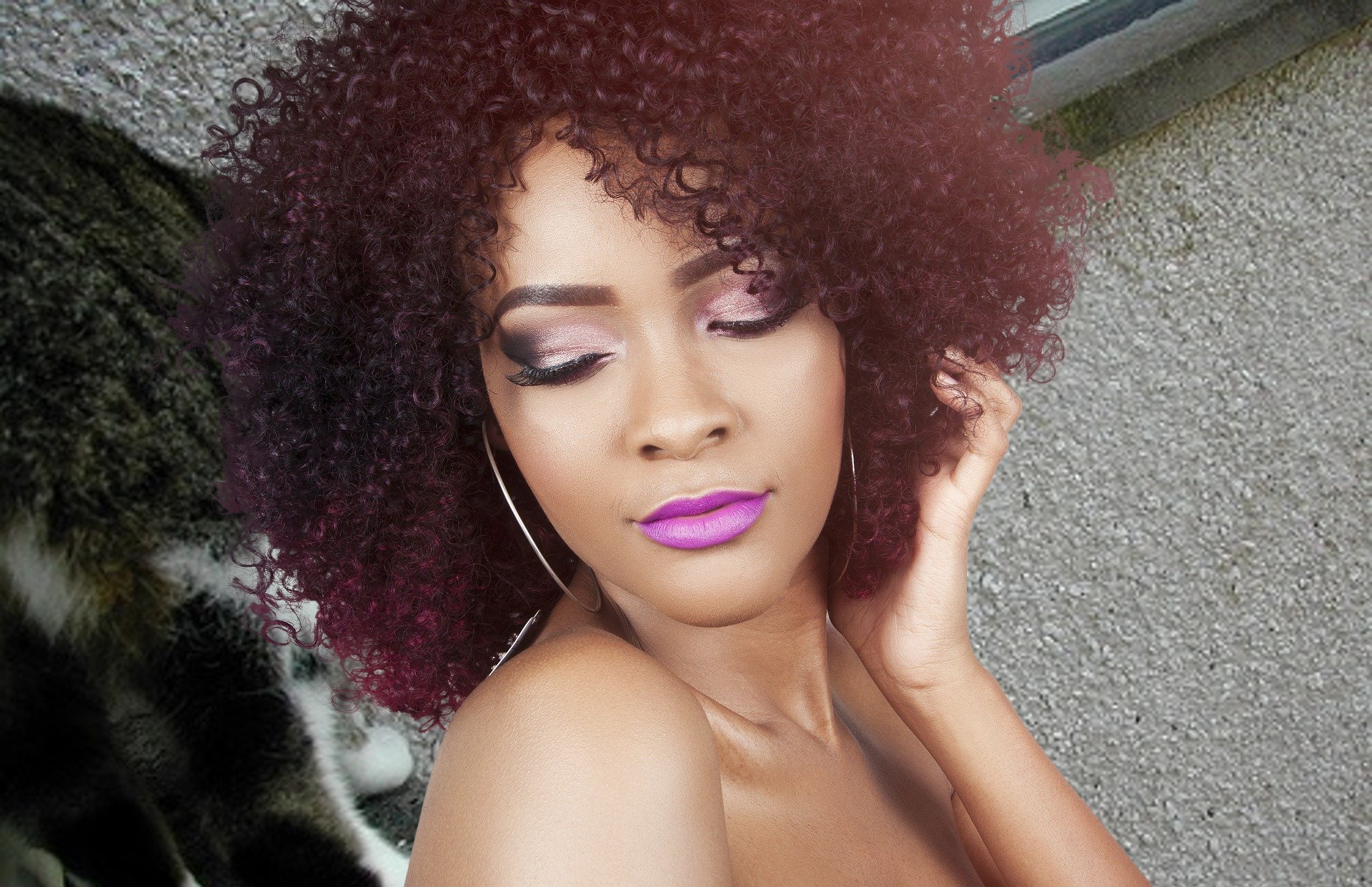}&\includegraphics[width=1.25cm]{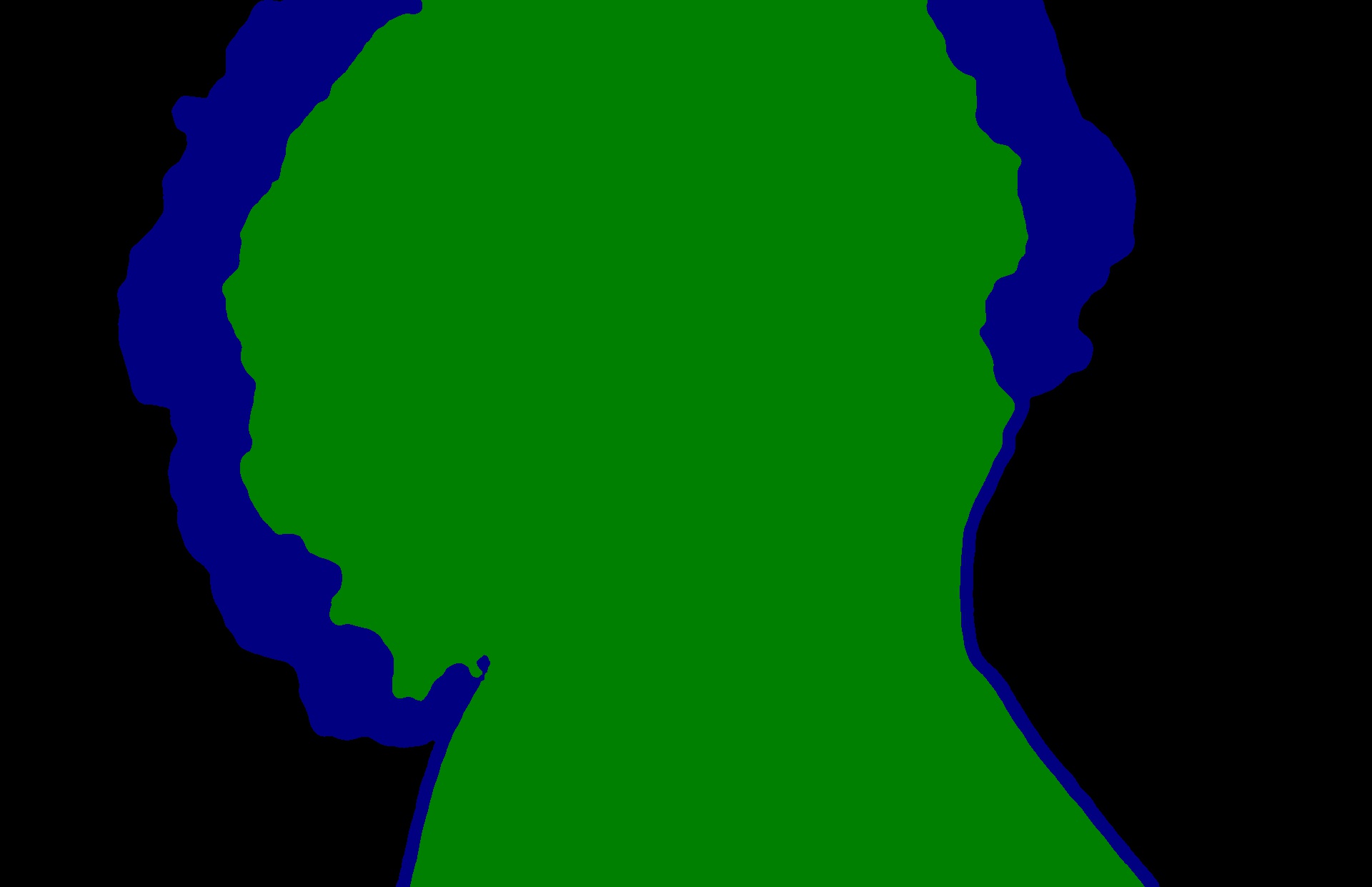}&\includegraphics[width=1.25cm]{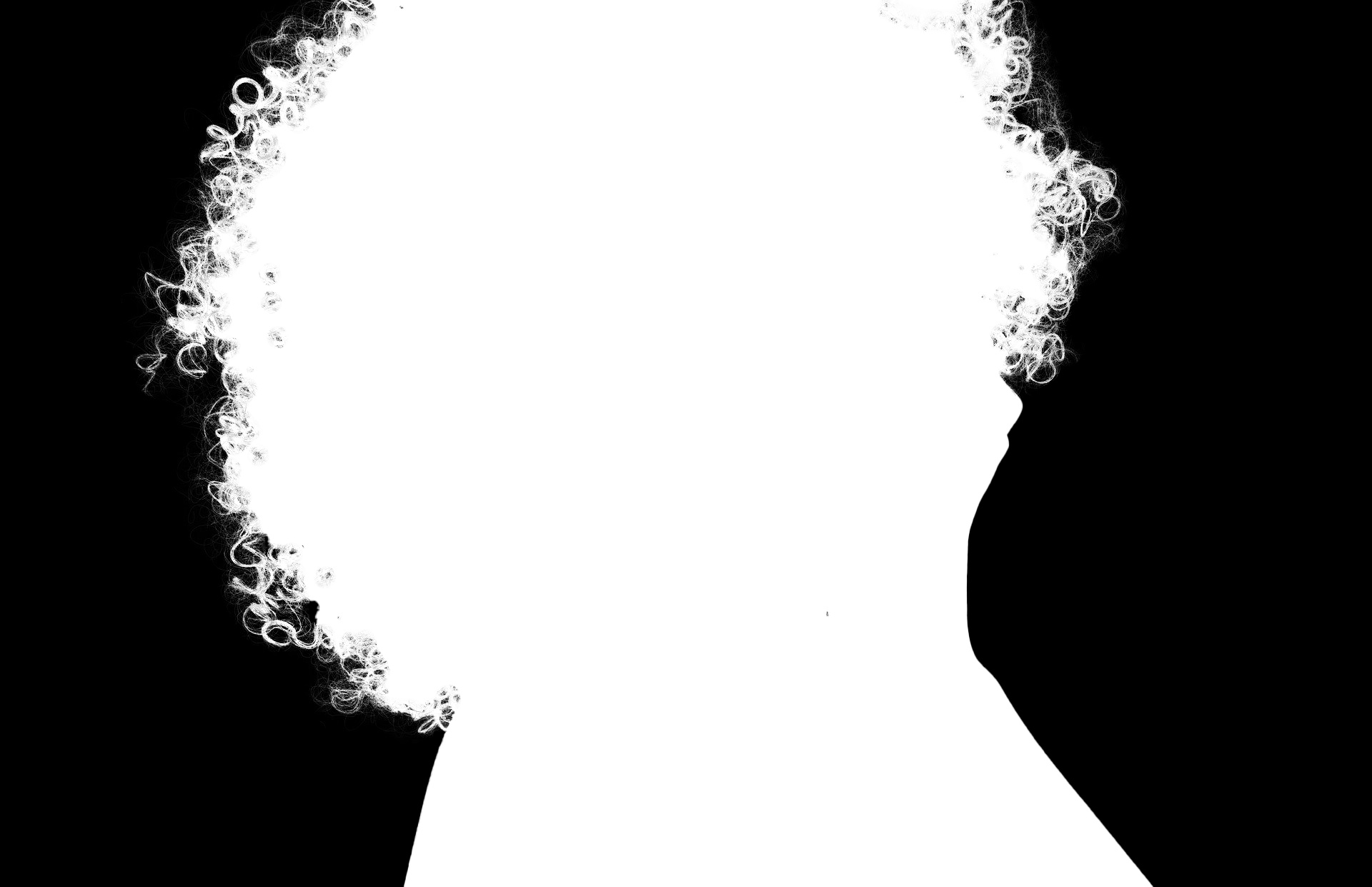}&\includegraphics[width=1.25cm]{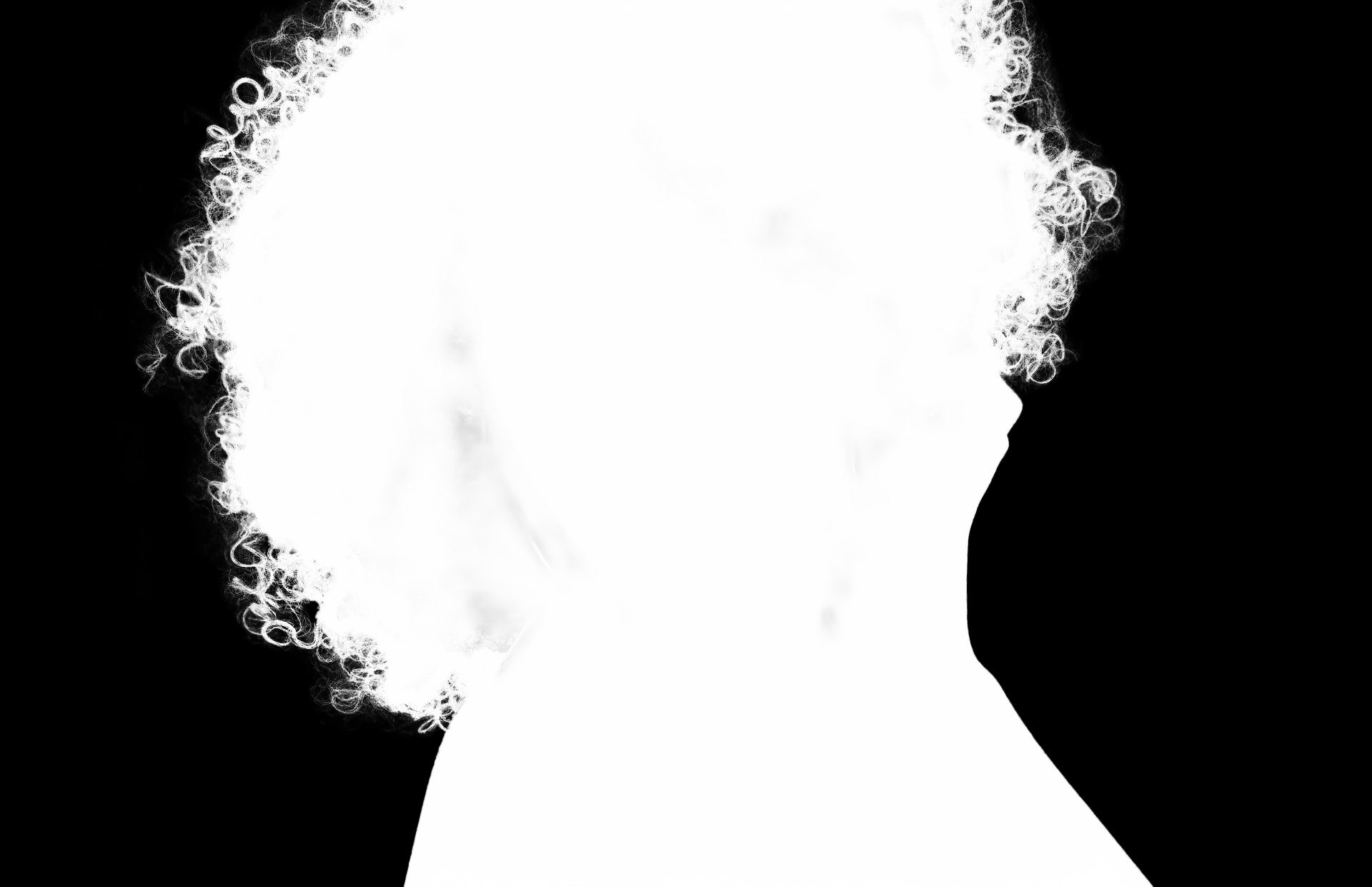}&\includegraphics[width=1.25cm]{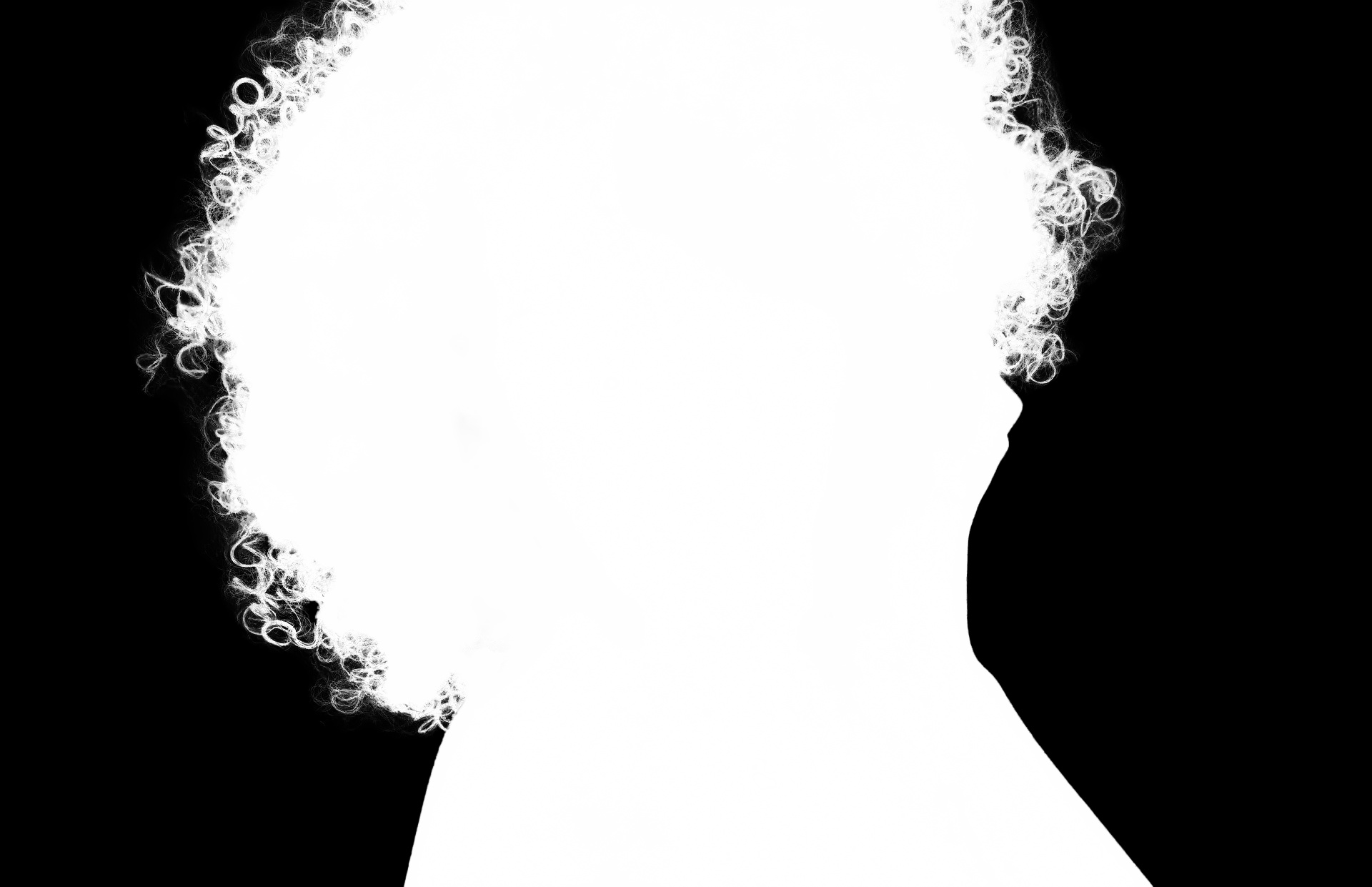}&\includegraphics[width=1.25cm]{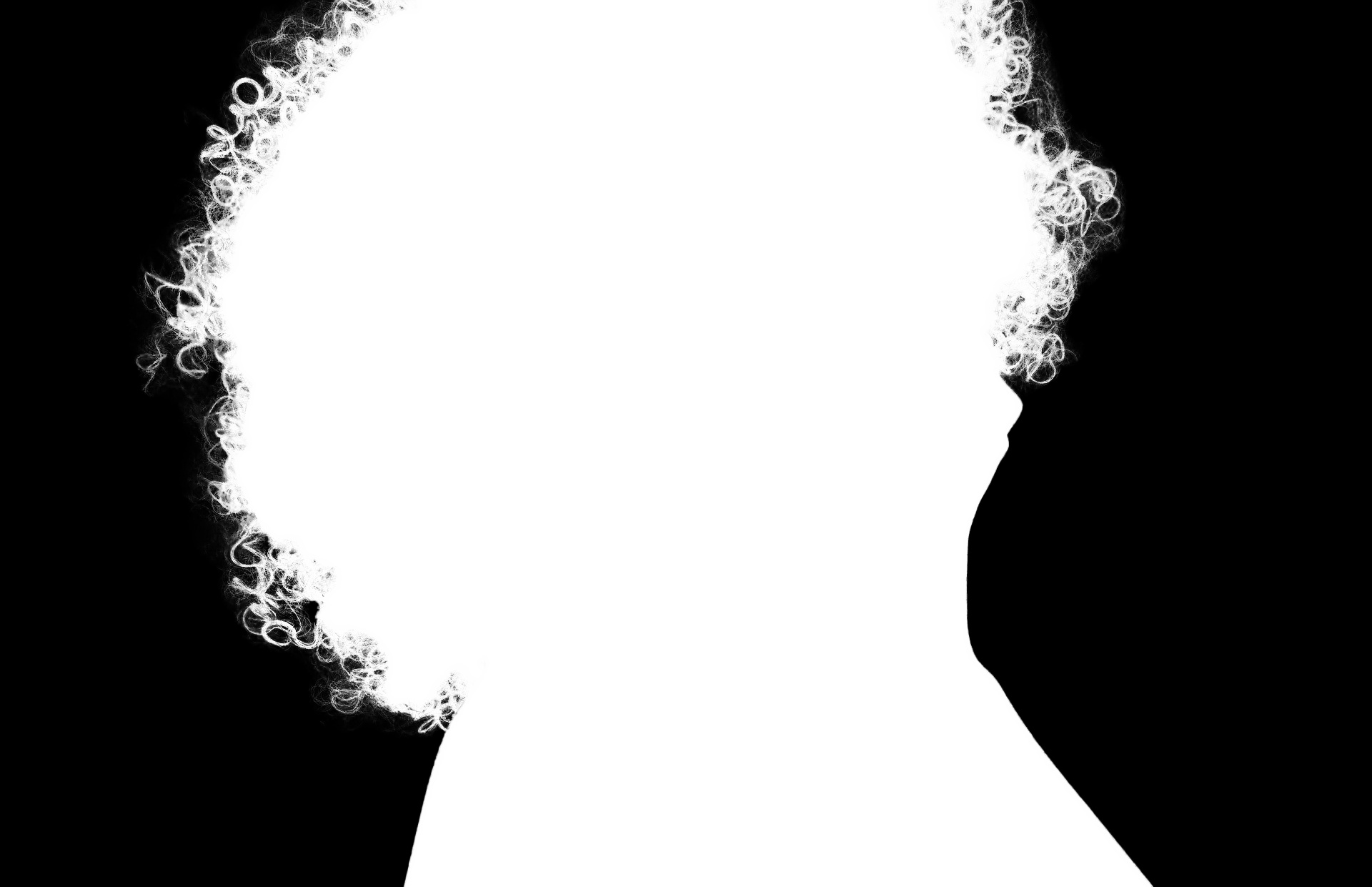}&\includegraphics[width=1.25cm]{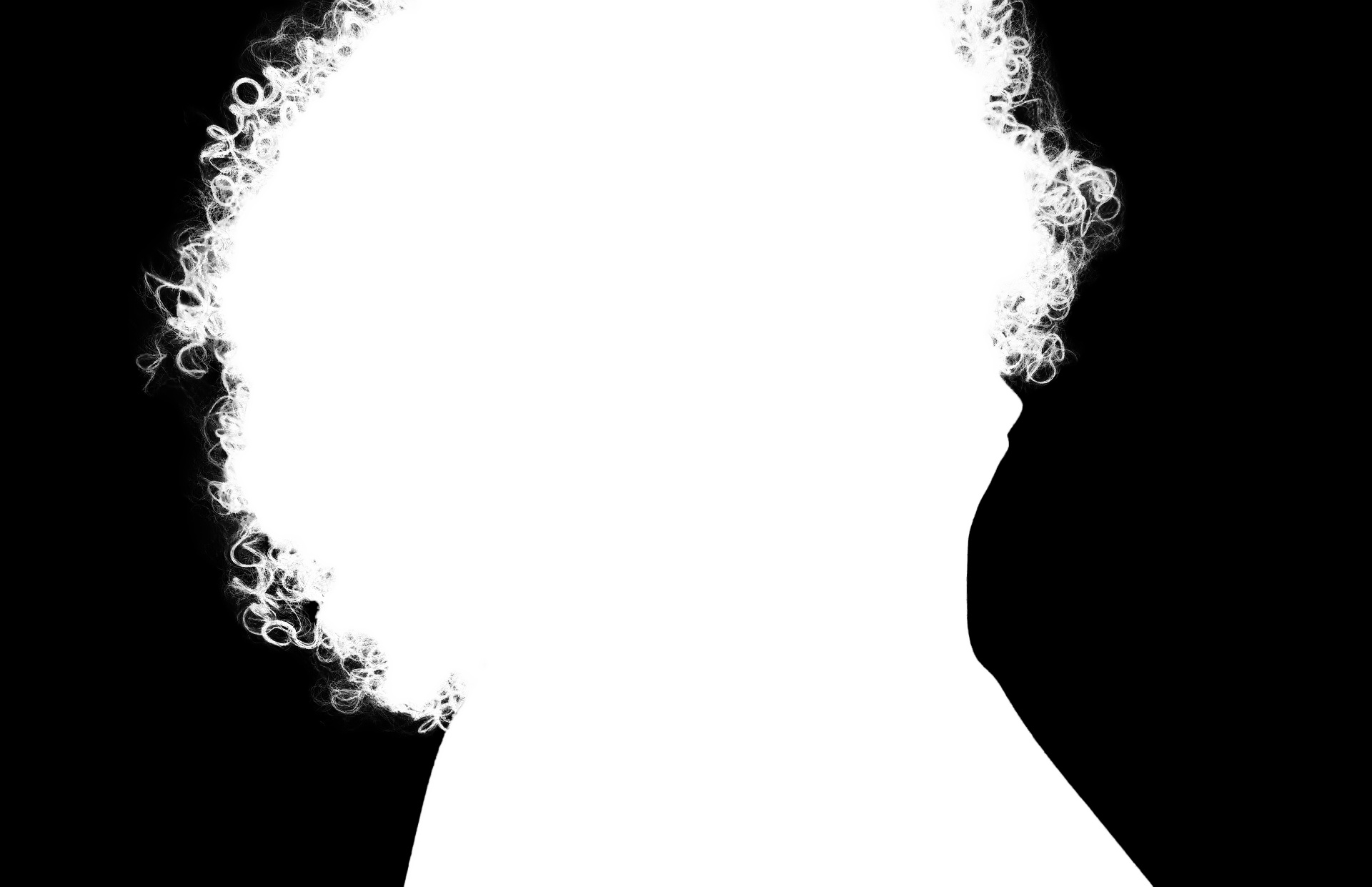}
%%%%%%%%%%%%%%%%%%%%%%%%%%%%
\end{tabular}}
\caption{Ablation comparison of visual results of Joint Inference with Fusion on Distinctions-646 benchmark. From left to right, image, trimap predicted by Net-T, GT, JIS-c, JIS, JIH-c, and JIH.}
\label{fig:joint} 
\end{center}
\end{figure}

\textbf{Adobe Image Matting Benchmark (AIM)}: We follow composition rules Xu et al. proposed~\cite{xu2017deep} to synthesize 43,100 training images and 1,000 testing images and compare our approach and its ablation study results with AlphaGAN~\cite{lutz2018alphagan}, Deep Image Matting~\cite{xu2017deep}, IndexNet Matting~\cite{lu2019indices}, AdaMatting~\cite{cai2019disentangled}, Learning Based Sampling~\cite{tang2019learning}, Context-Aware Matting~\cite{hou2019context}, GCA-Matting~\cite{li2020natural}, and HDMatt~\cite{yu2020high} in Table~\ref{tab:1ktest}. In perfect trimap provided situation, Net-M exhibits dominating performance on all four metrics compared with other methods. Also, from ablation study, Net-M possesses three out of four superior metrics except for Gradient error, compared to Net-M without hard mining loss (Net-M-nh) and Net-M without refinement (NLM), which demonstrates effectiveness of each part of our system. In trimap-free case, the comparison between our JI and approaches marked by $^\dagger$ reveals much more improvement of JI on MSE, Grad, and Conn metrics. In Fig.~\ref{fig:matting}, we represent qualitative comparison between our approaches and other state-of-the-art methods.\footnote{Please refer to supplementary material for visual samples of JI setting.} Given additional carefully-prepared trimap, our approach is capable of obtaining more fine-grained details of transparent objects than others do.

\textbf{Distinctions-646 Benchmark}: Qiao et al. establish Distinctions-646 dataset, consisting of 646 diversified foreground images~\cite{qiao2020attention}. Following the same composition rule as AIM, we synthesize 59,600 images for training and 1,000 images for testing. The Table~\ref{tab:distinction} compares our approach with HAttMatting~\cite{qiao2020attention} and Deep Image Matting (DIM). Our JI approach with either fusion technique exceeds HAttMatting and Deep Image Matting in SAD and Conn metrics by a solid margin. The Fig.~\ref{fig:joint} indicates that, after refinement module and the secondary fusion, much deterioration part of coarse matte is removed, which mainly benefits from our semantic-guided network architecture design. Considering a scenario where there is an image whose potential target foreground objects are non-salient or occluded with other equally-conspicuous objects in variegated backgrounds, common trimap-free matting~\cite{zhang2019late,qiao2020attention} that leverages a single RGB image as input is struggling for which object should be extracted. However, our JI strategy smartly decomposes automatic matting into two steps and circumvents this drawback.

\begin{table}[thpb]
\huge
  \begin{center}
    \subfloat[Net-T]{
  \resizebox{0.75\columnwidth}{!}{%
    % \label{tab:segcomp_nett]{%
     % \resizebox{\columnwidth}{!}{%
    % \resizebox{0.5\columnwidth}{!}{%
    \begin{tabular}{c|ccccc}
      \toprule % <-- Toprule here
      \multicolumn{6}{c}{\textbf{Composition-1k testset}}\\
    %   \midrule % <-- Midrule here
    %   \multirow{2}{*}{\textbf{Methods}} &\multicolumn{5}{c}{\textbf{Evaluation metrics}}\\
      \cmidrule{1-6}
      Methods & pixAcc & mIoU Bg  &mIoU Unk & mIoU Fg  & mIoU\\
      \midrule % <-- Midrule here
      Net-T-20&\textbf{96.41}&\textbf{93.53}&\textbf{82.78}&\textbf{70.64}&\textbf{82.32}\\
      Net-T-30&96.34&93.26&82.57&70.46&82.10\\
      Net-T-40&96.16&92.77&81.74&70.14&81.55\\
      Net-T-50&95.90&92.25&80.58&69.49&80.77\\
%       Net-T$^{\bigstar}$-20&0.9600&0.9275&0.8140&0.6956&0.8124\\
% Net-T$^{\bigstar}$-30&0.9585&0.9218&0.8067&0.6941&0.8075\\
% Net-T$^{\bigstar}$-40&0.9567&0.9169&0.7988&0.6913&0.8023\\
% Net-T$^{\bigstar}$-50&0.9547&0.9130&0.7897&0.6858&0.7962\\
      \bottomrule % <-- Bottomrule here
      \multicolumn{6}{c}{\textbf{Distinctions-646 testset}}\\
    %   \midrule % <-- Midrule here
    %   \multirow{2}{*}{\textbf{Methods}} & \multicolumn{5}{c}{\textbf{Evaluation metrics}}\\
      \cmidrule{1-6}
      Methods & pixAcc & mIoU Bg  & mIoU Unk & mIoU Fg  & mIoU\\
      \midrule % <-- Midrule here
      Net-T-20&\textbf{95.74}&\textbf{96.54}&\textbf{80.83}&66.72&\textbf{81.36}\\
      Net-T-30&95.66&96.20&80.22&66.82&81.08\\
      Net-T-40&95.57&96.00&79.60&\textbf{66.94}&80.84\\
      Net-T-50&95.31&95.68&78.34&66.72&80.25\\ %80.43
   \bottomrule % <-- Bottomrule here
   \end{tabular}}}

  % \subfloat[Net-T]{
  % \resizebox{\columnwidth}{!}{%
    % \label{tab:segcomp_nett]{%
     % \resizebox{\columnwidth}{!}{%
    % \resizebox{0.5\columnwidth}{!}{%
    % \begin{tabular}{c|c|c|c|c|c|c|c|c|c|c|c}
    %   \toprule % <-- Toprule here
    %   \multicolumn{6}{c}{\textbf{Composition-1k testset}}&\multicolumn{6}{c}{\textbf{Distinctions-646 testset}}\\
    %   \midrule % <-- Midrule here
    %   \multirow{2}{*}{\textbf{Methods}} & \multicolumn{5}{c}{\textbf{Evaluation metrics}}&\multirow{2}{*}{\textbf{Methods}} & \multicolumn{5}{c}{\textbf{Evaluation metrics}}\\
    %   \cmidrule{2-5}\\
      % & Accuracy & mean Bg IoU &mean Unknown IoU &mean Fg IoU & mIoU&& Accuracy & mean Bg IoU &mean Unknown IoU &mean Fg IoU & mIoU\\
      % \midrule % <-- Midrule here
      % Net-T-20&0.9641&0.9353&0.8278&0.7064&0.8232&Net-T-20&0.9574&0.9654&0.8083&0.6672&0.8136\\
      % Net-T-30&0.9634&0.9326&0.8257&0.7046&0.8210&Net-T-30&0.9566&0.9620&0.8022&0.6682&0.8108\\
      % Net-T-40&0.9616&0.9277&0.8174&0.7014&0.8155&Net-T-40&0.9557&0.9600&0.7960&0.6694&0.8084\\
      % Net-T-50&0.9590&0.9225&0.8058&0.6949&0.8077&Net-T-50&0.9531&0.9568&0.7834&0.6672&0.8025\\
%       Net-T$^{\bigstar}$-20&0.9600&0.9275&0.8140&0.6956&0.8124\\
% Net-T$^{\bigstar}$-30&0.9585&0.9218&0.8067&0.6941&0.8075\\
% Net-T$^{\bigstar}$-40&0.9567&0.9169&0.7988&0.6913&0.8023\\
% Net-T$^{\bigstar}$-50&0.9547&0.9130&0.7897&0.6858&0.7962\\
      % \bottomrule % <-- Bottomrule here
      
      % \midrule % <-- Midrule here
      
    %   \cmidrule{2-5}\\
      
      % \midrule % <-- Midrule here
   % \bottomrule % <-- Bottomrule here
    % \end{tabular}}}
    % \qquad
    
    \subfloat[JI]{%
    \resizebox{\columnwidth}{!}{%
    \begin{tabular}{c|cccccc|cccc}
    % \huge
      %\toprule % <-- Toprule here
      \cmidrule[\heavyrulewidth]{1-5}\cmidrule[\heavyrulewidth]{7-11}
      \multicolumn{5}{c}{\textbf{Composition-1k testset}}&&\multicolumn{5}{c}{\textbf{Distinctions-646 testset}} \\
      % \midrule % <-- Midrule here
     %\cmidrule{1-5}\cmidrule{7-11}
     % \multirow{1}{*}
    %   {\textbf{Methods}} &\multicolumn{4}{c}{\textbf{Evaluation metrics}}&&\multirow{2}{*}{\textbf{Methods}} & \multicolumn{4}{c}{\textbf{Evaluation metrics}}\\
      \cmidrule{1-5}\cmidrule{7-11}
      Methods & SAD$\downarrow$ & MSE$\downarrow$ &Grad$\downarrow$ & Conn$\downarrow$& & Methods & SAD$\downarrow$ & MSE$\downarrow$ &Grad$\downarrow$ & Conn$\downarrow$ \\
      \cmidrule{1-5}\cmidrule{7-11}
    %   \midrule % <-- Midrule here
      Late Fusion$^{\dagger}$&58.34&0.011&41.63&59.74&&DIM$^{\dagger}$&47.56&0.009&43.29&55.90\\
      HAttMatting$^{\dagger}$&44.01&0.007&29.26&46.41&&HAttMatting$^{\dagger}$&48.98&0.009&41.57&49.93\\
     % \midrule % <-- Midrule here
     \cmidrule{1-5}\cmidrule{7-11}
      JIS$^{\dagger}$-20&43.91&\textbf{0.0054}&18.64&39.82&& JIS$^{\dagger}$-20&41.91&0.0080&38.38&39.46\\
      JIS$^{\dagger}$-30&44.24&0.0055&18.75&40.10&& JIS$^{\dagger}$-30&41.96&0.0079&38.38&39.46\\
      JIS$^{\dagger}$-40&44.64&0.0056&18.92&40.43&& JIS$^{\dagger}$-40&42.40&0.0079&39.52&39.79\\
      JIS$^{\dagger}$-50&45.55&0.0059&19.47&41.24&& JIS$^{\dagger}$-50&43.92&0.0083&40.86&41.27\\
      JIH$^{\dagger}$-20&\textbf{41.85}&0.0055&\textbf{18.34}&\textbf{39.42}&&JIH$^{\dagger}$-20&38.52&0.0080&38.23&38.28\\
      JIH$^{\dagger}$-30&42.08&0.0056&18.42&39.66& &JIH$^{\dagger}$-30&\textbf{38.39}&\textbf{0.0078}&\textbf{37.77}&\textbf{38.17}\\
      JIH$^{\dagger}$-40&42.29&0.0057&18.53&39.87& &JIH$^{\dagger}$-40&38.58&0.0079&38.66&38.37\\
      JIH$^{\dagger}$-50&43.00&0.0060&19.05&40.54& &JIH$^{\dagger}$-50&39.88&0.0083&40.27&39.74\\
      \cmidrule[\heavyrulewidth]{1-5}\cmidrule[\heavyrulewidth]{7-11}
     %\cmidrule{1-5}\cmidrule{7-11}

    %\bottomrule % <-- Bottomrule here
    \end{tabular}}}
    \caption{Comparison of different segmentation inputs for Net-T and JI on different datasets. ($<$network$>$-X, like Net-T-X, JIS-X and JIH-X, means that segmentation is generated by erosion on \textit{initial segmentation} from \textit{synthesized ground-truth trimap} with X pixels and followed by a Gaussian Blur.)}
    \label{tab:segcomp}
  \end{center}
\end{table}

\begin{figure}[thpb]
\setlength\tabcolsep{0pt}
\renewcommand{\arraystretch}{0.25}
\begin{center}
\resizebox{0.9\columnwidth}{!}{
\begin{tabular}{ccc}
\includegraphics[height=2cm]{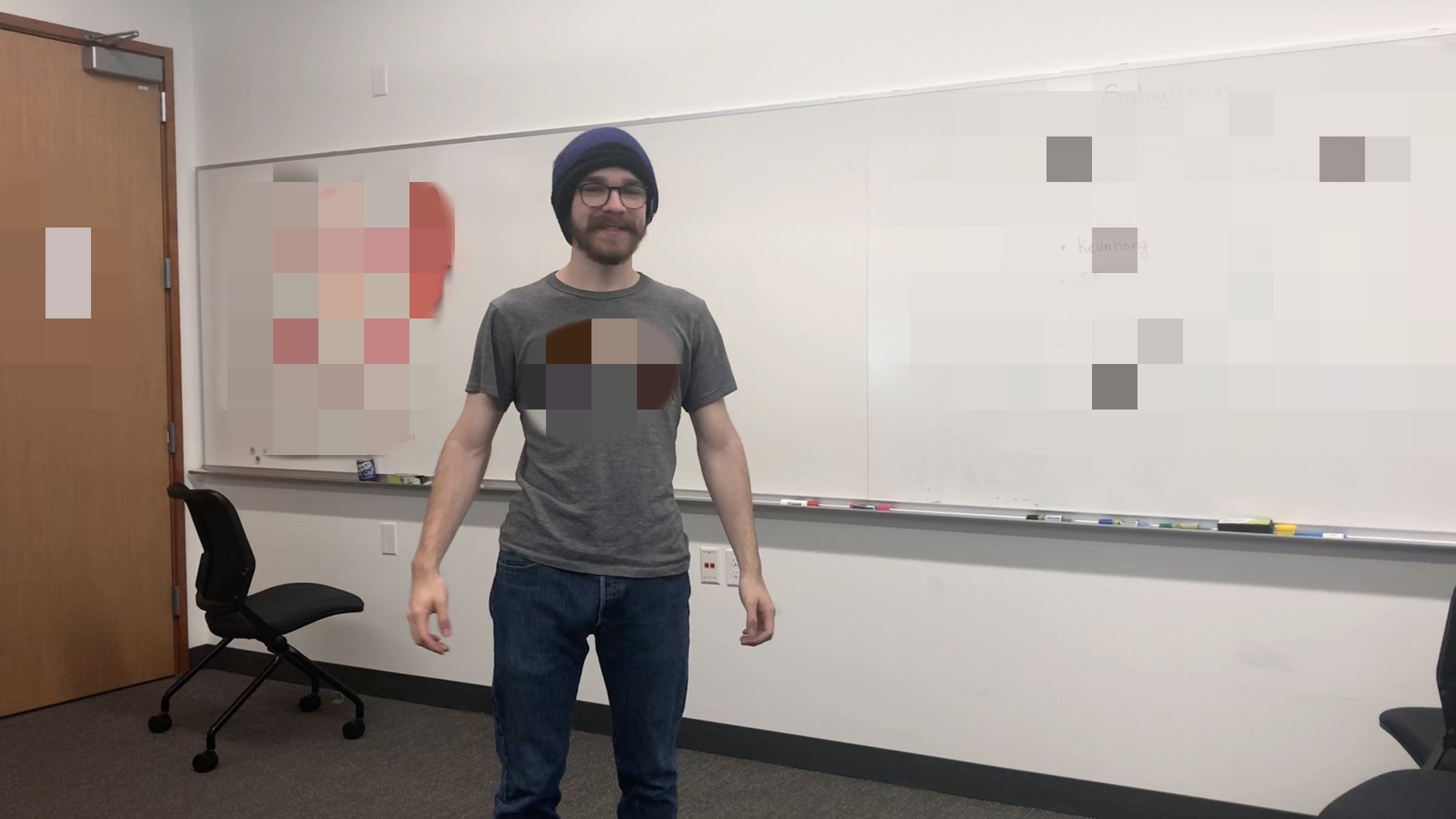}&\includegraphics[height=2cm]{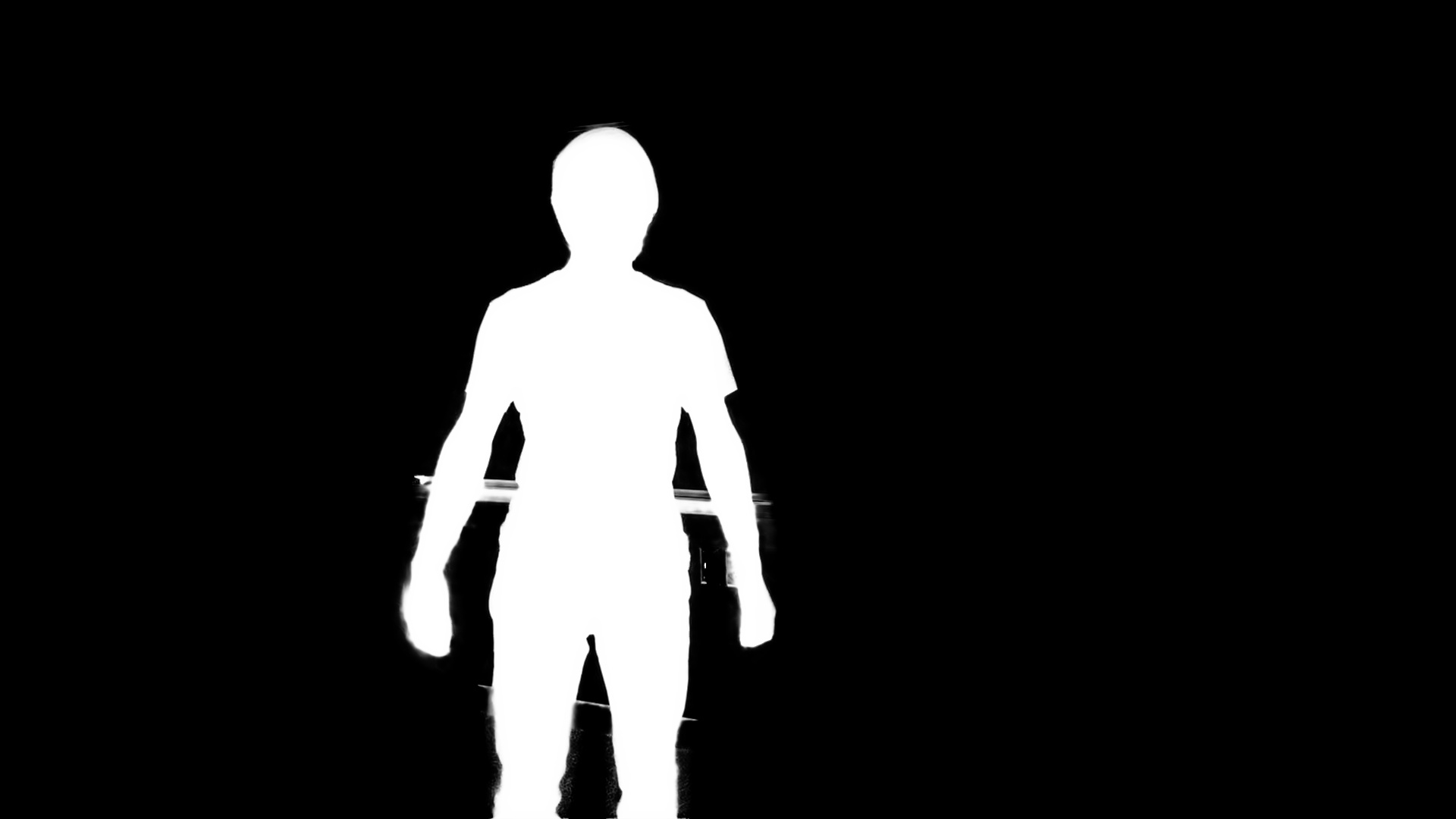}&\includegraphics[height=2cm]{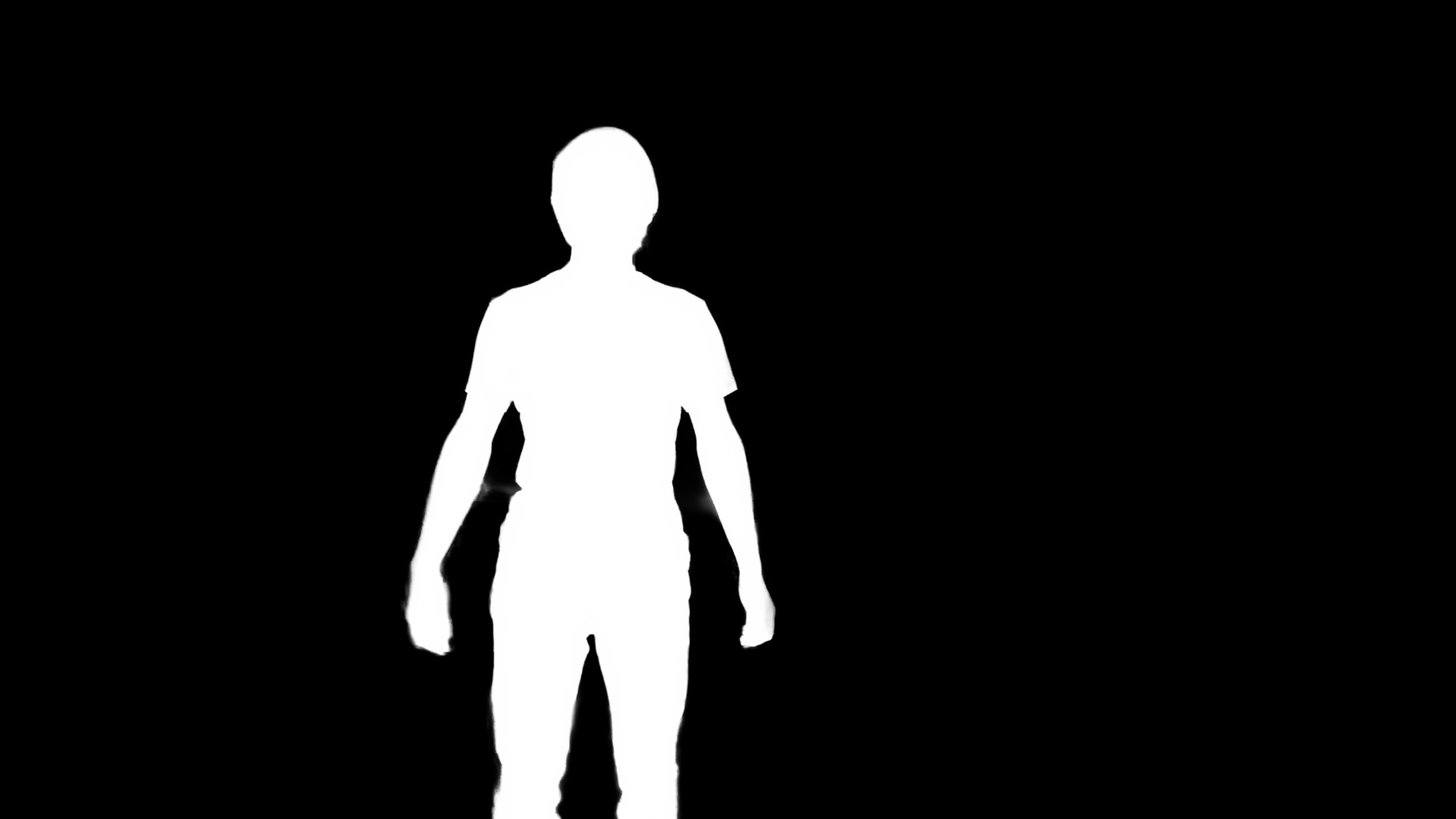}\\
\includegraphics[height=2cm]{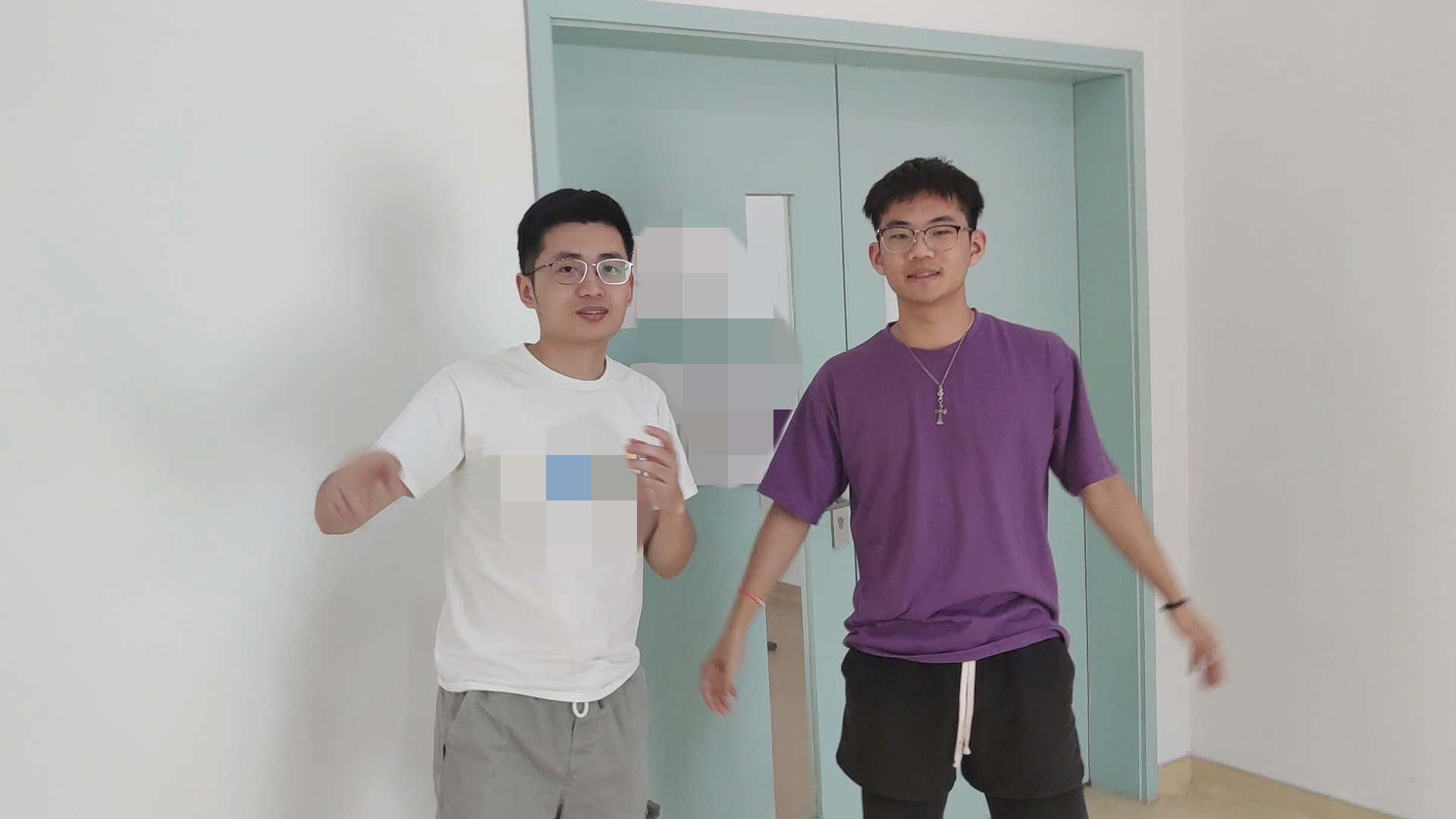}&\includegraphics[height=2cm]{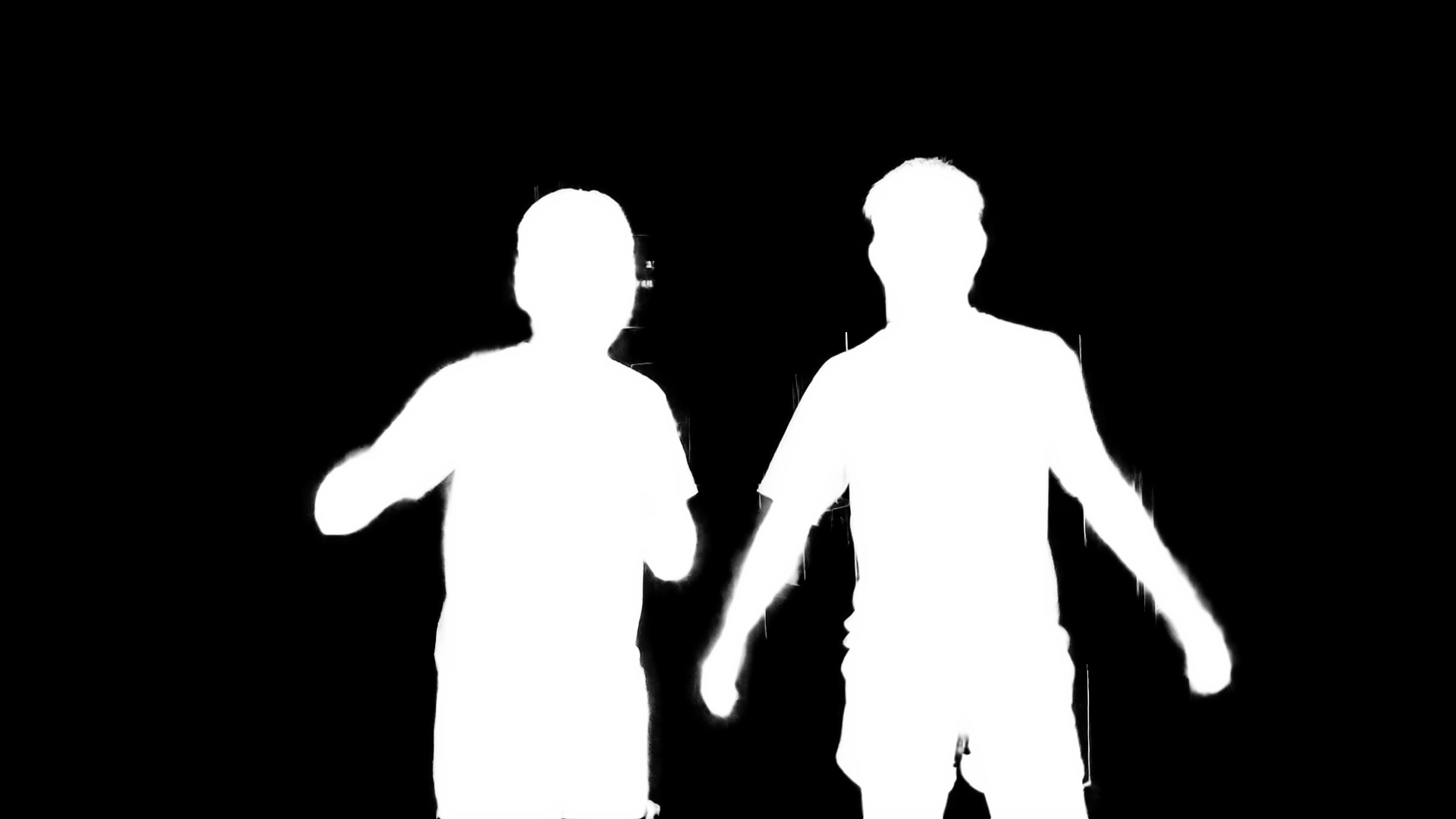}&\includegraphics[height=2cm]{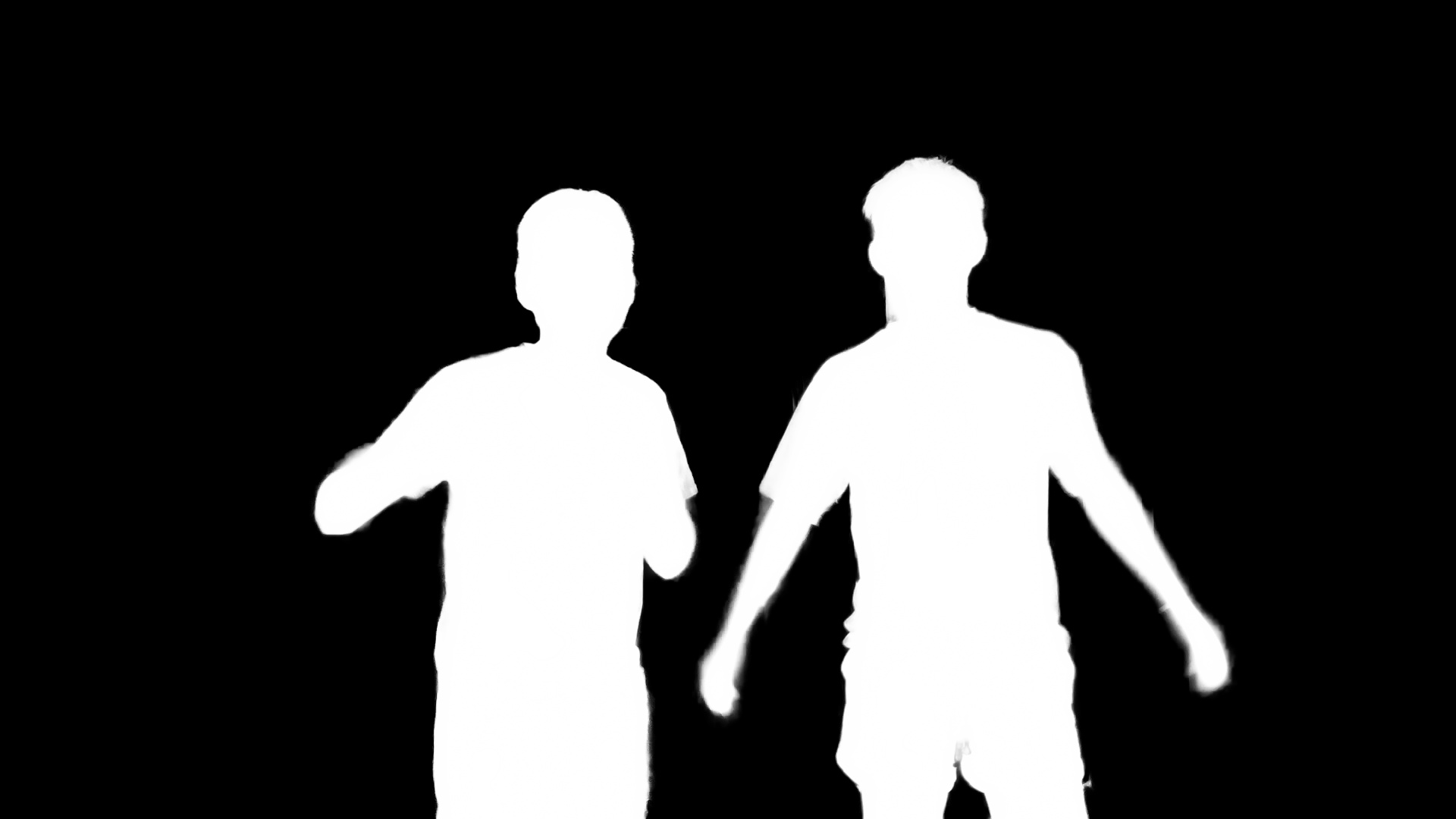}\\
% \includegraphics[width=2.5cm]{./images/real_data/images/VID_20201109_141618_0459_img_blurred.jpg}&\includegraphics[width=2.5cm]{./images/real_data/cam1/VID_20201109_141618_0459.jpg}&\includegraphics[width=2.5cm]{./images/real_data/ours/VID_20201109_141618_0459.jpg}\\
% Image&CAM-1&Net-T-M-Real\\
\end{tabular}}
\caption{Comparison of visual results of real data between CAM and Ours. From left to right, image, CAM and Ours.} 
\label{fig:real_data} 
\end{center}
\end{figure}

\label{realdatasec}
\textbf{Real World Human Data}: To demonstrate trimap-free application capability of our approach in real world, we adopt similar adversarial training for real data adaption as Sengupta et al.~\cite{sengupta2020background}, in which soft segmentation is generated by Parsing R-CNN~\cite{yang2019parsing}.\footnote{Please refer to supplementary materials for training detail.} We capture 37 handheld videos, in which subject is moving around and camera is moved randomly, and combine them with 10 real-world videos from Background Matting~\cite{sengupta2020background} for training (totally 22,144 real-world video frames). The testset consists of 100 images generated from 10 uniformly-sampled frames of each testing video (totally 10 testing videos, i.e. half self-captured, half Background Matting videos). In user study, we composite matte on black background and compare our approach with Context-Aware Matting. Each user was presented with one web page, showing 100 pairs of original image, composite of ours and CAM with random order of the last two. Participants are asked to rate the left composite relative to the right on three scales, i.e. better, similar, and worse. We survey 10 users and our method achieves 65.8\% better, 28.7\% similar but only 5.5\% worse than CAM, which implies competitive performance of our trimap-free pipeline compared to modest trimap-needed matting on real-world human images with diverse backgrounds. Fig.~\ref{fig:real_data} shows visual comparsion. This experiment validates that our approach can be trained to adapt to different domains of soft segmentation, therefore, we are confident about practial value of our pipeline. %-1\footnote{CAM with ME+CE+lap} 
% For testing, we combine other 5 self-captured videos and 5 Background Matting videos, and then testset consists of 100 images formed by 10 uniformly-sampled frames from each video.

% We retrain AIM-pretrained matting pipeline (JI-Real) end-to-end using soft fusion on 22,144 real-world video frames. Net-T of JI-Real is supervised by pseudo trimap generated by segmentation~\cite{yang2019parsing} and erosion/dilation, while human-finetuned Net-M serves as a teacher for Net-M of JI-Real.
% End-to-end training and interaction between generator (JI-Real) and discriminator help JI-Real jump out of the local minimum of human-finetuned Net-M.\footnote{Please refer to supplementary materials for training details.}

\subsubsection{Attention block and visualization}
The Table~\ref{tab:parameters} shows quantitative comparison of the number of parameters and GFLOPs of attention layer(s). Our attention layer is much lighter than that of GCA-Matting in both parameter number and GFLOPs aspects, which shows superiority of our attention. In Fig.~\ref{fig:att_vis}, we visualize global attention weight map of given image query patch denoted by red box and reconstructed alpha feature $A^{\prime}$. The brighter the color is, the larger attention weight (pixel value) the pixel holds. The weights of known and unknown part are shown in the top-left corner of each attention map. It is obvious that our attention module can not only select color-relevant pixels accurately but also capture long-distance pixel-to-pixel relationships. The reorganized alpha feature shown on the right of each row demonstrates that the encoder has already concentrated on unknown areas exquisitely which can promote better feature reconstruction in the decoder.
\subsubsection{Ablation study on coarse segmentation input}
In this section, we present ablation study about coarse foreground segmentation input on Net-T and JI.

\begin{table}[thpb]
\huge
  \begin{center}
    \resizebox{0.7\columnwidth}{!}{%
    \begin{tabular}{ccc}
      \toprule % <-- Toprule here
      Methods & \ Params & GFLOPs\\
      \midrule % <-- Midrule here
      % \cmidrule{3-4}
      GCA-Matting & 49,792&3.5621\\
      Non-local Matting(Ours) & \textbf{25,984} &  \textbf{0.1416}\\
      \bottomrule % <-- Bottomrule here
    \end{tabular}}
    \caption{Comparsion of the number of parameters and GFLOPs of attention block between our Non-local Matting and GCA-Matting~\cite{li2020natural}}
    \label{tab:parameters}
  \end{center}
\end{table}
% Ablation comparison of visual results of Joint I
\textbf{Net-T}: We conduct an ablation study to investigate the influence of quality of coarse segmentation on Net-T. We test Net-T with \textit{initial segmentation} eroded by 20, 30, 40, and 50 pixels, which simulates flawed applied circumstances, on Composition-1k and Distinctions-646 testsets as shown in Table~\ref{tab:segcomp} (a). The accuracy is all above 90\% and the mean IoU of unknown region is fluctuating around 0.80, which implies that Net-T is anti-jamming for unperfect foreground segmentation and capable of offering correct trimap estimation.

\textbf{JI}: To research how segmentation affects trimap-free JI pipeline, we test JIS and JIH with \textit{initial segmentation} eroded by 20, 30, 40, and 50 pixels on Composition-1k and Distinctions-646 testset. The results are illustrated in Table~\ref{tab:segcomp} (b). The deviation of each metric is quite slight and most metrics in their worst performance still set a state-of-the-art record, which reveals that our network can still attain robust property when soft segmentation is in an ill-posed state.%\footnote{Please refer to supplementary material for more detailed ablation study and visual samples.}

\section{Conclusion}
In this paper, we propose a novel two-stage trimap-free image matting network, which can predict accurate alpha matte from RGB image and its coarse foreground segmentation. Our matting pipeline can be easily integrated with other state-of-the-art semantic segmentation/salient object detection/matting methods to boost trimap-free matting in real world. Benefiting from semantic information provided by coarse foreground segmentation, our approach employs Trimap Generation Network to capture target objects roughly. The light-weight Non-local Matting Network with Refinement dedicates to transition region under predicted-trimap guidance. Comprehensive experiments indicate that our approach is comparable with state-of-the-art performance on Composition-1k, alphamatting.com, Distinctions-646 benchmarks, and real imagery in trimap-needed and trimap-free cases.
% We believe that this matting architecture should be more rational trimap-free matting framework, which has comprehensive integration ability with other semantic semgentation/salient object detection/matting approaches to solve automatic natural image matting.
% In the future, we plan to explore more powerful automantic matting approaches in order to promote more downstream applications of matting task, e.g. photo editing tools and group photo synthesis.
% group photo composition optimization. Experimental results show the  are generated with the proposed human composition rules.
% {\small
% \bibliographystyle{ieee_fullname}
% \bibliography{Ref}
% }

\section{Acknowledgments}
This paper was supported by funding 2019-INT007 from the Shenzhen Institute of Artificial Intelligence and Robotics for Society. Thank Zhixiang Wang, Junjie Hu, and Kangfu Mei for comments.

\bibliography{aaai22}

\newpage
\setcounter{section}{1}
\renewcommand{\thesection}{\Alph{section}}
\setcounter{table}{0}
\renewcommand{\thetable}{S\arabic{table}}%
\setcounter{figure}{0}
\renewcommand{\thefigure}{S\arabic{figure}}%

% \beginsupplement
\section{Appendices}

\subsection{Introduction}
In this supplementary material, we provide our network details, details of soft segmentation generation, and additional matting details and results.
\subsubsection{Network Structure Details}
The Fig. \ref{fig:nettstructure}, \ref{fig:nonlocalmattingstructure}, \ref{fig:ResBlock}, \ref{fig:ResBlockDown}, \ref{fig:ResBlockUp}, and \ref{fig:refinementstructure} show our networks in details.
\subsubsection{Soft Segmentation Generation}
\begin{figure}[thpb]
\centering
\includegraphics[scale=0.25]{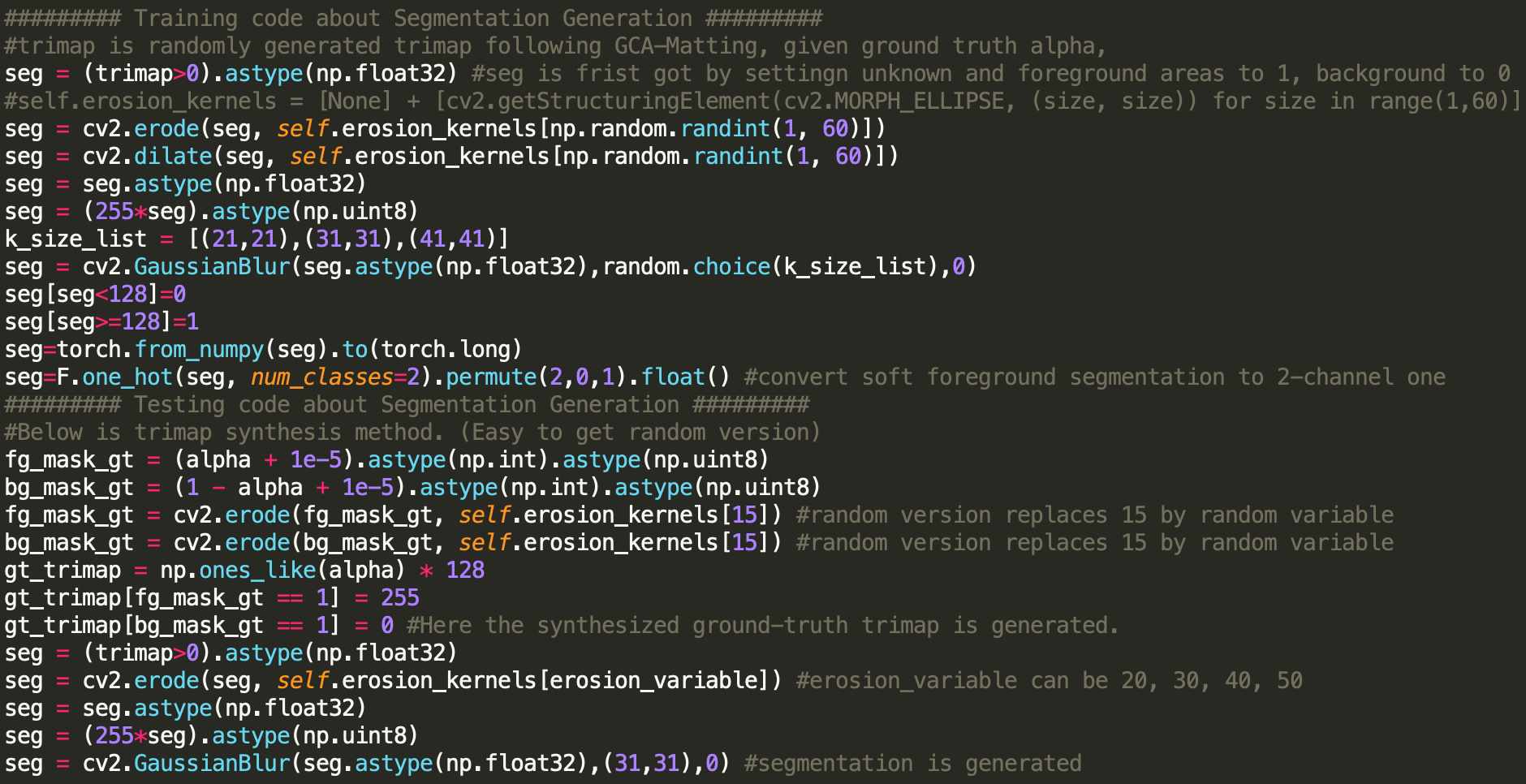}
\caption{Code Snapshot of Soft Foreground Segmentation Generation for Synthetic Datasets}
\label{code}
\end{figure}

Our soft foreground segmentation generation of synthetic datasets is similar to that in Background Matting~\cite{sengupta2020background}, which can be found by searching ``create\_seg'' in their github code. Fig.~\ref{code} shows code of our soft segmentation process. Besides, the code of our research will be publically available once the paper is accepted.
\subsubsection{Matting}
In this section, we provide matting results and training details.

\begin{figure}[thpb]
\setlength\tabcolsep{0pt}
\Large
\renewcommand{\arraystretch}{0.25}
\begin{center}
\resizebox{\columnwidth}{!}{
\begin{tabular}{ccccccc}
\includegraphics[width=1.75cm]{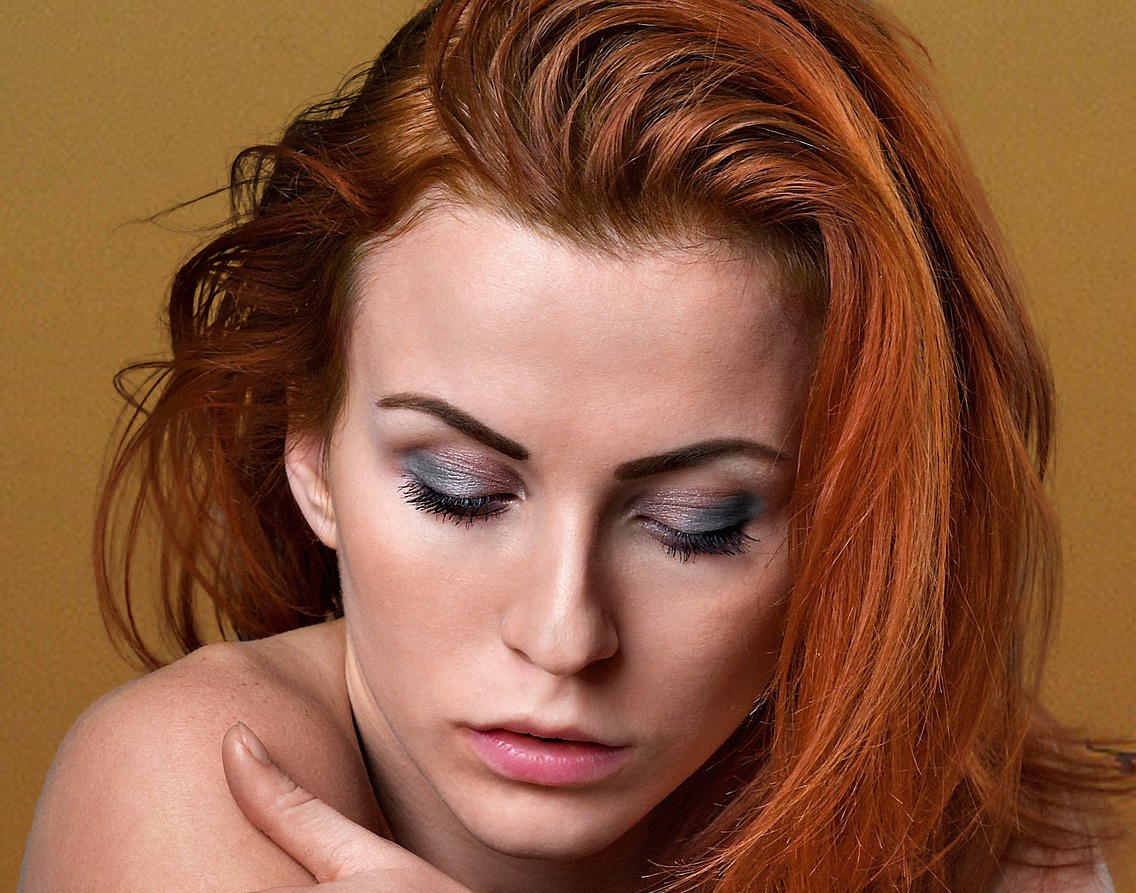}&\includegraphics[width=1.75cm]{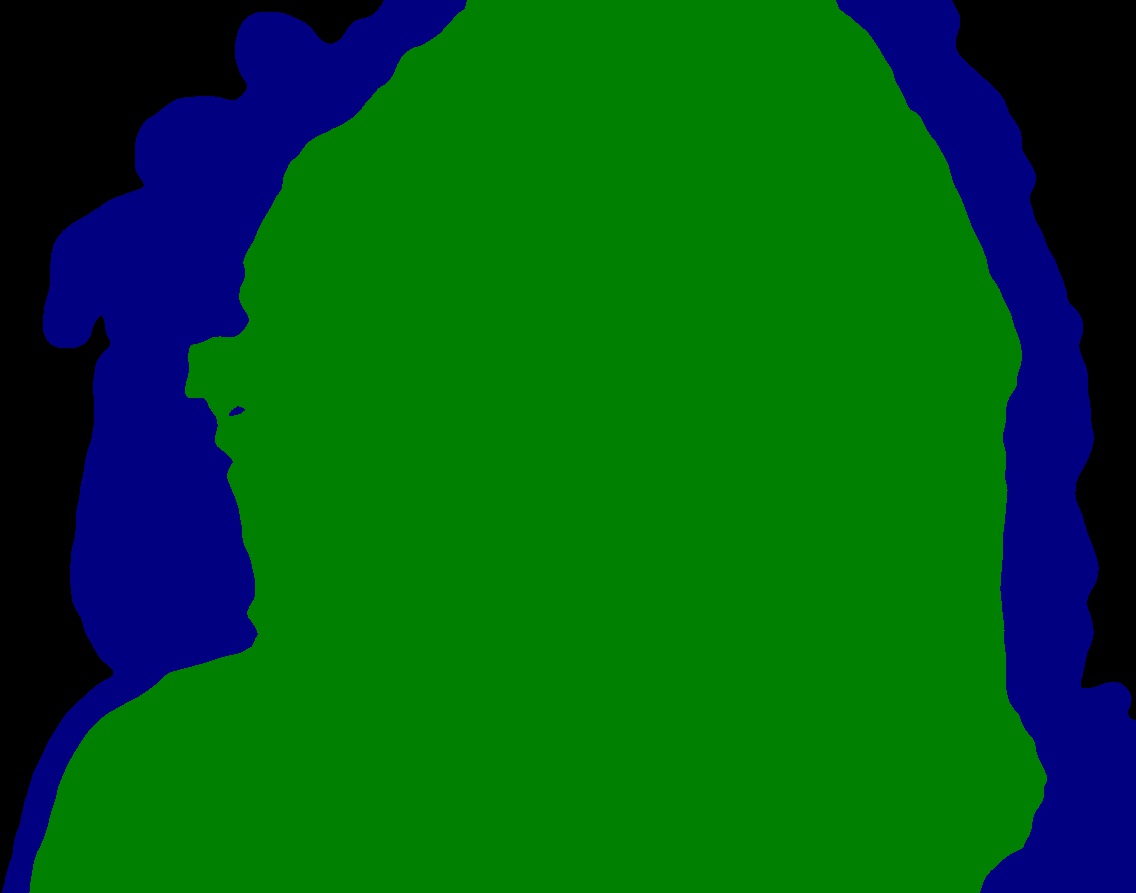}&\includegraphics[width=1.75cm]{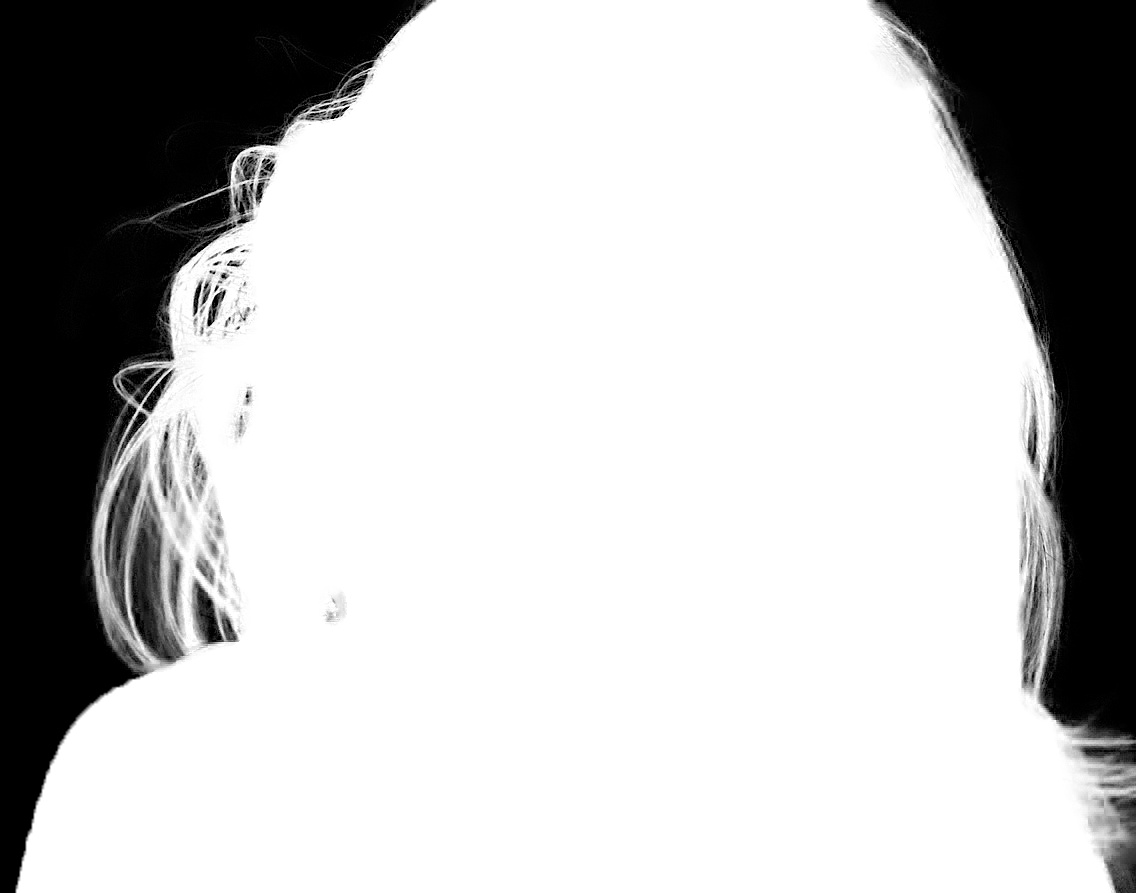}&\includegraphics[width=1.75cm]{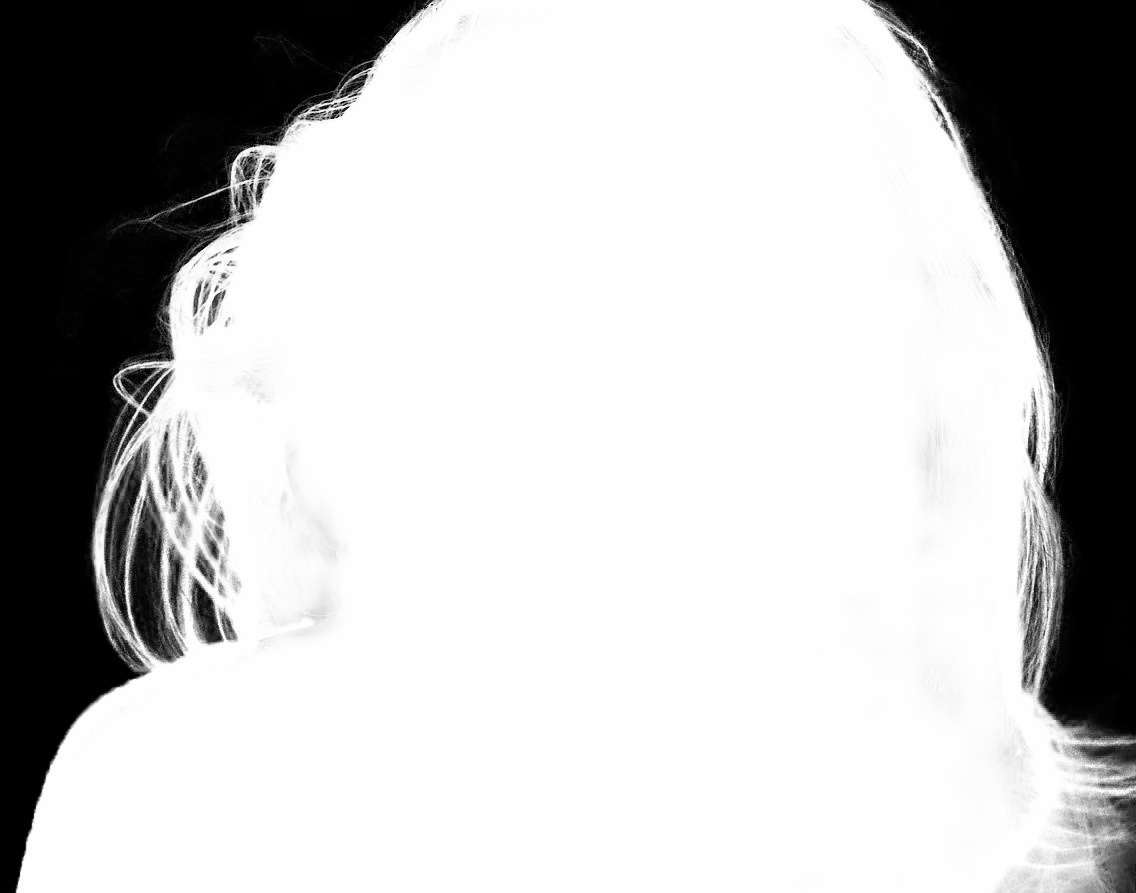}&\includegraphics[width=1.75cm]{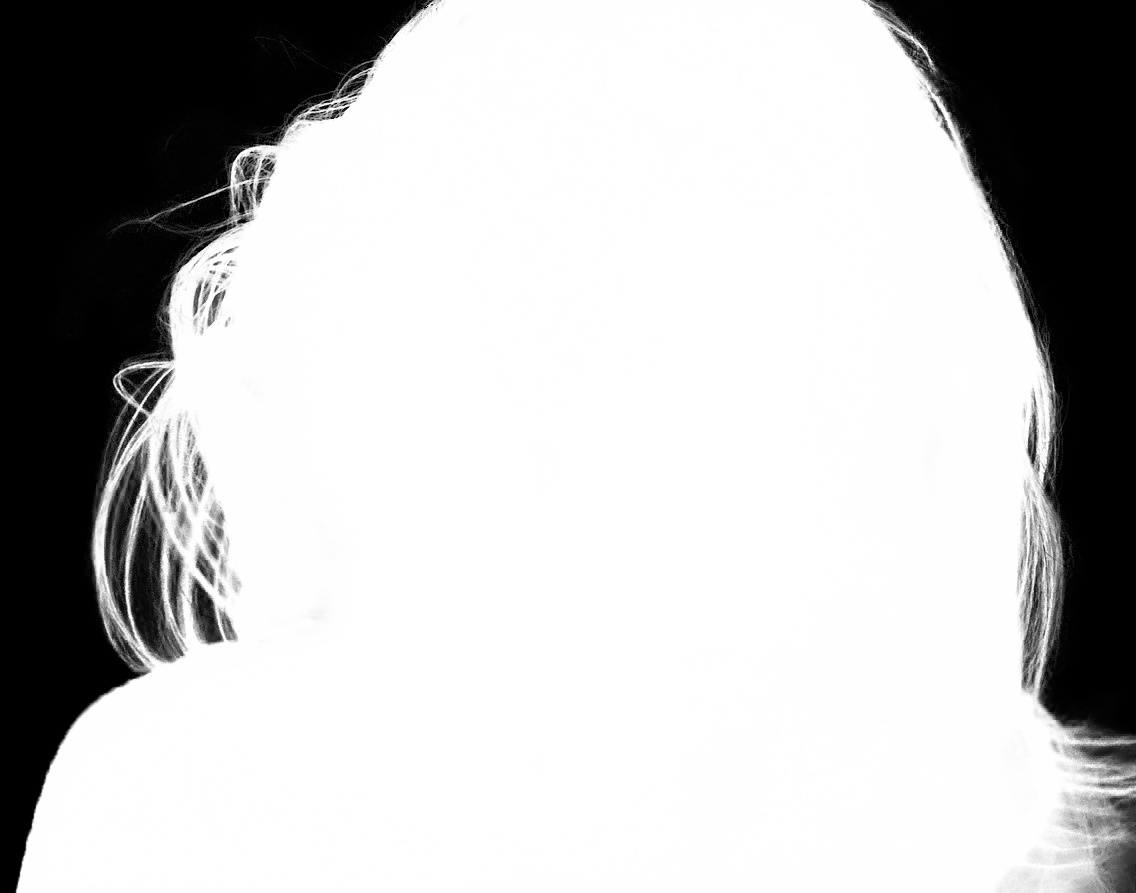}&\includegraphics[width=1.75cm]{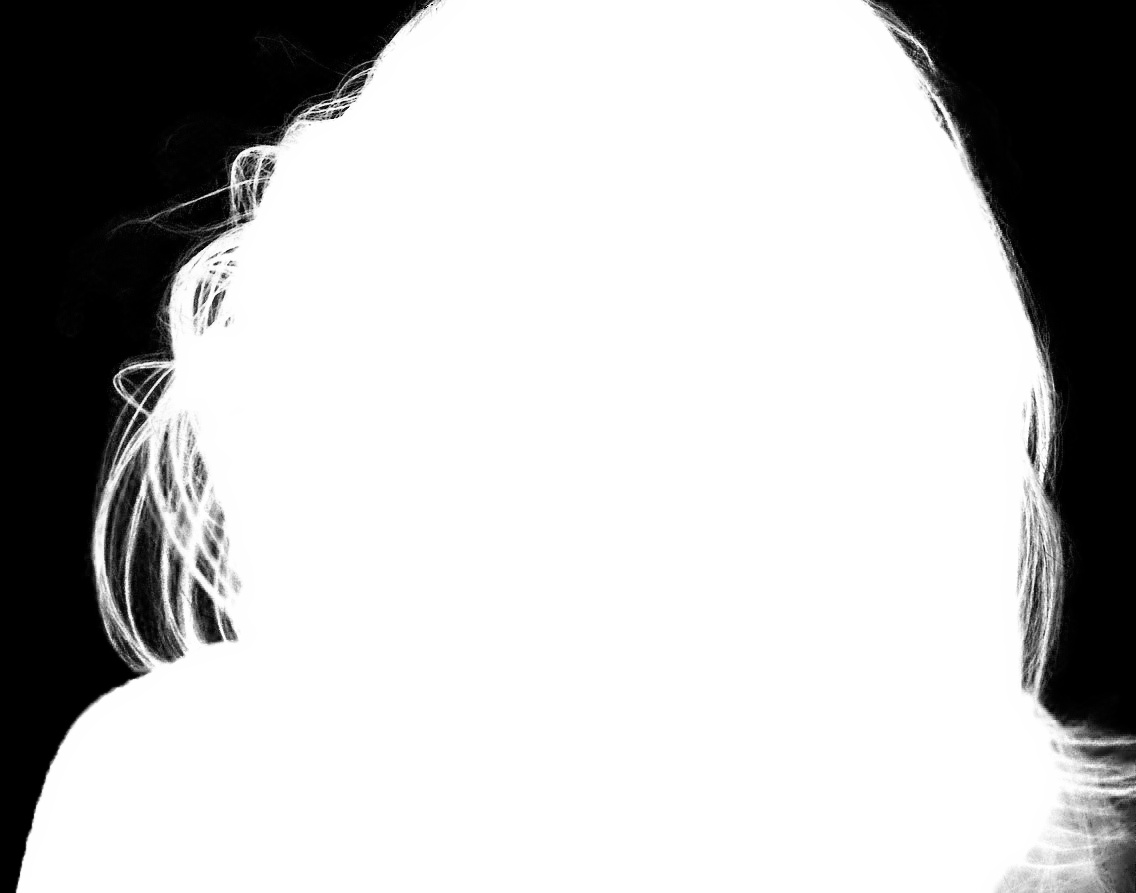}&\includegraphics[width=1.75cm]{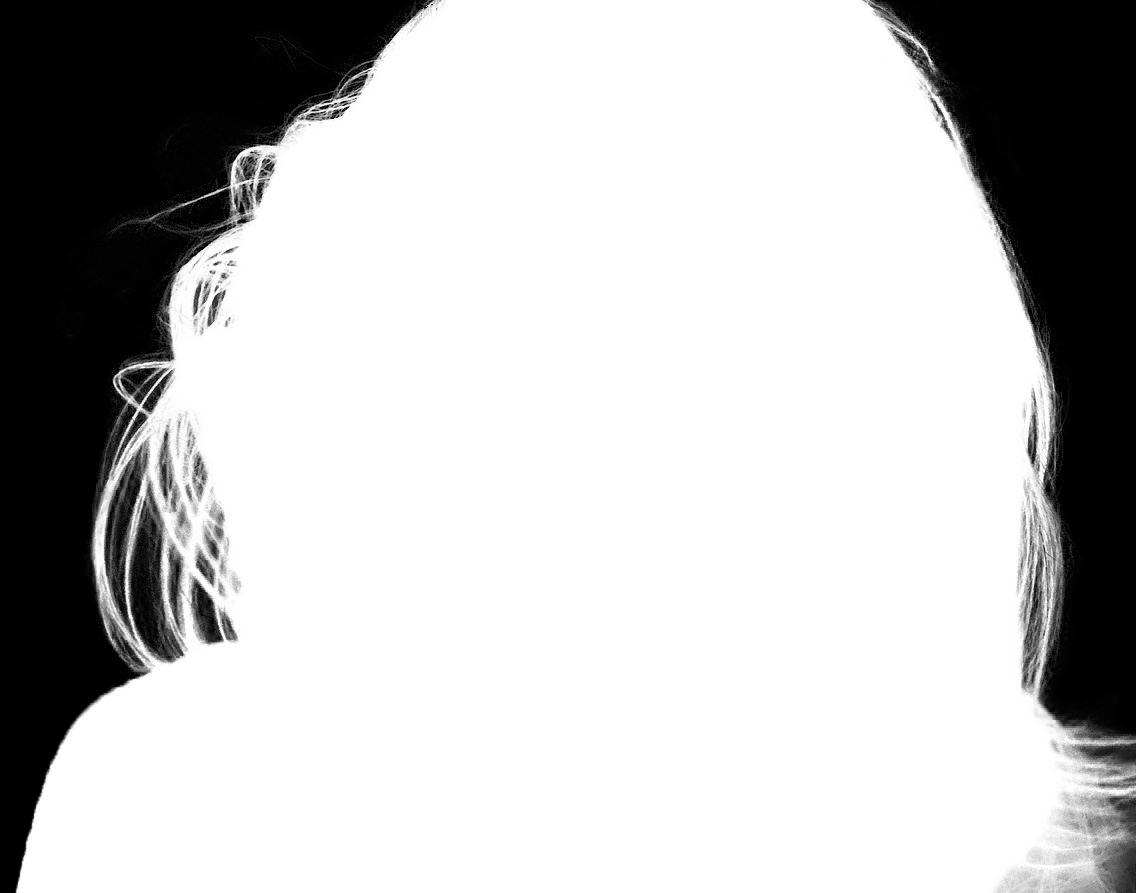}\\

\includegraphics[width=1.75cm]{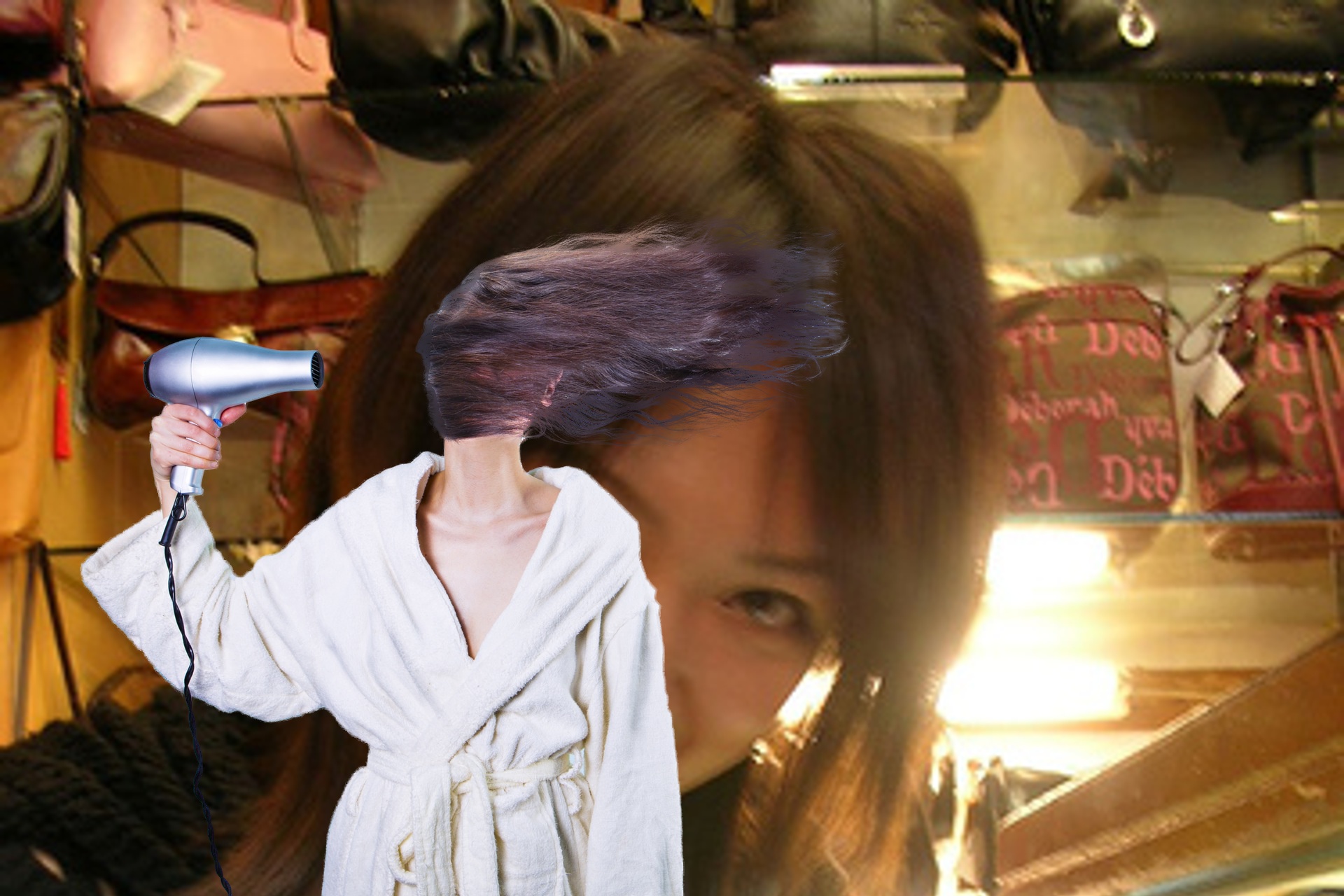}&\includegraphics[width=1.75cm]{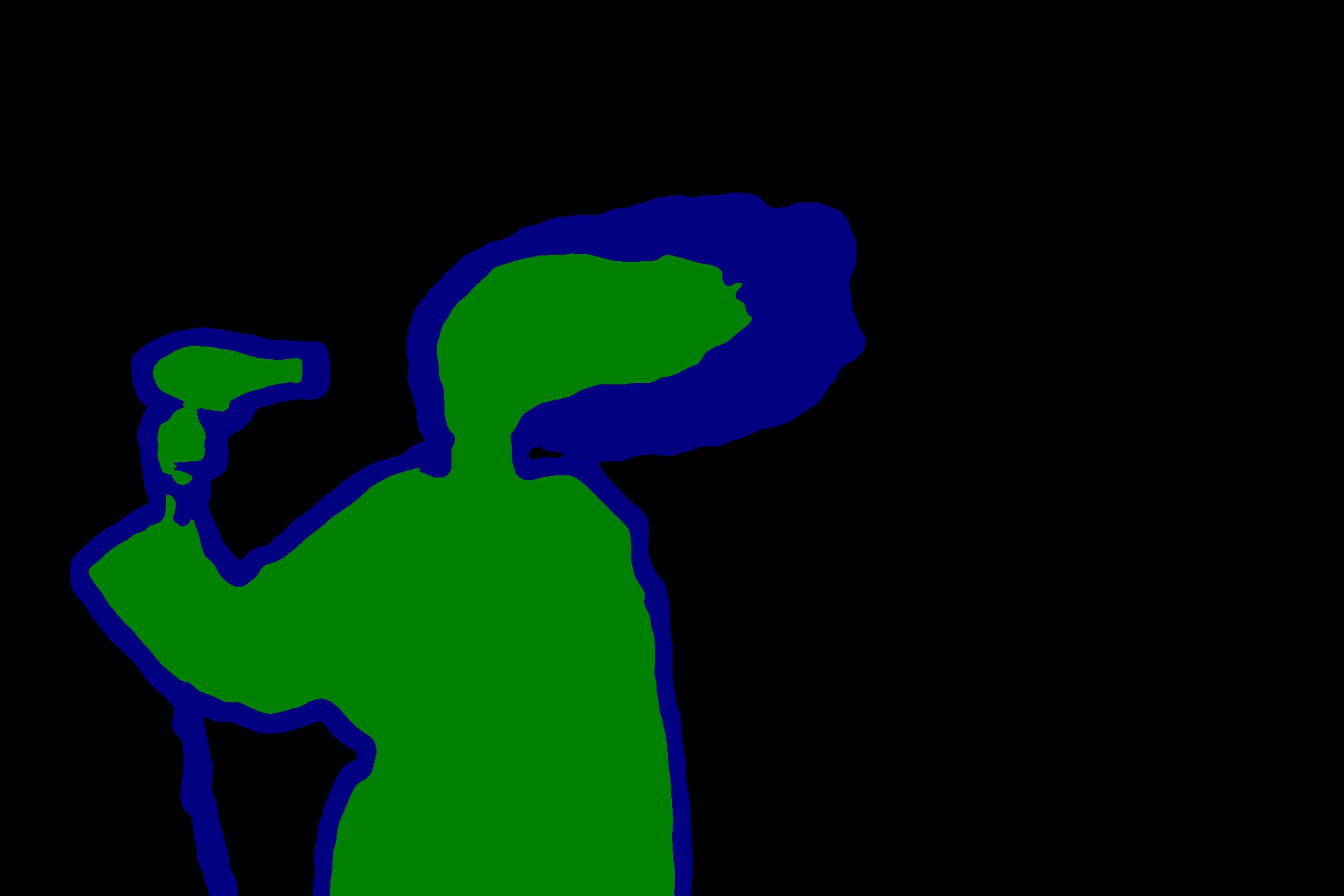}&\includegraphics[width=1.75cm]{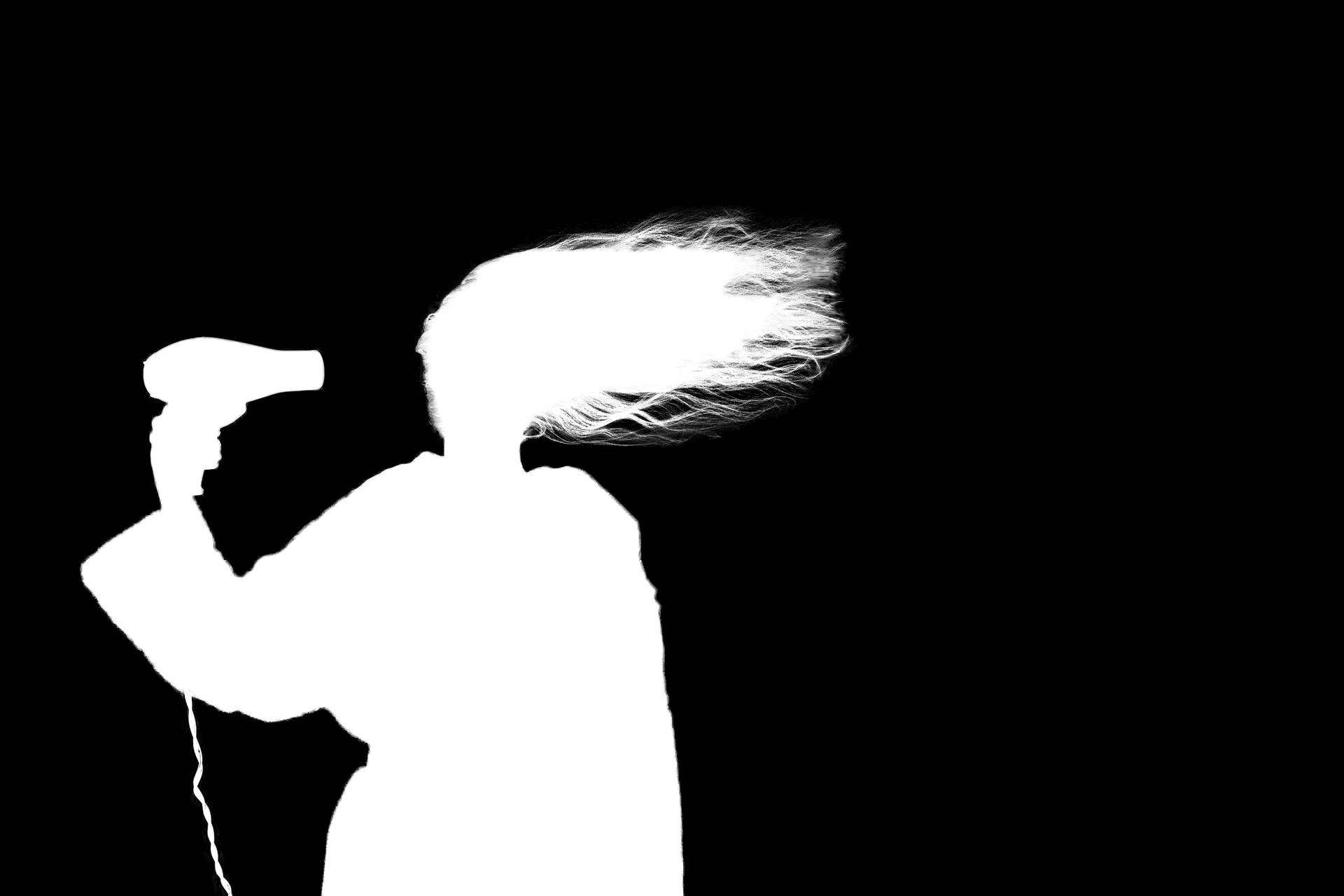}&\includegraphics[width=1.75cm]{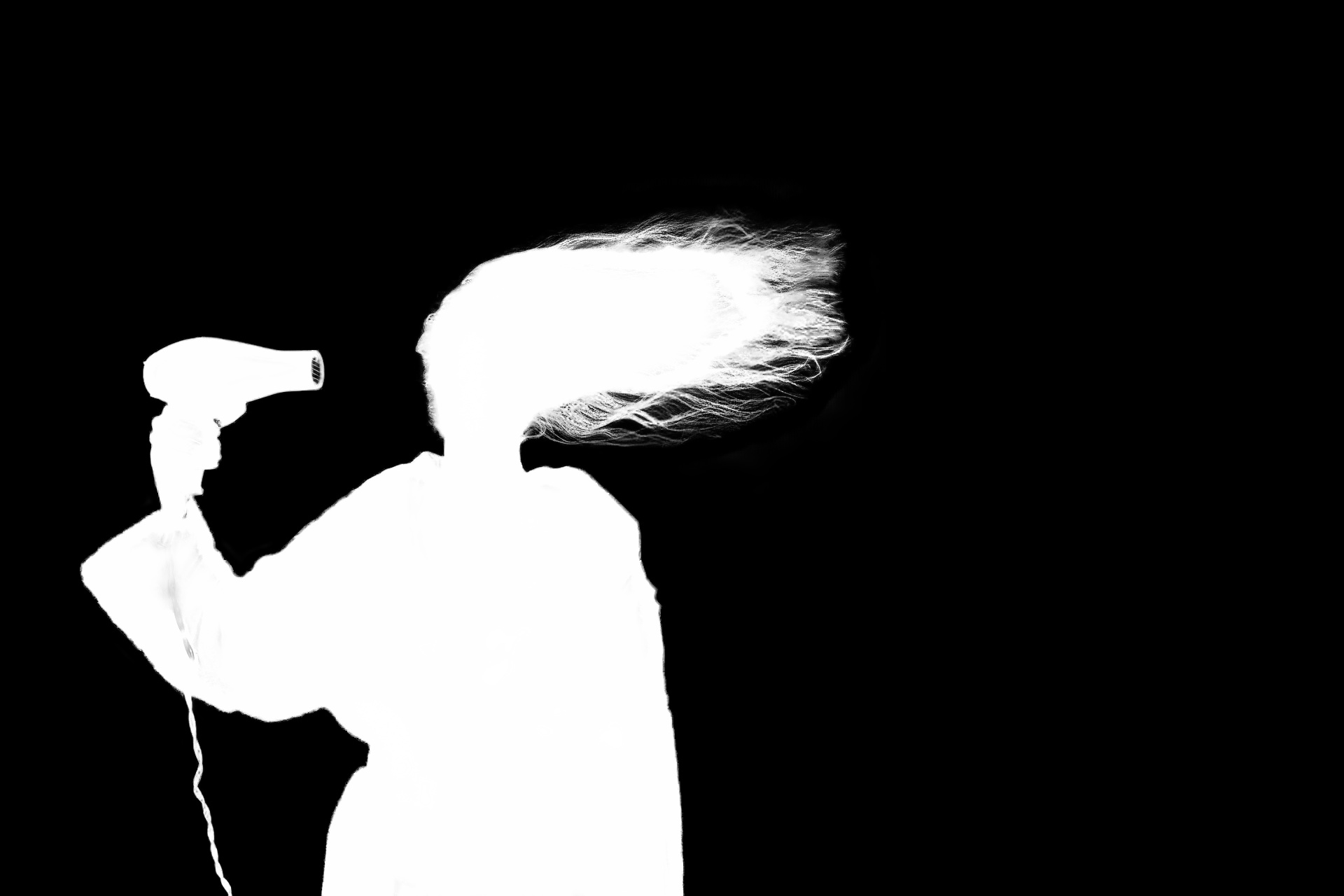}&\includegraphics[width=1.75cm]{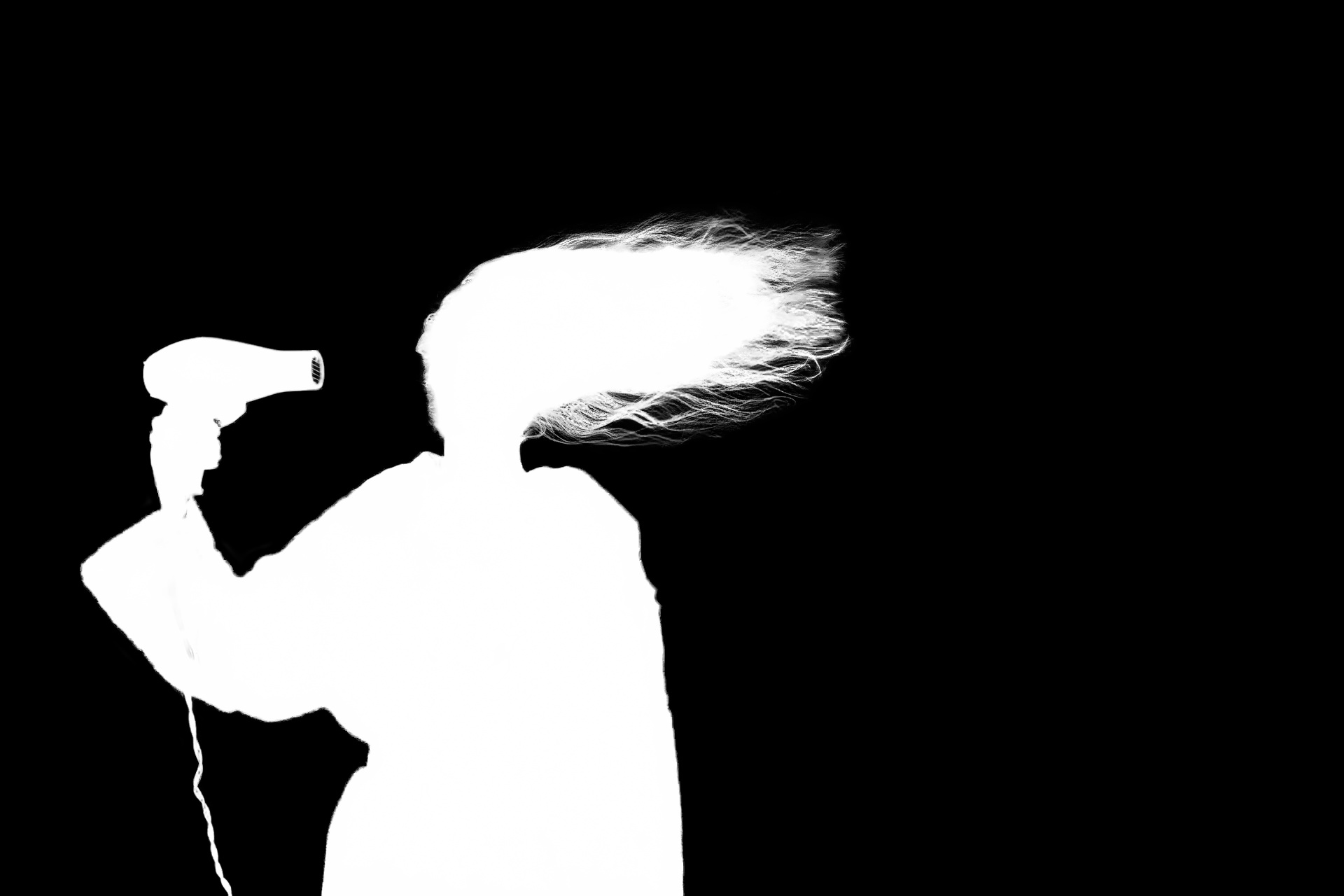}&\includegraphics[width=1.75cm]{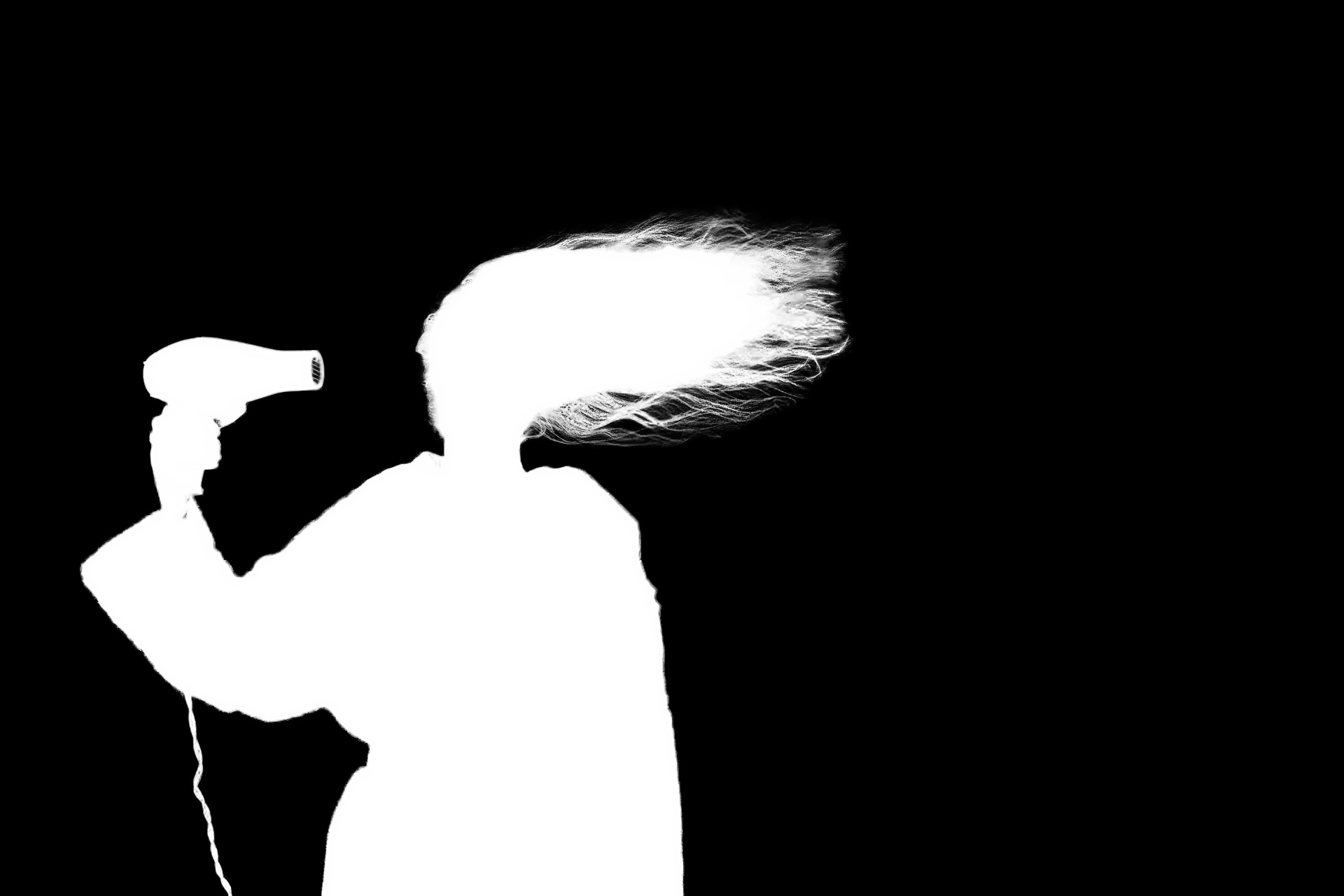}&\includegraphics[width=1.75cm]{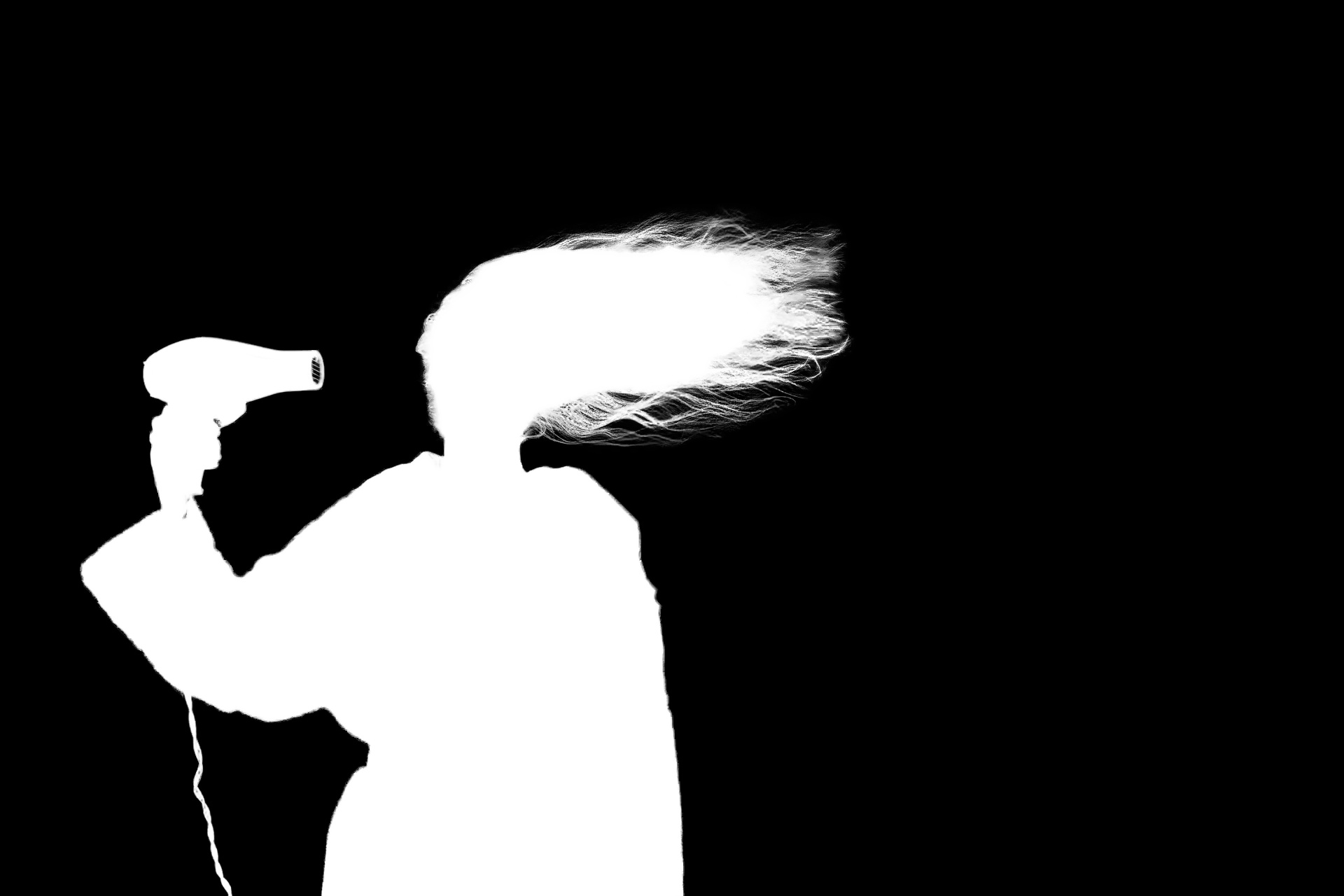}\\

\includegraphics[width=1.75cm]{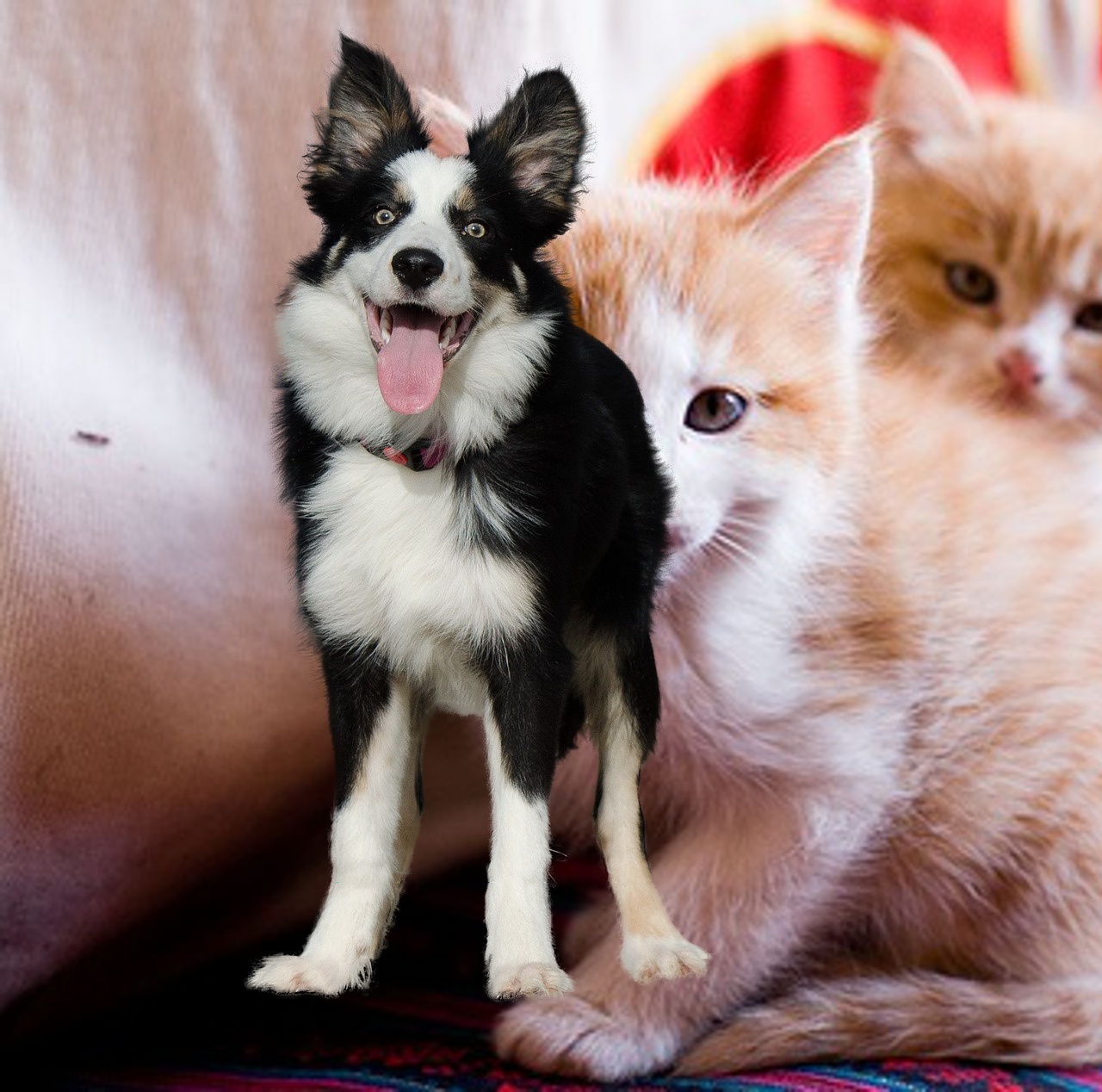}&\includegraphics[width=1.75cm]{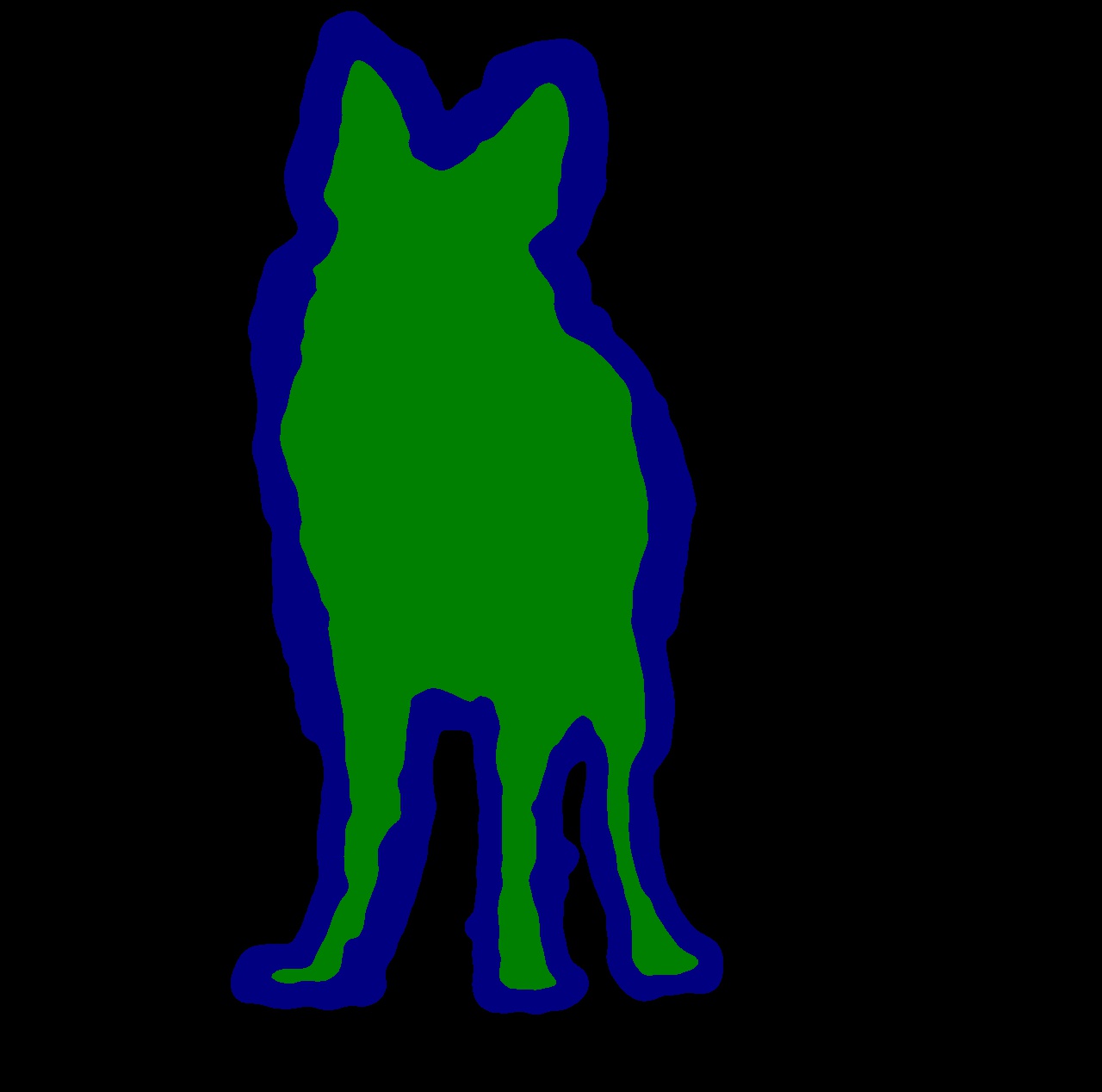}&\includegraphics[width=1.75cm]{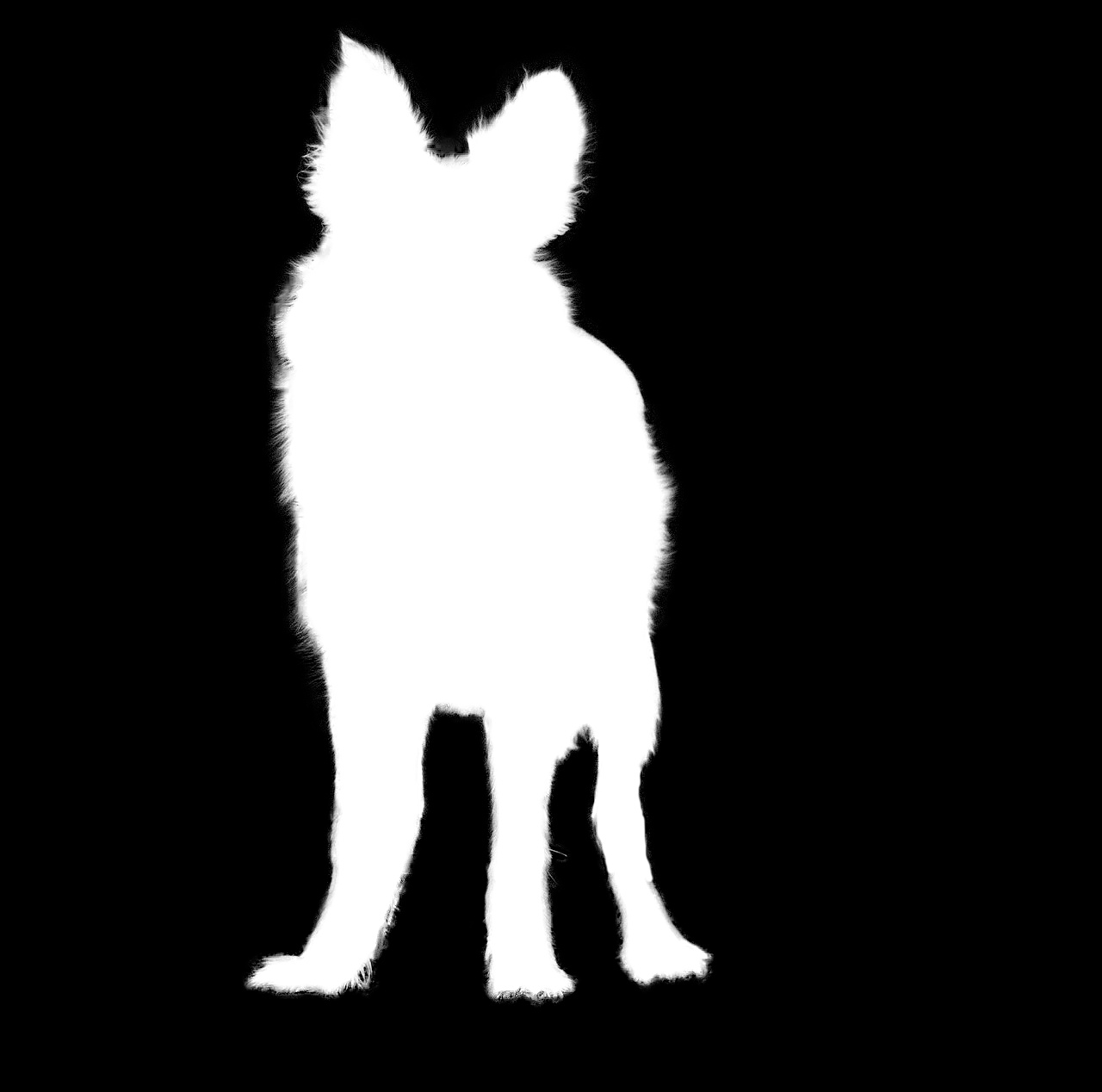}&\includegraphics[width=1.75cm]{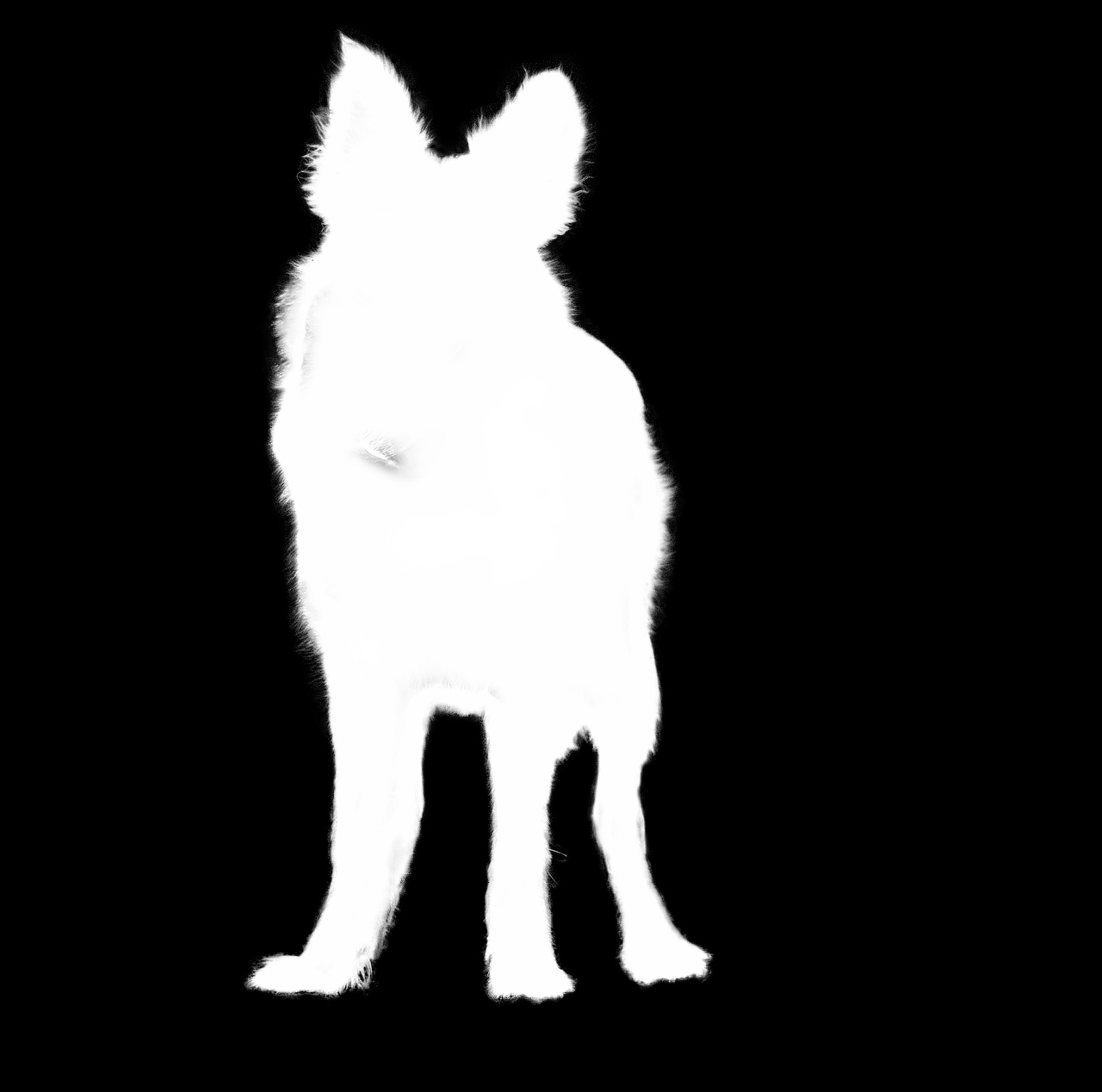}&\includegraphics[width=1.75cm]{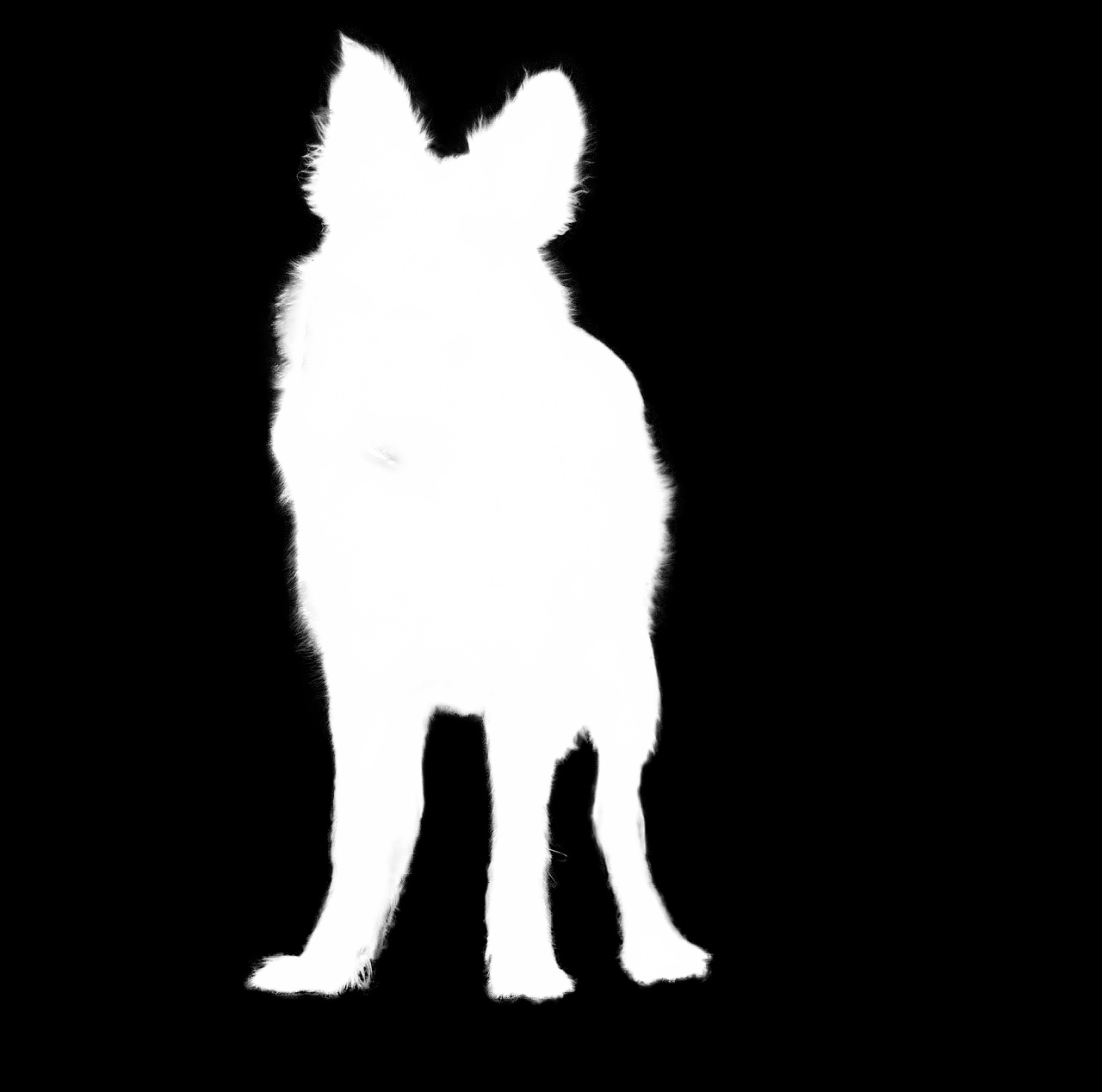}&\includegraphics[width=1.75cm]{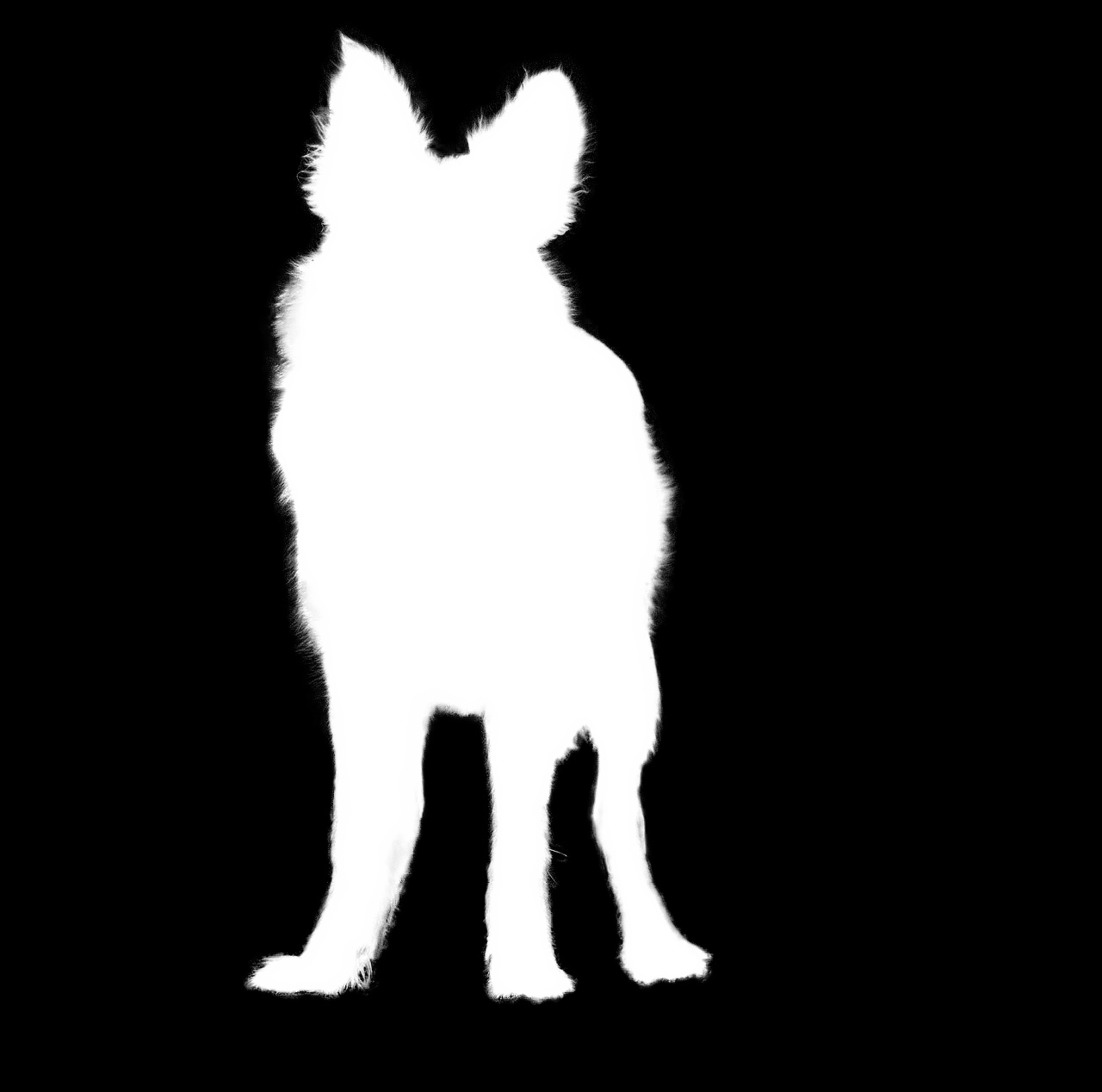}&\includegraphics[width=1.75cm]{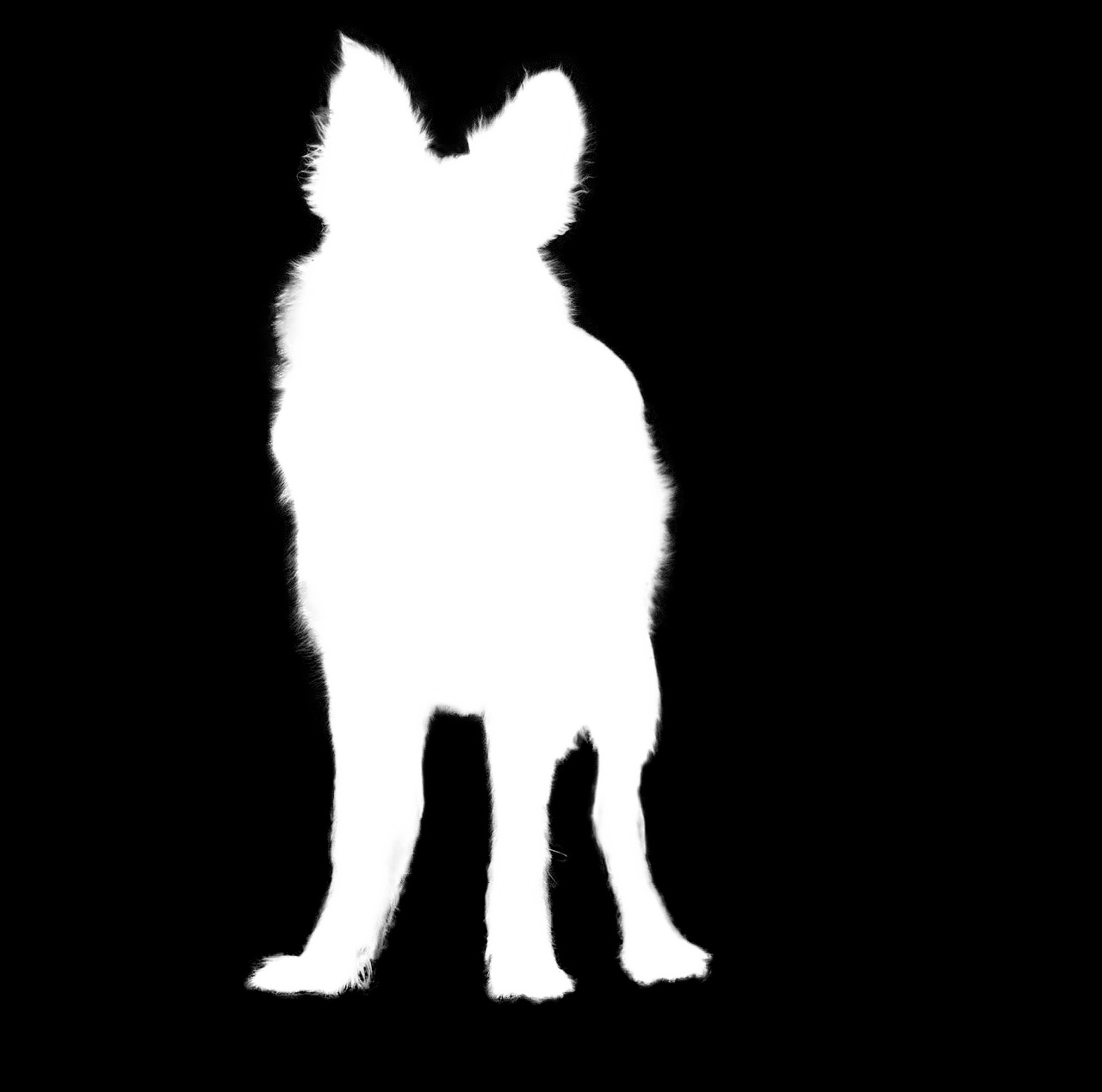}

\end{tabular}}
\caption{Ablation comparison of visual results of Joint Inference with Fusion on AIM benchmark. From left to right, image, trimap predicted by Net-T, GT, JIS-c, JIS, JIH-c, and JIH.}
\label{fig:joint_aim} 
\end{center}
\end{figure}

\textbf{Adobe Image Matting Benchmark (AIM)}
The Fig.~\ref{fig:joint_aim} shows the visual comparison of our appproach on JI setting with other matting methods.

\begin{figure*}[thpb]
\setlength\tabcolsep{0pt}
\renewcommand{\arraystretch}{0}
\begin{center}
\begin{tabular}{ccccccccc}
\includegraphics[width=2cm]{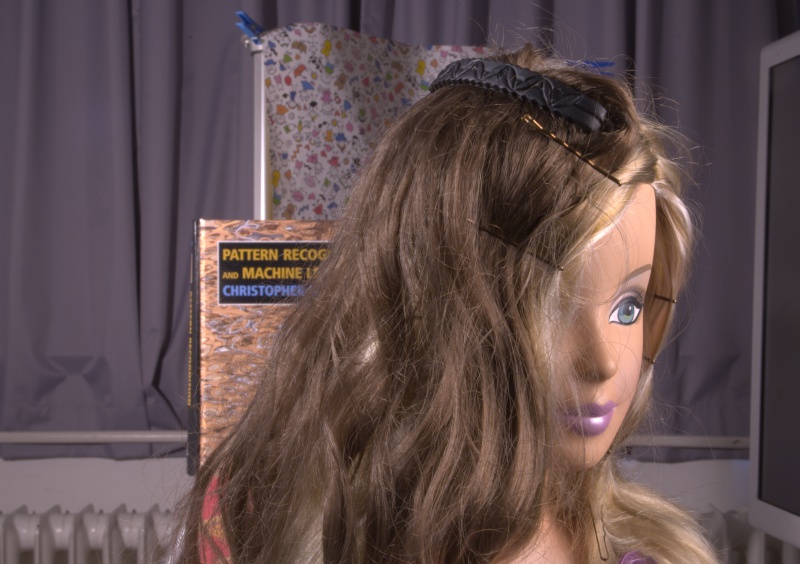}&\includegraphics[width=2cm]{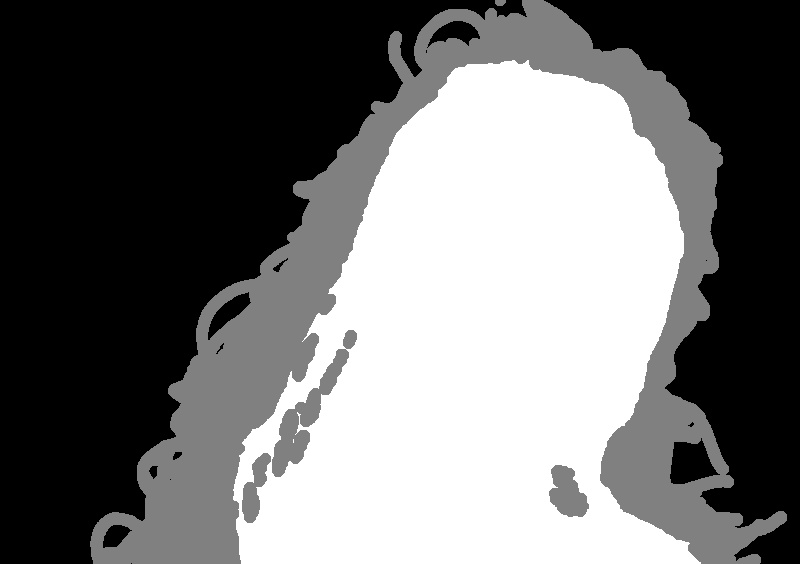}&\includegraphics[width=2cm]{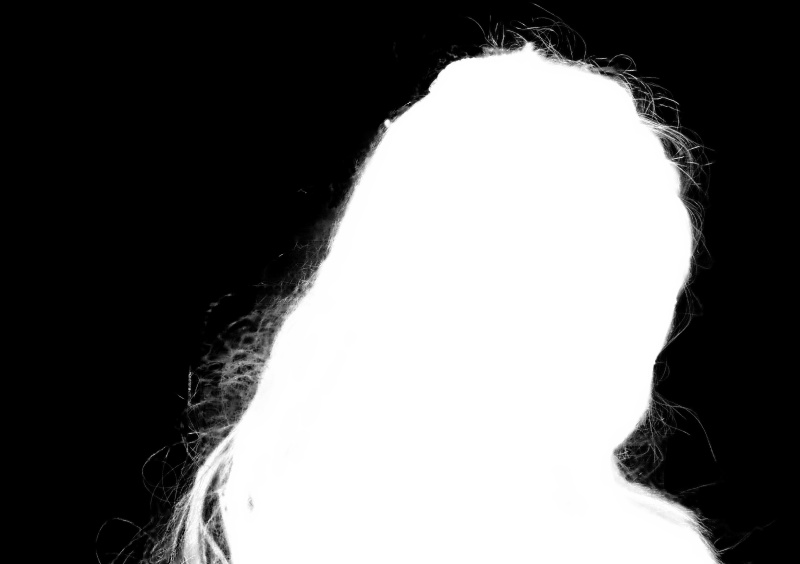}&\includegraphics[width=2cm]{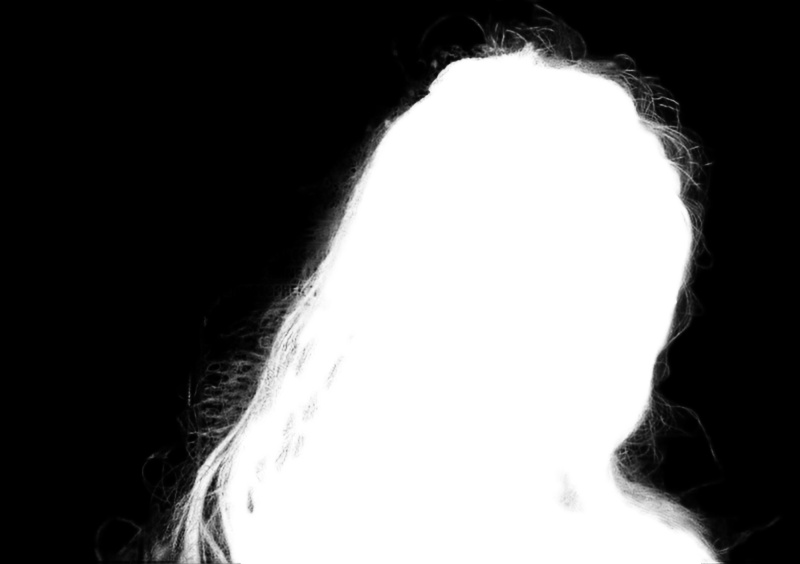}&\includegraphics[width=2cm]{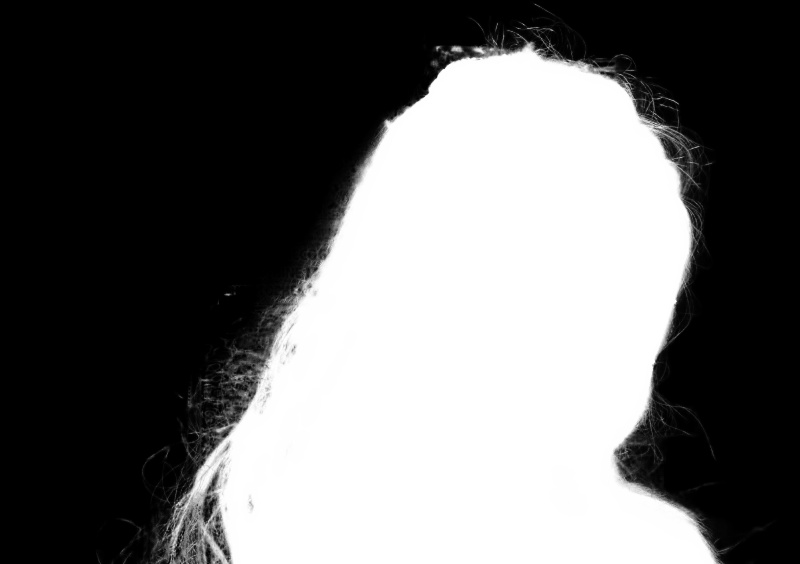}&\includegraphics[width=2cm]{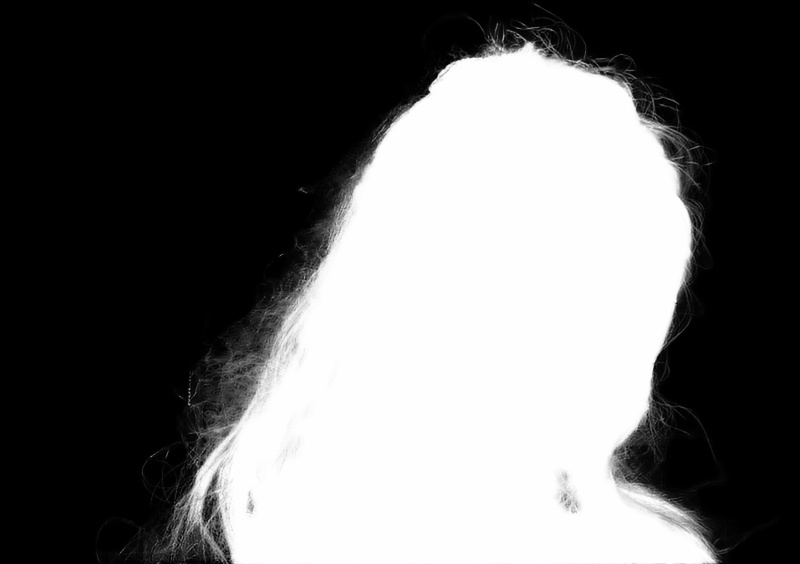}&\includegraphics[width=2cm]{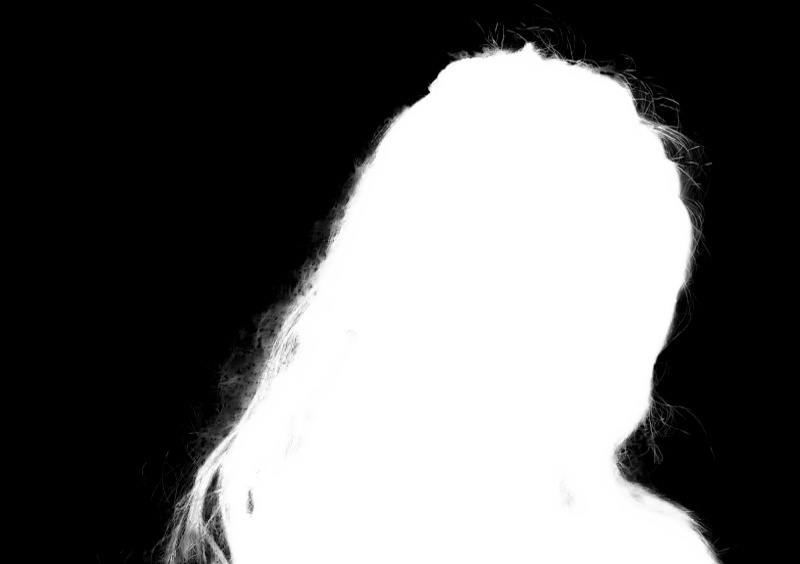}&\includegraphics[width=2cm]{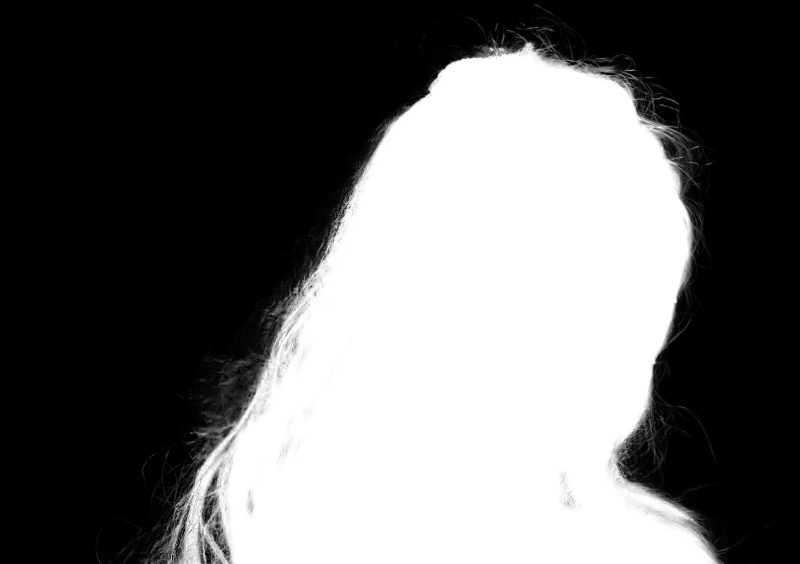}&\includegraphics[width=2cm]{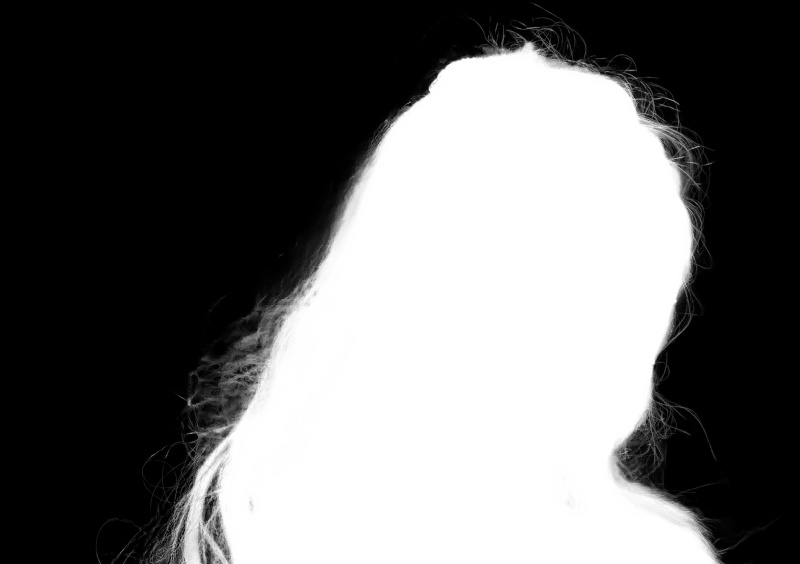}\\
\includegraphics[width=2cm]{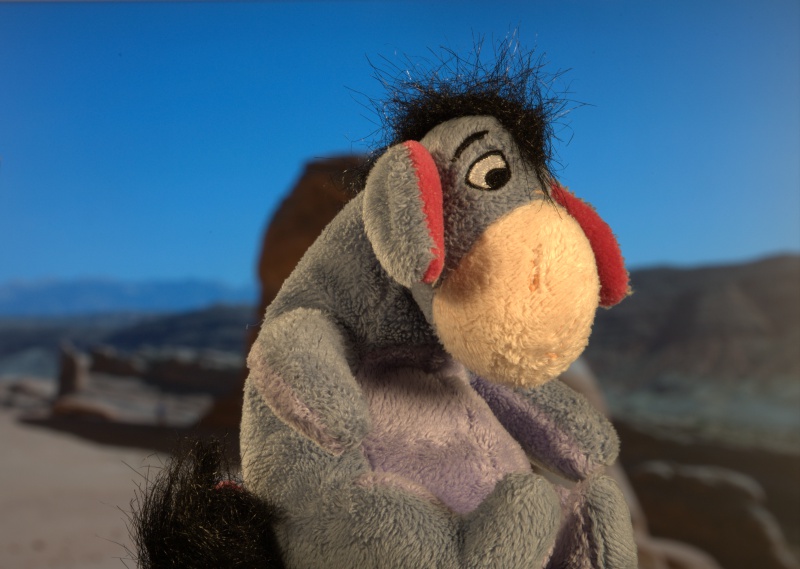}&\includegraphics[width=2cm]{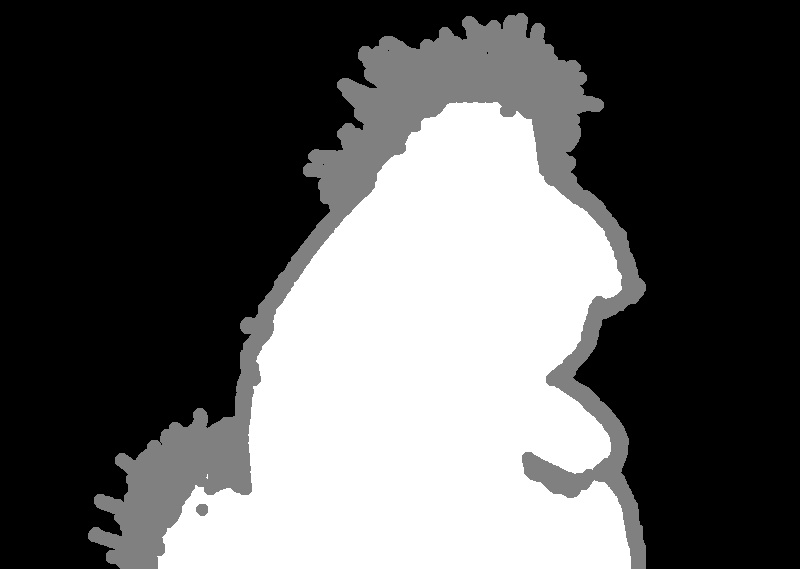}&\includegraphics[width=2cm]{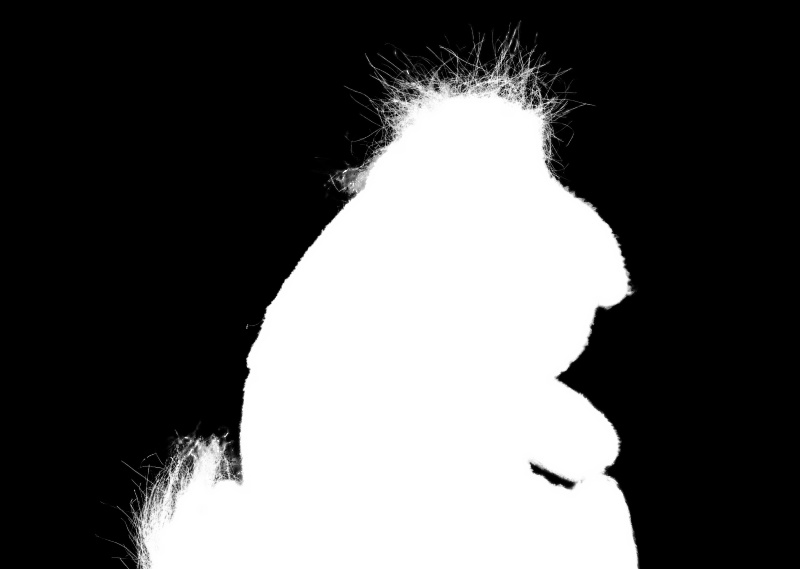}&\includegraphics[width=2cm]{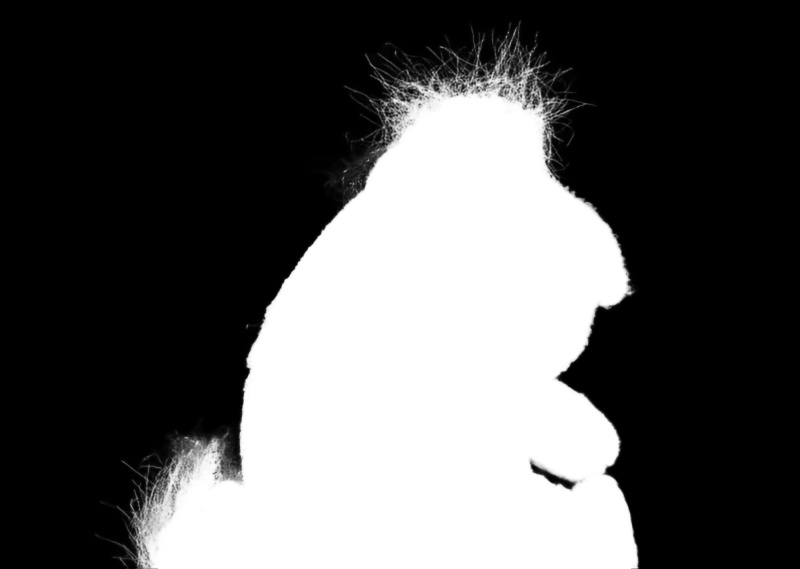}&\includegraphics[width=2cm]{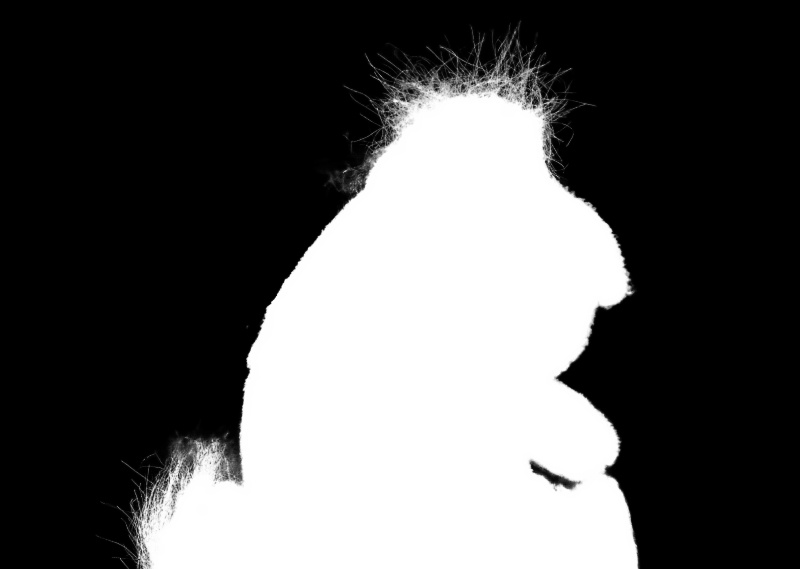}&\includegraphics[width=2cm]{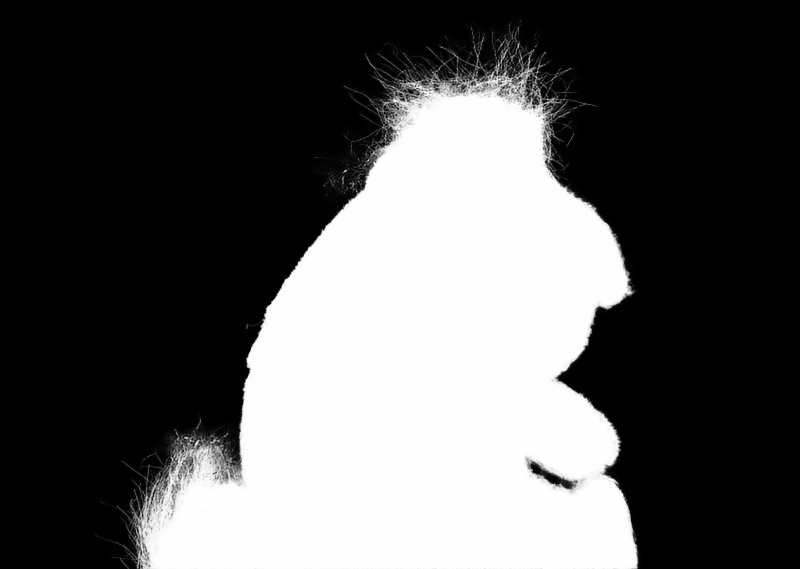}&\includegraphics[width=2cm]{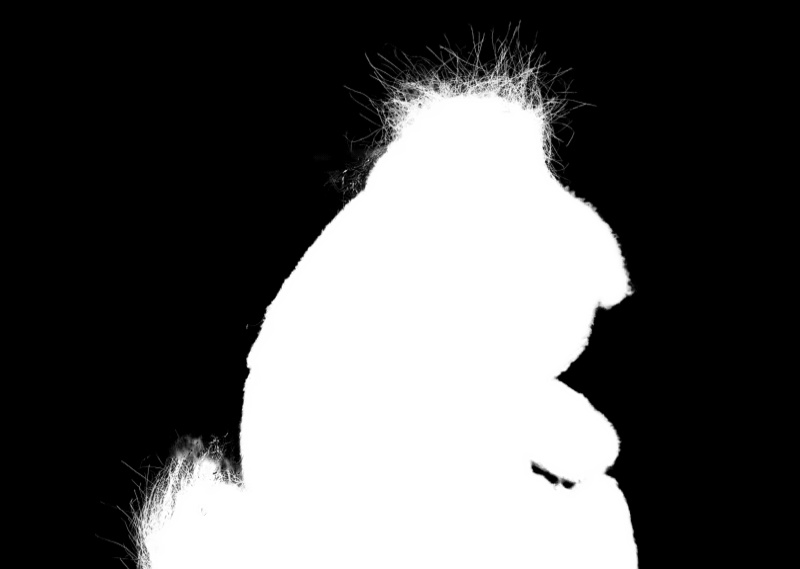}&\includegraphics[width=2cm]{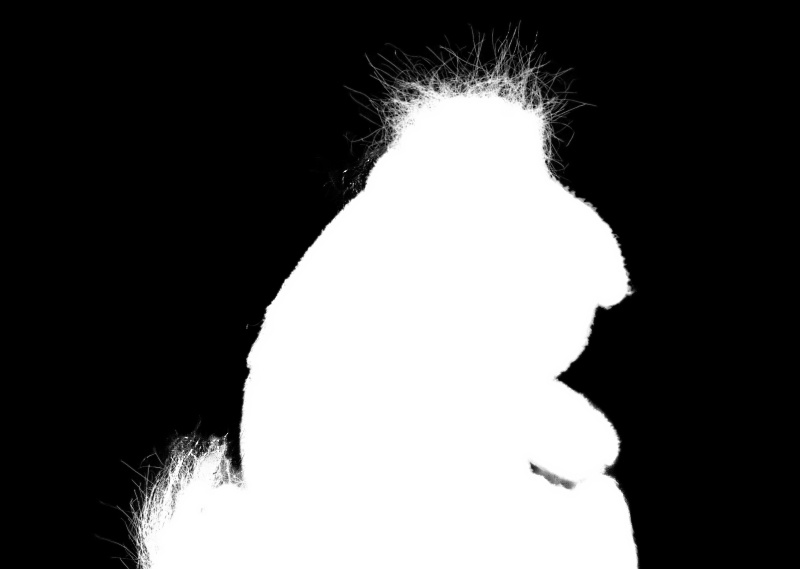}&\includegraphics[width=2cm]{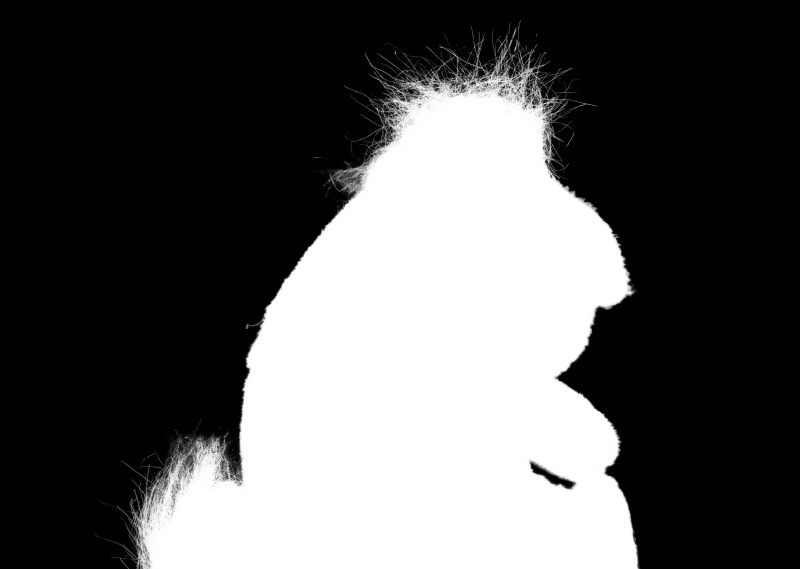}\\
\includegraphics[width=2cm]{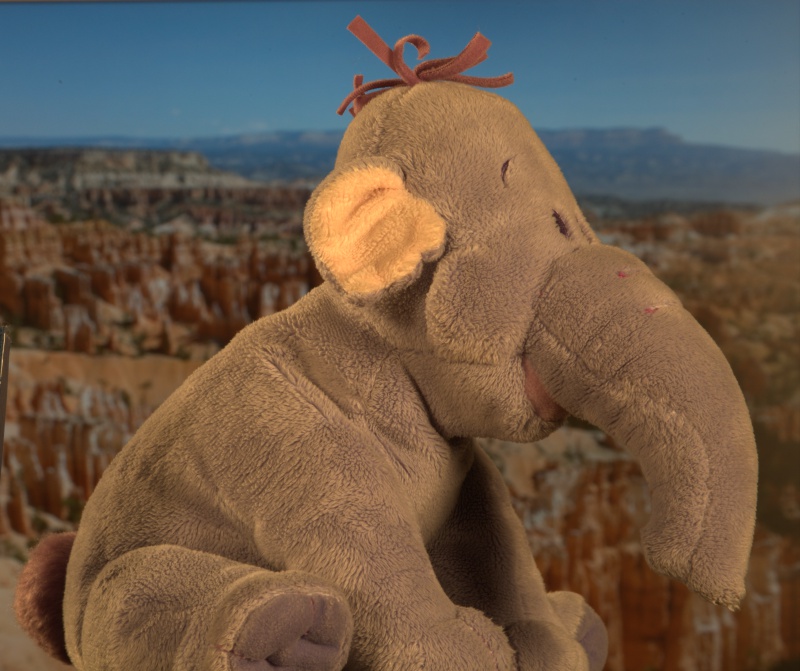}&\includegraphics[width=2cm]{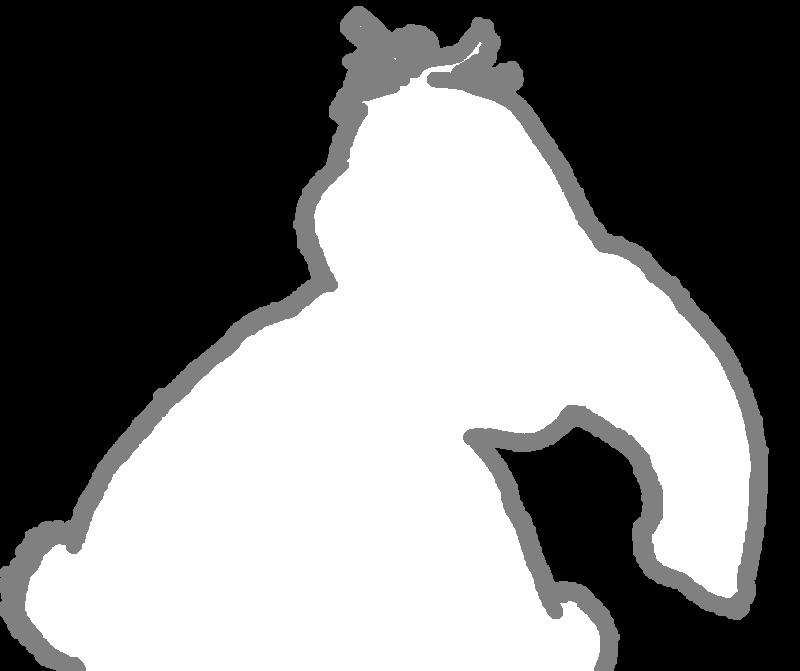}&\includegraphics[width=2cm]{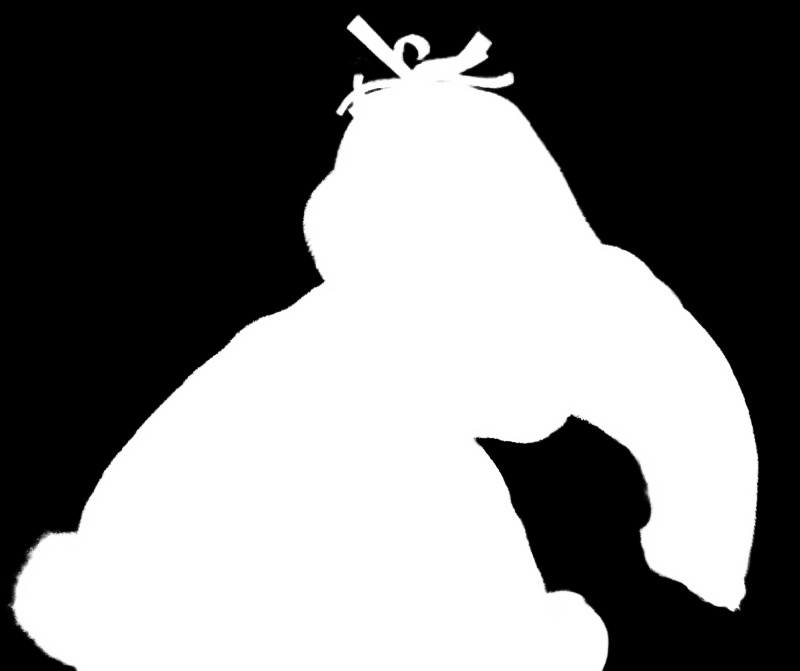}&\includegraphics[width=2cm]{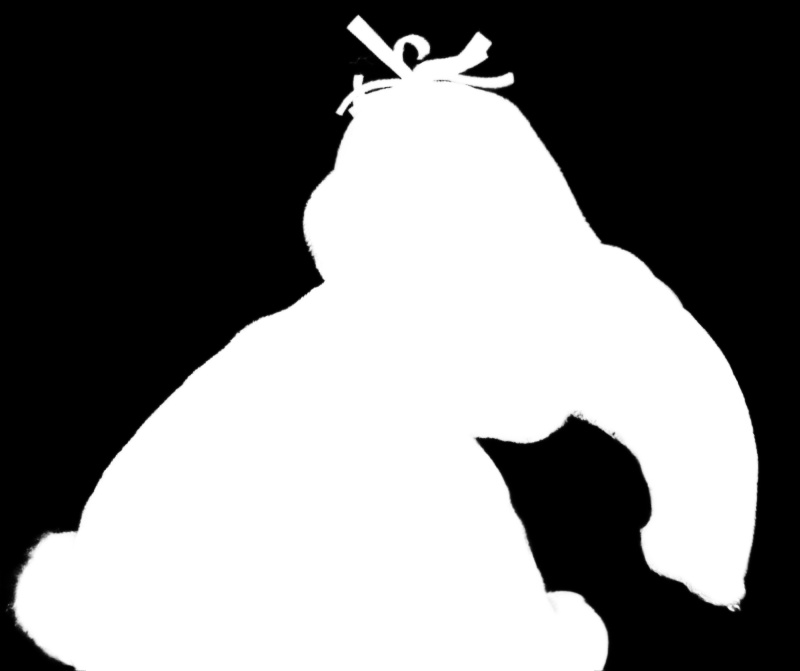}&\includegraphics[width=2cm]{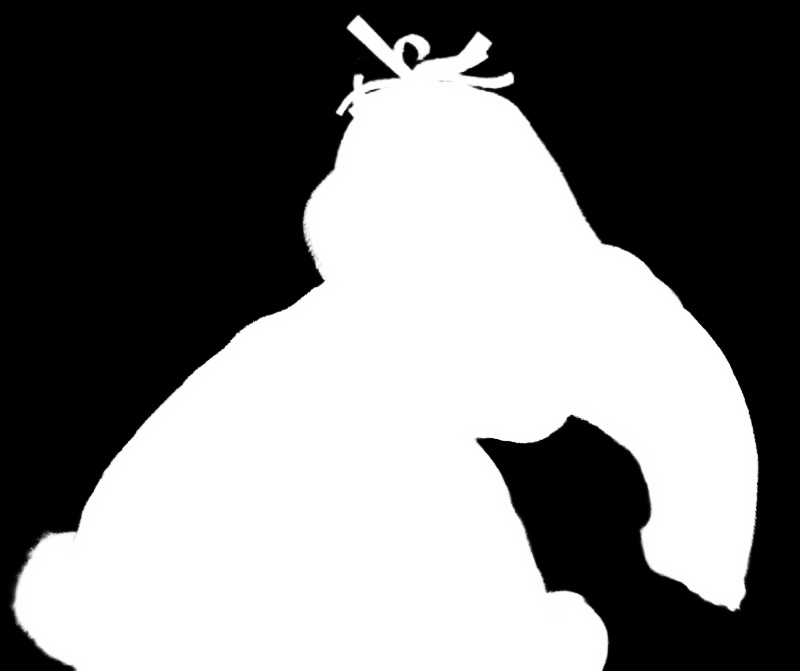}&\includegraphics[width=2cm]{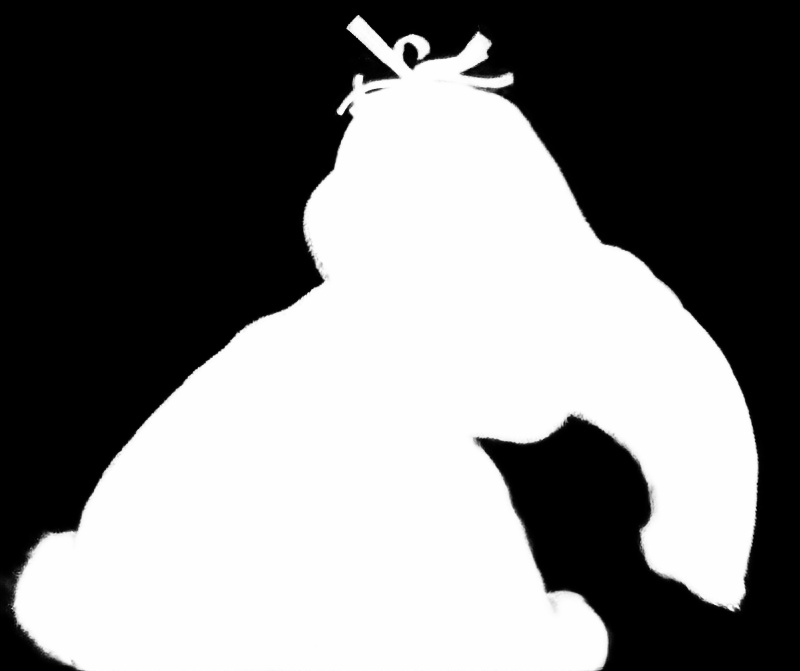}&\includegraphics[width=2cm]{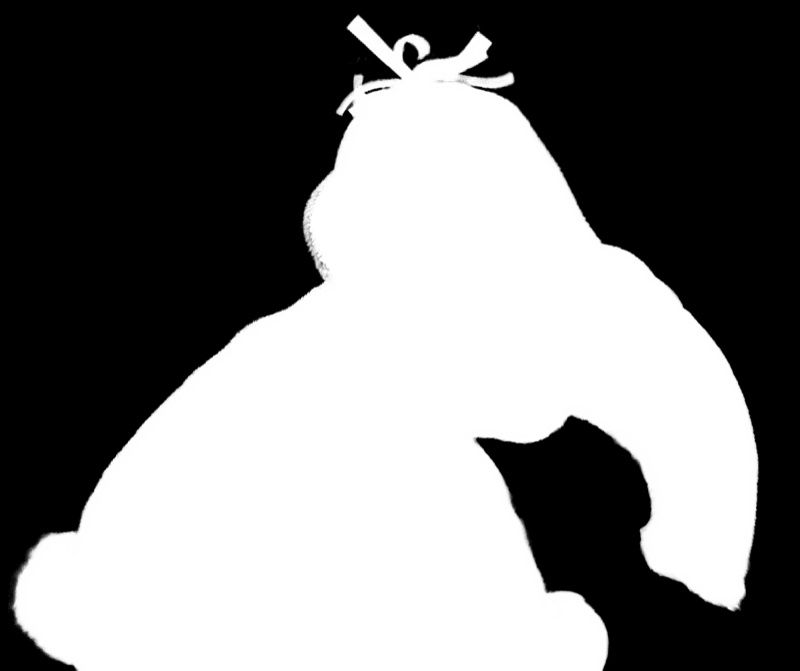}&\includegraphics[width=2cm]{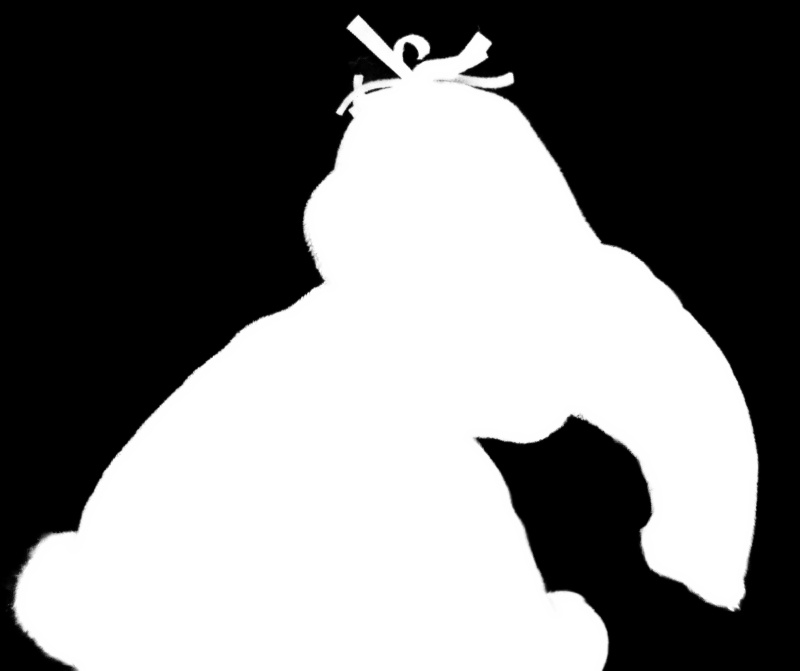}&\includegraphics[width=2cm]{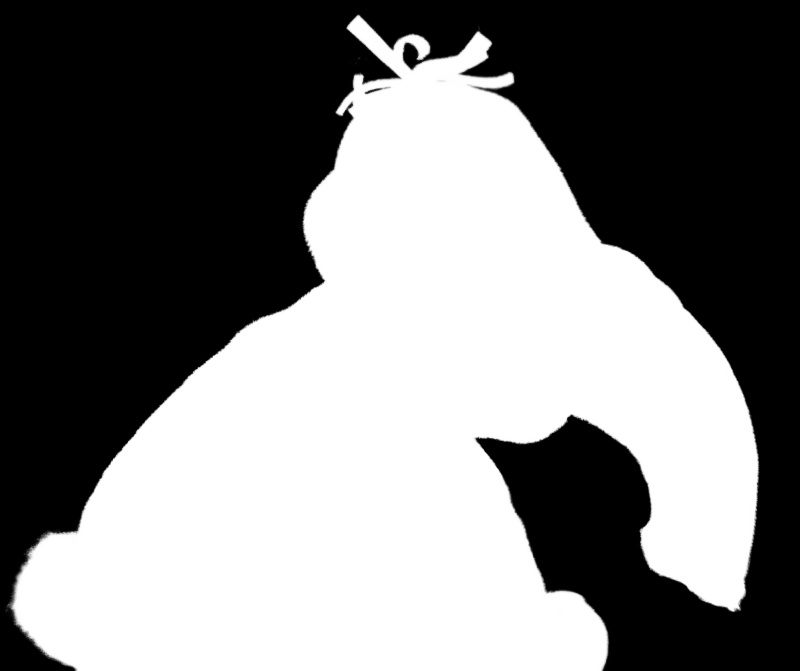}\\
\includegraphics[width=2cm]{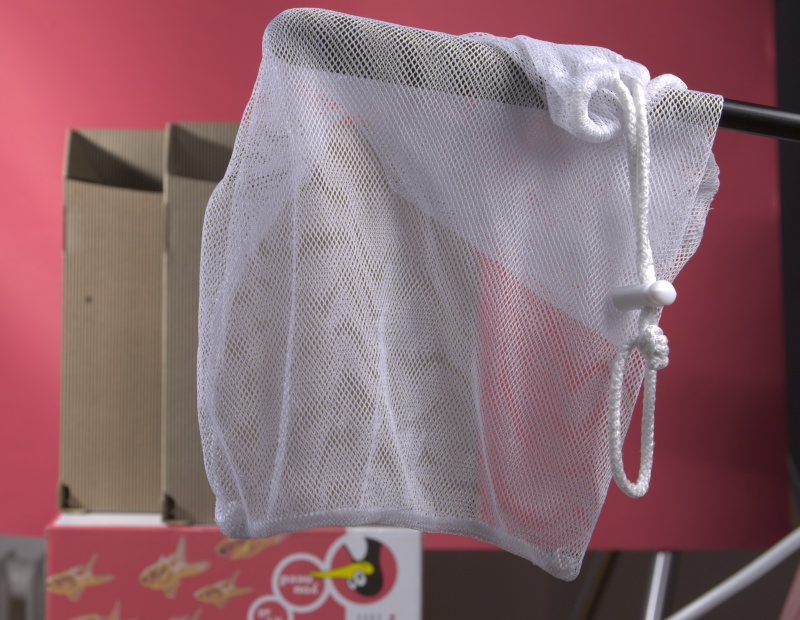}&\includegraphics[width=2cm]{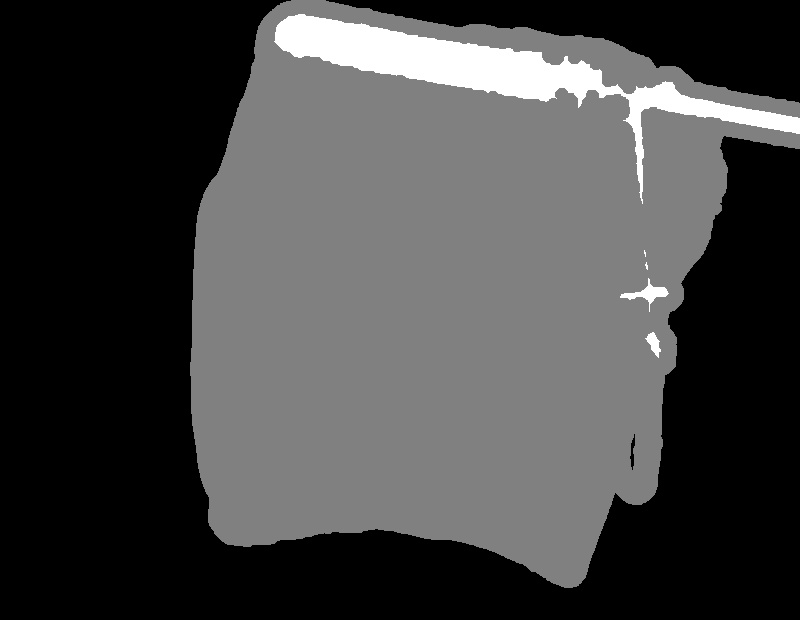}&\includegraphics[width=2cm]{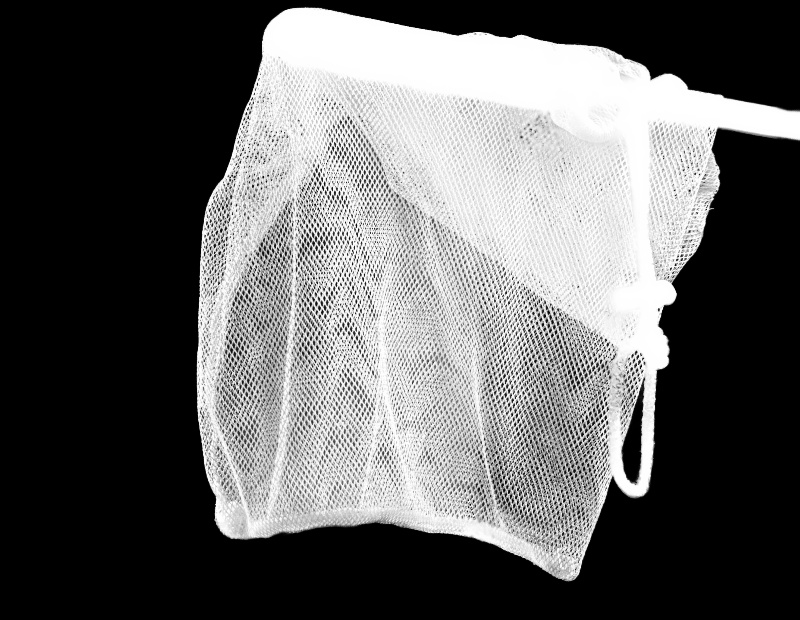}&\includegraphics[width=2cm]{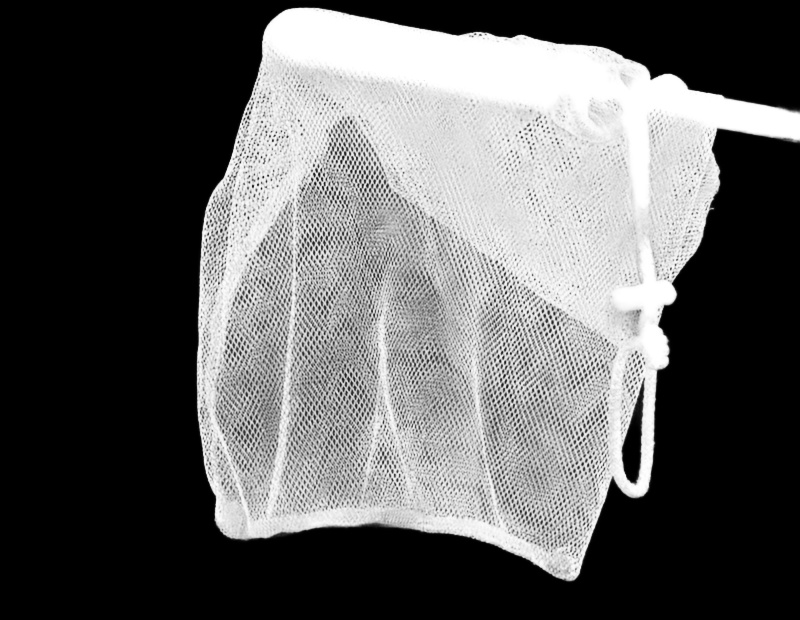}&\includegraphics[width=2cm]{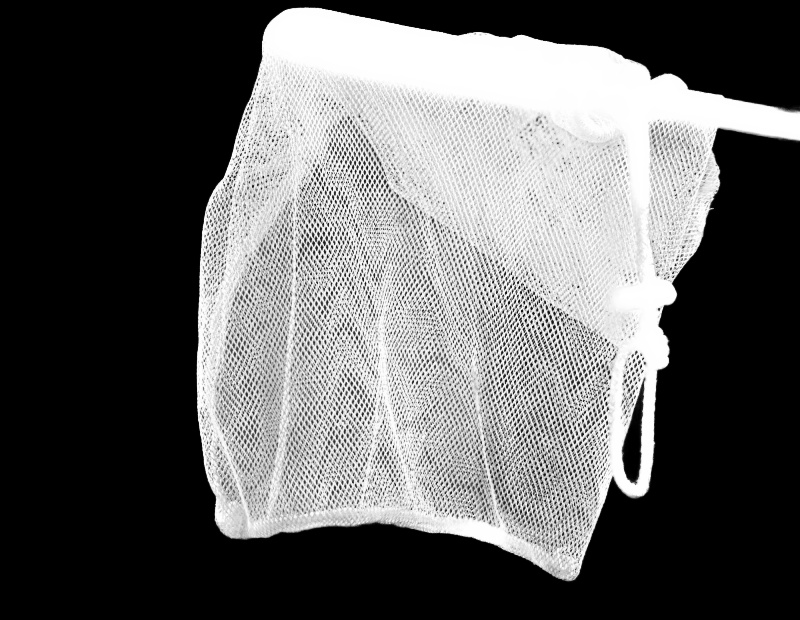}&\includegraphics[width=2cm]{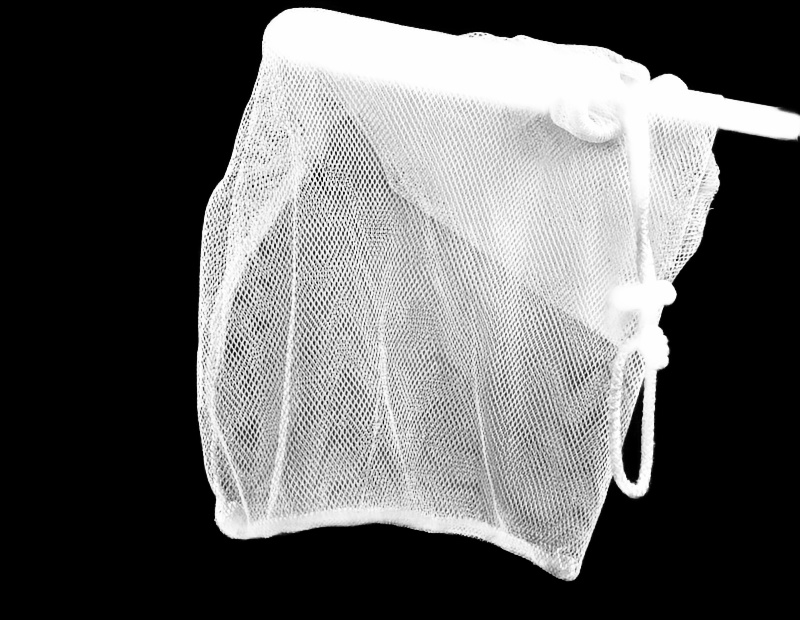}&\includegraphics[width=2cm]{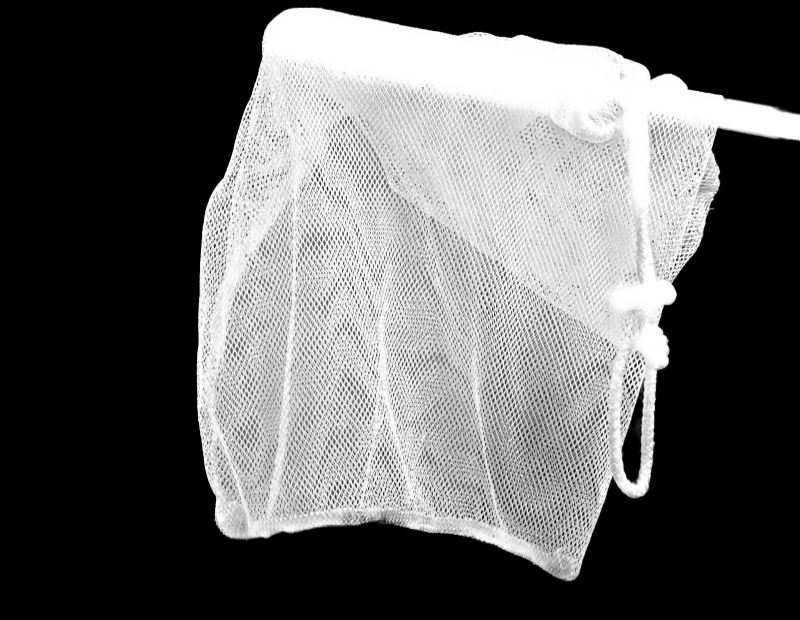}&\includegraphics[width=2cm]{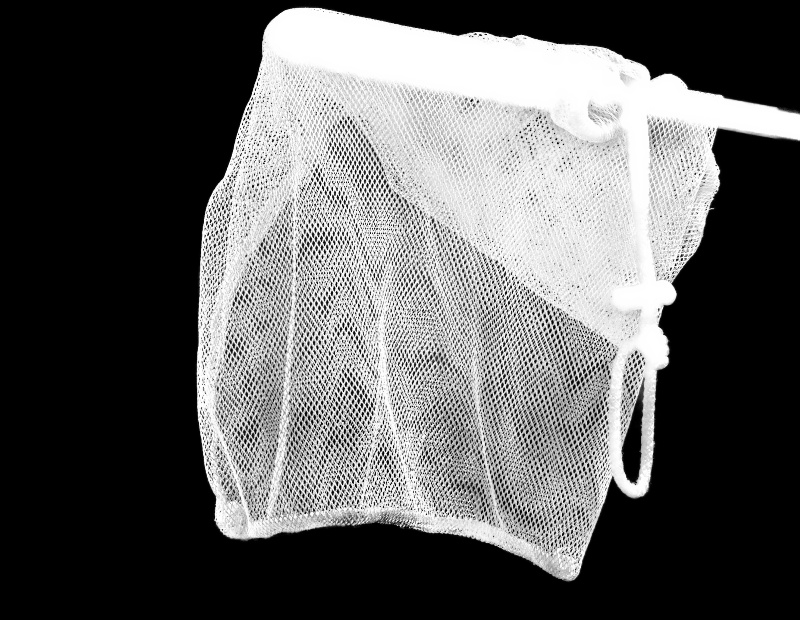}&\includegraphics[width=2cm]{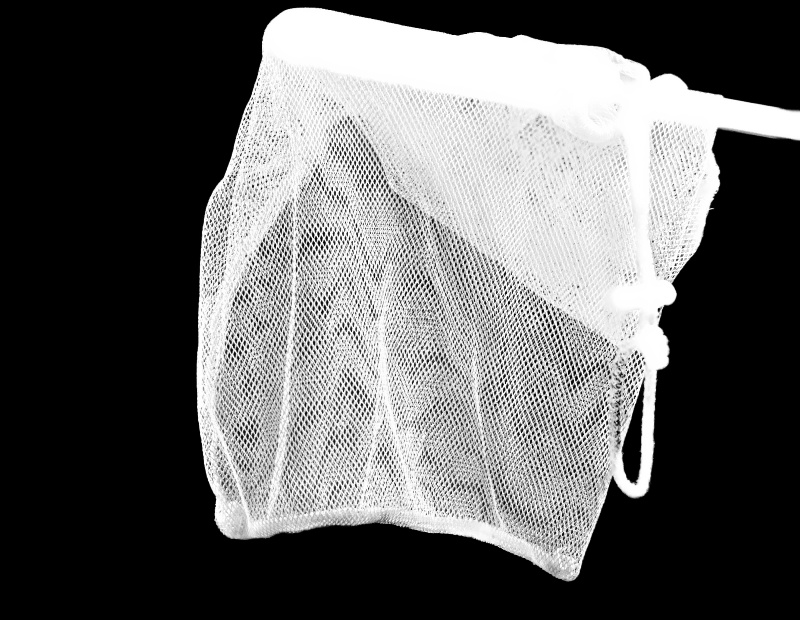}\\
\includegraphics[width=2cm]{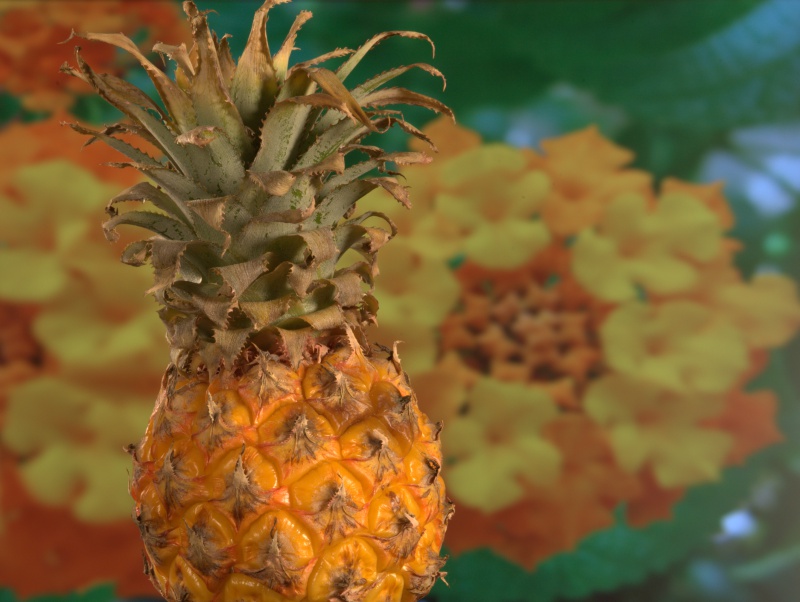}&\includegraphics[width=2cm]{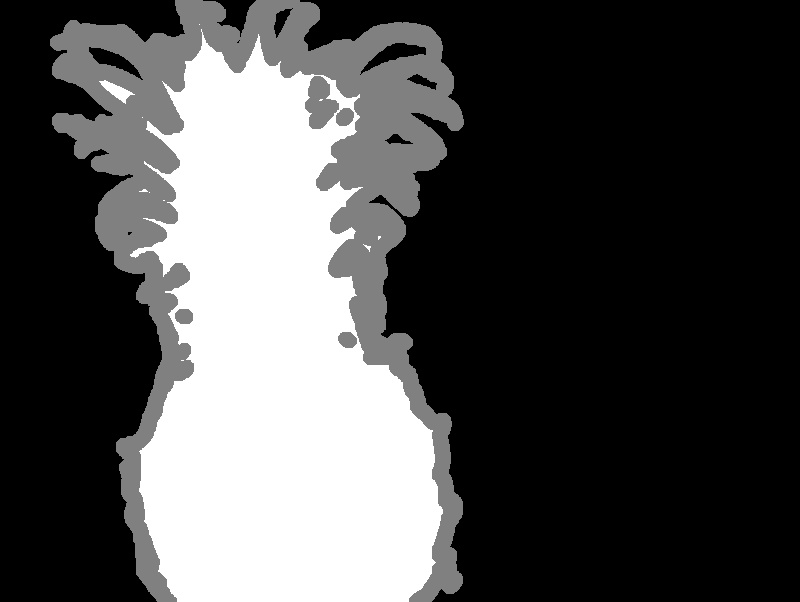}&\includegraphics[width=2cm]{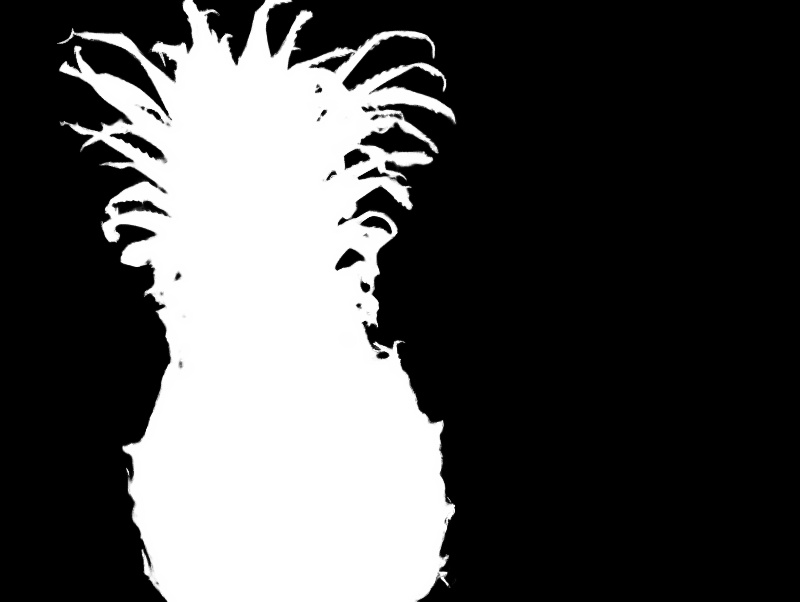}&\includegraphics[width=2cm]{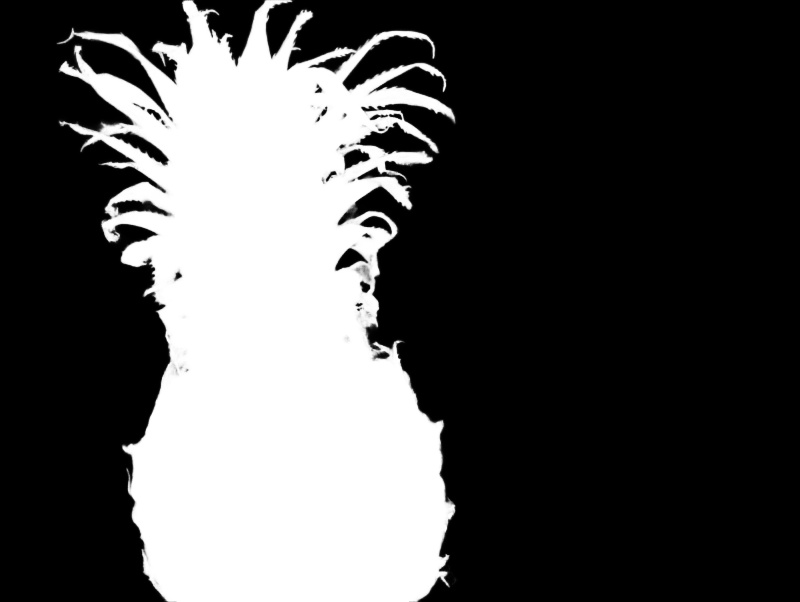}&\includegraphics[width=2cm]{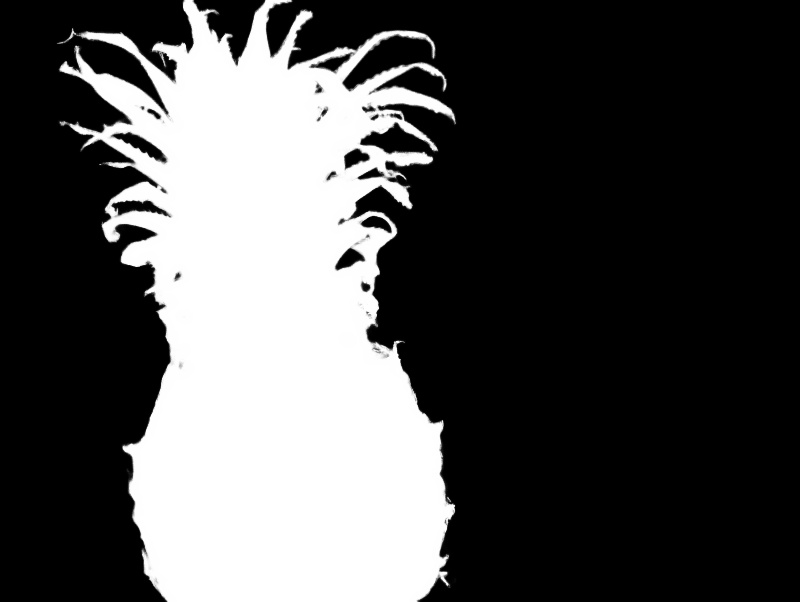}&\includegraphics[width=2cm]{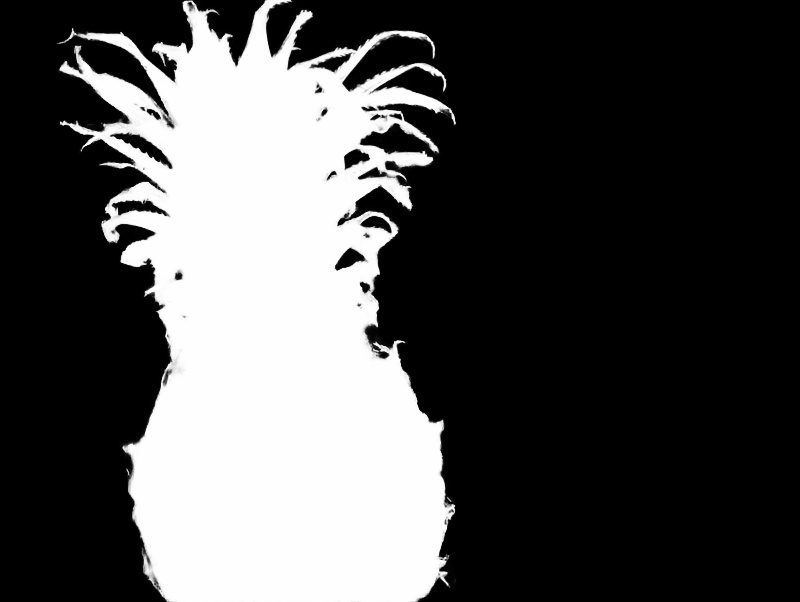}&\includegraphics[width=2cm]{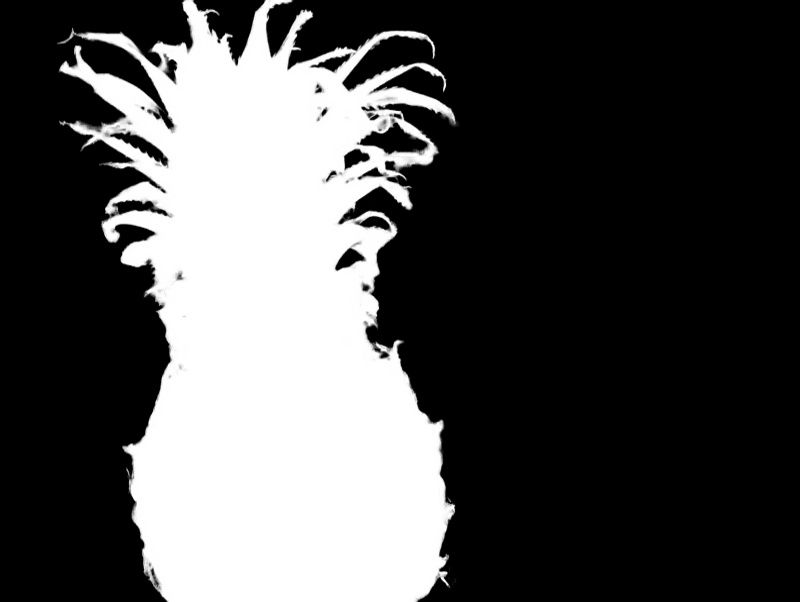}&\includegraphics[width=2cm]{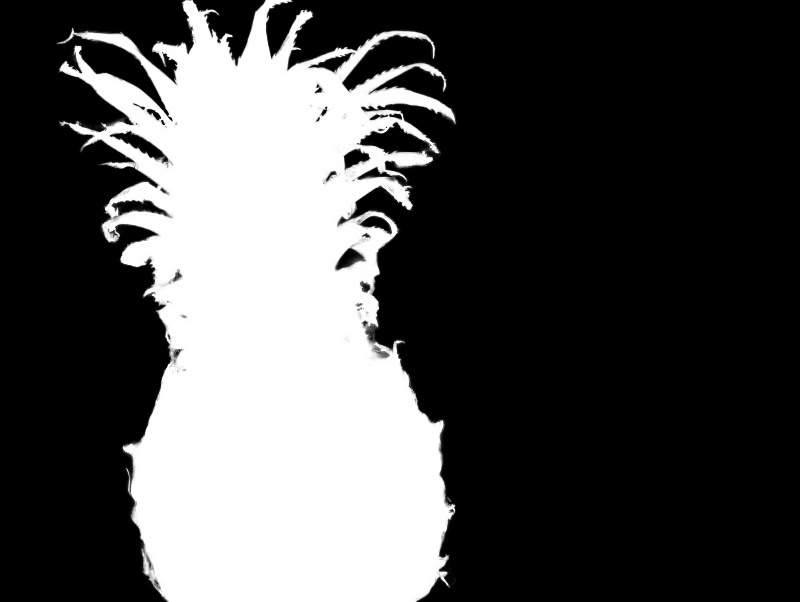}&\includegraphics[width=2cm]{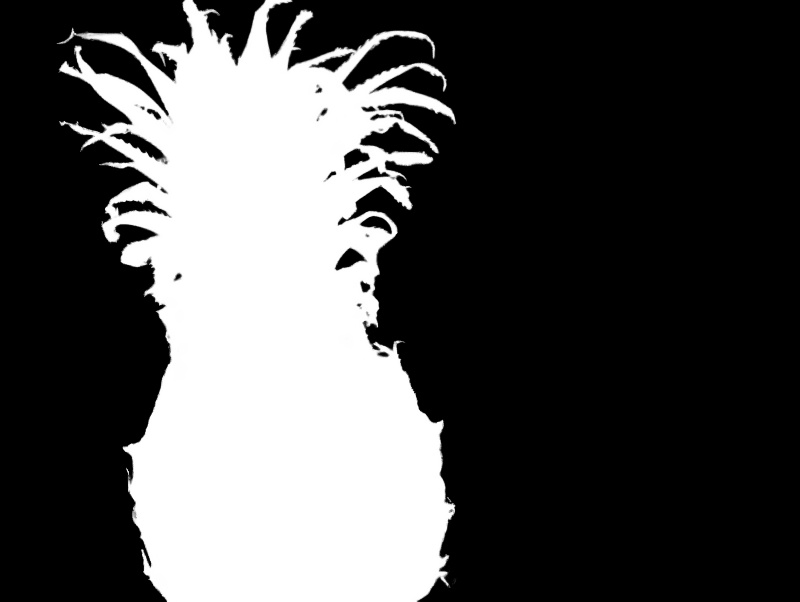}\\
\includegraphics[width=2cm]{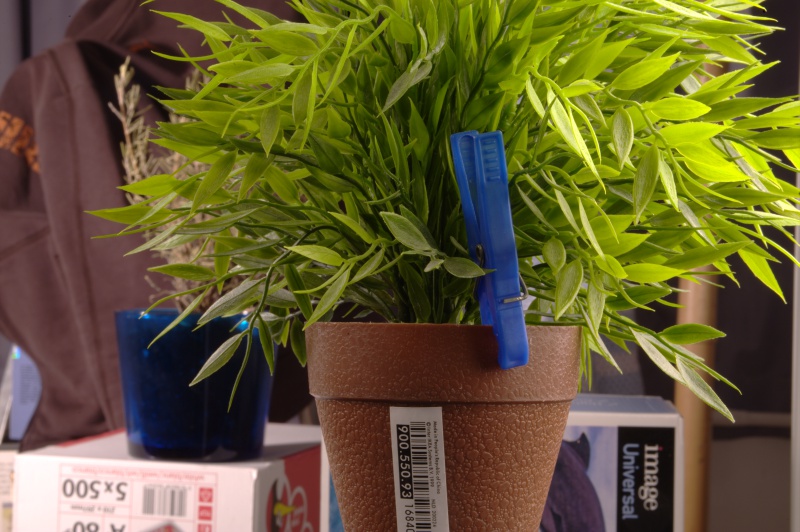}&\includegraphics[width=2cm]{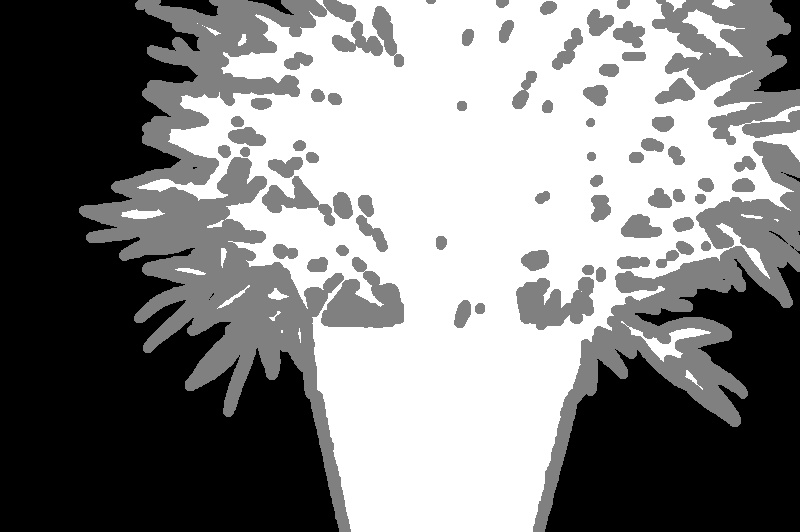}&\includegraphics[width=2cm]{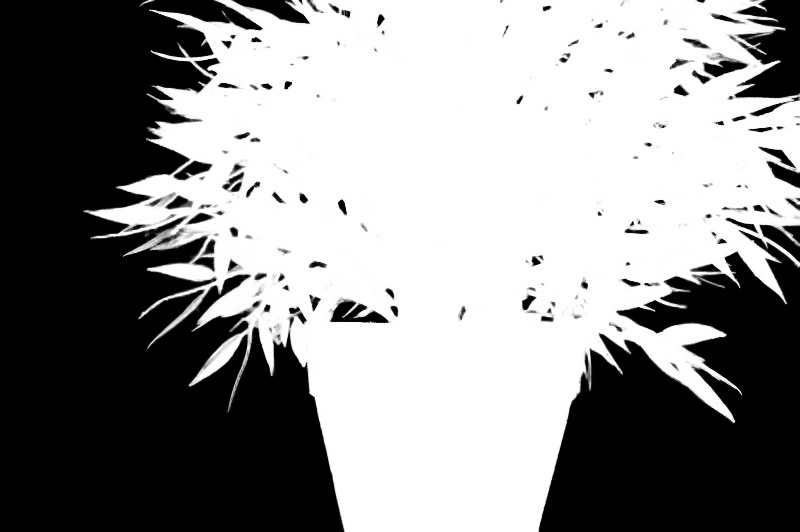}&\includegraphics[width=2cm]{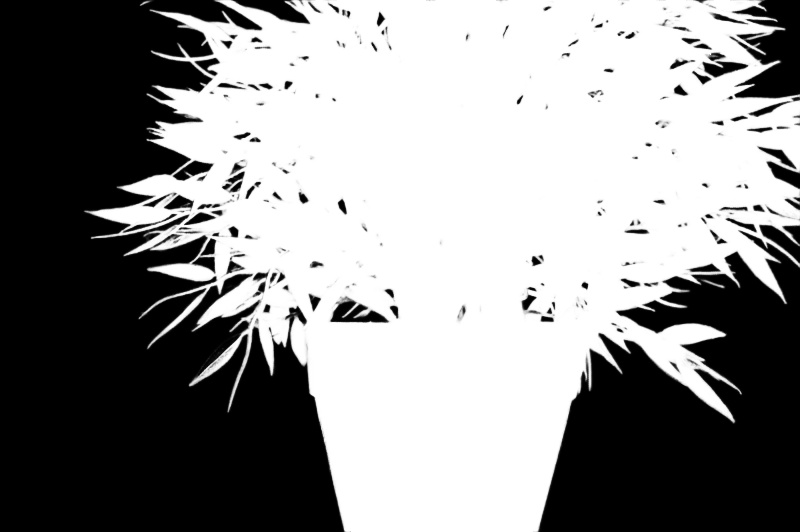}&\includegraphics[width=2cm]{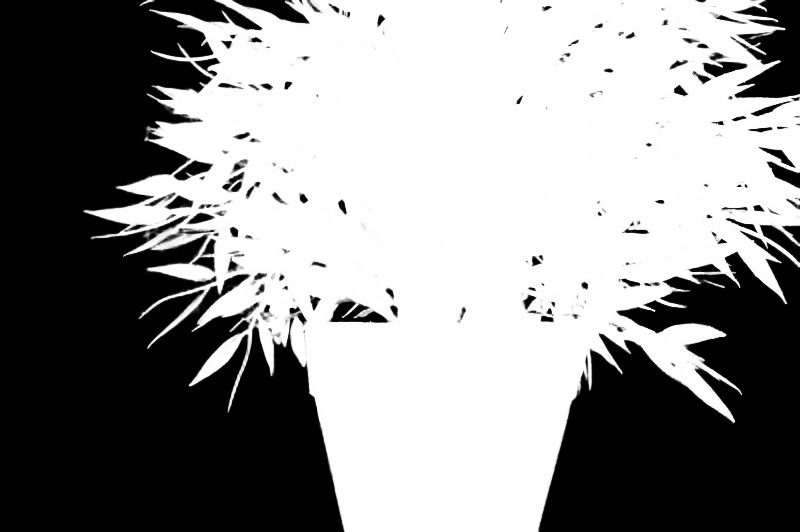}&\includegraphics[width=2cm]{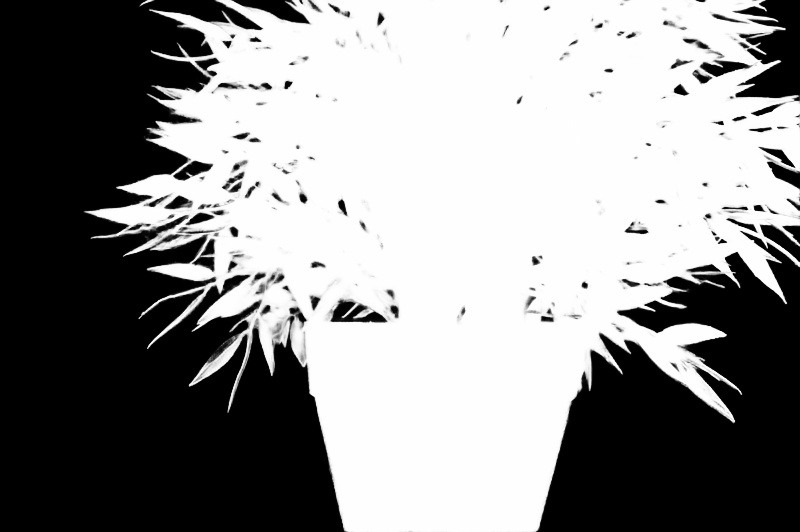}&\includegraphics[width=2cm]{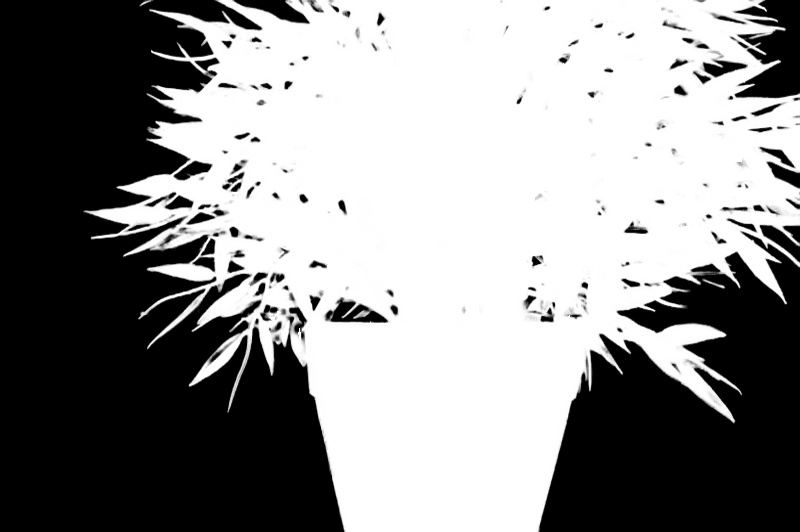}&\includegraphics[width=2cm]{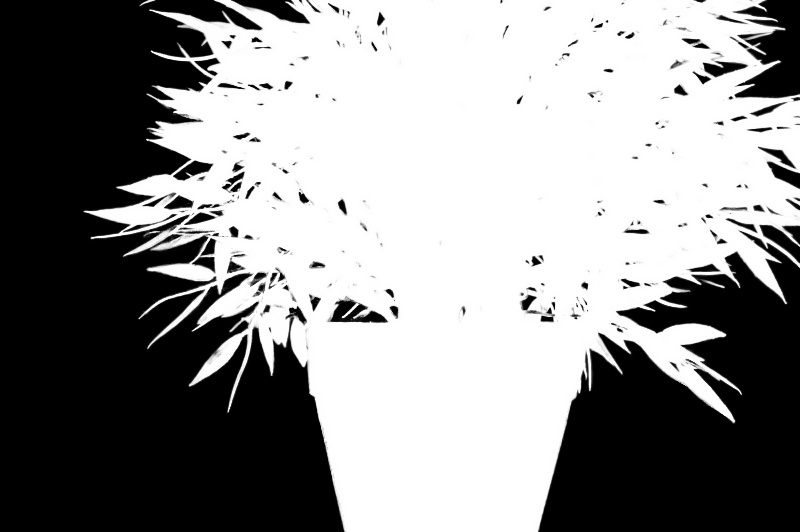}&\includegraphics[width=2cm]{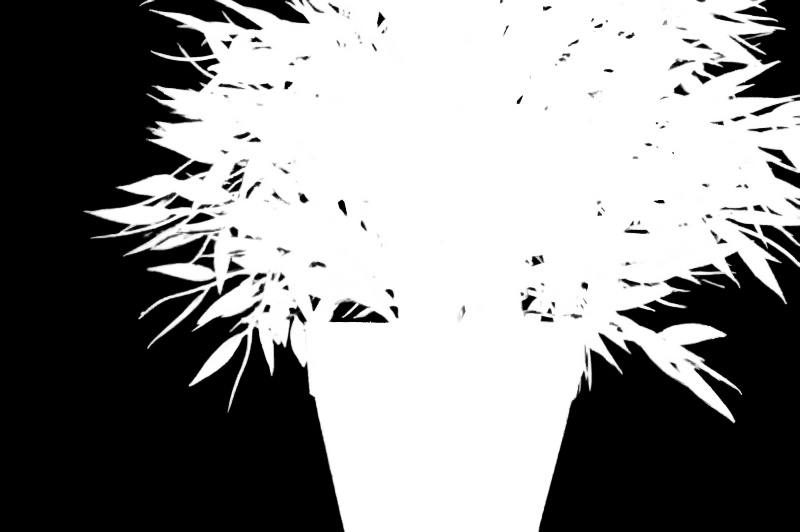}\\
\includegraphics[width=2cm]{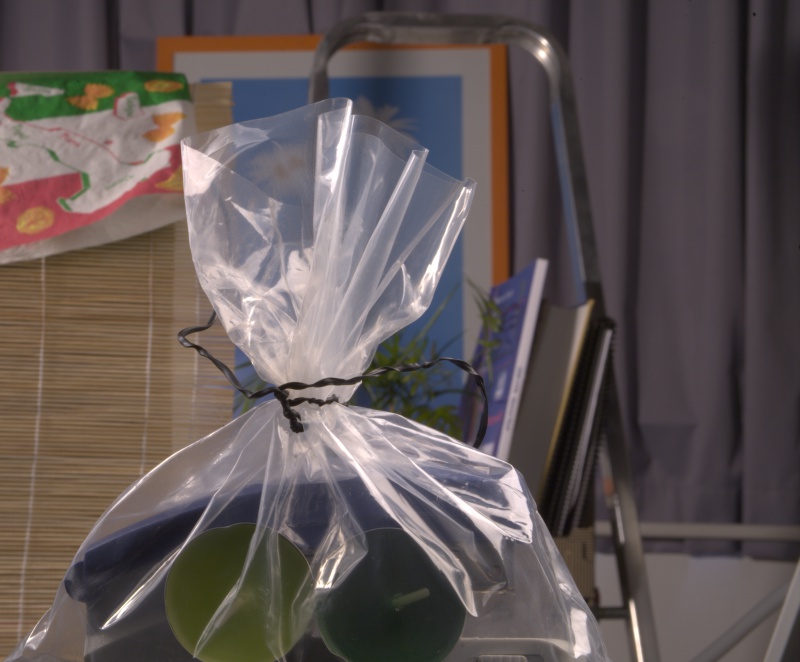}&\includegraphics[width=2cm]{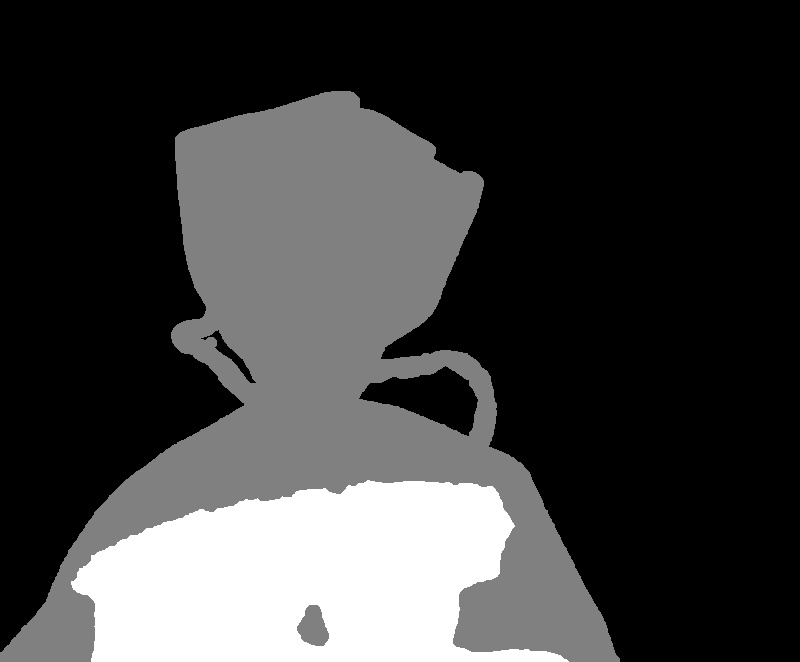}&\includegraphics[width=2cm]{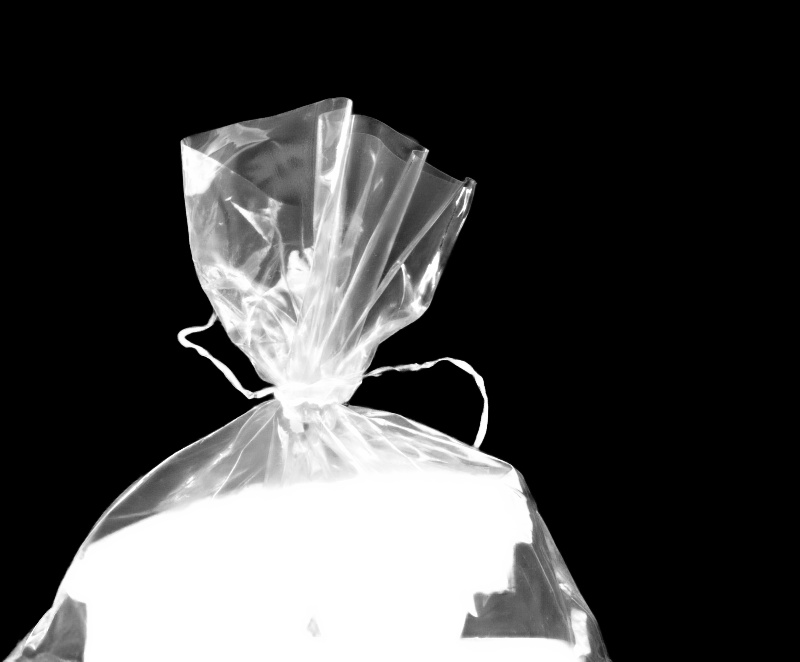}&\includegraphics[width=2cm]{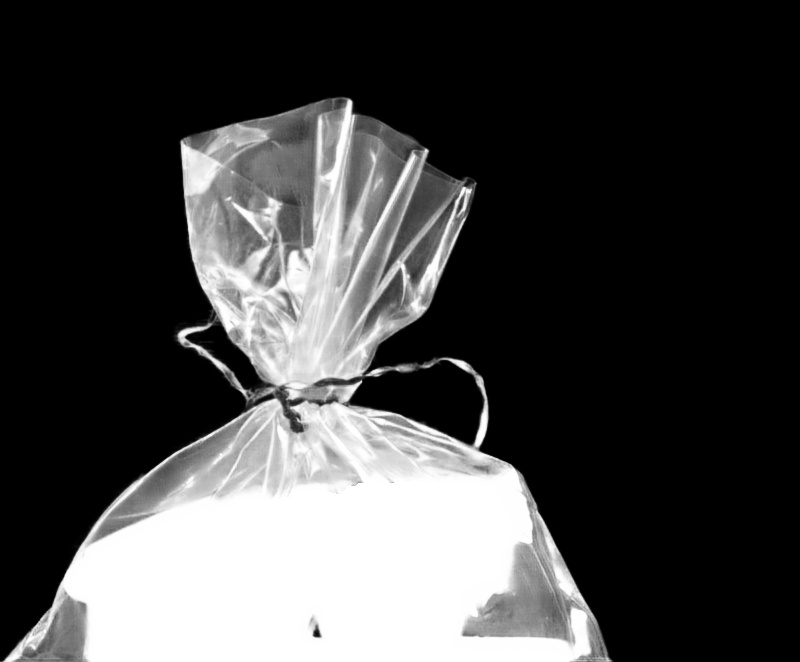}&\includegraphics[width=2cm]{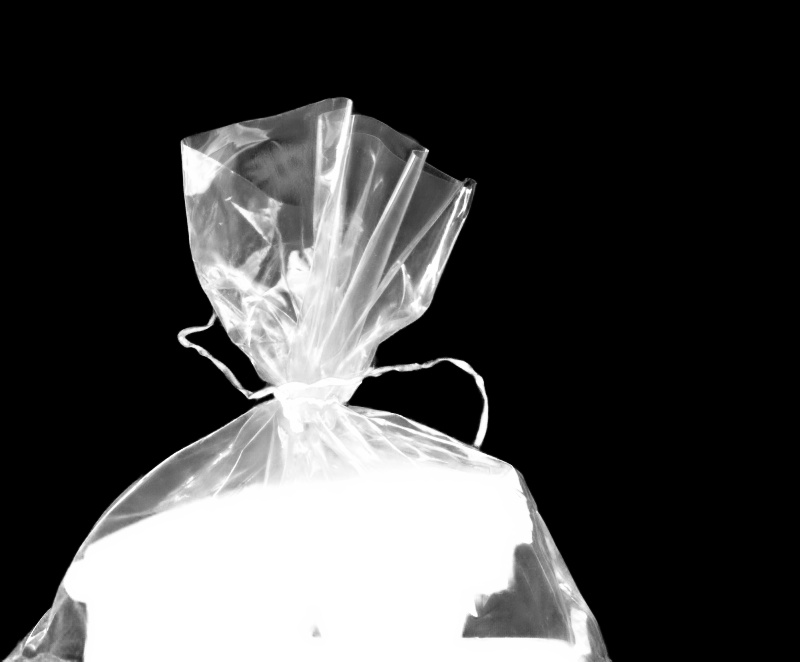}&\includegraphics[width=2cm]{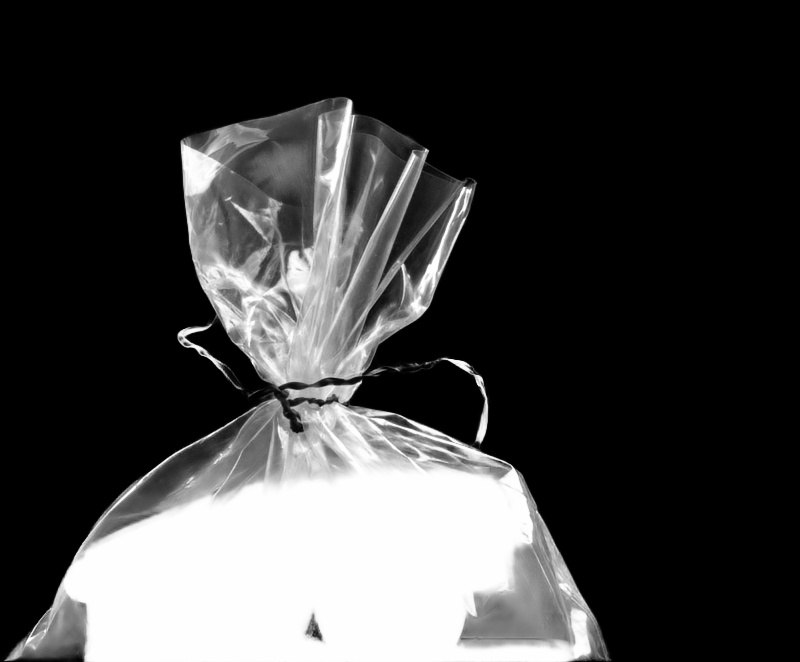}&\includegraphics[width=2cm]{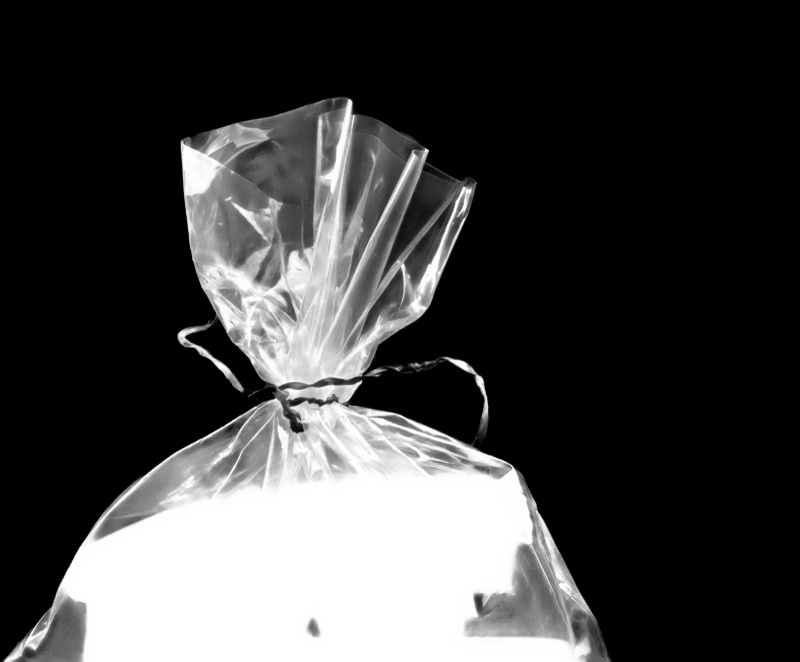}&\includegraphics[width=2cm]{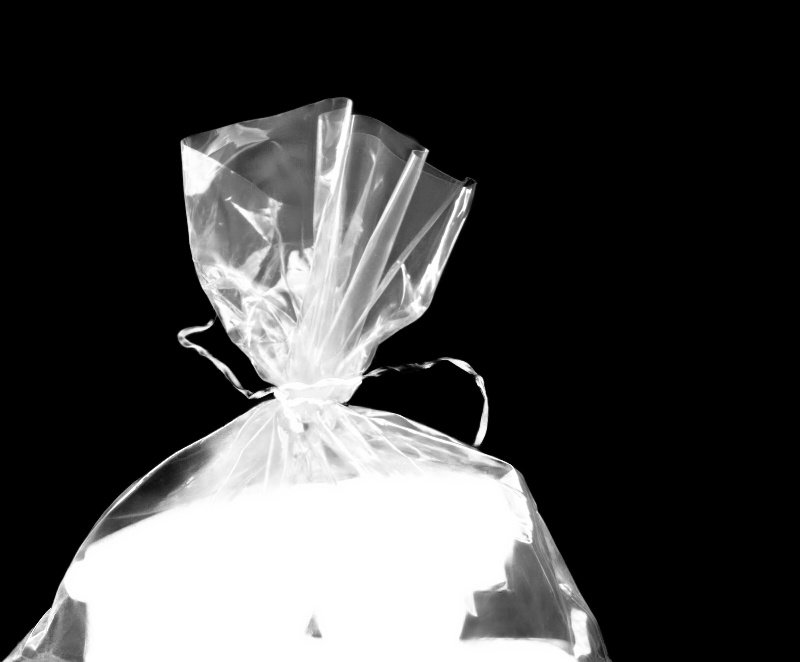}&\includegraphics[width=2cm]{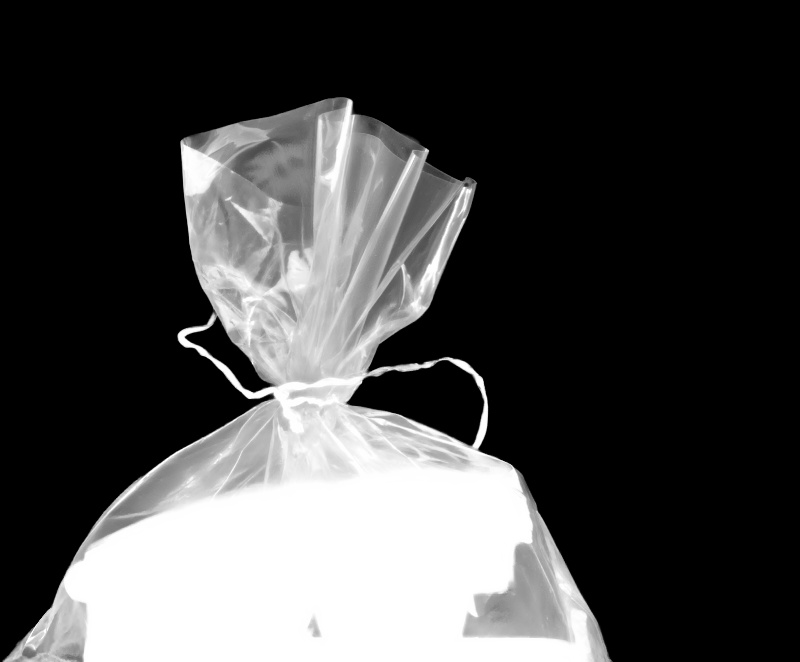}\\
\includegraphics[width=2cm]{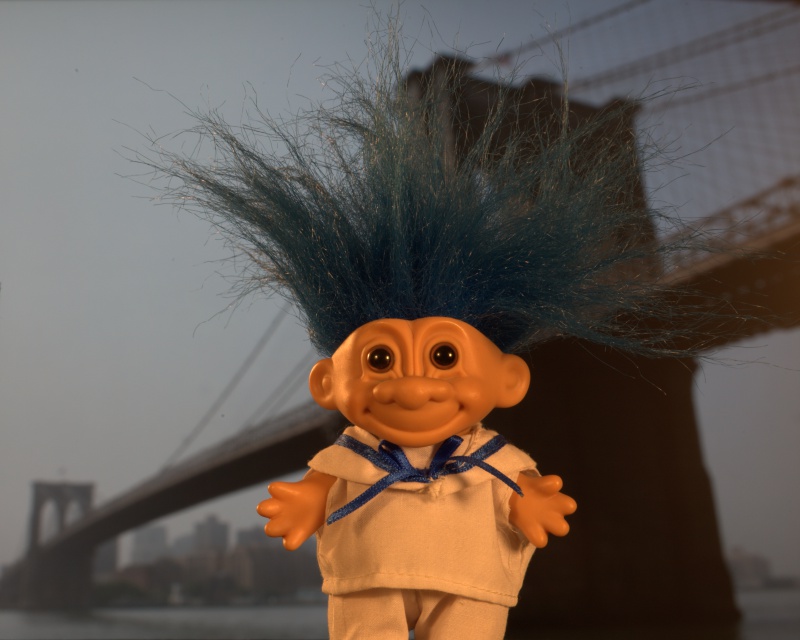}&\includegraphics[width=2cm]{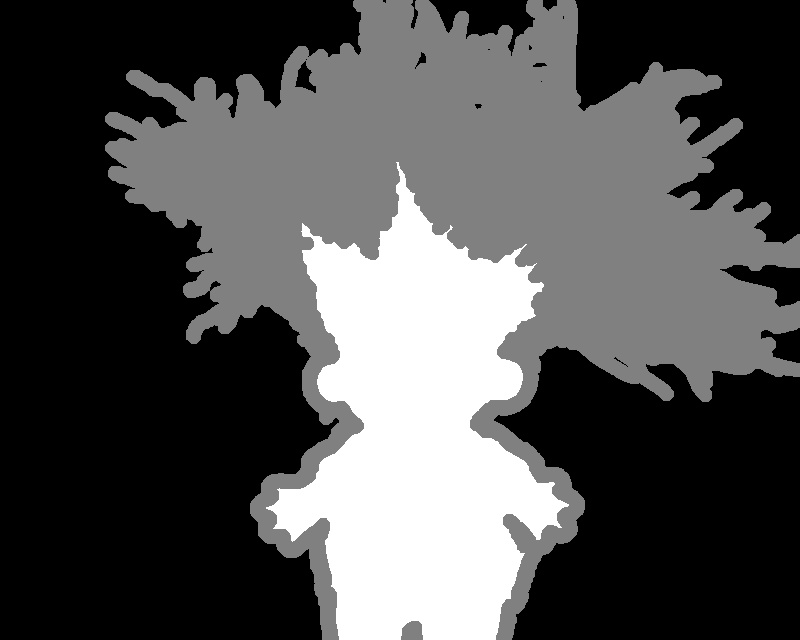}&\includegraphics[width=2cm]{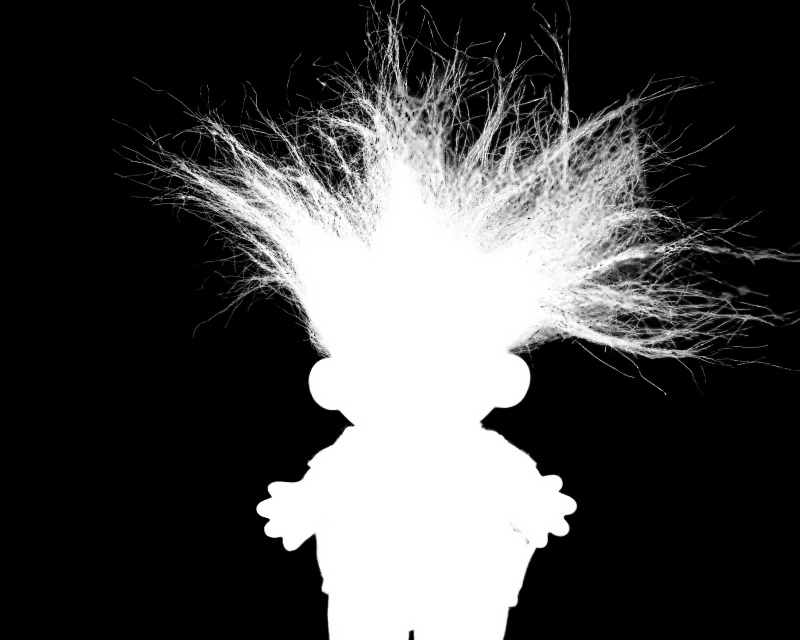}&\includegraphics[width=2cm]{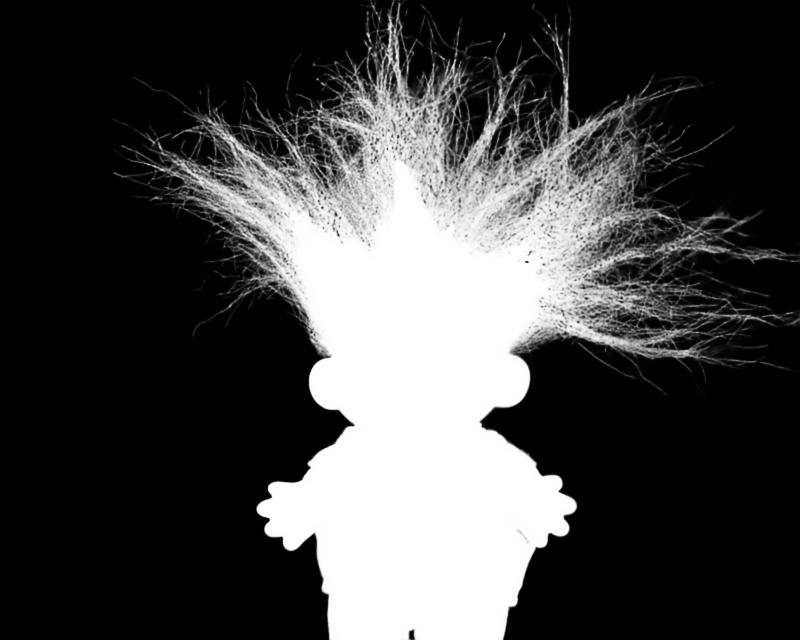}&\includegraphics[width=2cm]{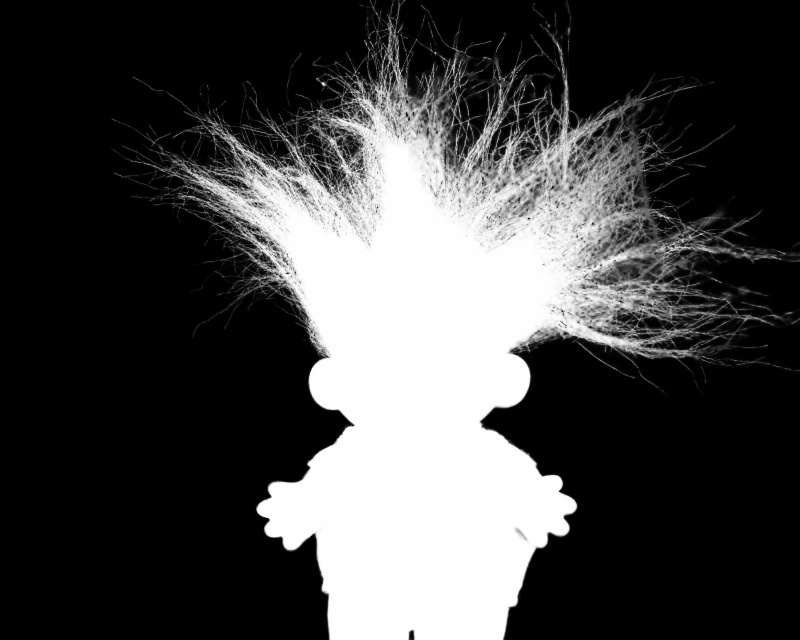}&\includegraphics[width=2cm]{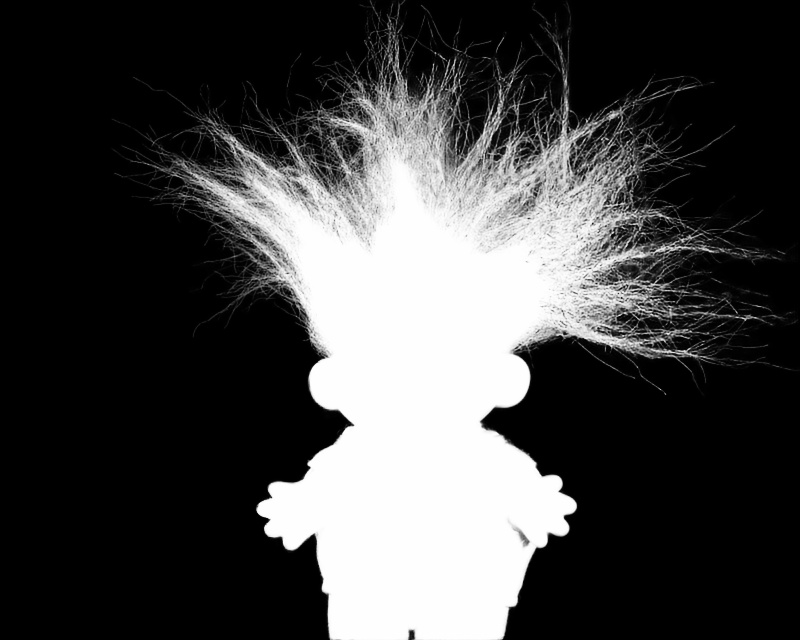}&\includegraphics[width=2cm]{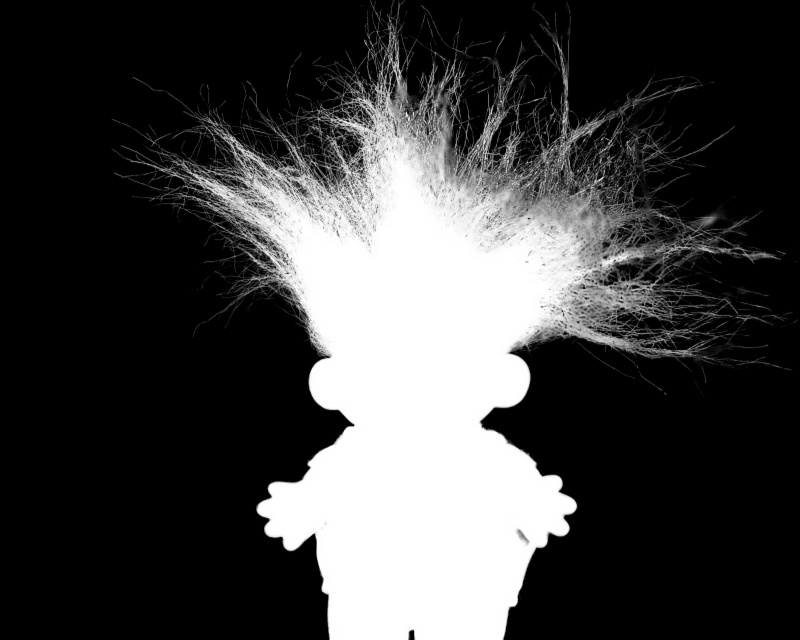}&\includegraphics[width=2cm]{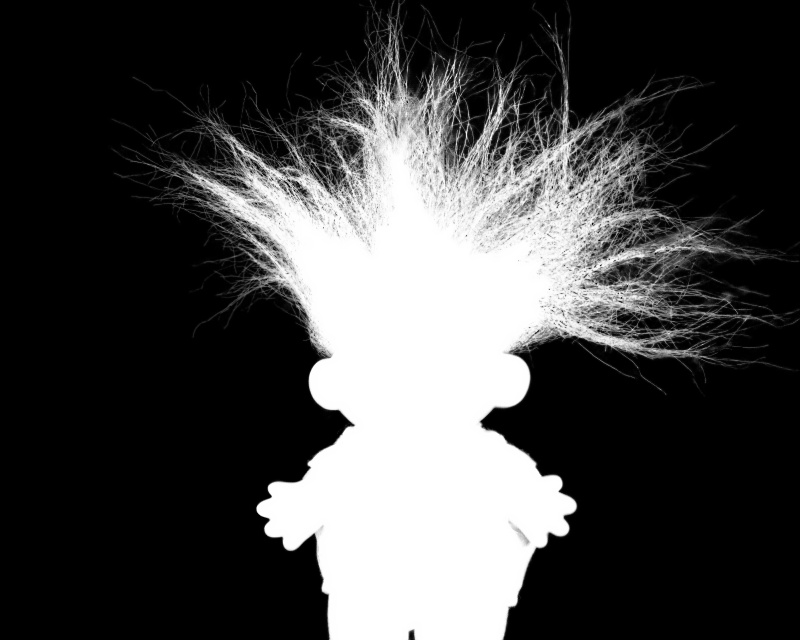}&\includegraphics[width=2cm]{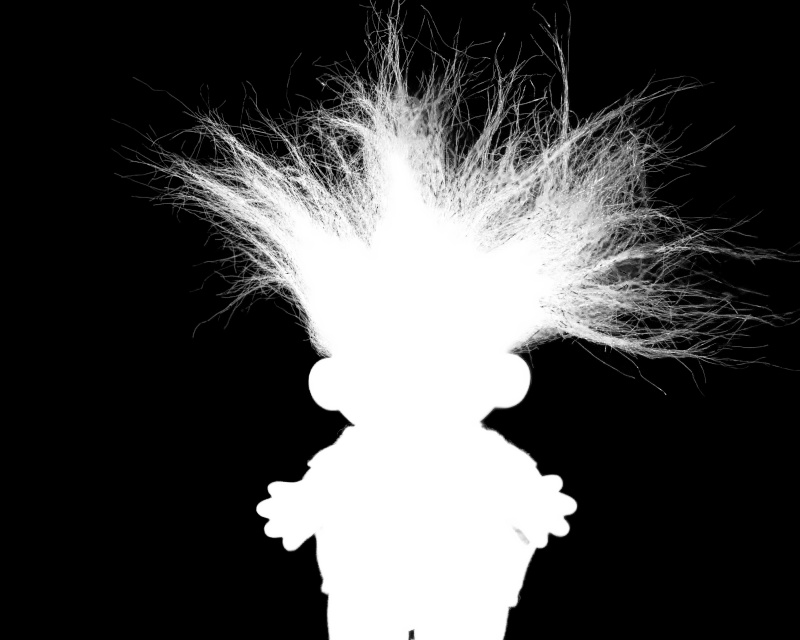}\\
\end{tabular}
\caption{Visual results on alphamatting.com benchmark. From left to right: Image, Trimap (S), HDMatt~\cite{yu2020high}, AdaMatting~\cite{cai2019disentangled}, GCA Matting~\cite{li2020natural}, Context-Aware Matting~\cite{hou2019context}, IndexNet Matting~\cite{lu2019indices}, SampleNet Matting~\cite{tang2019learning}, and Non-local Matting with Refinement (Ours).}
\label{fig:alphamatting} 
\end{center} 
\end{figure*}

\textbf{AlphaMatting.com Benchmark}
The submitted methods on this benchmark~\cite{rhemann2009perceptually} are ranked by averaged ranking over 24 alpha matte estimations according to different metrics. In Table~\ref{tab:alphamatting-sad}, our Non-local Matting with Refinement achieves the best SAD result among state-of-the-art methods on the alphamatting.com benchmark at the time of submission as well as satisfactory performance on the other three metrics at the same time. The Table~\ref{tab:alphamatting-mse}, \ref{tab:alphamatting-gradient} and \ref{tab:alphamatting-connectivity} show MSE, Gradient and Connectivity Error results of several popular methods including ours on alphamatting.com benchmark, which demonstrates promising performance of our approach. Fig.~\ref{fig:alphamatting} represents visual comparison of our approach and other methods on alphamatting.com benchmark.

\textbf{Real Data Adaption}
In this section, details about the real data adaption are provided.

We retrain AIM-pretrained matting pipeline (JI-Real) end-to-end using soft fusion on 22,144 real-world video frames. The Net-T of JI-Real is supervised by pseudo trimap. The pseudo trimap is generated by segmentation~\cite{yang2019parsing} and then erosion of 15 pixels on the foreground and 50 pixels on the background. The soft segmentation for training is produced by erosion on \textit{initial segmentation} of the pseudo trimap with 20 pixels and followed by a Gaussian Blur. The human-finetuned Net-M serves as a teacher for Net-M of JI-Real. The human-finetuned Net-M means finetuning the AIM-pretrained Net-M on combined human images from AIM and Distinctions-646 datasets by the same setting as Net-M except 0.00004 initial learning rate, 10 batch size, and 100,000 iterations. Both pseudo trimap and human-finetuned Net-M serve as teachers for AIM-pretrained Net-T and Net-M. Soft fusion aggregates these two originally separated networks to an end-to-end trimap-free matting network. The end-to-end training and interaction between generator (matting network) and discriminator help generator jump out of the local minimum of human-finetuned Net-M.

In the joint end-to-end training of JI-Real, we follow almost the same settings as Net-T and Net-M, but the learning rate of Net-T is initialized to 0.0001 and the initial learning rate of 0.00004 for Net-M with 6 batch size, 100,000 iterations, and $L_{coarse\_\alpha}+L_{refined\_\alpha}$ as the regression loss calculated on the whole image and combined with adversarial loss.

\begin{figure}[thpb]
\setlength\tabcolsep{1pt}
\renewcommand{\arraystretch}{1}
\begin{center}
\resizebox{\columnwidth}{!}{
\includegraphics[width=8cm]{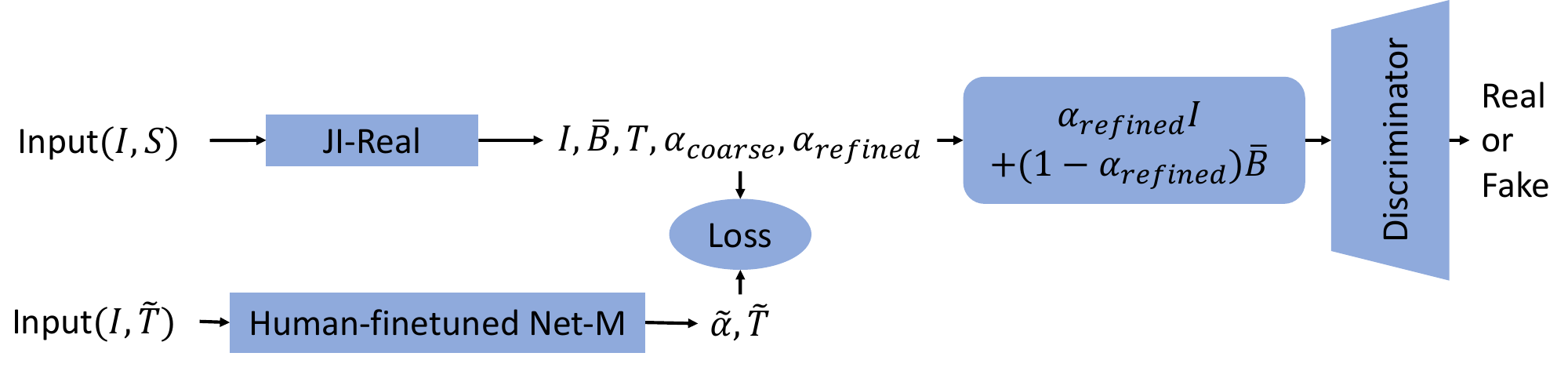}}
\caption{Real Data Adaption Architecture. $S$ is the coarse segmentation of foreground object of image $I$, $\widetilde{T}, \widetilde{\alpha}$ are the pseudo trimap and alpha, and $T,\alpha_{coarse},\alpha_{refined}$ are the predicted trimap, coarse alpha, and refined alpha.}
\label{fig:nettmreal} 
\end{center}
\end{figure}

27 background videos are collected, either self-captured or from Background Matting~\cite{sengupta2020background}, to provide background images $\bar{B}$ for image composition in training. We use the similar GAN framework and the same configuration of discriminator as Background Matting~\cite{mao2017least,sengupta2020background} to train our generator, JI-Real, and discriminator $D$. The overview of real data adaption architecture is shown in Fig.~\ref{fig:nettmreal}. For the generator, we minimize:
\begin{equation}
\begin{aligned}
    \min_{\theta_{Real}}{E_{{X,\bar{B}\sim{p_{X,\bar{B}}}}}[(D(\alpha I+(1-\alpha)\bar{B})-1)^2]}\\
    +{\lambda\{L_{ce}+L_{coarse\_\alpha}+L_{refined\_\alpha}\}},
\end{aligned}
\end{equation}
where $\alpha$=JI-Real$(X,\theta_{Real})$, $\alpha$ means the refined alpha matte, $X$ denotes input image $I$ and its coarse segmentation, $\theta_{Real}$ is the weights of the JI-Real network, $\bar{B}$ is the given composite background image, $\lambda$ is 0.5 and reduced by $\frac{1}{2}$ every 10,000 iterations during training, and $L_{ce}$ is the cross-entropy loss of Net-T. For the discriminator, the loss function is
\begin{equation}
\begin{aligned}
    \min_{\theta_{Disc}}{E_{{X,\bar{B}\sim{p_{X,\bar{B}}}}}[(D(\alpha I+(1-\alpha)\bar{B}))^2]}\\
    +{E_{I\in p_{data}}[(D(I)-1)^2]},
\end{aligned}
\end{equation}
where $\theta_{Disc}$ is the weights of the discriminator network.

For CAM testing, we use an image and its corresponding previous-mentioned pseudo trimap as input.

\begin{table*}[th]
\Huge
  \begin{center}
    \resizebox{0.9\textwidth}{!}{% 
    \begin{tabular}{c|c|ccc|ccc|ccc|ccc|ccc|ccc|ccc|ccc|ccc}
      \toprule % <-- Toprule here
      \multirow{2}{*}{SAD} & \multicolumn{4}{c}{Average Rank}& \multicolumn{3}{c}{Troll} &\multicolumn{3}{c}{Doll}& \multicolumn{3}{c}{Donkey}&\multicolumn{3}{c}{Elephant}&\multicolumn{3}{c}{Plant}&\multicolumn{3}{c}{Pineapple}&\multicolumn{3}{c}{Plastic bag}&\multicolumn{3}{c}{Net}\\
      &Overall&S&L&U&S&L&{U}&{S}&L&U&S&{L}&U&S&L&U&S&L&U&S&L&U&{S}&{L}&{U}&{S}&{L}&{U}\\
      \midrule % <-- Midrule here
      HDMatt\cite{yu2020high}&7.3&9.3&6&6.8&9.5&10&10.7&4.7&4.8&5.8&2.9&3&2.6&1.1&1.2&1.3&5.2&5.9&6.7&2.4&2.6&3.1&17.3&17.3&17&21.5&22.4&23.2\\
      \midrule % <-- Midrule here
      AdaMatting\cite{cai2019disentangled}&9.1&7.9&8.1&11.4&10.2&11.1&10.8&4.9&5.4&6.6&3.6&3.4&3.4&0.9&0.9&1.8&4.7&6.8&9.3&2.2&2.6&3.3&19.2&19.8&18.7&17.8&19.1&18.6\\
      \midrule % <-- Midrule here
      SampleNet Matting\cite{tang2019learning}&9.5&7.6&9.1&11.9&9.1&9.7&9.8&4.3&4.8&5.1&3.4&3.7&3.2&0.9&1.1&2&5.1&6.8&9.7&2.5&4&3.7&18.6&19.3&19.1&20&21.6&23.2\\
      \midrule % <-- Midrule here
      GCA Matting\cite{li2020natural}&10.7&11.8&7.9&12.5&8.8&9.5&11.1&4.9&4.8&5.8&3.4&3.7&3.2&1.1&1.2&1.3&5.7&6.9&7.6&2.8&3.1&4.5&18.3&19.2&18.5&20.8&21.7&24.7\\
      \midrule % <-- Midrule here
      Non-local Matting with Refinement(Ours)&\textbf{6.5}&\textbf{5.8}&\textbf{5.9}&{8}&{7}&{8.1}&{8.1}&{4.2}&{4.4}&{5.4}&{2.6}&{2.6}&{2.3}&{1}&1&{1.2}&{4.9}&{5.8}&{7.6}&2.1&{2.2}&3.2&19.7&22.5&{19.9}&{20.1}&{20.9}&{24.7}\\
      \bottomrule % <-- Bottomrule here
    \end{tabular}
    }
    \caption{Top-5 SAD results on alphamatting.com benchmark, where S, L, U represent the trimap type of small, large and user, and bold numbers show that our Non-local Matting with Refinement achieves the best SAD performance at the time of submission.}
    \label{tab:alphamatting-sad}
  \end{center}
\end{table*}
\begin{table*}[th]
\huge
  \begin{center}
    \resizebox{0.9\textwidth}{!}{% 
    \begin{tabular}{c|c|ccc|ccc|ccc|ccc|ccc|ccc|ccc|ccc|ccc}
      \toprule % <-- Toprule here
      \multirow{2}{*}{MSE} & \multicolumn{4}{c}{Average Rank}& \multicolumn{3}{c}{Troll} &\multicolumn{3}{c}{Doll}& \multicolumn{3}{c}{Donkey}&\multicolumn{3}{c}{Elephant}&\multicolumn{3}{c}{Plant}&\multicolumn{3}{c}{Pineapple}&\multicolumn{3}{c}{Plastic bag}&\multicolumn{3}{c}{Net}\\
      &Overall&S&L&U&S&L&{U}&{S}&L&U&S&{L}&U&S&L&U&S&L&U&S&L&U&{S}&{L}&{U}&{S}&{L}&{U}\\
      \midrule % <-- Midrule here
      HDMatt\cite{yu2020high}&7.5&10.1&6&6.5&0.3&0.3&0.4&0.2&0.2&0.3&0.1&0.1&0.1&0&0&0&0.4&0.4&0.6&0.1&0.2&0.2&0.9&0.9&0.9&0.8&0.8&0.8\\
      \midrule % <-- Midrule here
      AdaMatting\cite{cai2019disentangled}&9.9&7.6&8.9&13.1&0.3&0.4&0.4&0.2&0.2&0.3&0.2&0.2&0.2&0&0&0.1&0.4&0.6&1&0.1&0.2&0.3&1.1&1.2&1.1&0.6&0.6&0.6\\
      \midrule % <-- Midrule here
      SampleNet Matting\cite{tang2019learning}&10.5&7.1&10.6&13.8&0.3&0.3&0.3&0.1&0.2&0.2&0.2&0.2&0.2&0&0&0.1&0.4&0.6&1.2&0.1&0.3&0.3&1.1&1.1&1.2&0.7&0.8&0.8\\
      \midrule % <-- Midrule here
      GCA Matting\cite{li2020natural}&11.7&11.6&10.1&13.3&0.3&0.3&0.4&0.2&0.2&0.3&0.2&0.2&0.2&0&0&0.1&0.5&0.6&0.8&0.2&0.2&0.5&1&1.1&1.1&0.7&0.8&0.9\\
      \midrule % <-- Midrule here
      Non-local Matting with Refinement(Ours)&\textbf{7.5}&\textbf{6.1}&7.1&9.4&{0.2}&{0.2}&{0.3}&{0.1}&{0.2}&{0.2}&{0.1}&{0.1}&{0.1}&{0}&0&{0}&{0.4}&{0.5}&{0.8}&0.1&{0.1}&0.3&1.1&1.3&{1.1}&{0.7}&{0.8}&{0.9}\\
      \bottomrule % <-- Bottomrule here
    \end{tabular}
    }
    \caption{Top-5 MSE results on AlphaMatting benchmark, S, L, U represent the trimap type of small, large and user at the time of submission.}
    \label{tab:alphamatting-mse}
  \end{center}
\end{table*}
\begin{table*}[th]
\huge
  \begin{center}
    \resizebox{0.9\textwidth}{!}{% 
    \begin{tabular}{c|c|ccc|ccc|ccc|ccc|ccc|ccc|ccc|ccc|ccc}
      \toprule % <-- Toprule here
      \multirow{2}{*}{Gradient} & \multicolumn{4}{c}{Average Rank}& \multicolumn{3}{c}{Troll} &\multicolumn{3}{c}{Doll}& \multicolumn{3}{c}{Donkey}&\multicolumn{3}{c}{Elephant}&\multicolumn{3}{c}{Plant}&\multicolumn{3}{c}{Pineapple}&\multicolumn{3}{c}{Plastic bag}&\multicolumn{3}{c}{Net}\\
      &Overall&S&L&U&S&L&{U}&{S}&L&U&S&{L}&U&S&L&U&S&L&U&S&L&U&{S}&{L}&{U}&{S}&{L}&{U}\\
      \midrule % <-- Midrule here
      HDMatt\cite{yu2020high}&5.6&6.5&4&6.4&0.2&0.2&0.2&0.1&0.1&0.3&0.1&0.1&0.2&0.2&0.2&0.3&1.1&1.2&1.6&0.6&0.6&0.9&0.5&0.5&0.6&0.3&0.4&0.4\\
      \midrule % <-- Midrule here
      AdaMatting\cite{cai2019disentangled}&9.4&5.9&7&15.4&0.2&0.2&0.2&0.1&0.1&0.4&0.2&0.2&0.2&0.1&0.1&0.3&1.1&1.4&2.3&0.4&0.6&0.9&0.9&1&0.9&0.3&0.4&0.4\\
      \midrule % <-- Midrule here
      GCA Matting\cite{li2020natural}&9.7&9.9&8.1&11&0.1&0.1&0.2&0.1&0.1&0.3&0.2&0.2&0.2&0.2&0.2&0.3&1.3&1.6&1.9&0.7&0.8&1.4&0.6&0.7&0.6&0.4&0.4&0.4\\
      \midrule
      Context-aware Matting\cite{hou2019context}&10.9&12.4&11.6&8.6&0.2&0.2&0.2&0.1&0.2&0.2&0.2&0.2&0.2&0.2&0.4&0.4&1.4&1.5&1.8&0.8&1.3&1&1.1&1.1&0.9&0.4&0.4&0.4\\
      \midrule % <-- Midrule here
      Non-local Matting with Refinement(Ours)&6.7&{5.6}&5.5&8.9&{0.2}&{0.2}&{0.2}&{0.1}&{0.1}&{0.2}&{0.1}&{0.1}&{0.2}&{0.2}&0.2&{0.3}&{1}&{1.3}&{2}&0.5&{0.5}&1&0.5&0.5&{0.5}&{0.4}&{0.5}&{0.4}\\
      % \midrule % <-- Midrule here
      \bottomrule % <-- Bottomrule here
    \end{tabular}
    }
    \caption{Top-5 Gradient results on AlphaMatting benchmark, S, L, U represent the trimap type of small, large and user at the time of submission.}
    \label{tab:alphamatting-gradient}
  \end{center}
\end{table*}
\begin{table*}[th]
\huge
  \begin{center}
    \resizebox{0.9\textwidth}{!}{% 
    \begin{tabular}{c|c|ccc|ccc|ccc|ccc|ccc|ccc|ccc|ccc|ccc}
      \toprule % <-- Toprule here
      \multirow{2}{*}{Connectivity Error} & \multicolumn{4}{c}{Average Rank}& \multicolumn{3}{c}{Troll} &\multicolumn{3}{c}{Doll}& \multicolumn{3}{c}{Donkey}&\multicolumn{3}{c}{Elephant}&\multicolumn{3}{c}{Plant}&\multicolumn{3}{c}{Pineapple}&\multicolumn{3}{c}{Plastic bag}&\multicolumn{3}{c}{Net}\\
      &Overall&S&L&U&S&L&{U}&{S}&L&U&S&{L}&U&S&L&U&S&L&U&S&L&U&{S}&{L}&{U}&{S}&{L}&{U}\\
      
      \midrule % <-- Midrule here
      % Deep Matting\cite{yu2020high}&5.6&6.5&4&6.4&0.2&0.2&0.2&0.1&0.1&0.3&0.1&0.1&0.2&0.2&0.2&0.3&1.1&1.2&1.6&0.6&0.6&0.9&0.5&0.5&0.6&0.3&0.4&0.4\\
      % \midrule % <-- Midrule here
      % HDMatt\cite{yu2020high}&5.6&6.5&4&6.4&0.2&0.2&0.2&0.1&0.1&0.3&0.1&0.1&0.2&0.2&0.2&0.3&1.1&1.2&1.6&0.6&0.6&0.9&0.5&0.5&0.6&0.3&0.4&0.4\\
      % \midrule % <-- Midrule here
      GCA Matting\cite{li2020natural}&19.7&22.9&17.3&18.9&1.1&1.1&1&0.2&0.2&0.2&0.2&0.2&0.2&0&0&0&0.1&0.1&0.1&0&0.1&0.1&1.1&1.3&1.3&1.9&1.5&1.6\\
      \midrule
      AdaMatting\cite{cai2019disentangled}&21&19&23.1&20.8&1.1&1.1&1.1&0.1&0.2&0.2&0.2&0.2&0.2&0&0&0&0.1&0.1&0.1&0&0&0.1&6.8&13.3&1.4&1.3&1.3&1.3\\
      \midrule % <-- Midrule here
      IndexNet Matting\cite{lu2019indices}&23.1&21.7&24.5&23&1.3&1.3&1.3&0.1&0.1&0.1&0.2&0.2&0.2&0&0&0&0.1&0.1&0.3&0&0.1&0.2&10&10.7&1.8&1.6&1.7&1.6\\
      \midrule
      SampleNet Matting\cite{tang2019learning}&23.4&25&21&24.5&0.9&0.9&0.8&0.1&0.1&0.1&0.2&0.2&0.2&0&0&0&0.1&0.1&0.2&0&0.1&0.2&1.5&1.5&1.8&3.8&3.9&3.8\\
      \midrule % <-- Midrule here
      Non-local Matting with Refinement(Ours)&\textbf{17.3}&{19.7}&\textbf{14.7}&\textbf{17.4}&{0.9}&{0.9}&{0.9}&{0.1}&{0.1}&{0.1}&{0.2}&{0.2}&{0.2}&{0}&0&{0}&{0.1}&{0.1}&{0.2}&0&{0}&0&0.9&0.7&{0.9}&{3.2}&{3.3}&{3.2}\\
      % \midrule % <-- Midrule here
      \bottomrule % <-- Bottomrule here
    \end{tabular}
    }
    \caption{Popular matting approach results of Connectivity Error on AlphaMatting benchmark, S, L, U represent the trimap type of small, large and user at the time of submission.}
    \label{tab:alphamatting-connectivity}
  \end{center}
\end{table*}
% \subsubsection{Matting Results}
% \subsection{Additional Composition-1k Test Dataset Matting Results}
% Table~\ref{tab:extrahuman} shows results of our JI and Background Matting~\cite{sengupta2020background} on the Composition-240 test dataset. For Late Fusion~\cite{zhang2019late} and HAttMatting~\cite{qiao2020attention}, the authors did not release their pretrained models for the Composition-1k test dataset, so there is no comparison.

\begin{figure*}[thpb]
  \begin{center}
    \begin{tabular}{c|c|c}
      \toprule
      Block Name & Output Size & Details\\
      \midrule
      \multicolumn{3}{c}{Encoder}\\
      \midrule
      Conv+BN+ReLU & $256\times256\times64$ & $7\times7$, 64, stride 2, Bias=False\\
      \midrule
       Layer1 & $128\times128\times256$ & $3\times3$ max pool, stride=2; Block$\times3$, stride=[1,1,1], atrous=[1,1,1]\\
       \midrule
       Layer2 & $64\times64\times512$ & Block$\times4$, stride=[2,1,1,1], atrous=[1,1,1,1]\\
       \midrule
       Layer3 & $32\times32\times1024$ & Block$\times6$, stride=[2,1,1,1,1,1], atrous=[1,1,1,1,1,1]\\
       \midrule
       Layer4 & $32\times32\times2048$ & Block$\times3$, stride=[1,1,1], atrous=[1,2,1]\\
       \midrule
      \multicolumn{3}{c}{Decoder}\\
      \midrule
       ASPP+Dropout & $32\times32\times256$ & rate of ASPP=1, dropout ratio=0.5; input: output of Layer4\\
       \midrule
       Layer1-Shortcut(Conv2+BN+ReLU) & $128\times128\times48$ &$1\times1$, 48, stride=1, Bias=True; input: output of Layer1\\
      \midrule
       Layer2-Shortcut& $128\times128\times48$ &$1\times1$, 48, stride=1, Bias=True;\\
       (Conv3+BN+ReLU) & &input: $2\times2$ bilinear upsampling from Layer2\\
        \midrule
        Concatenation Conv& &$3\times3$, 256, stride=1, Bias=True; $3\times3$, 256, stride 1, \\
        (Conv4+BN+ReLU+& & Bias=True; Dropout ratio2=0.1; input: concatenation \\
        Dropout+Conv5+BN+& $128\times128\times48$& of $4\times4$ bilinear upsampling from ASPP+Dropout, \\
        ReLU+Dropout)& & output of Layer2-Shortcut \\
        & & and Layer1-Shortcut\\
       \midrule
       Output Conv& $512\times512\times3$ &$1\times1$, 3, stride=1, Bias=True; \\
       (Conv6)& & input: $4\times4$ bilinear upsampling from Concatenation Conv\\
       \bottomrule
    \end{tabular}
    \caption{Trimap Generation Network. Block is the Bottleneck of ResNet-atrous from \small{\url{https://download.pytorch.org/models/resnet50-19c8e357.pth}}}
    \label{fig:nettstructure}
  \end{center}
\end{figure*}

\begin{figure*}[thpb]
  \begin{center}
    \begin{tabular}{c|c|c}
      \toprule
      Block Name & Output Size & Details\\
      \midrule
      \multicolumn{3}{c}{Encoder}\\
      \midrule
      Stride Conv1+SN+BN & $256\times256\times32$ & $3\times3$, 32, stride 2, Bias=False\\
      \midrule
      ShortCut1(2*(Conv+SN+ReLU+BN)) & $512\times512\times32$ & $3\times3$, 32, stride=1, Bias=False; input: original image\\
      \midrule
       Conv1+SN+BN & $256\times256\times32$ & $3\times3$, 32, stride 1, Bias=False\\
       \midrule
      ShortCut2(2*(Conv+SN+ReLU+BN)) & $256\times256\times32$ & $3\times3$, 32, stride=1, Bias=False; input: output of Conv1+SN+BN\\
       \midrule
       Stride Conv2+SN+BN & $128\times128\times64$ & $3\times3$, 64, stride 2, Bias=False; input: output of Conv1+SN+BN\\
       \midrule
       ResBlocks1 & $128\times128\times64$ & Block$\times3$, stride=[1,1,1]\\
       \midrule
      ShortCut3(2*(Conv+SN+ReLU+BN)) & $128\times128\times64$ & $3\times3$, 64, stride=1, Bias=False; input: output of ResBlocks1\\
       \midrule
        Downsample ResBlock1& $64\times64\times128$ & Block$\times1$, stride=[2]\\
        \midrule
        ResBlocks2 & $64\times64\times128$ & Block$\times3$, stride=[1,1,1]\\
         \midrule
      Non-local Block & $64\times64\times128$ & input: image and alpha feature from ResBlocks2\\
        \midrule
      ShortCut4(2*(Conv+SN+ReLU+BN)) & $64\times64\times128$ & $3\times3$, 128, stride=1, Bias=False; input: output of Non-local Block\\
       \midrule
       Downsample ResBlock2& $32\times32\times256$ & Block$\times1$, stride=[2]\\
       \midrule
       ResBlocks2& $32\times32\times256$ & Block$\times3$, stride=[1,1,1]\\
       \midrule
      ShortCut5(2*(Conv+SN+ReLU+BN)) & $32\times32\times256$ & $3\times3$, 256, stride=1, Bias=False; input: output of ResBlocks2\\
      \midrule
       Downsample ResBlock3& $16\times16\times512$ & Block$\times1$, stride=[2]\\
       \midrule
       ResBlock3& $16\times16\times512$ & Block$\times1$, stride=[1]\\
       \midrule
       \multicolumn{3}{c}{Decoder}\\
      \midrule
       Upsample ResBlock1& $32\times32\times256$ & Block$\times1$, stride=[2]\\
       \midrule
       ResBlocks4& $32\times32\times256$ & Block$\times1$, stride=[1]\\
       \midrule
       Upsample ResBlock2& $64\times64\times128$ & Block$\times1$, stride=[2]; input: the\\
       & & summation of output of ShortCut5 and ResBlocks4\\
       \midrule
       ResBlocks5& $64\times64\times128$ & Block$\times2$, stride=[1,1]\\
       \midrule
       Upsample ResBlock3& $128\times128\times64$ & Block$\times1$, stride=[2]; input: the\\
       & & summation of output of ShortCut4 and ResBlocks5\\
       \midrule
       ResBlocks6& $128\times128\times64$ & Block$\times2$, stride=[1,1]\\
       \midrule
       Upsample ResBlock4& $256\times256\times32$ & Block$\times1$, stride=[2]; input: the\\
       & & summation of output of ShortCut3 and ResBlocks6\\
       \midrule
       ResBlocks7& $256\times256\times32$ & Block$\times1$, stride=[1]\\
       \midrule
       Deconv1+SN+BN& $512\times512\times32$ & $4\times4$, 32, stride=2, bias=False; input: the \\
       & & summation of output of ShortCut2 and ResBlocks7\\
       \midrule
       Conv2+SN+BN& $512\times512\times1$ & $3\times3$, 1, stride=1, bias=True; input: the \\
       & & summation of output of ShortCut1 and Deconv1+SN+BN\\
       \bottomrule
    \end{tabular}
    \caption{Non-local Matting. ResBlock, ResBlock with downsampling and upsampling are shown in following figures.}
    \label{fig:nonlocalmattingstructure}
  \end{center}
\end{figure*}

\begin{figure*}[thpb]
    \centering
    \begin{tabular}{c|c}
     \toprule
    Block Name & Details\\
    \midrule
    Conv1+SN+BN+ReLU& $3\times3$, stride=1, bias=False\\
    \midrule
    Conv2+SN+BN& $3\times3$, stride=1, bias=False\\
    \midrule
    ReLu&input: the summation of original block input and output of Conv2+SN+BN\\
    \bottomrule
    \end{tabular}
    \caption{ResBlock in Encode. In the decoder, ReLU is replaced with LeakyReLU.}
    \label{fig:ResBlock}
\end{figure*}

\begin{figure*}[thpb]
    \centering
    \begin{tabular}{c|c}
    \toprule
    Block Name & Details\\
    \midrule
    Conv1+SN+BN+ReLU& $3\times3$, stride=2, bias=False\\
    \midrule
    Conv2+SN+BN& $3\times3$, stride=1, bias=False\\
    \midrule
    Downsampling Layer& $2\times2$ Avg Pool; input: original block input\\
    \midrule
    ReLU&input: the summation of output of Downsampling Layer and Conv2+SN+BN\\
    \bottomrule
    \end{tabular}
    \caption{ResBlock with Downsampling}
    \label{fig:ResBlockDown}
\end{figure*}

\begin{figure*}[thpb]
    \centering
    \begin{tabular}{c|c}
    \toprule
    Block Name & Details\\
    \midrule
    DeConv1+SN+BN+LeakyReLU& $4\times4$, stride=2, bias=False\\
    \midrule
    Conv2+SN+BN& $3\times3$, stride=1, bias=False\\
    \midrule
    Upsampling Layer& Nearest Upsampling; input: original block input\\
    \midrule
    LeakyReLU&input: the summation of output of Upsampling Layer and Conv2+SN+BN\\
    \bottomrule
    \end{tabular}
    \caption{ResBlock with Upsampling}
    \label{fig:ResBlockUp}
\end{figure*}

\begin{figure*}[thpb]
  \begin{center}
    \begin{tabular}{c|c|c}
      \toprule
      Block Name & Output Size & Details\\
      \midrule
      \multicolumn{3}{c}{Encoder}\\
      \midrule
      Conv0 & $512\times512\times1$ & $3\times3$, 1, stride=1, Bias=True\\
      \midrule
       Conv1+BN+ReLU+MaxPool & $256\times256\times64$ & $3\times3$, 64, stride=1, Bias=True; $2\times2$ Max Pool\\
       \midrule
        Conv2+BN+ReLU+MaxPool & $128\times128\times64$ & $3\times3$, 64, stride=1, Bias=True; $2\times2$ Max Pool\\
       \midrule
        Conv3+BN+ReLU+MaxPool & $64\times64\times64$ & $3\times3$, 64, stride=1, Bias=True; $2\times2$ Max Pool\\
       \midrule
         Conv4+BN+ReLU+MaxPool & $32\times32\times64$ & $3\times3$, 64, stride=1, Bias=True; $2\times2$ Max Pool\\
         \midrule
       \multicolumn{3}{c}{Decoder}\\
        \midrule
         Conv5+BN+ReLU+Upsample & $64\times64\times64$ & $3\times3$, 64, stride=1, Bias=True; $2\times2$ bilinear upsampling;  \\
          &  & input: output of Conv4+BN+ReLU+MaxPool\\
       \midrule
       &  & $3\times3$, 64, stride=1, Bias=True; $2\times2$ bilinear upsampling; \\
          Conv6+BN+ReLU+Upsample& $128\times128\times64$ & input: concatnation of output of \\
          & & Conv5+BN+ReLU+Upsample and Conv4+BN+ReLU+MaxPool\\
       \midrule
       &  & $3\times3$, 64, stride=1, Bias=True; $2\times2$ bilinear upsampling; \\
        Conv7+BN+ReLU+Upsample  & $256\times256\times64$ & input: concatnation of output of\\
          & & Conv6+BN+ReLU+Upsample and Conv3+BN+ReLU+MaxPool\\
       \midrule
       & & $3\times3$, 64, stride=1, Bias=True; $2\times2$ bilinear upsampling; \\
        Conv8+BN+ReLU+Upsample& $512\times512\times64$ & input: concatnation of output of\\
          & & Conv7+BN+ReLU+Upsample and Conv2+BN+ReLU+MaxPool\\
       \midrule
       & & $3\times3$, 64, stride=1, Bias=True; $2\times2$ bilinear upsampling;\\
         Conv9+BN+ReLU &$512\times512\times64$  & input: concatnation of output of\\
          & & Conv8+BN+ReLU+Upsample and Conv1+BN+ReLU+MaxPool\\
        \midrule
       Conv10& $512\times512\times1$ & $3\times3$, stride=1, bias=True; input: output of Conv9+BN+ReLU\\
       \midrule
       Residual& $512\times512\times1$ & summation of output of Conv10 and original network input coarse alpha\\
       \bottomrule
    \end{tabular}
    \caption{Refinement Module}
    \label{fig:refinementstructure}
  \end{center}
\end{figure*}

\end{document}